\documentclass[preprint,12pt]{elsarticle}
\usepackage[margin=2.5cm]{geometry}
\usepackage{newtxtext,newtxmath}
\usepackage{graphicx}
\usepackage{setspace}
\usepackage{url}
\usepackage{lineno}
\usepackage{titlesec} % optional
\usepackage{longtable}
\usepackage{tabularx}
\usepackage{multirow}
\usepackage{makecell}
\usepackage{booktabs}
\usepackage{array}
\usepackage{float}
\usepackage{ragged2e}
\usepackage{caption}
\usepackage{hyperref}
\usepackage{amsmath}
\usepackage{ltablex}
\usepackage{capt-of}
\usepackage{appendix}
\usepackage{chngcntr} % for figure/table numbering
\usepackage{etoolbox} % for patching commands
\usepackage[resetlabels,labeled]{multibib}

\newcites{Sup}{Supplementary References}

\newcommand{\opus}[1]{%
	\begingroup
	\spaceskip=\fontdimen2\font plus \fontdimen3\font minus \fontdimen4\font
	\xspaceskip=\fontdimen7\font\relax
	\ttfamily
	#1%
	\endgroup
}

\makeatletter
\@addtoreset{table}{section}
\makeatother

\makeatletter
\newcommand{\phantomlabel}[1]{%
	\begingroup
	\let\@currentlabel\thetable
	\label{#1}%
	\endgroup
}
\makeatother

\newcolumntype{P}[1]{>{\RaggedRight\arraybackslash}p{#1}}

\makeatletter
\renewcommand{\@biblabelSup}[1]{[#1]}
\makeatother

\newcommand{\autorefrange}[3]{%
	Supplementary~\autoref{#1}--\ref{#2}%
}

\newcounter{supsection}
\renewcommand{\thesupsection}{\Alph{supsection}}

\newcommand{\suppsection}[1]{%
	\refstepcounter{supsection}%
	\section*{Supplementary Materials \thesupsection{} - #1}%
	\addcontentsline{toc}{section}{Supplementary Materials \thesupsection{} - #1}%
	\setcounter{figure}{0}%
	\setcounter{table}{0}%
	\renewcommand{\thefigure}{\thesupsection\arabic{figure}}%
	\renewcommand{\thetable}{\thesupsection\arabic{table}}%
	\renewcommand{\thesubsection}{\thesupsection.\arabic{subsection}}
	\renewcommand{\thesubsubsection}{\thesupsection.\arabic{subsection}.\arabic{subsubsection}}
}

\title{\textbf{Evaluating Open-Weight Large Language Models for Structured Data Extraction from Narrative Medical Reports Across Multiple Use Cases and Languages}}

\author[1,*]{Douwe J. Spaanderman}
\author[1,2]{Karthik Prathaban}
\author[3]{Petr Zelina}
\author[4]{Kaouther Mouheb}
\author[5]{Lukáš Hejtmánek}
\author[1,6,7]{Matthew Marzetti}
\author[8]{Antonius W. Schurink}
\author[8]{Damian Chan}
\author[1]{Ruben Niemantsverdriet}
\author[1]{Frederik Hartmann}
\author[1,8]{Zhen Qian}
\author[1]{Maarten G.J. Thomeer}
\author[4,9]{Petr Holub}
\author[2]{Farhan Akram}
\author[4,10]{Frank J. Wolters}
\author[4,10]{Meike W. Vernooij}
\author[8]{Cornelis Verhoef}
\author[4]{Esther E. Bron}
\author[3,11]{Vít Nováček}
\author[8]{Dirk J. Grünhagen}
\author[12]{Wiro J. Niessen}
\author[1,2]{Martijn P.A. Starmans}
\author[1]{Stefan Klein}

\address[1]{Department of Radiology and Nuclear Medicine, Erasmus MC Cancer Institute, University Medical Center Rotterdam, the Netherlands}
\address[2]{Department of Pathology and Clinical Bioinformatics, Erasmus MC Cancer Institute, University Medical Center Rotterdam, the Netherlands}
\address[3]{Faculty of Informatics, Masaryk University}
\address[4]{Department of Radiology and Nuclear Medicine, Alzheimer Center Erasmus MC, University Medical Center Rotterdam, the Netherlands}
\address[5]{Institute of Computer Science, Masaryk University}
\address[6]{Department of Medical Physics, Leeds Teaching Hospitals NHS Trust, UK; Leeds Biomedical Research Centre, University of Leeds, UK}
\address[7]{Biomedical Imaging, University of Leeds, Leeds, United Kingdom}
\address[8]{Department of Surgical Oncology and Gastrointestinal Surgery, Erasmus MC Cancer Institute, University Medical Center Rotterdam, the Netherlands}
\address[9]{BBMRI-ERIC, Graz, Austria}
\address[10]{Department of Epidemiology, Erasmus MC University Medical Center, Rotterdam, the Netherlands}
\address[11]{Bioinformatics Research Group, Masaryk Memorial Cancer Institute}
\address[12]{Faculty of Medical Sciences, University of Groningen, Groningen, the Netherlands}

\cortext[*]{Corresponding author: Douwe J. Spaanderman, \texttt{d.spaanderman@erasmusmc.nl}}

\date{}

\begin{document}

\begin{abstract}
	\textbf{Background:} Large language models (LLMs) are increasingly explored for extracting structured information from free-text clinical records. However, most studies focus on single tasks, limited models, and English-language reports, leaving gaps in understanding performance across diseases, languages, prompting strategies and report types.\\
	\textbf{Methods:} We retrospectively evaluated 15 open-weight LLMs for structured information extraction from pathology and radiology reports across six use cases at three institutes in the Netherlands, United Kingdom, and Czech Republic, covering colorectal liver metastases, liver tumours, neurodegenerative diseases, soft-tissue tumours, melanomas, and sarcomas. Reports were manually annotated by one or more raters, with consensus annotations where applicable. Models included general-purpose and medical-specialised LLMs across four scales: large, medium, small, and tiny. Six prompting strategies were compared: zero-shot, one-shot, few-shot, chain-of-thought, self-consistency, and prompt graph. Performance was assessed using metrics appropriate for each variable type, summarised via macro-averages, and where available compared with inter-rater agreement. Consensus rank aggregation integrated performance across metrics and linear mixed-effects models quantified sources of variance. Open-source software was provided to facilitate application to new datasets.\\
	\textbf{Findings:} Across use cases, top-ranked models achieved macro-average scores closely aligned with inter-rater agreement (IRA) where available: colorectal liver metastases 0.89, liver tumours 0.95, neurodegenerative diseases 0.85 (IRA: 0.88), soft-tissue tumours English 0.76 (IRA: 0.73), Dutch 0.70 (IRA: 0.68), melanomas 0.81 (IRA: 0.76), and sarcomas 0.76 (IRA: 0.78). Summarised across use cases, small (0.77) and medium (0.77) general-purpose models performed comparably to large models (0.78), while tiny models (0.57) and specialised medical models (0.73) performed worse. Prompt Graph and few-shot prompting yielded the largest improvements (~13\%) over zero-shot. Task-specific factors, including variable complexity and annotation variability, influenced performance more than model size or prompting strategy.\\
	\textbf{Interpretation:} Open-weight LLMs can extract structured data from clinical reports across diseases, languages, and institutions. Small-to-medium general-purpose models with optimised prompting achieve performance comparable to expert annotators, providing a practical approach for scalable clinical data curation.\\
	\textbf{Funding:} Supported by Stichting Hanarth Fonds and NWO (AiNed NGF.1607.22.025); computations via SURF (EINF-9173). Additional funding was obtained for individual use cases.
\end{abstract}

\maketitle

\newpage
\noindent\textbf{Panel 1}
\begin{quote}
	\textbf{Evidence before this study}\\
	We searched PubMed for articles published till Aug 10, 2025, without language restriction, using the search terms: (“open-source” OR “open-weight” OR “open access”) AND (“large language model” OR “LLM”) AND (“medical report” OR “radiology report” OR “pathology report” OR “clinical note” OR “electronic health record”), and reviewed 25 studies. Most prior studies focused on single tasks (e.g., medication extraction or clinical code assignment) and a limited set of models, often in English only. Multilingual capabilities, comparisons across model sizes, and domain transfer between different medical report types were rarely examined.\\
	
	\textbf{Added value of this study}\\
	To our knowledge, this is the first systematic evaluation of state-of-the-art open-weight LLMs for structured clinical information extraction across multiple medical domains (pathology, radiology) and languages, without model fine-tuning. We evaluate a balanced set of models spanning different sizes (1.5B–650B parameters), domain specialisation (general-purpose vs medical-specific), and developing organisations. We compare the results across multiple prompting strategies to assess their strengths and limitations for extracting structured data from free-text medical reports. Additionally, we provide an open-source framework that makes these models easy to use, enabling researchers to apply them directly to their own datasets with minimal setup costs.\\
	
	\textbf{Implications of all the available evidence}\\
	Open-weight LLMs can extract structured data from clinical reports with performance approaching expert annotations, without task-specific fine-tuning. Smaller and medium models often perform as well as larger ones, providing a practical option for institutes with limited computing resources. Performance depends primarily on caul task specification and high-quality reference annotations rather than model size or prompting strategy, emphasizing the importance of local validation and pilot testing. Together, these findings indicate that open-weight LLMs are ready for scaling clinical data structuring, provided that tasks are clearly defined, prompts are well-specified, and reference annotations are robust and reviewed before deployment.
\end{quote}

\newpage
\section{Introduction}
Medical health record notes are typically recorded in narrative form, whether through dictation, direct entry by healthcare providers, or speech recognition systems \cite{meystre_extracting_2018}. Before narrative text can be used for research, decision support, or large-scale analysis, it must be converted into structured information \cite{jensen_mining_2012}. Conventional Natural Language Processing (NLP) methods have been developed for this purpose, but they usually depend on task-specific rules or models that require substantial manual work to design, adapt, and maintain \cite{sheikhalishahi_natural_2019}.  

Large Language Models (LLMs) offer a new approach. Unlike conventional NLP methods, LLMs can structure narrative reports with minimal customisation, making them a more scalable option for different clinical tasks. Early studies have applied LLMs to a range of problems, including structuring radiology reports \cite{woznicki_automatic_2025,adams_leveraging_2023,dehdab_llm-based_2025,di_palma_structured_2025,choubey_data_2025}, extracting diagnostic and staging information from pathology reports \cite{truhn_extracting_2024,grothey_comprehensive_2025}, processing admission, progress, and consultation notes \cite{menezes_potential_2025}, and standardizing neurology clinic notes \cite{chiang_large_2023}.  

However, most of these studies focus on a single type of data, a specific task, or a single language, which limits their generalisability. Evaluating models in such restricted settings provides little insight into their robustness or adaptability to different reporting styles, clinical domains, or languages. Another important limitation is that most published work has relied on commercial, cloud-based models such as ChatGPT \cite{adams_leveraging_2023,truhn_extracting_2024,menezes_potential_2025,menezes_potential_2025,choubey_data_2025}.  Keeping data under local control and ensuring patient privacy are critical when working with clinical records. Therefore, recently, open-weight LLMs, which provide full access to their underlying parameters, have become more popular. These models allow users to run, fine-tune, or adapt the model on local infrastructures without relying on external services, making them a promising alternative for clinical applications where privacy is a concern. 

Despite the promise of open-weight LLMs, several practical questions remain. First, it is unclear which models perform best across different clinical domains and languages. Second, it is unclear how different prompting strategies affect performance. Addressing these questions is essential to guide researchers and clinicians in selecting and applying LLMs effectively for structured information extraction from narrative medical records. Additionally, as the popularity of LLMs drives continuous innovation, establishing a standardised benchmark is crucial for consistent model comparison. 

In this study, we benchmark fifteen open-weight LLMs with different prompting strategies on a variety of clinical use cases for structuring medical records. We evaluate fifteen LLMs that vary in size and developing organization, including general-purpose and medical-specific ones. We evaluate their performance on six clinical use cases across three languages, covering three different countries and institutions, from two modalities, radiology and pathology reports, and compare six different prompting strategies to identify practical ways of applying these models in diverse clinical settings. 

\section{Methods}
\subsection{Study design}
\begin{figure}[htbp]
	\centering
	\includegraphics[width=0.9\textwidth]{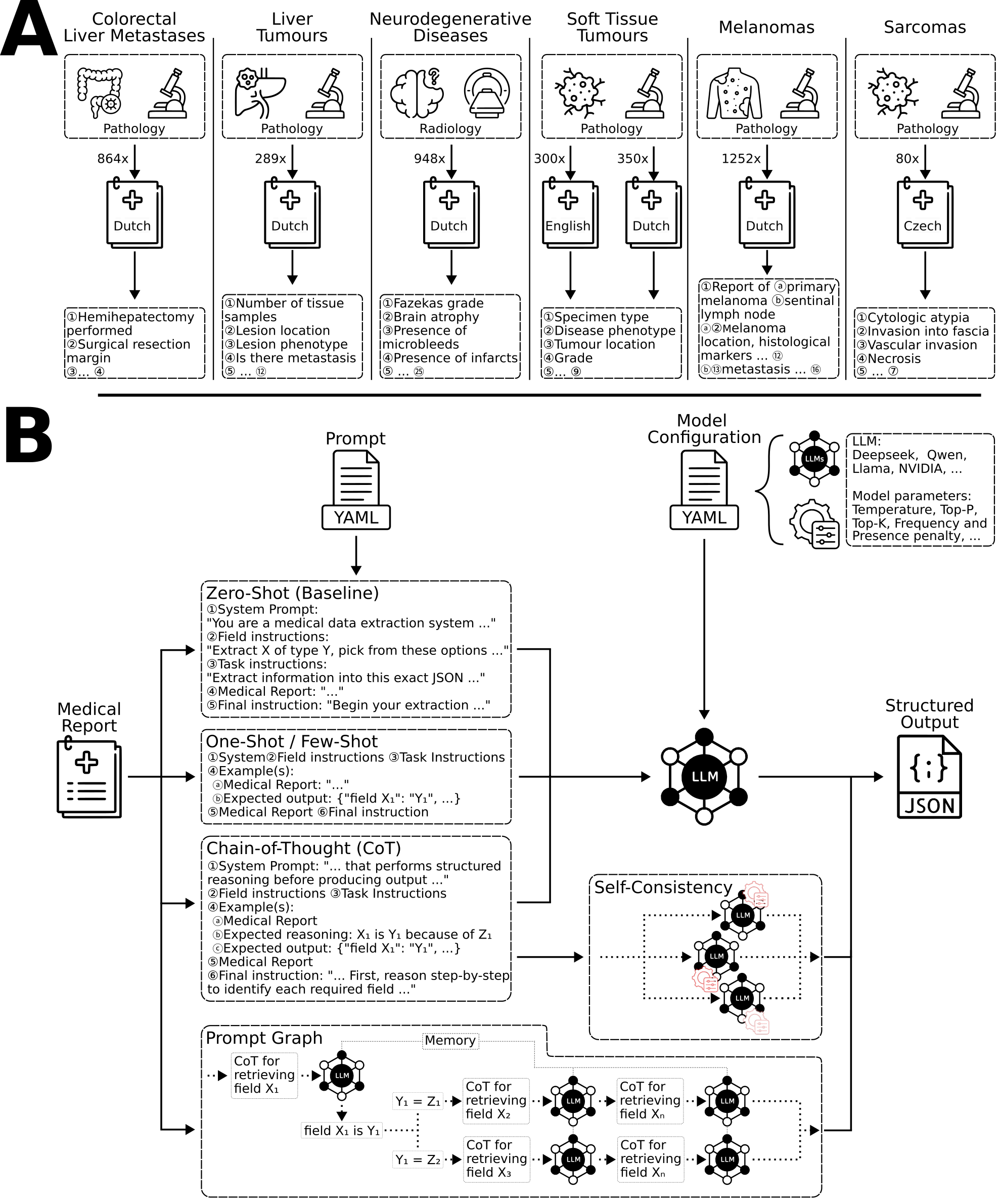}
	\caption{Study design. (\textbf{A}) Clinical use cases across different diseases, report languages, and extraction targets. (\textbf{B}) Evaluation pipeline: medical reports are processed through different prompting strategies (zero-shot, few-shot, chain-of-thought, self-consistency, and prompt graphs) defined by configuration files in YAML format (a human-readable format for structured data). Model outputs are converted into JSON (JavaScript Object Notation) files, a widely used format for structured data.}
	\label{fig:fig1}
\end{figure}

For this retrospective LLM evaluation study, we collected six use cases from pathology and radiology records across three university hospitals in different countries: Erasmus University Medical Center (The Netherlands), Leeds Teaching Hospitals NHS Trust (United Kingdom), and Masaryk Memorial Cancer Institute (Czech Republic) (\autoref{fig:fig1}). Each site obtained institutional review board approval from their own institutions. 

The use cases covered diverse clinical contexts. For colorectal liver metastases (Erasmus MC, The Netherlands), we extracted tumour characteristics and histopathological findings from 864 Dutch pathology reports annotated by one rater. For liver tumours (Erasmus MC, The Netherlands), we extracted surgical and histopathological information from 289 Dutch pathology reports annotated by one rater. For neurodegenerative diseases (Alzheimer Centre Erasmus MC), we extracted imaging markers, including vascular lesion severity, and brain atrophy scores, from 948 Dutch radiology reports annotated by two raters. For soft-tissue tumours (Erasmus MC, The Netherlands, and Leeds Teaching Hospitals, United Kingdom), which consisted of two sub-use cases, we extracted tumour phenotype, location, and grade from 627 reports in total (300 in English and 327 in Dutch), annotated by two raters. For melanomas (Erasmus MC, The Netherlands), we extracted tumour characteristics, presence of recurrence, and sentinel lymph node status from 1,252 Dutch pathology reports annotated by two raters. Finally, for sarcomas (Masaryk Memorial Cancer Institute, Czech Republic), we extracted tumour morphology and histological features from 80 Czech pathology reports annotated by nine raters. 

All reports were manually annotated by domain experts. When multiple raters evaluated the same report, discrepancies were resolved either by involving an additional rater or through discussion among raters to reach consensus. Details of each use case, including the specific extraction tasks, inclusion and exclusion criteria, annotation procedures, variable distributions, and inter-rater agreement metrics are provided in \autoref{sec:LLMSup1}. 

\subsection{Study design}
Four groups were defined based on LLM model size: large (>150B parameters), medium (40B–150B), small (4B–40B), tiny (<4B), and specialised for medical domain (\autoref{tab:tab1}A). From each group, three models were included from different developers, chosen based on open-weight availability (downloadable via Hugging Face \cite{wolf_huggingfaces_2020}). General-purpose models were selected based on their average ranking across two complementary leaderboards (as of June 25, 2025): ArtificialAnalysis.ai \cite{noauthor_ai_nodate}, which reports aggregate benchmark performance by model size, and LMArena.ai \cite{chiang_chatbot_2024}, which provides crowd-sourced human preference rankings. Within each size category, the top three open-weight models according to this combined ranking were selected, ensuring representation from distinct developers. Medically specialised models were similarly selected from the Hugging Face OpenMedical leaderboard (as of June 25, 2025), based on their mean performance across domain-specific benchmarks (PubMedQA, MedMCQA, and BioASQ \cite{singhal_large_2023}), also ensuring three models per group from different developers and open-weight availability. 

We hosted the LLMs locally on high-performance GPUs, using vLLM (v0.10.0), with the number of NVIDIA H100 GPUs varying by model size (1–16 GPUs) for experiments on the Erasmus MC and Leeds Teaching Hospitals NHS Trust datasets \cite{kwon_efficient_2023}. For the Masaryk Memorial Cancer Institute dataset, models were hosted using SGlang (v0.5.1) on an NVIDIA DGX B200 \cite{zheng_sglang_2024}. For all experiments, models were prompted via LangChain (v0.3.27) in Python (v3.12.3), with parameters set according to the developers’ recommendations \cite{chase_langchain_2022}. Resource usage and per-report generation time were recorded for each prompting strategy to quantify computational demands. Hyperparameter details are provided in \autoref{tab:llm_hyperparameters}, all code is made publicly available \cite{spaanderman_douwe-spaandermanmedicalrecordllm_2025}. 

\subsection{Prompting strategies}
To evaluate different prompting strategies for structured clinical information extraction, we implemented a range of established methods (\autoref{fig:fig1}B). In the zero-shot approach, the model is provided only with a task description and the input text, without any examples. The one-shot approach adds a single annotated input–output example to guide the model. The few-shot setting extends this by including three annotated examples to demonstrate task variability. In both prompting strategies, examples were randomly selected. Extending upon the few-shot setting, chain-of-thought prompting explicitly enables model reasoning by instructing the model to “think step by step” and by providing examples where a defined reasoning process is shown before the final answer \cite{wei_chain--thought_2023}. Self-consistency further builds on this by sampling multiple reasoning paths at slightly varied temperatures (original ±0.1), with the final output determined by majority vote or by selecting the response most semantically representative of the paths using cosine similarity \cite{wang_self-consistency_2023}. Finally, graph prompting decomposes the task into a directed graph of smaller, interdependent subtasks, with conditional paths allowing subsequent prompts to depend on earlier answers \cite{sun_prompt_2024}. The graph is defined manually, and while outputs are not directly passed between prompts, conversational memory is retained to maintain contextual consistency during sequential execution. Detailed prompting templates for each strategy and use case are provided in \autoref{sec:LLMSup1}. 

\begin{table}[htbp]
	\centering
	\caption{Selected LLMs.}
	\label{tab:tab1}
	\small
	\begin{tabularx}{\textwidth}{
			>{\RaggedRight\arraybackslash}p{3.0cm}  % Group
			>{\RaggedRight\arraybackslash}X         % Model Name
			>{\RaggedRight\arraybackslash}p{2.6cm}  % Parameters
			>{\RaggedRight\arraybackslash}X          % Repository
		}
		\toprule
		\textbf{Group} & \textbf{Model Name} & \textbf{Parameters} & \textbf{Hugging Face Repository} \\
		\midrule
		
		\multirow[t]{3}{*}{\parbox[t]{3.0cm}{\raggedright General -- Large\\($>$150B)}} 
		& DeepSeek R1 0528 \cite{deepseek-ai_deepseek-r1_2025} & 685B (45B activated) & deepseek-ai/DeepSeek-R1-0528 \\
		& Llama-4-Maverick \cite{ai_llama_2025} & 402B (17B activated) & meta-llama/Llama-4-Maverick-17B-128E-Instruct \\
		& Qwen3 235B A22B \cite{yang_qwen3_2025} & 235B (22B activated) & Qwen/Qwen3-235B-A22B \\
		\midrule
		
		\multirow[t]{3}{*}{\parbox[t]{3.0cm}{\raggedright General -- Medium\\(40B--150B)}} 
		& Llama 4 Scout \cite{ai_llama_2025-1} & 109B (17B activated) & meta-llama/Llama-4-Scout-17B-16E \\
		& Qwen2.5 72B Instruct \cite{qwen_qwen25_2025} & 72B & Qwen/Qwen2.5-72B-Instruct \\
		& NVIDIA Nemotron-Super-49B \cite{bercovich_llama-nemotron_2025} & 49B & nvidia/Llama-3\_3-Nemotron-Super-49B-v1 \\
		\midrule
		
		\multirow[t]{3}{*}{\parbox[t]{3.0cm}{\raggedright General -- Small\\(4B--40B)}} 
		& Gemma 3 27B \cite{team_gemma_2025} & 27B & google/gemma-3-27b-it \\
		& Mistral Small 3.1 24B Instruct 2503 \cite{ai_mistral_2025} & 24B & mistralai/Mistral-Small-3.1-24B-Instruct-2503 \\
		& DeepSeek R1 0528 Qwen3 8B \cite{deepseek-ai_deepseek-r1_2025} & 8B & deepseek-ai/DeepSeek-R1-0528-Qwen3-8B \\
		\midrule
		
		\multirow[t]{3}{*}{\parbox[t]{3.0cm}{\raggedright General -- Tiny\\($<$4B)}} 
		& Gemma3 4B \cite{team_gemma_2025} & 4B & google/gemma-3-4b-it \\
		& Qwen3 1.7B \cite{team_gemma_2025} & 1.7B & Qwen/Qwen3-1.7B \\
		& DeepSeek R1 Distill Qwen 1.5B \cite{deepseek-ai_deepseek-r1_2025} & 1.5B & deepseek-ai/DeepSeek-R1-Distill-Qwen-1.5B \\
		\midrule
		
		\multirow[t]{3}{*}{\parbox[t]{3.0cm}{\raggedright Specialised -- Medical}} 
		& Llama3-Med42-70B \cite{christophe_med42-v2_2024} & 70B & m42-health/Llama3-Med42-70B \\
		& Llama3-OpenBioLLM-70B \cite{ankit_pal_openbiollms_2024} & 70B & aaditya/Llama3-OpenBioLLM-70B \\
		& MedGemma 3 27B \cite{sellergren_medgemma_2025} & 27B & google/medgemma-27b-text-it \\
		\bottomrule
	\end{tabularx}
\end{table}

\subsection{Statistical analysis}
Full methodological details and equations are provided in \autoref{sec:LLMSup2}. 

Model predictions generated by the LLMs were compared against consensus annotations from trained annotators. Evaluation metrics were chosen according to variable type: balanced accuracy for categorical variables (binary: tumour present/absent; ordinal: 0–5 scores), accuracy for numeric variables (e.g., lab measurements), cosine similarity for free-text fields (e.g., disease type), and symmetric cosine similarity for list-type variables (e.g., reported symptoms). Performance across variables was summarised using the macro-average (the mean of per-variable scores), ensuring equal contribution of each variable. 

Inter-rater agreement among human annotators was assessed before consensus formation using the same variable-type–specific metrics as above. Agreements for all possible pairs of annotators were averaged to obtain a per-variable measure, and a macro-average across variables was computed to yield an overall agreement score. 

To identify the best-performing model and prompting strategy within each use case, we applied the Kemeny–Young rank aggregation method, integrating performance rankings across variables into a single consensus ordering \cite{kemeny_mathematics_1959}. The same rank aggregation procedure was repeated across all use cases to identify the overall best-performing LLM configuration. 

To evaluate performance in relation to model size, we also computed the group wise mean macro-average performance for the tiny, small, medium, and large models, as well as for specialised domain-specific variants. Finally, the mean of the macro-averages across all use cases for each LLM was reported as a complementary measure of overall performance alongside the consensus rankings. 

Uncertainty for all performance estimates was quantified using 1000 bootstrap resampling to derive 95\% confidence intervals (CIs). The same procedure was applied consistently across analyses, including inter-rater agreement, rank aggregation, and group wise evaluation. 

To understand what drives differences in performance, we used linear mixed-effects models, treating the LLM identity as a random effect and the prompting strategy as a fixed effect, while modelling the macro-average performance as the dependent variable \cite{lindstrom_newton-raphson_1988}. This allowed us to estimate how much of the observed performance variation was attributable to the model itself, the prompting strategy, or other unexplained factors. 

\section{Results}
\subsection{Overall performance across models and prompting strategies}
\begin{figure}[htbp]
	\centering
	\includegraphics[width=\textwidth]{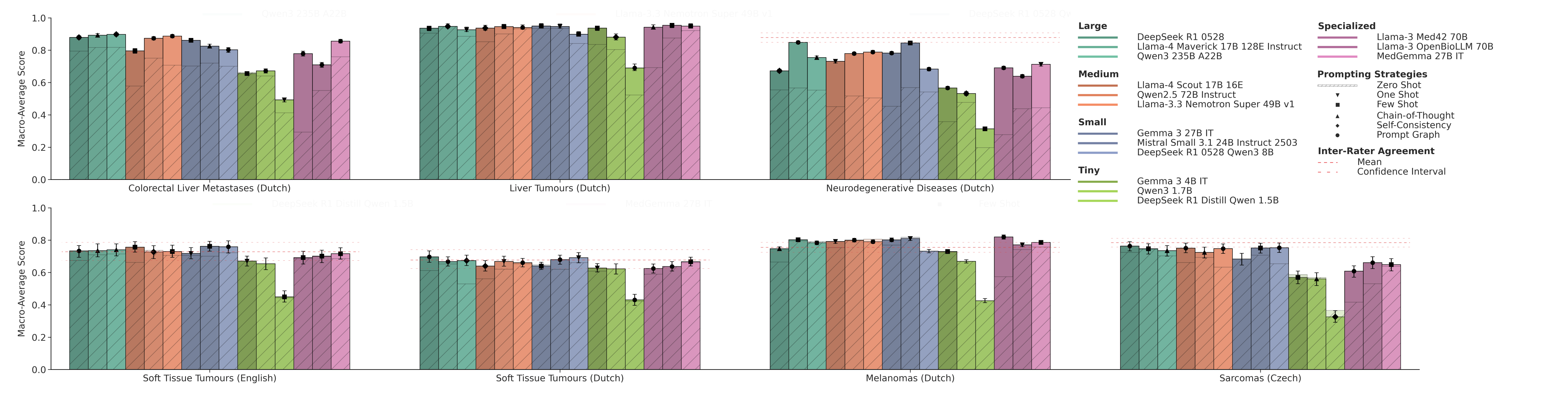}
	\caption{Performance of LLMs across clinical information extraction tasks. Results are shown for large, medium, small, tiny, and specialised models on six tasks in multiple languages, evaluated with different prompting strategies. Bars represent macro-average scores with error bars indicating 95\% confidence intervals from bootstrapping. For each model and task, zero shot prompting (diagonal hatching) is included alongside the best performing strategy determined by Kemeny–Young rank aggregation, which is indicated with the corresponding marker. Where available, inter-rater agreement values are displayed to provide a human benchmark for model performance.}
	\label{fig:fig2}
\end{figure}

\begin{figure}[htbp]
	\centering
	\includegraphics[width=\textwidth]{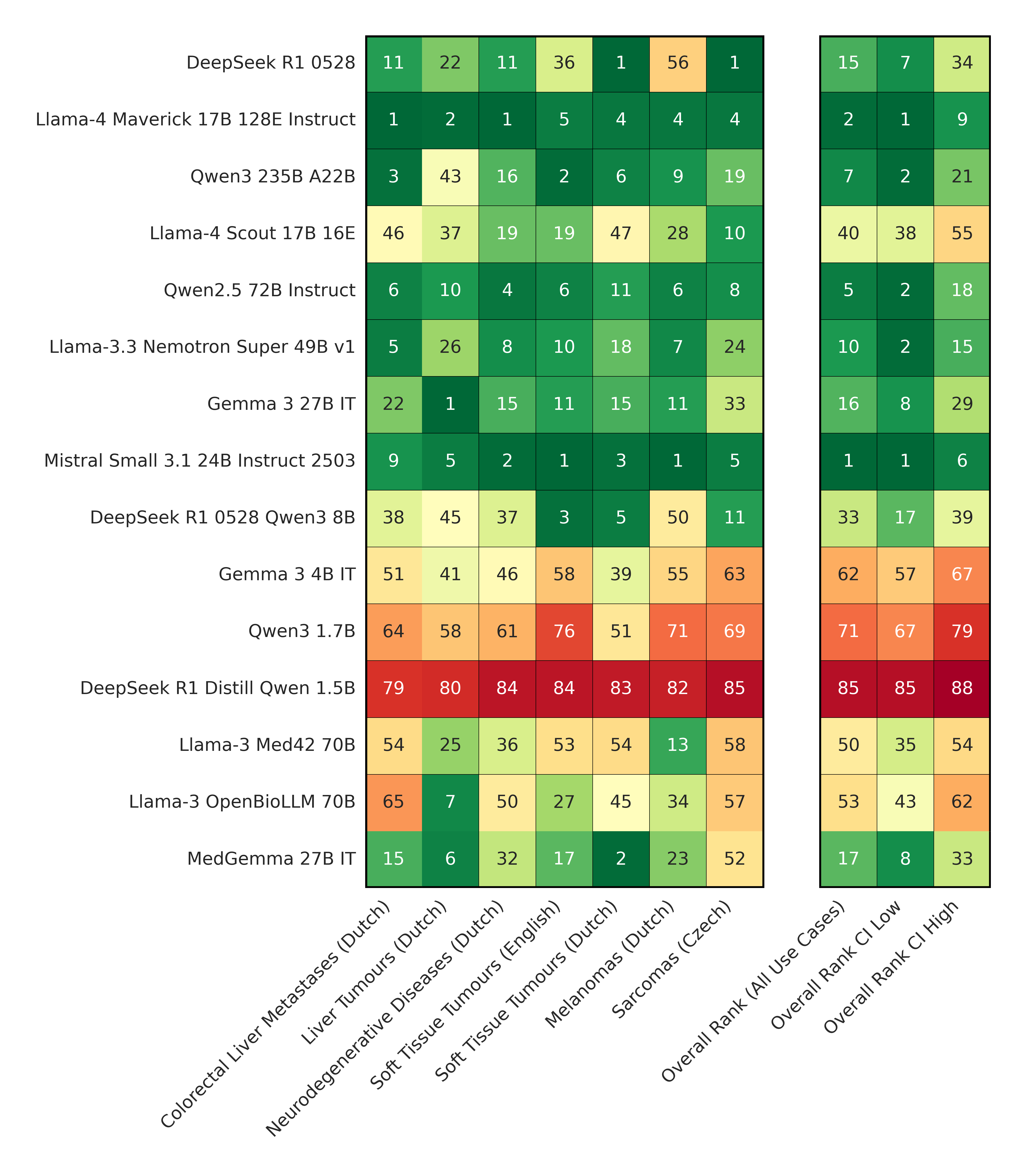}
	\caption{Kemeny-Young aggregated ranking of LLMs across clinical information extraction tasks. For each LLM, only the rank corresponding to its best-performing prompting strategy is reported. The final rank represents the consensus ordering obtained via the Kemeny-Young method across all evaluated use cases.}
	\label{fig:fig3}
\end{figure}

The combination of 15 LLMs, 6 prompting strategies, and 6 use cases (one of which comprised two separate sub–use cases) resulted in a total of 630 experiments. The overall performance of all models and their highest-ranked prompting strategies is shown in \autoref{fig:fig2}, with the aggregated ranking across models presented in \autoref{fig:fig3}. Performance metrics for each use case, comparing the top model-prompting strategy combination against inter-rater agreement, are detailed in \autoref{tab:tab2}. Additionally, \autorefrange{fig:LLMfigS1}{fig:LLMfigS7}{Figures} feature heatmaps depicting performance and rankings per use case for each LLM and prompting strategy. 

Across use cases, the Kemeny–Young ranking identified Mistral Small as the highest-ranked model overall. DeepSeek R1, Llama-4 Maverick, and Gemma 3 also achieved top rankings in specific tasks: Llama-4 Maverick performed best in the neurodegenerative diseases (0.85; 95\% CI: [0.84–0.86]) and colorectal liver metastases (0.89 [0.88–0.90]) use cases; DeepSeek R1 excelled in sarcomas (0.76 [0.73–0.79]) and soft-tissue tumours (Dutch) (0.70 [0.66–0.73]); Mistral Small led in melanomas (0.81 [0.80–0.82]) and soft-tissue tumours (English) (0.76 [0.73–0.79]); and Gemma 3 performed best in liver tumours (0.95 [0.94–0.96]). When averaged across use cases, performance remained closely grouped, with mean macro-average scores of 0.81 [0.72–0.90] for Llama-4 Maverick, 0.80 [0.73–0.88] for Mistral Small, 0.79 [0.70–0.87] for Qwen-2.5, and 0.78 [0.69–0.86] for DeepSeek R1. Due to the substantial overlap in confidence intervals, no single model demonstrated a significant advantage, indicating that multiple LLMs can achieve comparable performance across diverse medical information extraction tasks. 

When considering only the best prompting strategy per model, large models achieved a group wise mean macro-average performance of 0.78 [0.74–0.83], comparable to medium models (0.77 [0.73–0.82]) and small models (0.77 [0.73–0.81]). Tiny models performed substantially worse (0.57 [0.50–0.65]), and specialised medical models did not outperform general-purpose models (0.73 [0.68–0.78]). 

\subsection{Prompting Strategy and Variance Analysis}
Prompting strategies had a measurable but modest effect on performance. Across 105 model–use-case combinations, prompt graph was the most frequently top-ranked strategy, achieving the top rank in 37 instances, followed by few-shot (28), one-shot (13), self-consistency (11), chain-of-thought (10), and zero-shot (6). On average, using the best-performing prompting strategy for each model–use-case combination yielded a substantial improvement compared with the corresponding zero-shot baseline: +6.5\% [0.8–12.3] for one-shot, +5.1\% [2.4–7.8] for few-shot, +8.2\% [2.2–14.2] for chain-of-thought, +6.1\% [2.4–9.9] for self-consistency, and +12.7\% [8.6–16.8] for prompt graph. 

Mixed-effects modelling showed that the choice of LLM explained 18–37\% of the performance variance, prompting strategy contributed 0.2–35\%, and residual variability accounted for 45–80\%, depending on the use case (\autoref{tab:variance_partitioning}). 

\subsection{Comparison with Inter-Rater Agreement}
Inter-rater agreement was available for four of the six main use cases. The highest-ranked LLM macro-averages were generally close to inter-rater agreement (IRA): neurodegenerative diseases (Dutch) LLM: 0.85 [0.84–0.86] vs. IRA: 0.88 [0.85–0.91]; English soft-tissue tumours LLM: 0.76 [0.73–0.79] vs. IRA: 0.73 [0.67–0.79]; Dutch soft-tissue tumours LLM: 0.70 [0.66–0.73] vs. IRA: 0.68 [0.62–0.74]; melanomas (Dutch) LLM: 0.81 [0.80–0.82] vs. IRA: 0.76 [0.73–0.79]; and sarcomas (Czech) LLM: 0.76 [0.73–0.79] vs. IRA: 0.78 [0.76–0.81]. 

At the variable level, model performance closely matched inter-rater agreement for highly structured variables. Examples include SLNB tumour burden in melanomas (LLM: 0.98 [0.97–0.99] vs. IRA: 0.97 [0.94–1.00]), presence of  brain infarct in neurodegenerative diseases (LLM: 0.91 [0.89–0.92] vs. IRA: 0.88 [0.83–0.93]), and  tumour size of sarcomas (LLM: 0.86 [0.79–0.92] vs. IRA: 0.84 [0.75–0.92]). Similarly, tumour location and disease type in soft-tissue tumours achieved high metric scores (LLM: 0.90–0.94 vs. IRA: 0.89–0.92). In contrast, fields requiring contextual interpretation or synthesis of dispersed information showed lower model performance.  Examples include pathology request reason in Dutch soft-tissue tumours was LLM: 0.48 [0.29–0.72] vs. IRA: 0.47 [0.26–0.63], report revision in melanomas LLM: 0.30 [0.26–0.33] vs. IRA: 0.59 [0.43–0.77], SWI abnormalities in neurodegenerative diseases LLM: 0.59 [0.52–0.65] vs. IRA: 0.77 [0.61–0.88], and number of cerebellar infarcts LLM: 0.65 [0.62–0.68] vs. IRA: 0.83 [0.76–0.90].

\subsection{Language Variability}
Use cases were available in three languages (English, Dutch, Czech). In the soft-tissue tumours use case, macro-average model performance closely tracked inter-rater agreement across languages: 0.76 [0.73–0.79] for English (IRA: 0.73 [0.67–0.79]) and 0.70 [0.66–0.73] for Dutch (IRA: 0.68 [0.62–0.74]). For the Czech sarcomas use case, macro-average performance was 0.76 [0.73–0.79], again closely matching inter-rater agreement (0.78 [0.76–0.81]). These results indicate that, while absolute scores differ between languages, model performance aligns with human annotation variability.

\scriptsize
\begin{longtable}{P{3cm} P{3cm} P{3cm} P{6cm}}
	\caption{Performance metrics for the highest-ranked LLM and prompting strategy combination for each clinical use case. Where available, inter-rater agreement values (in parentheses) indicate the agreement between human annotators for the corresponding field. Metrics are reported as mean values with 95\% confidence intervals.}
	\label{tab:tab2} \\
	
	\toprule
	\textbf{Clinical use case} & 
	\textbf{Highest Ranked Model \& Prompting Strategy} & 
	\textbf{Macro-Average (95\% CI) [inter-rater-agreement (95\% CI)]} & 
	\textbf{Per-variable metric (95\% CI) [inter-rater-agreement (95\% CI)]} \\
	\endfirsthead
	
	\multicolumn{4}{@{}l}{\textit{Table \thetable\ continued from previous page}} \\
	\toprule
	\textbf{Clinical use case} & 
	\textbf{Highest Ranked Model \& Prompting Strategy} & 
	\textbf{Macro-Average (95\% CI) [inter-rater-agreement (95\% CI)]} & 
	\textbf{Per-variable metric (95\% CI) [inter-rater-agreement (95\% CI)]} \\
	\midrule
	\endhead
	
	\bottomrule
	
	\textbf{Colorectal Liver Metastases (Dutch)} & 
	Llama-4 Maverick 17B 128E Instruct + Chain-of-Thought & 
	0.89 [0.88–0.90] & 
	\begin{minipage}[t]{\hsize}
		Liver segments resected during first surgery: 0.95 [0.93–0.96] \\
		Hemihepatectomy: 0.97 [0.95–0.98]\\
		Surgical resection margin: 0.73 [0.70–0.77]\\
		Radicality surgery: 0.92 [0.90–0.94]\\
	\end{minipage} \\[6pt]
	\midrule
	
	\textbf{Liver Tumours (Dutch)} & 
	Gemma 3 27B IT + Few Shot & 
	0.95 [0.94–0.96] &
	\begin{minipage}[t]{\hsize}
		Number tissue samples: 0.93 [0.90–0.96] \\
		Organs: 1.00 [0.99–1.00] \\
		Method tissue retrieval: 0.69 [0.58–0.87] \\
		Explant: 0.99 [0.98–0.99] \\
		Frozen section: 0.97 [0.95–0.98] \\
		Number liver lesion types: 0.94 [0.92–0.97] \\
		Locations tissue samples: 0.98 [0.97–0.99] \\
		Lesion phenotype: 0.93 [0.92–0.94] \\
		Metastasis: 0.96 [0.93–0.98] \\
		Primary site metastasis: 0.99 [0.99–1.00] \\
		Steatosis: 1.00 [1.00–1.00] \\
		Presence fibrosis: 0.96 [0.95–0.98] \\
		Inflammation: 1.00 [1.00–1.00]
	\end{minipage} \\[6pt]
	\midrule
	
	\textbf{Neurodegenerative Diseases (Dutch)} & 
	Llama-4 Maverick 17B 128E Instruct + Prompt Graph & 
	0.85 [0.84–0.86]; (0.88 [0.85–0.91]) &
	\begin{minipage}[t]{\hsize}
		Fazekas: 0.95 [0.94–0.96]; (0.99 [0.97–1.00]) \\
		MTA left: 0.91 [0.87–0.98]; (0.99 [0.97–1.00]) \\
		MTA right: 0.97 [0.96–0.98]; (0.99 [0.96–1.00]) \\
		GCA frontal: 0.91 [0.88–0.93]; (0.92 [0.83–0.99]) \\
		GCA temporal: 0.88 [0.85–0.91]; (0.94 [0.87–0.99]) \\
		GCA occipital: 0.73 [0.66–0.79]; (0.94 [0.86–1.00]) \\
		GCA parietal: 0.89 [0.86–0.92]; (0.93 [0.86–0.99]) \\
		GCA overall: 0.91 [0.89–0.93]; (0.95 [0.88–1.00]) \\
		Any brain infarct: 0.91 [0.89–0.92]; (0.88 [0.81–0.93]) \\
		Total number brain infarcts: 0.84 [0.82–0.86]; (0.83 [0.75–0.90]) \\
		Cortical infarcts: 0.85 [0.81–0.88]; (0.87 [0.81–0.91]) \\
		Number cortical infarcts: 0.83 [0.80–0.85]; (0.78 [0.69–0.86]) \\
		Lacunes: 0.92 [0.89–0.93]; (0.92 [0.88–0.96]) \\
		Number lacunar: 0.88 [0.86–0.90]; (0.86 [0.79–0.92]) \\
		Cerebellar infarcts: 0.76 [0.74–0.78]; (0.80 [0.69–0.89]) \\
		Number cerebellar infarcts: 0.65 [0.62–0.68]; (0.83 [0.76–0.90]) \\
		Splinter infarcts: 0.67 [0.67–0.67]; (0.86 [0.71–0.95]) \\
		DWI abnormalities: 0.88 [0.83–0.93]; (0.82 [0.59–0.95]) \\
		SWI abnormalities: 0.59 [0.52–0.65]; (0.77 [0.61–0.88]) \\
		Presence microbleeds: 0.94 [0.93–0.95]; (0.85 [0.79–0.91]) \\
		Total number microbleeds: 0.92 [0.90–0.94]; (0.81 [0.74–0.88]) \\
		Presence siderosis: 0.89 [0.80–0.96]; (0.81 [0.72–1.00])
	\end{minipage} \\[6pt]
	\midrule
	
	\textbf{Soft Tissue Tumours (English)} &
	Mistral Small 3.1 24B Instruct 2503 + Few Shot &
	0.76 [0.73–0.79]; (0.73 [0.67–0.79]) &
	\begin{minipage}[t]{\hsize}
		Specimen type: 0.38 [0.30–0.54]; (0.66 [0.50–0.78]) \\
		Pathology request reason: 0.58 [0.44–0.70]; (0.38 [0.29–0.65]) \\
		Disease type: 0.90 [0.89–0.91]; (0.89 [0.85–0.92]) \\
		Disease type differential diagnosis: 0.79 [0.74–0.83]; (0.69 [0.55–0.81]) \\
		Soft tissue tumour: 0.93 [0.90–0.95]; (0.71 [0.52–0.97]) \\
		Suspected or confirmed: 0.66 [0.49–0.79]; (0.72 [0.55–0.85]) \\
		Benign or malignant: 0.88 [0.77–0.96]; (0.85 [0.73–0.99]) \\
		Tumour grade: 0.83 [0.75–0.90]; (0.73 [0.56–0.88]) \\
		Tumour location: 0.92 [0.91–0.92]; (0.92 [0.91–0.94])
	\end{minipage} \\[6pt]
	\midrule
	
	\textbf{Soft Tissue Tumours (Dutch)} &
	DeepSeek R1 0528 + Prompt Graph &
	0.70 [0.66–0.73]; (0.68 [0.62–0.74]) &
	\begin{minipage}[t]{\hsize}
		Specimen type: 0.53 [0.39–0.66]; (0.41 [0.28–0.56]) \\
		Pathology request reason: 0.48 [0.29–0.72]; (0.47 [0.26–0.63]) \\
		Disease type: 0.94 [0.93–0.94]; (0.91 [0.88–0.94]) \\
		Disease type differential diagnosis: 0.85 [0.81–0.89]; (0.86 [0.76–0.95]) \\
		Soft tissue tumour: 0.58 [0.52–0.64]; (0.52 [0.31–0.83]) \\
		Suspected or confirmed: 0.67 [0.52–0.87]; (0.69 [0.62–0.81]) \\
		Benign or malignant: 0.72 [0.66–0.77]; (0.61 [0.53–0.78]) \\
		Tumour grade: 0.61 [0.50–0.70]; (0.72 [0.56–0.83]) \\
		Tumour location: 0.90 [0.90–0.91]; (0.89 [0.87–0.91])
	\end{minipage} \\[6pt]
	\midrule
	
	\textbf{Melanomas (Dutch)} &
	Mistral Small 3.1 24B Instruct 2503 + One Shot &
	0.81 [0.80–0.82]; (0.76 [0.73–0.79]) &
	\begin{minipage}[t]{\hsize}
		Primary melanoma: 0.73 [0.69–0.77]; (0.76 [0.69–0.83]) \\
		Report revision: 0.30 [0.26–0.33]; (0.59 [0.43–0.77]) \\
		Multiple primary melanoma: 0.57 [0.47–0.67]; (0.80 [0.70–0.93]) \\
		In situ melanoma: 0.87 [0.82–0.91]; (0.80 [0.70–0.92]) \\
		Breslow thickness: 0.96 [0.95–0.97]; (0.85 [0.78–0.91]) \\
		Ulceration: 0.97 [0.96–0.98]; (0.80 [0.66–0.90]) \\
		Location melanoma: 0.84 [0.82–0.87]; (0.82 [0.73–0.89])
		Subtype melanoma: 0.94 [0.90–0.96]; (0.91 [0.81–0.97]) \\
		Microsatellites: 0.91 [0.86–0.96]; (0.72 [0.61–0.95]) \\
		Mitosis described: 0.78 [0.75–0.81]; (0.52 [0.40–0.64]) \\
		Mitosis count: 0.98 [0.97–0.98]; (0.99 [0.97–1.00]) \\
		Lymphatic invasion: 0.75 [0.66–0.82]; (0.75 [0.68–0.89]) \\
		Specimen type: 0.65 [0.61–0.68]; (0.48 [0.38–0.56]) \\
		SLNB metastases: 0.92 [0.90–0.94]; (0.60 [0.46–0.72]) \\
		SLNB tumour burden: 0.98 [0.97–0.99]; (0.97 [0.94–1.00]) \\
	\end{minipage} \\[6pt]
	\midrule
	
	\textbf{Sarcomas (Czech)} &
	DeepSeek R1 0528 + Prompt Graph &
	0.76 [0.73–0.79]; (0.78 [0.76–0.81]) &
	\begin{minipage}[t]{\hsize}
		Cytologic atypia: 0.54 [0.44–0.64]; (0.57 [0.46–0.69])
		Invasion into fascia: 0.44 [0.35–0.53]; (0.64 [0.56–0.73]) \\
		Vascular invasion: 0.99 [0.97–1.00]; (0.87 [0.80–0.93]) \\
		Necrosis: 0.62 [0.52–0.71]; (0.88 [0.83–0.93]) \\
		Tumour size: 0.86 [0.79–0.92]; (0.84 [0.75–0.92]) \\
		Mitotic HPF: 0.95 [0.90–0.99]; (0.76 [0.65–0.86]) \\
		Mitotic mm: 0.94 [0.88–0.99]; (0.93 [0.88–0.97]) \\
	\end{minipage} \\[6pt]
	\bottomrule
\end{longtable}
\normalsize

\subsection{Resource requirements}
\begin{figure}[htbp]
	\centering
	\includegraphics[width=\textwidth]{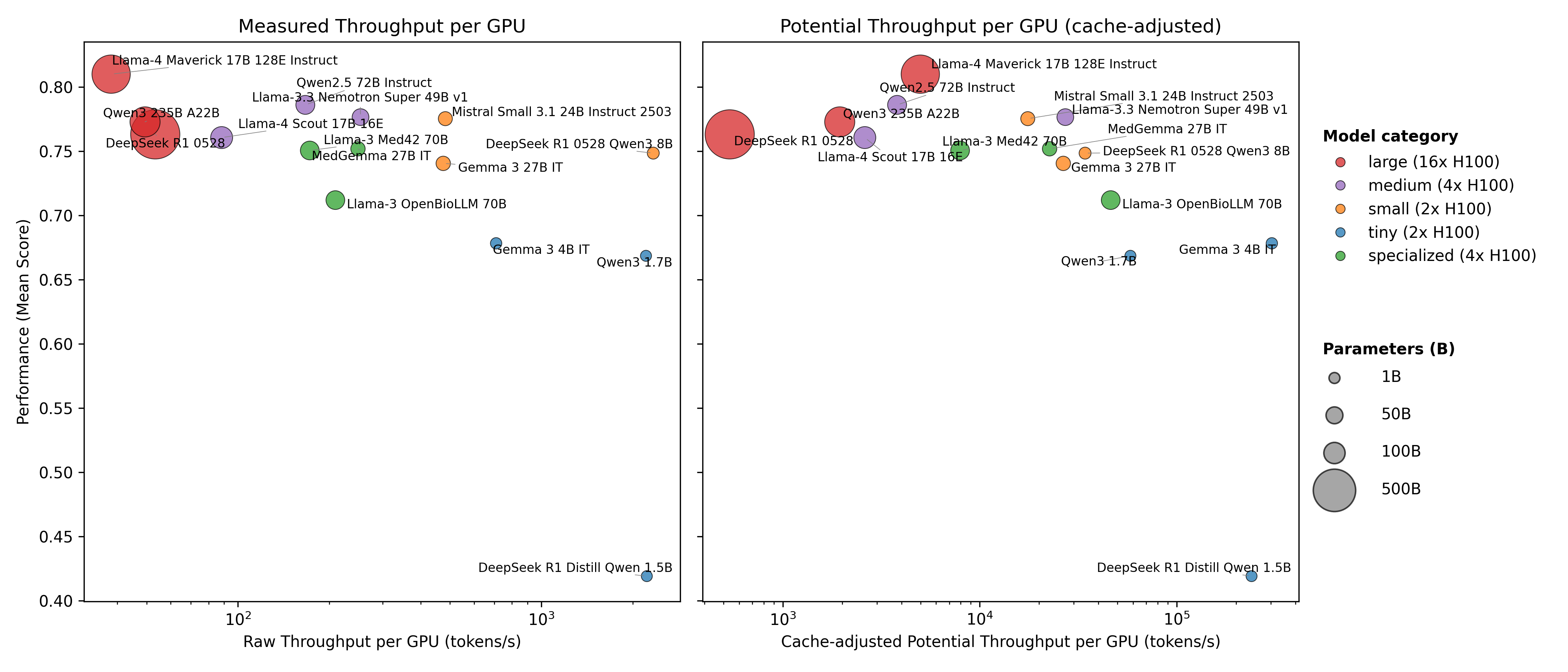}
	\caption{Model performance versus per-GPU throughput over all use cases. The y-axis shows the mean macro-average score across use cases, calculated using Few-Shot prompting only. Per-GPU throughput is defined as the number of tokens processed per second, normalized by the number of GPUs used to allow fair comparison across models. Bubble size represents model parameter size and colour indicates model category. Left: measured throughput; right: potential throughput per GPU corrected for cache utilisation (accounting for how much of the model’s cached key-value memory is effectively used). The Sarcoma use case was excluded from this analysis since it was performed on different hardware.}
	\label{fig:fig4}
\end{figure}

The LLM performance versus per-GPU throughput is illustrated in \autoref{fig:fig4}. Larger models generally achieved slightly higher performance but demanded substantially more GPU resources. few-shot inference offered a favourable balance between speed and performance, with an average generation time of 4.5~s per report. In contrast, more complex prompting strategies, such as self-consistency and prompt graph, required considerably more GPU resources, requiring 15.9~s and 37.6~s per report, respectively. 

\section{Discussion}
Recent studies have explored large language models (LLMs) for extracting structured data from radiology and pathology reports, demonstrating promising results but often focusing on specific clinical context or limited model comparisons \cite{woznicki_automatic_2025,adams_leveraging_2023,truhn_extracting_2024,menezes_potential_2025,dehdab_llm-based_2025,di_palma_structured_2025,choubey_data_2025}. Few have systematically examined multiple clinical domains, prompting strategies, and languages. To address this gap, we conducted a multicentre, multi-language evaluation across six clinically distinct use cases, comparing 15 open-weight models and six prompting strategies within a unified and transparent benchmarking framework. The framework includes inter-observer agreement analysis, quantitative ranking of models as well as prompting methods, and public release of all code to promote reproducibility and future benchmarking \cite{spaanderman_douwe-spaandermanmedicalrecordllm_2025}. 

Several key findings emerged. First, inter-rater agreement serves as a critical benchmark for evaluating model performance. Across the neurodegenerative diseases, melanomas, soft-tissue tumours, and sarcomas use cases, LLM performance closely mirrored human agreement, highlighting that task subjectivity and ambiguity, rather than model limitations, often dictate performance ceilings. Second, small and medium-sized general-purpose LLMs achieved performance comparable to much larger models, suggesting that compact architectures already encode much of the clinically relevant knowledge for these tasks. This aligns with our observation that no single model demonstrated a significant advantage across use cases as evidenced by overlapping confidence intervals in macro-average scores and ranking. Third, prompting strategies had a measurable but modest effect: methods such as prompt graph and few-shot often outperformed zero-shot baselines, but absolute gains were limited, and differences between models were generally smaller than differences between use cases. Together, these findings indicate that task characteristics, including information complexity, report structure, and clinical context, play a more decisive role than model size or prompting strategy. 

Performance variability across use cases further underscored this point. Structured fields, such as tumour location, histopathological measurements, or laboratory values, were extracted with high metric scores, closely following inter-rater agreement. In contrast, context-dependent fields, such as pathology request reasons or report revisions, showed lower performance, reflecting the inherent variability and subjectivity observed among human annotators. These findings underscore the importance of clearly defining tasks and variables before  deploying the model, to ensure reliable and clinically meaningful extraction. 

Overall, LLM performance was largely consistent across English, Dutch, and Czech reports. In the soft-tissue tumours use case, performance on Dutch reports was slightly lower than English, mirroring lower inter-rater agreement. This difference likely reflects multiple factors, including inherent variability in free-text structure between languages, and differences in clinical populations or report characteristics. While a direct comparison was not possible for the Czech sarcomas dataset, model performance aligned closely with inter-rater agreement. Together, these results suggest that LLMs are well calibrated across languages, and that observed differences are more likely attributable to annotation or dataset characteristics than to model limitations. 

For researchers applying LLMs to new clinical extraction tasks, we have the following recommendation. Careful prompt definition is critical and often drives more performance gains than testing multiple large models or complex prompting strategies. In our study, for example, the liver tumour use case required iterative refinement of the reference annotations and prompt formulation: initial annotations informed prompt design, which then revealed gaps or ambiguities that were addressed through annotation updates. Based on these experiences, we recommend specifying prompts with unambiguous instructions, clearly defined answer categories, and illustrative examples where possible. With a small set of high-quality reference annotations, ideally reviewed by multiple experts, a few targeted experiments using small-to-medium-sized LLMs (including newer models beyond those assessed here) and selected prompting strategies (e.g., few-shot, prompt graph) are usually sufficient to identify well-performing configurations. To this end, our open-source framework allows this entire process to be run with minimal setup: users define the prompts and the combination of models and prompting strategies to test, and the software handles experiments and reporting automatically \cite{spaanderman_douwe-spaandermanmedicalrecordllm_2025}. 

Since high-quality extraction can be achieved with small-to-medium-sized LLMs, particularly when combined with few-shot or single-run prompt graph prompting, competitive performance is possible with lower computational demands than larger models. Nevertheless, even these models were evaluated on high-performance GPUs (NVIDIA H100), so adequate infrastructure remains essential, and resource planning should be considered when deploying LLMs in clinical environments \cite{dennstadt_implementing_2025}. 

This study has several limitations. First, although we included a range of small-to-large and medically specialised LLMs, we did not systematically explore distillation, compression or quantisation strategies that could reduce computational requirements, which may be important for deployment in more resource-constrained environments. Second, our evaluation focused on quantitative metrics, which capture accuracy on structured fields but may not fully reflect clinical usability or the nuanced interpretation of free-text outputs; a complementary qualitative assessment using frameworks such as QUEST could provide further insight but was beyond the scope of this study \cite{tam_framework_2024}. Third, while our experiments covered two modalities, five of the six use cases were based on pathology reports and only one on radiology reports, which limits the strength of conclusions regarding cross-modality generalization and applicability to imaging-related text. Finally, inter-rater agreement was not available for all use cases, and where assessed, it was often based on a small subset of reports and a limited number of raters, which may affect the precision of performance benchmarks and the generalizability of findings. 

In summary, the results demonstrate that across multiple institutions, languages, and clinical domains, LLMs can achieve performance levels comparable expert annotations. Advances in model architectures and prompting strategies have enabled this high level of accuracy, particularly for well-defined extraction tasks. Careful task specification and high-quality reference annotations remain key to realizing these gains. Overall, these findings suggest that LLMs are now practical tools for structuring information from pathology and radiology reports, opening a pathway toward large-scale, automated curation of clinical data to support research, quality monitoring, and patient care. 

\section{Acknowledgments}
The colorectal liver metastases use case was supported by Erasmus MC research funding. The liver tumour use case was supported by the Dutch Research Council (NWO) through The Liver Artificial Intelligence (LAI) consortium (20431), and made use of Palga, the Dutch Automated National Pathological Anatomy Archive, for collecting data. The neurodegenerative diseases use case was supported by ZonMw (10510032120003) within the Onderzoeksprogramma Dementie as part of the Dutch National neurodegenerative diseases Strategy. We thank Harro Seelaar, Francesco Mattace-Raso for establishing the ACE cohort, Romy de Haar for data management and Amos Pomp, Joy Martens, and Farog Faghir for their contributions to data collection and annotation on the neurodegenerative diseases use case. The soft-tissue tumours use case was supported by Health Education England (HEE), the National Institute for Health Research (NIHR), and an unrestricted grant from Stichting Hanarth Fonds, The Netherlands. The views expressed are those of the authors and do not necessarily reflect those of the NIHR, NHS, or the UK Department of Health and Social Care. The sarcomas use case was supported by European Union Horizon Research and Innovation Programme (grant no. 101057048, IDEA4RC). Grant Agency of Masaryk University (grant no. MUNI/A/1638/2024, SV25-AI4Data), Ministry of Health of the Czech Republic (grant no. NW25-09-00465, EMPOWER), and made use of AI infrastructure developed within BioMedAI and e-INFRA CZ project (ID: 90254), supported by the Ministry of Education, Youth and Sports of the Czech Republic. 

\bibliography{references}

\begin{thebibliography}{1}
\expandafter\ifx\csname url\endcsname\relax
  \def\url#1{\texttt{#1}}\fi
\expandafter\ifx\csname urlprefix\endcsname\relax\def\urlprefix{URL }\fi
\expandafter\ifx\csname href\endcsname\relax
  \def\href#1#2{#2} \def\path#1{#1}\fi

\bibitem{maaten_visualizing_2008}
L.~v.~d. Maaten, G.~Hinton, Visualizing {Data} using t-{SNE}, Journal of
  Machine Learning Research 9~(86) (2008) 2579--2605.

\bibitem{reimers_sentence-bert_2019}
N.~Reimers, I.~Gurevych,
  \href{http://arxiv.org/abs/1908.10084}{Sentence-{BERT}: {Sentence}
  {Embeddings} using {Siamese} {BERT}-{Networks}}, arXiv:1908.10084 [cs] (Aug.
  2019).
\newblock \href {https://doi.org/10.48550/arXiv.1908.10084}
  {\path{doi:10.48550/arXiv.1908.10084}}.
\newline\urlprefix\url{http://arxiv.org/abs/1908.10084}

\bibitem{gao2021scaling}
L.~Gao, Y.~Zhang, J.~Han, J.~Callan, Scaling deep contrastive learning batch
  size under memory limited setup (2021).
\newblock \href {http://arxiv.org/abs/2101.06983} {\path{arXiv:2101.06983}}.

\bibitem{kemeny_mathematics_1959_sup}
J.~G. Kemeny, \href{https://www.jstor.org/stable/20026529}{Mathematics without
  {Numbers}}, Daedalus 88~(4) (1959) 577--591, publisher: The MIT Press.
\newline\urlprefix\url{https://www.jstor.org/stable/20026529}

\bibitem{kirkpatrick_optimization_1983}
S.~Kirkpatrick, C.~D. Gelatt, M.~P. Vecchi, Optimization by {Simulated}
  {Annealing}, Science 220~(4598) (1983) 671--680, publisher: American
  Association for the Advancement of Science.
\newblock \href {https://doi.org/10.1126/science.220.4598.671}
  {\path{doi:10.1126/science.220.4598.671}}.

\bibitem{johnson_optimization_1989}
D.~S. Johnson, C.~R. Aragon, L.~A. McGeoch, C.~Schevon, Optimization by
  {Simulated} {Annealing}: {An} {Experimental} {Evaluation}; {Part} {I},
  {Graph} {Partitioning}, Operations Research 37~(6) (1989) 865--892,
  publisher: INFORMS.
\newblock \href {https://doi.org/10.1287/opre.37.6.865}
  {\path{doi:10.1287/opre.37.6.865}}.

\end{thebibliography}


\begin{thebibliography}{10}
\expandafter\ifx\csname url\endcsname\relax
  \def\url#1{\texttt{#1}}\fi
\expandafter\ifx\csname urlprefix\endcsname\relax\def\urlprefix{URL }\fi
\expandafter\ifx\csname href\endcsname\relax
  \def\href#1#2{#2} \def\path#1{#1}\fi

\bibitem{meystre_extracting_2018}
S.~M. Meystre, G.~K. Savova, K.~C. Kipper-Schuler, J.~F. Hurdle,
  \href{https://www.thieme-connect.com/products/ejournals/abstract/10.1055/s-0038-1638592}{Extracting
  {Information} from {Textual} {Documents} in the {Electronic} {Health}
  {Record}: {A} {Review} of {Recent} {Research}}, Yearbook of Medical
  Informatics 17 (2018) 128--144, publisher: Georg Thieme Verlag KG.
\newblock \href {https://doi.org/10.1055/s-0038-1638592}
  {\path{doi:10.1055/s-0038-1638592}}.
\newline\urlprefix\url{https://www.thieme-connect.com/products/ejournals/abstract/10.1055/s-0038-1638592}

\bibitem{jensen_mining_2012}
P.~B. Jensen, L.~J. Jensen, S.~Brunak,
  \href{https://www.nature.com/articles/nrg3208}{Mining electronic health
  records: towards better research applications and clinical care}, Nature
  Reviews Genetics 13~(6) (2012) 395--405, publisher: Nature Publishing Group.
\newblock \href {https://doi.org/10.1038/nrg3208} {\path{doi:10.1038/nrg3208}}.
\newline\urlprefix\url{https://www.nature.com/articles/nrg3208}

\bibitem{sheikhalishahi_natural_2019}
S.~Sheikhalishahi, R.~Miotto, J.~T. Dudley, A.~Lavelli, F.~Rinaldi, V.~Osmani,
  Natural {Language} {Processing} of {Clinical} {Notes} on {Chronic}
  {Diseases}: {Systematic} {Review}, JMIR medical informatics 7~(2) (2019)
  e12239.
\newblock \href {https://doi.org/10.2196/12239} {\path{doi:10.2196/12239}}.

\bibitem{woznicki_automatic_2025}
P.~Woźnicki, C.~Laqua, I.~Fiku, A.~Hekalo, D.~Truhn, S.~Engelhardt, J.~Kather,
  S.~Foersch, T.~A. D’Antonoli, D.~Pinto~dos Santos, B.~Baeßler, F.~C.
  Laqua, \href{https://doi.org/10.1007/s00330-024-11074-y}{Automatic
  structuring of radiology reports with on-premise open-source large language
  models}, European Radiology 35~(4) (2025) 2018--2029.
\newblock \href {https://doi.org/10.1007/s00330-024-11074-y}
  {\path{doi:10.1007/s00330-024-11074-y}}.
\newline\urlprefix\url{https://doi.org/10.1007/s00330-024-11074-y}

\bibitem{adams_leveraging_2023}
L.~C. Adams, D.~Truhn, F.~Busch, A.~Kader, S.~M. Niehues, M.~R. Makowski, K.~K.
  Bressem,
  \href{https://pubs.rsna.org/doi/full/10.1148/radiol.230725}{Leveraging
  {GPT}-4 for {Post} {Hoc} {Transformation} of {Free}-text {Radiology}
  {Reports} into {Structured} {Reporting}: {A} {Multilingual} {Feasibility}
  {Study}}, Radiology 307~(4) (2023) e230725, publisher: Radiological Society
  of North America.
\newblock \href {https://doi.org/10.1148/radiol.230725}
  {\path{doi:10.1148/radiol.230725}}.
\newline\urlprefix\url{https://pubs.rsna.org/doi/full/10.1148/radiol.230725}

\bibitem{dehdab_llm-based_2025}
R.~Dehdab, F.~Mankertz, J.~M. Brendel, N.~Maalouf, K.~Kaya, S.~Afat,
  S.~Kolahdoozan, A.~R. Radmard, {LLM}-{Based} {Extraction} of {Imaging}
  {Features} from {Radiology} {Reports}: {Automating} {Disease} {Activity}
  {Scoring} in {Crohn}'s {Disease}, Academic Radiology 32~(10) (2025)
  5869--5877.
\newblock \href {https://doi.org/10.1016/j.acra.2025.07.041}
  {\path{doi:10.1016/j.acra.2025.07.041}}.

\bibitem{di_palma_structured_2025}
L.~Di~Palma, F.~Darvizeh, M.~Alì, D.~Fazzini, Structured {Transformation} of
  {Unstructured} {Prostate} {MRI} {Reports} {Using} {Large} {Language}
  {Models}, Tomography (Ann Arbor, Mich.) 11~(6) (2025) 69.
\newblock \href {https://doi.org/10.3390/tomography11060069}
  {\path{doi:10.3390/tomography11060069}}.

\bibitem{choubey_data_2025}
A.~P. Choubey, E.~Eguia, A.~Hollingsworth, S.~Chatterjee, M.~I. D'Angelica,
  W.~R. Jarnagin, A.~C. Wei, M.~A. Schattner, R.~K.~G. Do, K.~C. Soares, {MSKCC
  Pancreas Cyst Collaborative}, Data {Extraction} and {Curation} from
  {Radiology} {Reports} for {Pancreatic} {Cyst} {Surveillance} {Using} {Large}
  {Language} {Models}, Journal of the American College of Surgeons (Jul. 2025).
\newblock \href {https://doi.org/10.1097/XCS.0000000000001478}
  {\path{doi:10.1097/XCS.0000000000001478}}.

\bibitem{truhn_extracting_2024}
D.~Truhn, C.~M. Loeffler, G.~Müller-Franzes, S.~Nebelung, K.~J. Hewitt,
  S.~Brandner, K.~K. Bressem, S.~Foersch, J.~N. Kather, Extracting structured
  information from unstructured histopathology reports using generative
  pre-trained transformer 4 ({GPT}-4), The Journal of Pathology 262~(3) (2024)
  310--319.
\newblock \href {https://doi.org/10.1002/path.6232}
  {\path{doi:10.1002/path.6232}}.

\bibitem{grothey_comprehensive_2025}
B.~Grothey, J.~Odenkirchen, A.~Brkic, B.~Schömig-Markiefka, A.~Quaas,
  R.~Büttner, Y.~Tolkach,
  \href{https://www.nature.com/articles/s43856-025-00808-8}{Comprehensive
  testing of large language models for extraction of structured data in
  pathology}, Communications Medicine 5~(1) (2025) 96, publisher: Nature
  Publishing Group.
\newblock \href {https://doi.org/10.1038/s43856-025-00808-8}
  {\path{doi:10.1038/s43856-025-00808-8}}.
\newline\urlprefix\url{https://www.nature.com/articles/s43856-025-00808-8}

\bibitem{menezes_potential_2025}
M.~C.~S. Menezes, A.~F. Hoffmann, A.~L.~M. Tan, M.~Nalbandyan, G.~S. Omenn,
  D.~R. Mazzotti, A.~Hernández-Arango, S.~Visweswaran, S.~Venkatesh, K.~D.
  Mandl, F.~T. Bourgeois, J.~W.~K. Lee, A.~Makmur, D.~A. Hanauer, M.~G.
  Semanik, L.~T. Kerivan, T.~Hill, J.~Forero, C.~Restrepo, M.~Vigna,
  P.~Ceriana, N.~Abu-El-Rub, P.~Avillach, R.~Bellazzi, T.~Callaci,
  A.~Gutiérrez-Sacristán, A.~Malovini, J.~P. Mathew, M.~Morris, V.~L. Murthy,
  T.~M. Buonocore, E.~Parimbelli, L.~P. Patel, C.~Sáez, M.~J. Samayamuthu,
  J.~A. Thompson, V.~Tibollo, Z.~Xia, I.~S. Kohane, {Consortium for Clinical
  Characterization of COVID-19 by Electronic Health Records}, The potential of
  {Generative} {Pre}-trained {Transformer} 4 ({GPT}-4) to analyse medical notes
  in three different languages: a retrospective model-evaluation study, The
  Lancet. Digital Health 7~(1) (2025) e35--e43.
\newblock \href {https://doi.org/10.1016/S2589-7500(24)00246-2}
  {\path{doi:10.1016/S2589-7500(24)00246-2}}.

\bibitem{chiang_large_2023}
C.-C. Chiang, M.~Luo, G.~Dumkrieger, S.~Trivedi, Y.-C. Chen, C.-J. Chao, T.~J.
  Schwedt, A.~Sarker, I.~Banerjee, A {Large} {Language} {Model}-{Based}
  {Generative} {Natural} {Language} {Processing} {Framework} {Finetuned} on
  {Clinical} {Notes} {Accurately} {Extracts} {Headache} {Frequency} from
  {Electronic} {Health} {Records}, medRxiv: The Preprint Server for Health
  Sciences (2023) 2023.10.02.23296403\href
  {https://doi.org/10.1101/2023.10.02.23296403}
  {\path{doi:10.1101/2023.10.02.23296403}}.

\bibitem{wolf_huggingfaces_2020}
T.~Wolf, L.~Debut, V.~Sanh, J.~Chaumond, C.~Delangue, A.~Moi, P.~Cistac,
  T.~Rault, R.~Louf, M.~Funtowicz, J.~Davison, S.~Shleifer, P.~v. Platen,
  C.~Ma, Y.~Jernite, J.~Plu, C.~Xu, T.~L. Scao, S.~Gugger, M.~Drame, Q.~Lhoest,
  A.~M. Rush, \href{http://arxiv.org/abs/1910.03771}{{HuggingFace}'s
  {Transformers}: {State}-of-the-art {Natural} {Language} {Processing}},
  arXiv:1910.03771 [cs] (Jul. 2020).
\newblock \href {https://doi.org/10.48550/arXiv.1910.03771}
  {\path{doi:10.48550/arXiv.1910.03771}}.
\newline\urlprefix\url{http://arxiv.org/abs/1910.03771}

\bibitem{noauthor_ai_nodate}
\href{https://artificialanalysis.ai}{{AI} {Model} \& {API} {Providers}
  {Analysis} {\textbar} {Artificial} {Analysis}}.
\newline\urlprefix\url{https://artificialanalysis.ai}

\bibitem{chiang_chatbot_2024}
W.-L. Chiang, L.~Zheng, Y.~Sheng, A.~N. Angelopoulos, T.~Li, D.~Li, H.~Zhang,
  B.~Zhu, M.~Jordan, J.~E. Gonzalez, I.~Stoica,
  \href{http://arxiv.org/abs/2403.04132}{Chatbot {Arena}: {An} {Open}
  {Platform} for {Evaluating} {LLMs} by {Human} {Preference}}, arXiv:2403.04132
  [cs] (Mar. 2024).
\newblock \href {https://doi.org/10.48550/arXiv.2403.04132}
  {\path{doi:10.48550/arXiv.2403.04132}}.
\newline\urlprefix\url{http://arxiv.org/abs/2403.04132}

\bibitem{singhal_large_2023}
K.~Singhal, S.~Azizi, T.~Tu, S.~S. Mahdavi, J.~Wei, H.~W. Chung, N.~Scales,
  A.~Tanwani, H.~Cole-Lewis, S.~Pfohl, P.~Payne, M.~Seneviratne, P.~Gamble,
  C.~Kelly, A.~Babiker, N.~Schärli, A.~Chowdhery, P.~Mansfield,
  D.~Demner-Fushman, B.~Agüera~y Arcas, D.~Webster, G.~S. Corrado, Y.~Matias,
  K.~Chou, J.~Gottweis, N.~Tomasev, Y.~Liu, A.~Rajkomar, J.~Barral, C.~Semturs,
  A.~Karthikesalingam, V.~Natarajan,
  \href{https://www.nature.com/articles/s41586-023-06291-2}{Large language
  models encode clinical knowledge}, Nature 620~(7972) (2023) 172--180,
  publisher: Nature Publishing Group.
\newblock \href {https://doi.org/10.1038/s41586-023-06291-2}
  {\path{doi:10.1038/s41586-023-06291-2}}.
\newline\urlprefix\url{https://www.nature.com/articles/s41586-023-06291-2}

\bibitem{kwon_efficient_2023}
W.~Kwon, Z.~Li, S.~Zhuang, Y.~Sheng, L.~Zheng, C.~H. Yu, J.~E. Gonzalez,
  H.~Zhang, I.~Stoica, \href{http://arxiv.org/abs/2309.06180}{Efficient
  {Memory} {Management} for {Large} {Language} {Model} {Serving} with
  {PagedAttention}}, arXiv:2309.06180 [cs] (Sep. 2023).
\newblock \href {https://doi.org/10.48550/arXiv.2309.06180}
  {\path{doi:10.48550/arXiv.2309.06180}}.
\newline\urlprefix\url{http://arxiv.org/abs/2309.06180}

\bibitem{zheng_sglang_2024}
L.~Zheng, L.~Yin, Z.~Xie, C.~Sun, J.~Huang, C.~H. Yu, S.~Cao, C.~Kozyrakis,
  I.~Stoica, J.~E. Gonzalez, C.~Barrett, Y.~Sheng,
  \href{http://arxiv.org/abs/2312.07104}{{SGLang}: {Efficient} {Execution} of
  {Structured} {Language} {Model} {Programs}}, arXiv:2312.07104 [cs] (Jun.
  2024).
\newblock \href {https://doi.org/10.48550/arXiv.2312.07104}
  {\path{doi:10.48550/arXiv.2312.07104}}.
\newline\urlprefix\url{http://arxiv.org/abs/2312.07104}

\bibitem{chase_langchain_2022}
H.~Chase, \href{https://github.com/langchain-ai/langchain}{{LangChain}},
  original-date: 2022-10-17T02:58:36Z (Oct. 2022).
\newline\urlprefix\url{https://github.com/langchain-ai/langchain}

\bibitem{spaanderman_douwe-spaandermanmedicalrecordllm_2025}
D.~Spaanderman,
  \href{https://zenodo.org/records/17131185}{Douwe-{Spaanderman}/{MedicalRecordLLM}:
  v0.1} (Sep. 2025).
\newblock \href {https://doi.org/10.5281/zenodo.17131185}
  {\path{doi:10.5281/zenodo.17131185}}.
\newline\urlprefix\url{https://zenodo.org/records/17131185}

\bibitem{wei_chain--thought_2023}
J.~Wei, X.~Wang, D.~Schuurmans, M.~Bosma, B.~Ichter, F.~Xia, E.~Chi, Q.~Le,
  D.~Zhou, \href{http://arxiv.org/abs/2201.11903}{Chain-of-{Thought}
  {Prompting} {Elicits} {Reasoning} in {Large} {Language} {Models}},
  arXiv:2201.11903 [cs] (Jan. 2023).
\newblock \href {https://doi.org/10.48550/arXiv.2201.11903}
  {\path{doi:10.48550/arXiv.2201.11903}}.
\newline\urlprefix\url{http://arxiv.org/abs/2201.11903}

\bibitem{wang_self-consistency_2023}
X.~Wang, J.~Wei, D.~Schuurmans, Q.~Le, E.~Chi, S.~Narang, A.~Chowdhery,
  D.~Zhou, \href{http://arxiv.org/abs/2203.11171}{Self-{Consistency} {Improves}
  {Chain} of {Thought} {Reasoning} in {Language} {Models}}, arXiv:2203.11171
  [cs] (Mar. 2023).
\newblock \href {https://doi.org/10.48550/arXiv.2203.11171}
  {\path{doi:10.48550/arXiv.2203.11171}}.
\newline\urlprefix\url{http://arxiv.org/abs/2203.11171}

\bibitem{sun_prompt_2024}
S.~Sun, R.~Yuan, Z.~Cao, W.~Li, P.~Liu,
  \href{http://arxiv.org/abs/2406.00507}{Prompt {Chaining} or {Stepwise}
  {Prompt}? {Refinement} in {Text} {Summarization}}, arXiv:2406.00507 [cs]
  (Jun. 2024).
\newblock \href {https://doi.org/10.48550/arXiv.2406.00507}
  {\path{doi:10.48550/arXiv.2406.00507}}.
\newline\urlprefix\url{http://arxiv.org/abs/2406.00507}

\bibitem{deepseek-ai_deepseek-r1_2025}
DeepSeek-AI, D.~Guo, D.~Yang, H.~Zhang, J.~Song, R.~Zhang, R.~Xu, Q.~Zhu,
  S.~Ma, P.~Wang, X.~Bi, X.~Zhang, X.~Yu, Y.~Wu, Z.~F. Wu, Z.~Gou, Z.~Shao,
  Z.~Li, Z.~Gao, A.~Liu, B.~Xue, B.~Wang, B.~Wu, B.~Feng, C.~Lu, C.~Zhao,
  C.~Deng, C.~Zhang, C.~Ruan, D.~Dai, D.~Chen, D.~Ji, E.~Li, F.~Lin, F.~Dai,
  F.~Luo, G.~Hao, G.~Chen, G.~Li, H.~Zhang, H.~Bao, H.~Xu, H.~Wang, H.~Ding,
  H.~Xin, H.~Gao, H.~Qu, H.~Li, J.~Guo, J.~Li, J.~Wang, J.~Chen, J.~Yuan,
  J.~Qiu, J.~Li, J.~L. Cai, J.~Ni, J.~Liang, J.~Chen, K.~Dong, K.~Hu, K.~Gao,
  K.~Guan, K.~Huang, K.~Yu, L.~Wang, L.~Zhang, L.~Zhao, L.~Wang, L.~Zhang,
  L.~Xu, L.~Xia, M.~Zhang, M.~Zhang, M.~Tang, M.~Li, M.~Wang, M.~Li, N.~Tian,
  P.~Huang, P.~Zhang, Q.~Wang, Q.~Chen, Q.~Du, R.~Ge, R.~Zhang, R.~Pan,
  R.~Wang, R.~J. Chen, R.~L. Jin, R.~Chen, S.~Lu, S.~Zhou, S.~Chen, S.~Ye,
  S.~Wang, S.~Yu, S.~Zhou, S.~Pan, S.~S. Li, S.~Zhou, S.~Wu, S.~Ye, T.~Yun,
  T.~Pei, T.~Sun, T.~Wang, W.~Zeng, W.~Zhao, W.~Liu, W.~Liang, W.~Gao, W.~Yu,
  W.~Zhang, W.~L. Xiao, W.~An, X.~Liu, X.~Wang, X.~Chen, X.~Nie, X.~Cheng,
  X.~Liu, X.~Xie, X.~Liu, X.~Yang, X.~Li, X.~Su, X.~Lin, X.~Q. Li, X.~Jin,
  X.~Shen, X.~Chen, X.~Sun, X.~Wang, X.~Song, X.~Zhou, X.~Wang, X.~Shan, Y.~K.
  Li, Y.~Q. Wang, Y.~X. Wei, Y.~Zhang, Y.~Xu, Y.~Li, Y.~Zhao, Y.~Sun, Y.~Wang,
  Y.~Yu, Y.~Zhang, Y.~Shi, Y.~Xiong, Y.~He, Y.~Piao, Y.~Wang, Y.~Tan, Y.~Ma,
  Y.~Liu, Y.~Guo, Y.~Ou, Y.~Wang, Y.~Gong, Y.~Zou, Y.~He, Y.~Xiong, Y.~Luo,
  Y.~You, Y.~Liu, Y.~Zhou, Y.~X. Zhu, Y.~Xu, Y.~Huang, Y.~Li, Y.~Zheng, Y.~Zhu,
  Y.~Ma, Y.~Tang, Y.~Zha, Y.~Yan, Z.~Z. Ren, Z.~Ren, Z.~Sha, Z.~Fu, Z.~Xu,
  Z.~Xie, Z.~Zhang, Z.~Hao, Z.~Ma, Z.~Yan, Z.~Wu, Z.~Gu, Z.~Zhu, Z.~Liu, Z.~Li,
  Z.~Xie, Z.~Song, Z.~Pan, Z.~Huang, Z.~Xu, Z.~Zhang, Z.~Zhang,
  \href{http://arxiv.org/abs/2501.12948}{{DeepSeek}-{R1}: {Incentivizing}
  {Reasoning} {Capability} in {LLMs} via {Reinforcement} {Learning}},
  arXiv:2501.12948 [cs] (Jan. 2025).
\newblock \href {https://doi.org/10.48550/arXiv.2501.12948}
  {\path{doi:10.48550/arXiv.2501.12948}}.
\newline\urlprefix\url{http://arxiv.org/abs/2501.12948}

\bibitem{ai_llama_2025}
M.~AI,
  \href{https://huggingface.co/meta-llama/Llama-4-Maverick-17B-128E-Instruct}{Llama
  4 {Maverick}: {A} {Multimodal} {Mixture}-of-{Experts} {Model}}, publication
  Title: Hugging Face repository (2025).
\newline\urlprefix\url{https://huggingface.co/meta-llama/Llama-4-Maverick-17B-128E-Instruct}

\bibitem{yang_qwen3_2025}
A.~Yang, A.~Li, B.~Yang, B.~Zhang, B.~Hui, B.~Zheng, B.~Yu, C.~Gao, C.~Huang,
  C.~Lv, C.~Zheng, D.~Liu, F.~Zhou, F.~Huang, F.~Hu, H.~Ge, H.~Wei, H.~Lin,
  J.~Tang, J.~Yang, J.~Tu, J.~Zhang, J.~Yang, J.~Yang, J.~Zhou, J.~Zhou,
  J.~Lin, K.~Dang, K.~Bao, K.~Yang, L.~Yu, L.~Deng, M.~Li, M.~Xue, M.~Li,
  P.~Zhang, P.~Wang, Q.~Zhu, R.~Men, R.~Gao, S.~Liu, S.~Luo, T.~Li, T.~Tang,
  W.~Yin, X.~Ren, X.~Wang, X.~Zhang, X.~Ren, Y.~Fan, Y.~Su, Y.~Zhang, Y.~Zhang,
  Y.~Wan, Y.~Liu, Z.~Wang, Z.~Cui, Z.~Zhang, Z.~Zhou, Z.~Qiu,
  \href{http://arxiv.org/abs/2505.09388}{Qwen3 {Technical} {Report}},
  arXiv:2505.09388 [cs] (May 2025).
\newblock \href {https://doi.org/10.48550/arXiv.2505.09388}
  {\path{doi:10.48550/arXiv.2505.09388}}.
\newline\urlprefix\url{http://arxiv.org/abs/2505.09388}

\bibitem{ai_llama_2025-1}
M.~AI, \href{https://huggingface.co/meta-llama/Llama-4-Scout-17B-16E}{Llama 4
  {Scout}: {A} {Multimodal} {Mixture}-of-{Experts} {Model}}, publication Title:
  Hugging Face repository (2025).
\newline\urlprefix\url{https://huggingface.co/meta-llama/Llama-4-Scout-17B-16E}

\bibitem{qwen_qwen25_2025}
Qwen, A.~Yang, B.~Yang, B.~Zhang, B.~Hui, B.~Zheng, B.~Yu, C.~Li, D.~Liu,
  F.~Huang, H.~Wei, H.~Lin, J.~Yang, J.~Tu, J.~Zhang, J.~Yang, J.~Yang,
  J.~Zhou, J.~Lin, K.~Dang, K.~Lu, K.~Bao, K.~Yang, L.~Yu, M.~Li, M.~Xue,
  P.~Zhang, Q.~Zhu, R.~Men, R.~Lin, T.~Li, T.~Tang, T.~Xia, X.~Ren, X.~Ren,
  Y.~Fan, Y.~Su, Y.~Zhang, Y.~Wan, Y.~Liu, Z.~Cui, Z.~Zhang, Z.~Qiu,
  \href{http://arxiv.org/abs/2412.15115}{Qwen2.5 {Technical} {Report}},
  arXiv:2412.15115 [cs] (Jan. 2025).
\newblock \href {https://doi.org/10.48550/arXiv.2412.15115}
  {\path{doi:10.48550/arXiv.2412.15115}}.
\newline\urlprefix\url{http://arxiv.org/abs/2412.15115}

\bibitem{bercovich_llama-nemotron_2025}
A.~Bercovich, I.~Levy, I.~Golan, M.~Dabbah, R.~El-Yaniv, O.~Puny, I.~Galil,
  Z.~Moshe, T.~Ronen, N.~Nabwani, I.~Shahaf, O.~Tropp, E.~Karpas,
  R.~Zilberstein, J.~Zeng, S.~Singhal, A.~Bukharin, Y.~Zhang, T.~Konuk,
  G.~Shen, A.~S. Mahabaleshwarkar, B.~Kartal, Y.~Suhara, O.~Delalleau, Z.~Chen,
  Z.~Wang, D.~Mosallanezhad, A.~Renduchintala, H.~Qian, D.~Rekesh, F.~Jia,
  S.~Majumdar, V.~Noroozi, W.~U. Ahmad, S.~Narenthiran, A.~Ficek, M.~Samadi,
  J.~Huang, S.~Jain, I.~Gitman, I.~Moshkov, W.~Du, S.~Toshniwal, G.~Armstrong,
  B.~Kisacanin, M.~Novikov, D.~Gitman, E.~Bakhturina, P.~Varshney,
  M.~Narsimhan, J.~P. Scowcroft, J.~Kamalu, D.~Su, K.~Kong, M.~Kliegl, R.~K.
  Mahabadi, Y.~Lin, S.~Satheesh, J.~Parmar, P.~Gundecha, B.~Norick,
  J.~Jennings, S.~Prabhumoye, S.~N. Akter, M.~Patwary, A.~Khattar,
  D.~Narayanan, R.~Waleffe, J.~Zhang, B.-Y. Su, G.~Huang, T.~Kong, P.~Chadha,
  S.~Jain, C.~Harvey, E.~Segal, J.~Huang, S.~Kashirsky, R.~McQueen,
  I.~Putterman, G.~Lam, A.~Venkatesan, S.~Wu, V.~Nguyen, M.~Kilaru, A.~Wang,
  A.~Warno, A.~Somasamudramath, S.~Bhaskar, M.~Dong, N.~Assaf, S.~Mor, O.~U.
  Argov, S.~Junkin, O.~Romanenko, P.~Larroy, M.~Katariya, M.~Rovinelli,
  V.~Balas, N.~Edelman, A.~Bhiwandiwalla, M.~Subramaniam, S.~Ithape,
  K.~Ramamoorthy, Y.~Wu, S.~V. Velury, O.~Almog, J.~Daw, D.~Fridman,
  E.~Galinkin, M.~Evans, S.~Ghosh, K.~Luna, L.~Derczynski, N.~Pope, E.~Long,
  S.~Schneider, G.~Siman, T.~Grzegorzek, P.~Ribalta, M.~Katariya, C.~Alexiuk,
  J.~Conway, T.~Saar, A.~Guan, K.~Pawelec, S.~Prayaga, O.~Kuchaiev,
  B.~Ginsburg, O.~Olabiyi, K.~Briski, J.~Cohen, B.~Catanzaro, J.~Alben,
  Y.~Geifman, E.~Chung,
  \href{http://arxiv.org/abs/2505.00949}{Llama-{Nemotron}: {Efficient}
  {Reasoning} {Models}}, arXiv:2505.00949 [cs] (Sep. 2025).
\newblock \href {https://doi.org/10.48550/arXiv.2505.00949}
  {\path{doi:10.48550/arXiv.2505.00949}}.
\newline\urlprefix\url{http://arxiv.org/abs/2505.00949}

\bibitem{team_gemma_2025}
G.~Team, A.~Kamath, J.~Ferret, S.~Pathak, N.~Vieillard, R.~Merhej, S.~Perrin,
  T.~Matejovicova, A.~Ramé, M.~Rivière, L.~Rouillard, T.~Mesnard, G.~Cideron,
  J.-b. Grill, S.~Ramos, E.~Yvinec, M.~Casbon, E.~Pot, I.~Penchev, G.~Liu,
  F.~Visin, K.~Kenealy, L.~Beyer, X.~Zhai, A.~Tsitsulin, R.~Busa-Fekete,
  A.~Feng, N.~Sachdeva, B.~Coleman, Y.~Gao, B.~Mustafa, I.~Barr, E.~Parisotto,
  D.~Tian, M.~Eyal, C.~Cherry, J.-T. Peter, D.~Sinopalnikov, S.~Bhupatiraju,
  R.~Agarwal, M.~Kazemi, D.~Malkin, R.~Kumar, D.~Vilar, I.~Brusilovsky, J.~Luo,
  A.~Steiner, A.~Friesen, A.~Sharma, A.~Sharma, A.~M. Gilady, A.~Goedeckemeyer,
  A.~Saade, A.~Feng, A.~Kolesnikov, A.~Bendebury, A.~Abdagic, A.~Vadi,
  A.~György, A.~S. Pinto, A.~Das, A.~Bapna, A.~Miech, A.~Yang, A.~Paterson,
  A.~Shenoy, A.~Chakrabarti, B.~Piot, B.~Wu, B.~Shahriari, B.~Petrini, C.~Chen,
  C.~L. Lan, C.~A. Choquette-Choo, C.~J. Carey, C.~Brick, D.~Deutsch,
  D.~Eisenbud, D.~Cattle, D.~Cheng, D.~Paparas, D.~S. Sreepathihalli, D.~Reid,
  D.~Tran, D.~Zelle, E.~Noland, E.~Huizenga, E.~Kharitonov, F.~Liu,
  G.~Amirkhanyan, G.~Cameron, H.~Hashemi, H.~Klimczak-Plucińska, H.~Singh,
  H.~Mehta, H.~T. Lehri, H.~Hazimeh, I.~Ballantyne, I.~Szpektor, I.~Nardini,
  J.~Pouget-Abadie, J.~Chan, J.~Stanton, J.~Wieting, J.~Lai, J.~Orbay,
  J.~Fernandez, J.~Newlan, J.-y. Ji, J.~Singh, K.~Black, K.~Yu, K.~Hui,
  K.~Vodrahalli, K.~Greff, L.~Qiu, M.~Valentine, M.~Coelho, M.~Ritter,
  M.~Hoffman, M.~Watson, M.~Chaturvedi, M.~Moynihan, M.~Ma, N.~Babar, N.~Noy,
  N.~Byrd, N.~Roy, N.~Momchev, N.~Chauhan, N.~Sachdeva, O.~Bunyan, P.~Botarda,
  P.~Caron, P.~K. Rubenstein, P.~Culliton, P.~Schmid, P.~G. Sessa, P.~Xu,
  P.~Stanczyk, P.~Tafti, R.~Shivanna, R.~Wu, R.~Pan, R.~Rokni, R.~Willoughby,
  R.~Vallu, R.~Mullins, S.~Jerome, S.~Smoot, S.~Girgin, S.~Iqbal, S.~Reddy,
  S.~Sheth, S.~Põder, S.~Bhatnagar, S.~R. Panyam, S.~Eiger, S.~Zhang, T.~Liu,
  T.~Yacovone, T.~Liechty, U.~Kalra, U.~Evci, V.~Misra, V.~Roseberry,
  V.~Feinberg, V.~Kolesnikov, W.~Han, W.~Kwon, X.~Chen, Y.~Chow, Y.~Zhu,
  Z.~Wei, Z.~Egyed, V.~Cotruta, M.~Giang, P.~Kirk, A.~Rao, K.~Black, N.~Babar,
  J.~Lo, E.~Moreira, L.~G. Martins, O.~Sanseviero, L.~Gonzalez, Z.~Gleicher,
  T.~Warkentin, V.~Mirrokni, E.~Senter, E.~Collins, J.~Barral, Z.~Ghahramani,
  R.~Hadsell, Y.~Matias, D.~Sculley, S.~Petrov, N.~Fiedel, N.~Shazeer,
  O.~Vinyals, J.~Dean, D.~Hassabis, K.~Kavukcuoglu, C.~Farabet, E.~Buchatskaya,
  J.-B. Alayrac, R.~Anil, Dmitry, Lepikhin, S.~Borgeaud, O.~Bachem, A.~Joulin,
  A.~Andreev, C.~Hardin, R.~Dadashi, L.~Hussenot,
  \href{http://arxiv.org/abs/2503.19786}{Gemma 3 {Technical} {Report}},
  arXiv:2503.19786 [cs] (Mar. 2025).
\newblock \href {https://doi.org/10.48550/arXiv.2503.19786}
  {\path{doi:10.48550/arXiv.2503.19786}}.
\newline\urlprefix\url{http://arxiv.org/abs/2503.19786}

\bibitem{ai_mistral_2025}
M.~AI,
  \href{https://huggingface.co/mistralai/Mistral-Small-3.1-24B-Instruct-2503}{Mistral
  {Small} 3.1 {24B} {Instruct} 2503}, publication Title: Hugging Face
  repository (2025).
\newline\urlprefix\url{https://huggingface.co/mistralai/Mistral-Small-3.1-24B-Instruct-2503}

\bibitem{christophe_med42-v2_2024}
C.~Christophe, P.~K. Kanithi, T.~Raha, S.~Khan, M.~A. Pimentel,
  \href{http://arxiv.org/abs/2408.06142}{Med42-v2: {A} {Suite} of {Clinical}
  {LLMs}}, arXiv:2408.06142 [cs] (Aug. 2024).
\newblock \href {https://doi.org/10.48550/arXiv.2408.06142}
  {\path{doi:10.48550/arXiv.2408.06142}}.
\newline\urlprefix\url{http://arxiv.org/abs/2408.06142}

\bibitem{ankit_pal_openbiollms_2024}
M.~S. Ankit~Pal,
  \href{https://huggingface.co/aaditya/OpenBioLLM-Llama3-70B}{{OpenBioLLMs}:
  {Advancing} {Open}-{Source} {Large} {Language} {Models} for {Healthcare} and
  {Life} {Sciences}}, publication Title: Hugging Face repository (2024).
\newline\urlprefix\url{https://huggingface.co/aaditya/OpenBioLLM-Llama3-70B}

\bibitem{sellergren_medgemma_2025}
A.~Sellergren, S.~Kazemzadeh, T.~Jaroensri, A.~Kiraly, M.~Traverse,
  T.~Kohlberger, S.~Xu, F.~Jamil, C.~Hughes, C.~Lau, J.~Chen, F.~Mahvar,
  L.~Yatziv, T.~Chen, B.~Sterling, S.~A. Baby, S.~M. Baby, J.~Lai,
  S.~Schmidgall, L.~Yang, K.~Chen, P.~Bjornsson, S.~Reddy, R.~Brush,
  K.~Philbrick, M.~Asiedu, I.~Mezerreg, H.~Hu, H.~Yang, R.~Tiwari, S.~Jansen,
  P.~Singh, Y.~Liu, S.~Azizi, A.~Kamath, J.~Ferret, S.~Pathak, N.~Vieillard,
  R.~Merhej, S.~Perrin, T.~Matejovicova, A.~Ramé, M.~Riviere, L.~Rouillard,
  T.~Mesnard, G.~Cideron, J.-b. Grill, S.~Ramos, E.~Yvinec, M.~Casbon,
  E.~Buchatskaya, J.-B. Alayrac, D.~Lepikhin, V.~Feinberg, S.~Borgeaud,
  A.~Andreev, C.~Hardin, R.~Dadashi, L.~Hussenot, A.~Joulin, O.~Bachem,
  Y.~Matias, K.~Chou, A.~Hassidim, K.~Goel, C.~Farabet, J.~Barral,
  T.~Warkentin, J.~Shlens, D.~Fleet, V.~Cotruta, O.~Sanseviero, G.~Martins,
  P.~Kirk, A.~Rao, S.~Shetty, D.~F. Steiner, C.~Kirmizibayrak, R.~Pilgrim,
  D.~Golden, L.~Yang, \href{http://arxiv.org/abs/2507.05201}{{MedGemma}
  {Technical} {Report}}, arXiv:2507.05201 [cs] (Jul. 2025).
\newblock \href {https://doi.org/10.48550/arXiv.2507.05201}
  {\path{doi:10.48550/arXiv.2507.05201}}.
\newline\urlprefix\url{http://arxiv.org/abs/2507.05201}

\bibitem{kemeny_mathematics_1959}
J.~G. Kemeny, \href{https://www.jstor.org/stable/20026529}{Mathematics without
  {Numbers}}, Daedalus 88~(4) (1959) 577--591, publisher: The MIT Press.
\newline\urlprefix\url{https://www.jstor.org/stable/20026529}

\bibitem{lindstrom_newton-raphson_1988}
M.~J. Lindstrom, D.~M. Bates,
  \href{https://www.jstor.org/stable/2290128}{Newton-{Raphson} and {EM}
  {Algorithms} for {Linear} {Mixed}-{Effects} {Models} for
  {Repeated}-{Measures} {Data}}, Journal of the American Statistical
  Association 83~(404) (1988) 1014--1022, publisher: [American Statistical
  Association, Taylor \& Francis, Ltd.].
\newblock \href {https://doi.org/10.2307/2290128} {\path{doi:10.2307/2290128}}.
\newline\urlprefix\url{https://www.jstor.org/stable/2290128}

\bibitem{dennstadt_implementing_2025}
F.~Dennstädt, J.~Hastings, P.~M. Putora, M.~Schmerder, N.~Cihoric,
  \href{https://pmc.ncbi.nlm.nih.gov/articles/PMC11885444/}{Implementing large
  language models in healthcare while balancing control, collaboration, costs
  and security}, NPJ Digital Medicine 8 (2025) 143.
\newblock \href {https://doi.org/10.1038/s41746-025-01476-7}
  {\path{doi:10.1038/s41746-025-01476-7}}.
\newline\urlprefix\url{https://pmc.ncbi.nlm.nih.gov/articles/PMC11885444/}

\bibitem{tam_framework_2024}
T.~Y.~C. Tam, S.~Sivarajkumar, S.~Kapoor, A.~V. Stolyar, K.~Polanska, K.~R.
  McCarthy, H.~Osterhoudt, X.~Wu, S.~Visweswaran, S.~Fu, P.~Mathur, G.~E.
  Cacciamani, C.~Sun, Y.~Peng, Y.~Wang,
  \href{https://pmc.ncbi.nlm.nih.gov/articles/PMC11437138/}{A framework for
  human evaluation of large language models in healthcare derived from
  literature review}, NPJ Digital Medicine 7 (2024) 258.
\newblock \href {https://doi.org/10.1038/s41746-024-01258-7}
  {\path{doi:10.1038/s41746-024-01258-7}}.
\newline\urlprefix\url{https://pmc.ncbi.nlm.nih.gov/articles/PMC11437138/}

\end{thebibliography}

\clearpage
\begin{appendices}
\section*{Supplementary Materials}
	
	\suppsection{Use Cases}\label{sec:LLMSup1}
	In this section, we describe all use cases investigated in this study. An overview is provided in \autoref{tab:use-case-summary}, which summarizes the report type, language, number of questions, number of patients, and task characteristics investigated in this study. Distributions are visualized using histograms for both categorical and numerical variables, and t-SNE plots of embeddings generated with \opus{embeddinggemma-300m-medical} for free-text variables \citeSup{maaten_visualizing_2008,reimers_sentence-bert_2019,gao2021scaling}. More details on this embedding function are described in \autoref{sec:text_embedding}. For each use case, we describe the specific extraction tasks, inclusion and exclusion criteria, annotation procedures, and the distributions of ground-truth variables in the following sections.
	
	\begin{table}[h]
		\centering
		\caption{Summary of use cases for information extraction from medical reports across diverse clinical contexts.}
		\scriptsize
		\setlength{\tabcolsep}{4pt}
		\begin{tabular}{@{}
				>{\raggedright\arraybackslash}p{0.18\textwidth} 
				>{\raggedright\arraybackslash}p{0.14\textwidth} 
				>{\raggedright\arraybackslash}p{0.10\textwidth} 
				>{\raggedright\arraybackslash}p{0.10\textwidth} 
				>{\raggedright\arraybackslash}p{0.10\textwidth} 
				>{\raggedright\arraybackslash}p{0.28\textwidth} @{}
			}
			\toprule
			\textbf{Use Case (Name)} & \textbf{Report Type} & \textbf{Language} & \textbf{Number of Questions} & \textbf{Number of Patients} & \textbf{Short Description of Task} \\
			\midrule
			Colorectal Liver Metastases & Pathology Report & Dutch & 4 & 864 & Extract liver resection characteristics. \\
			\midrule
			Liver Tumours & Pathology Report & Dutch & 12 & 289 & Extract disease phenotypes from the conclusion sections. \\
			\midrule
			Neurodegenerative Diseases & Radiology Report & Dutch & 25 & 948 & Extract imaging markers, including vascular lesion severity, and brain atrophy scores. \\
			\midrule
			Soft Tissue Tumours & Pathology Report & English \& Dutch & 9 & 627 (300 in English and 327 in Dutch) & Extract phenotype, tumour location, and grade. \\
			\midrule
			Melanoma & Pathology Report & Dutch & 16 & 1,252 & Extract data on the primary tumour and sentinel lymph node involvement. \\
			\midrule\
			Sarcoma & Pathology Report & Czech & 7 & 80 & Extract histopathological markers. \\
			\bottomrule
		\end{tabular}
		\label{tab:use-case-summary}
	\end{table}
	
	\newpage
	\subsection{Use Case - Colorectal Liver Metastases (Dutch)}
	
	\subsubsection{Overview}
	This use case focuses on extracting structured information from pathology reports of patients undergoing resection of colorectal liver metastases and/or primary colorectal resection \autoref{tab:use-case-CRLM}. The objective is to accurately extract key variables of interest from these reports, including:  
	\begin{itemize}
		\item List of liver segments resected during first surgery  
		\item Whether a hemihepatectomy was performed  
		\item Surgical resection margin (mm)  
		\item Radicality of liver surgery (R classification)  
	\end{itemize}
	No additional special considerations were applicable for this use case.
	
	\subsubsection{Inclusion and Exclusion Criteria}
	\begin{itemize}
		\item Inclusion criteria: Patients surgically treated for colorectal liver metastases in a tertiary center between 2000–2019.  
		\item Exclusion criteria: Patients without a surgical resection.  
		\item Dataset size: 865 reports.  
	\end{itemize}
	
	\subsubsection{Annotation}
	Annotations were performed by medical master’s students and PhD students. In total, more than 10 annotators were involved. Inter-annotator agreement was reached through discussions.
	
	\subsubsection{Ethical Considerations and Funding}
	Ethical approval was granted by the Medical Ethics Committee of the Erasmus Medical Center (MEC-2018-1743). Funding for this study was provided by Erasmus MC research funding.
	
	\begin{scriptsize}
		\setlength{\tabcolsep}{4pt}
		\begin{longtable}{@{}
				>{\scriptsize\raggedright\arraybackslash}p{0.12\textwidth} % Variable name
				>{\scriptsize\raggedright\arraybackslash}p{0.21\textwidth} % Description
				>{\scriptsize\raggedright\arraybackslash}p{0.12\textwidth} % Variable type
				>{\scriptsize\raggedright\arraybackslash}p{0.15\textwidth} % Variable options
				>{\scriptsize\centering\arraybackslash}p{0.30\textwidth} @{}} % Distribution with image
			\multicolumn{5}{c}{\parbox{\textwidth}{
					\normalsize \tablename~\thetable{} -- Colorectal liver metastases pathology report variable definitions with reference standard distribution.\\}} \\
			\toprule
			\textbf{Variable Name} & 
			\textbf{Description} & 
			\textbf{Type (Metric)} & 
			\textbf{Variable Options} & 
			\textbf{Reference Standard Distribution} \\
			\midrule
			\endfirsthead
			
			\multicolumn{5}{c}{\parbox{\textwidth}{
					\normalsize \tablename~\thetable{} -- Continued\\}} \\
			\toprule
			\textbf{Variable Name} & 
			\textbf{Description} & 
			\textbf{Type (Metric)} & 
			\textbf{Variable Options} & 
			\textbf{Reference Standard Distribution} \\
			\midrule
			\endhead
			
			\midrule
			\endfoot
			
			\bottomrule
			\endlastfoot
			\phantomlabel{tab:use-case-CRLM}
			List of liver segments resected during first surgery & Liver segments removed in the first surgery & List (symmetric similarity) & - & 
			\raisebox{-\totalheight}{\includegraphics[width=\linewidth]{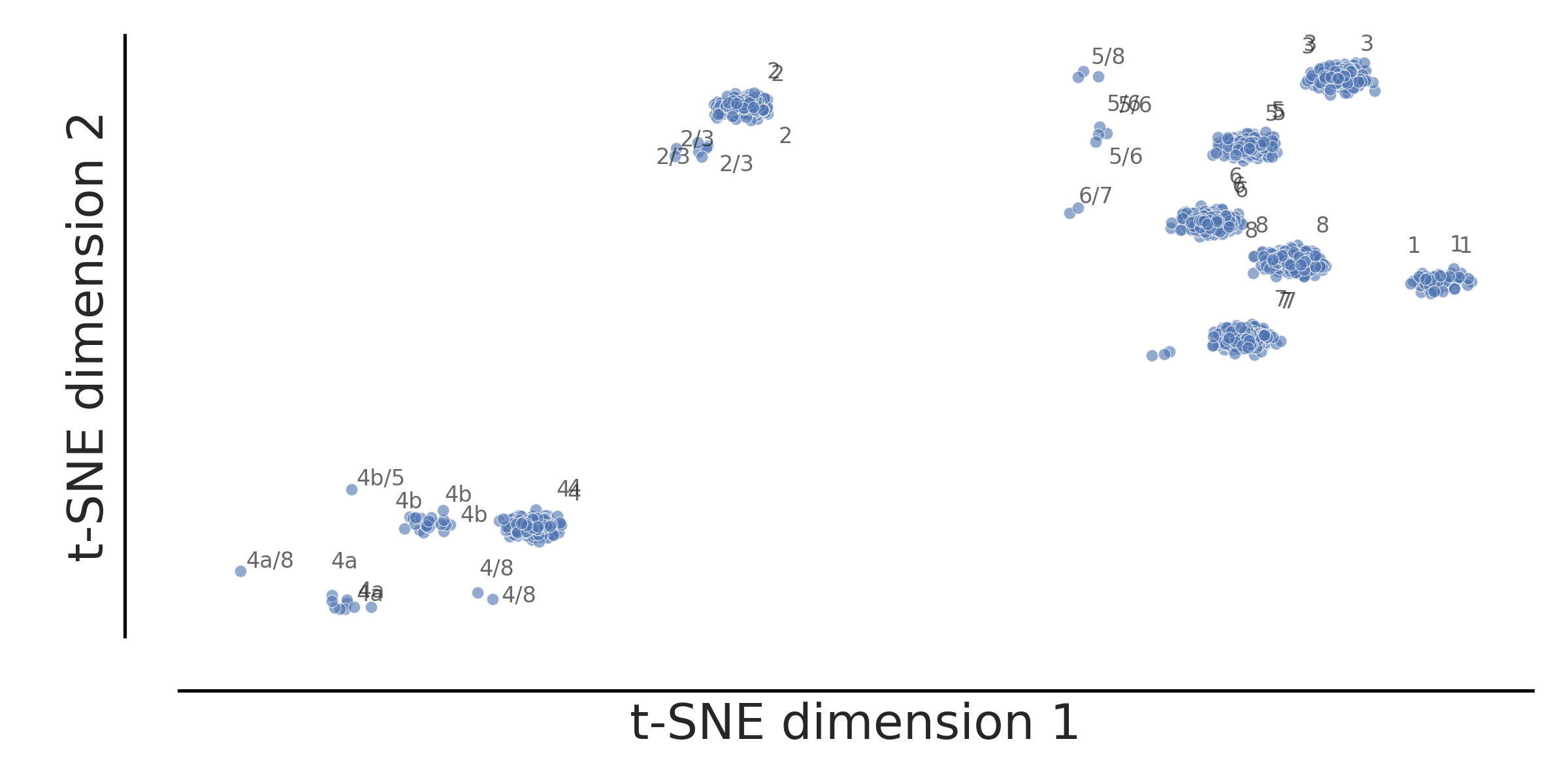}} \\
			
			Was a hemihepatectomy performed? & Indicates whether a right or left hemihepatectomy was performed during the first surgery & Binary (balanced accuracy) & Yes, Not specified & 
			\raisebox{-\totalheight}{\includegraphics[width=\linewidth]{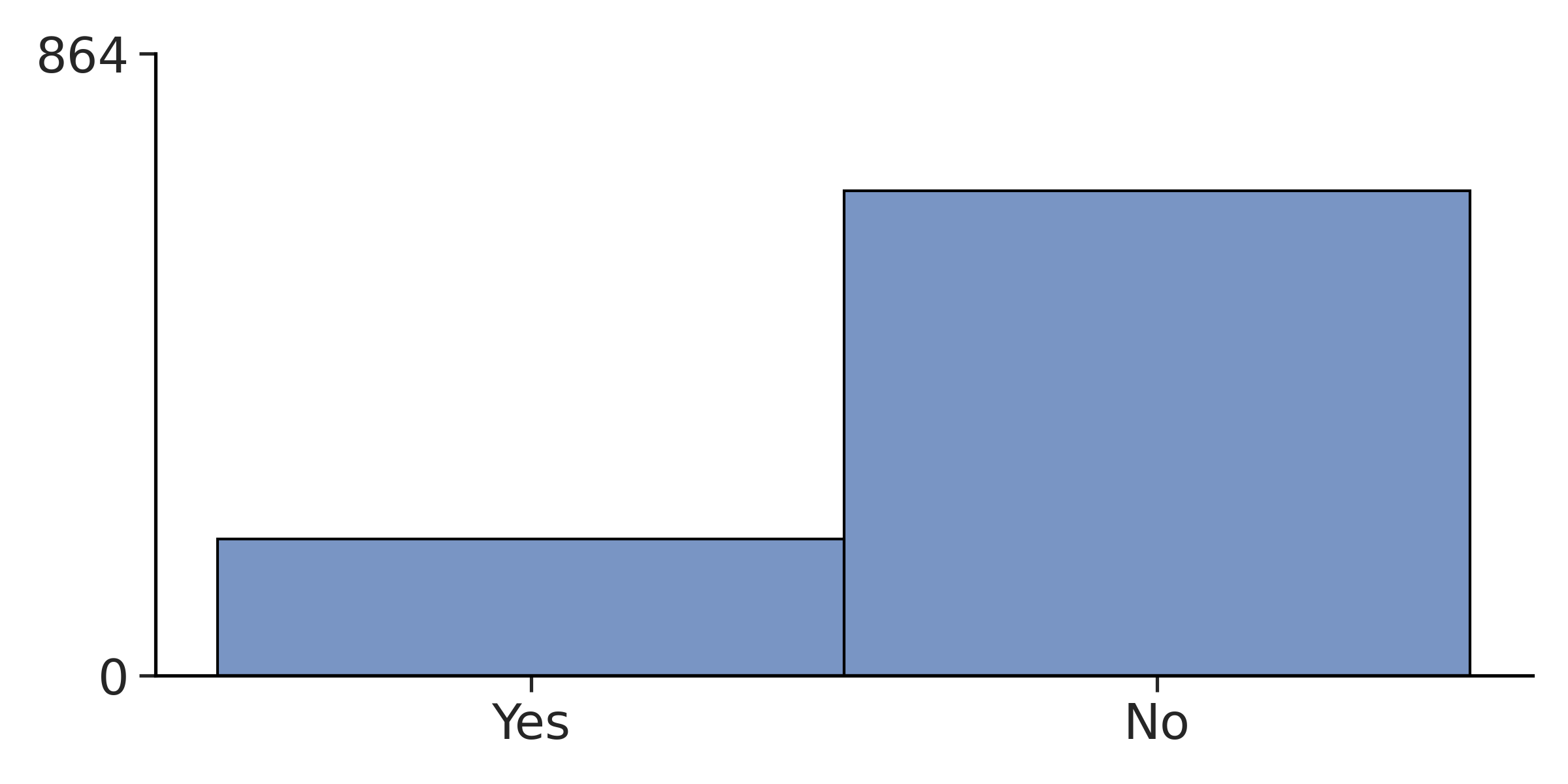}} \\
			
			Surgical resection margin (mm) & Closest distance in millimetres between tumour and resection margin & Numeric (accuracy) & Any non-negative integer value (e.g., 2, 5, 10) & 
			\raisebox{-\totalheight}{\includegraphics[width=\linewidth]{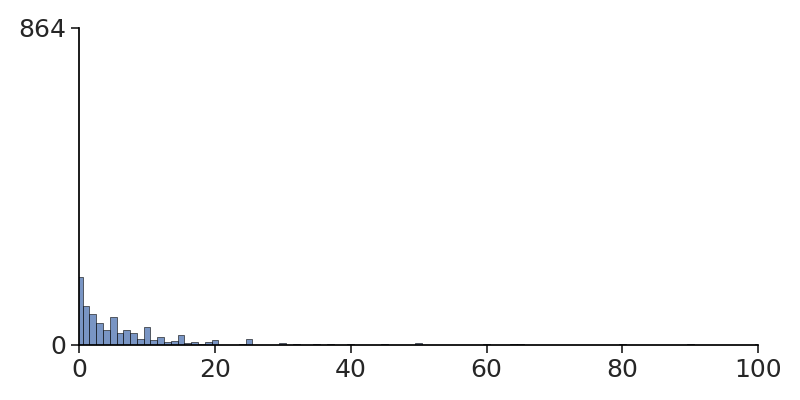}} \\
			
			Radicality of liver surgery (R classification) & Radicality classification of liver resection & Categorical (balanced accuracy) & R0, R1, Not specified & 
			\raisebox{-\totalheight}{\includegraphics[width=\linewidth]{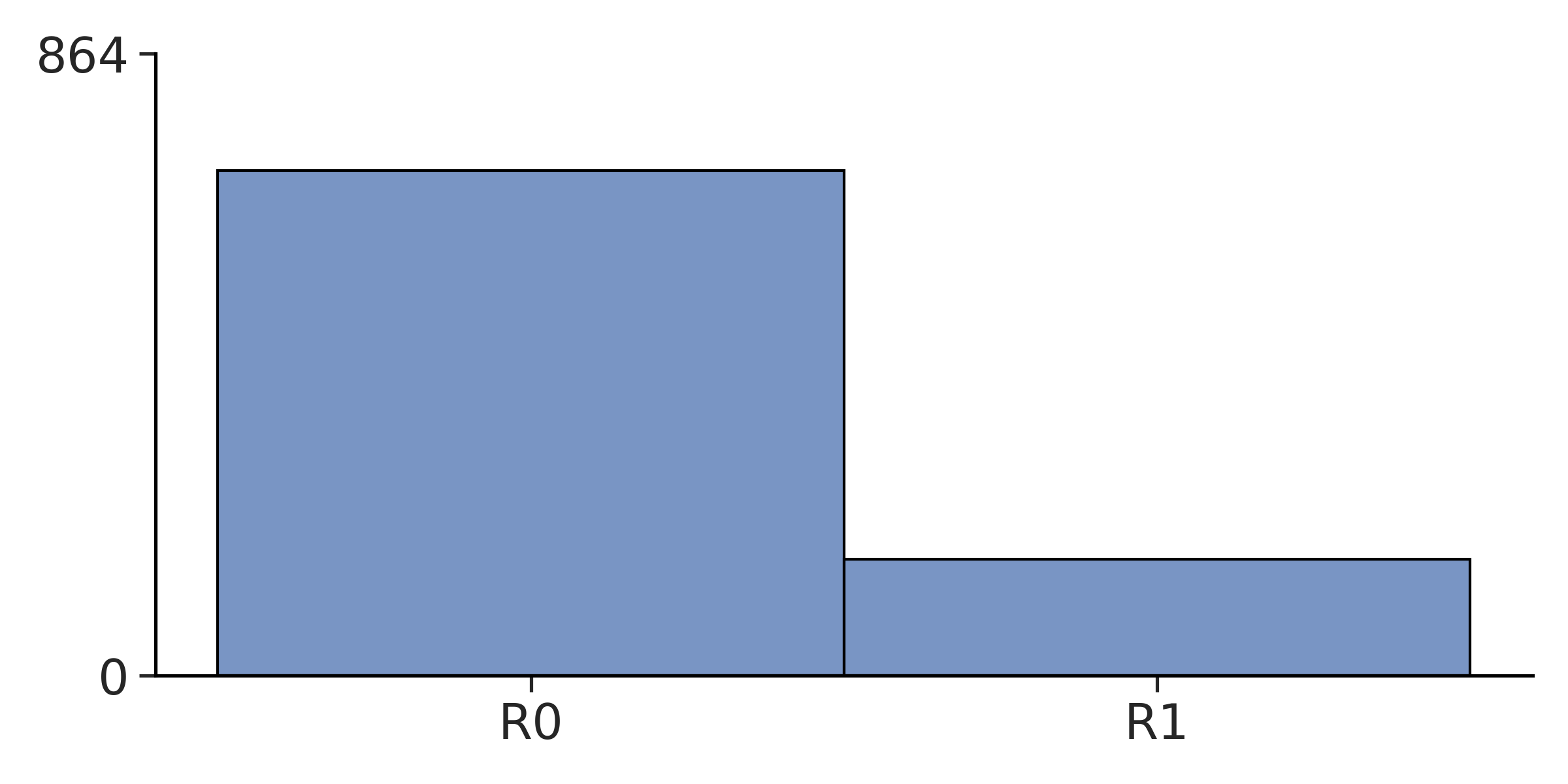}} \\
		\end{longtable}
	\end{scriptsize}
	
	\subsubsection{Prompt}
	For this use case, a structured prompt was created to ensure consistent extraction and enforce strict JSON formatting. 
	The prompt is divided into role-based instructions (System, Human, AI). A full example of the exact prompt used is shown in \autoref{tab:prompt-CRLM}.
	
	\begin{scriptsize}
		\setlength{\tabcolsep}{4pt}
		\begin{longtable}{>{\scriptsize\raggedright\arraybackslash}m{0.2\linewidth} >{\scriptsize\raggedright\arraybackslash}m{0.37\linewidth} >{\scriptsize\raggedright\arraybackslash}m{0.37\linewidth}}
			\multicolumn{3}{c}{\parbox{\textwidth}{
					\normalsize \tablename~\thetable{} -- Full structured prompt used for the colorectal liver metastases extraction task. The items in brackets indicate the role of the message (System, Human, AI), while the text provides the corresponding content.\\}} \\
			\toprule
			\textbf{Section} & \multicolumn{2}{c}{\textbf{Content} (Depending on prompt strategy)} \\
			& Zero-Shot, One-Shot and Few-Shot & CoT, Self-Consistency and Graph\\
			\midrule
			\endfirsthead
			
			\multicolumn{3}{c}{\parbox{\textwidth}{
					\normalsize \tablename~\thetable{} -- Continued\\}} \\
			\toprule
			\textbf{Section} & \multicolumn{2}{c}{\textbf{Content}} \\
			\midrule
			\endhead
			
			\phantomlabel{tab:prompt-CRLM}
			\textbf{[System] -- System instructions} & 
			You are a medical data extraction system that ONLY outputs valid JSON. Maintain strict compliance with these rules: \newline
			1. ALWAYS begin and end your response with \verb|```json| markers \newline
			2. Use EXACT field names and structure provided \newline
			3. If a value is missing or not mentioned, use the specified default for that field. \newline
			4. NEVER add commentary, explanations, or deviate from the output structure & 
			You are a medical data extraction system that performs structured reasoning before producing output. Follow these strict rules: \newline
			1. First, reason step-by-step to identify and justify each extracted field. \newline
			2. After reasoning, output ONLY valid JSON in the exact structure provided. \newline
			3. ALWAYS begin and end the final output with \verb|```json| markers — do not include reasoning within these markers. \newline
			4. Use EXACT field names and structure as specified. \newline
			5. If a value is missing or not mentioned, use the specified default for that field. \newline
			6. NEVER include commentary, explanations, or deviate from the specified format in the final JSON. \newline \\
			\midrule
			\textbf{[Human] -- Field instructions} & \multicolumn{2}{c}{\parbox{0.75\linewidth}{
					1. \opus{"List of liver segments resected during first surgery"}: \newline
					- Type: list \newline
					- List of liver segments resected during first surgery in exact format reported (e.g., "2", "4a", "5/6", "7/8") and in ascending order (e.g., ['2', '3', '4a', '5/6', '7/8']). Preserve multiple mentions if repeated. Exclude extrahepatic organs. Leave empty if not specified. \newline
					- Default: empty list (\opus{[]}) \newline
					2. \opus{"Was a hemihepatectomy performed?"}: \newline
					- Type: string \newline
					- Indicate "Yes" if a right or left hemihepatectomy was performed during first surgery, else "". \newline
					- Options: ["Yes", ""] \newline
					- Default: "" \newline
					3. \opus{"Surgical resection margin (mm)"}: \newline
					- Type: number \newline
					- Distance in mm between tumour and closest resection margin as described in the pathology report. Use numeric values only (e.g., 2, 5, 10). Leave as "" if not specified. \newline
					- Default: "" \newline
					4. \opus{"Radicality of liver surgery (R classification)"}: \newline
					- Type: string \newline
					- Radicality based on resection margin. "R0" if margins $>$1 mm or explicitly radical, "R1" if microscopic tumour at margin or $\leq$1 mm. Leave as "" if not mentioned. \newline
					- Options: ["R0", "R1"] \newline
					- Default: "" \newline
			}} \\
			\midrule
			\textbf{[Human] -- Task instructions} & \multicolumn{2}{c}{\parbox{0.75\linewidth}{
					Extract information into this exact JSON structure:
					\opus{```json} \newline
					\opus{\{} \newline
					\opus{\quad "List of liver segments resected during first surgery": [],} \newline
					\opus{\quad "Was a hemihepatectomy performed?": "",} \newline
					\opus{\quad "Surgical resection margin (mm)": "",} \newline
					\opus{\quad "Radicality of liver surgery (R classification)": "",} \newline
					\opus{\}} \newline
					\opus{```}%
				}%
			} \\
			\midrule
			\textbf{[Human] -- Example intro} & \multicolumn{2}{c}{\parbox{0.75\linewidth}{Below are 1 example of expected input and output, followed by a new task.}} \\
			\midrule
			\textbf{[Human] -- Example user} & \multicolumn{2}{c}{\parbox{0.75\linewidth}{\opus{[...CENSORED...]}}} \\
			\midrule
			\textbf{[AI] -- Example assistant reasoning} & - & 
			- \opus{List of liver segments resected during first surgery} - "wedge resection of segment IVB" implies ["4b"] \newline
			- \opus{Was a hemihepatectomy performed?} - "only a wedge, not a hemihepatectomy" implies "" \newline
			- \opus{Surgical resection margin (mm)} - "narrowest surgical margin is explicitly stated as 1 cm" implies 10 \newline
			- \opus{Radicality of liver surgery (R classification)} - "resection margin is free (‘resectievlak vrij’)" implies "R0" \newline \\
			\midrule
			\textbf{[AI] -- Example assistant output} & \multicolumn{2}{c}{\parbox{0.75\linewidth}{
					\opus{```json} \newline
					\opus{\{} \newline
					\opus{\quad "List of liver segments resected during first surgery": ["4b"],} \newline
					\opus{\quad "Was a hemihepatectomy performed?": "",} \newline
					\opus{\quad "Surgical resection margin (mm)": "10",} \newline
					\opus{\quad "Radicality of liver surgery (R classification)": "R0",} \newline
					\opus{\}} \newline
					\opus{```}%
				}%
			} \newline \\
			\midrule
			\textbf{[Human] -- Report instructions} & \multicolumn{2}{c}{\parbox{0.75\linewidth}{[file name]: \opus{[...CENSORED...]} \newline \opus{[...CENSORED...]}}} \\
			\midrule
			\textbf{[Human] -- Final instructions} &
			Begin the extraction now. Your response must contain only a single valid JSON block, enclosed in triple backticks and prefixed with \verb|`json`|, like this: \verb|```json  ... ```|& 
			Begin the extraction now. First, reason step-by-step to identify and justify the value for each required field, enclosed within \verb|<think>...</think>| tags. Then, output only the final structured data as a single valid JSON block, starting with \verb|```json| and ending with \verb|```|.
			\\
			\bottomrule
		\end{longtable}
	\end{scriptsize}
	
	\subsubsection{Prompt Graph}
	The dependencies, conditional branches, and extraction order of variables are represented as a directed acyclic graph. This graph reflects how the extraction task is decomposed into smaller, sequential subtasks for the Prompt Graph prompting strategy. 
	
	\begin{figure}[htbp]
		\centering
		\includegraphics[width=\linewidth, height=0.5\textheight, keepaspectratio]{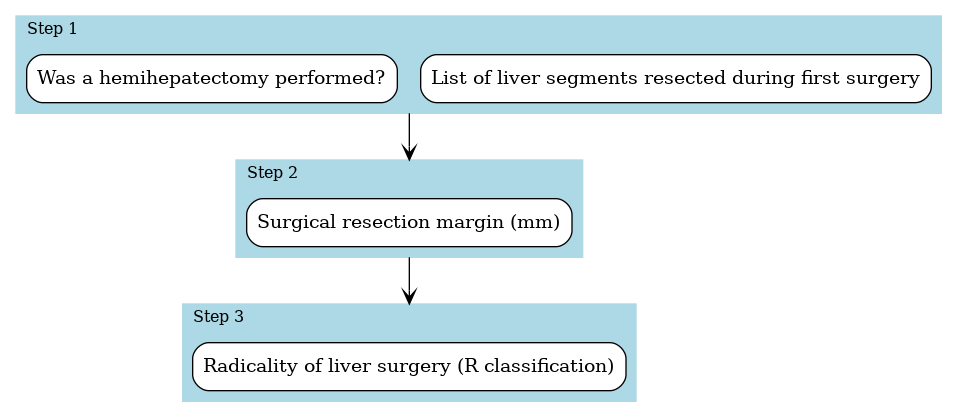}
		\caption{Directed acyclic graph showing sequential extraction order of variable extraction for colorectal liver metastases use case.}
	\end{figure}
	
	\newpage
	\subsection{Use Case – Liver Tumours (Dutch)}
	
	\subsubsection{Overview}
	This use case focuses on extracting structured information from pathology reports of patients with liver tumours. Specifically, the task involves extracting relevant information from the conclusion fields of Dutch liver tumour pathology reports \autoref{tab:use-case-Liver}. The following information was targeted: number of tissue samples, organs, method of tissue retrieval, whether the sample is an explant, whether it is a frozen section, number of liver lesion types, locations of tissue samples extracted from the liver, lesion phenotype, presence of metastasis, primary site of metastasis, steatosis, presence of fibrosis, and inflammation.
	
	\subsubsection{Inclusion and Exclusion Criteria}
	The dataset consists of two subsets:  
	\begin{itemize}
		\item One subset contains a selection of liver tumour pathology reports from the Erasmus MC from 2023. This subset was created with the goal of achieving a relatively balanced distribution of liver tumour phenotypes.  
		\item The other subset contains a selection of liver tumour pathology reports from the Erasmus MC from 2015–2022. This subset mainly contains one specific phenotype, hepatocellular adenoma.  
	\end{itemize}
	
	\subsubsection{Annotation}
	Annotations were performed by one PhD student.
	
	\subsubsection{Ethical Considerations and Funding}
	Ethical approval was granted by the Medical Ethics Committee of the Erasmus Medical Center on September 4th, 2024 (METC number: MEC-2024-0493). Funding for this study was provided by the Dutch Research Council (NWO) for the project ``The Liver Artificial Intelligence (LAI) consortium: a benchmark dataset and optimized machine learning methods for MRI-based diagnosis of solid appearing liver lesions'' (file number: 20431).

	\begin{scriptsize}
		\setlength{\tabcolsep}{4pt}
		\begin{longtable}{@{}
				>{\scriptsize\raggedright\arraybackslash}p{0.12\textwidth} % Variable name
				>{\scriptsize\raggedright\arraybackslash}p{0.21\textwidth} % Description
				>{\scriptsize\raggedright\arraybackslash}p{0.12\textwidth} % Variable type
				>{\scriptsize\raggedright\arraybackslash}p{0.15\textwidth} % Variable options
				>{\scriptsize\centering\arraybackslash}p{0.30\textwidth} @{}} % Distribution with image
			\multicolumn{5}{c}{\parbox{\textwidth}{
					\normalsize \tablename~\thetable{} -- Liver tumours pathology report variable definitions with reference standard distribution.\\}} \\
			\toprule
			\textbf{Variable Name} & 
			\textbf{Description} & 
			\textbf{Type (Metric)} & 
			\textbf{Variable Options} & 
			\textbf{Reference Standard Distribution} \\
			\midrule
			\endfirsthead
			
			\multicolumn{5}{c}{\parbox{\textwidth}{
					\normalsize \tablename~\thetable{} -- Continued\\}} \\
			\toprule
			\textbf{Variable Name} & 
			\textbf{Description} & 
			\textbf{Type (Metric)} & 
			\textbf{Variable Options} & 
			\textbf{Reference Standard Distribution} \\
			\midrule
			\endhead
			
			\midrule
			\endfoot
			
			\bottomrule
			\endlastfoot
			
			\phantomlabel{tab:use-case-Liver}
			Number of tissue samples & Total number of tissue samples described in the report & Numeric (accuracy) & Any non-negative integer ($\geq$1) & 
			\raisebox{-\totalheight}{\includegraphics[width=\linewidth]{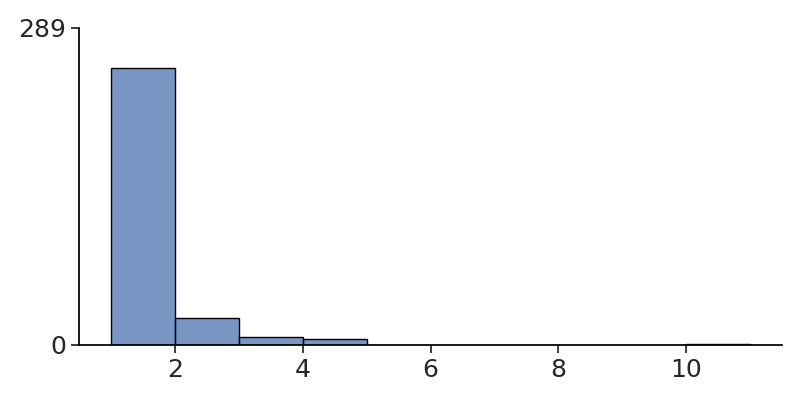}} \\
			
			Organs & Organs from which the tissue samples were obtained & List (symmetric similarity) & - & 
			\raisebox{-\totalheight}{\includegraphics[width=\linewidth]{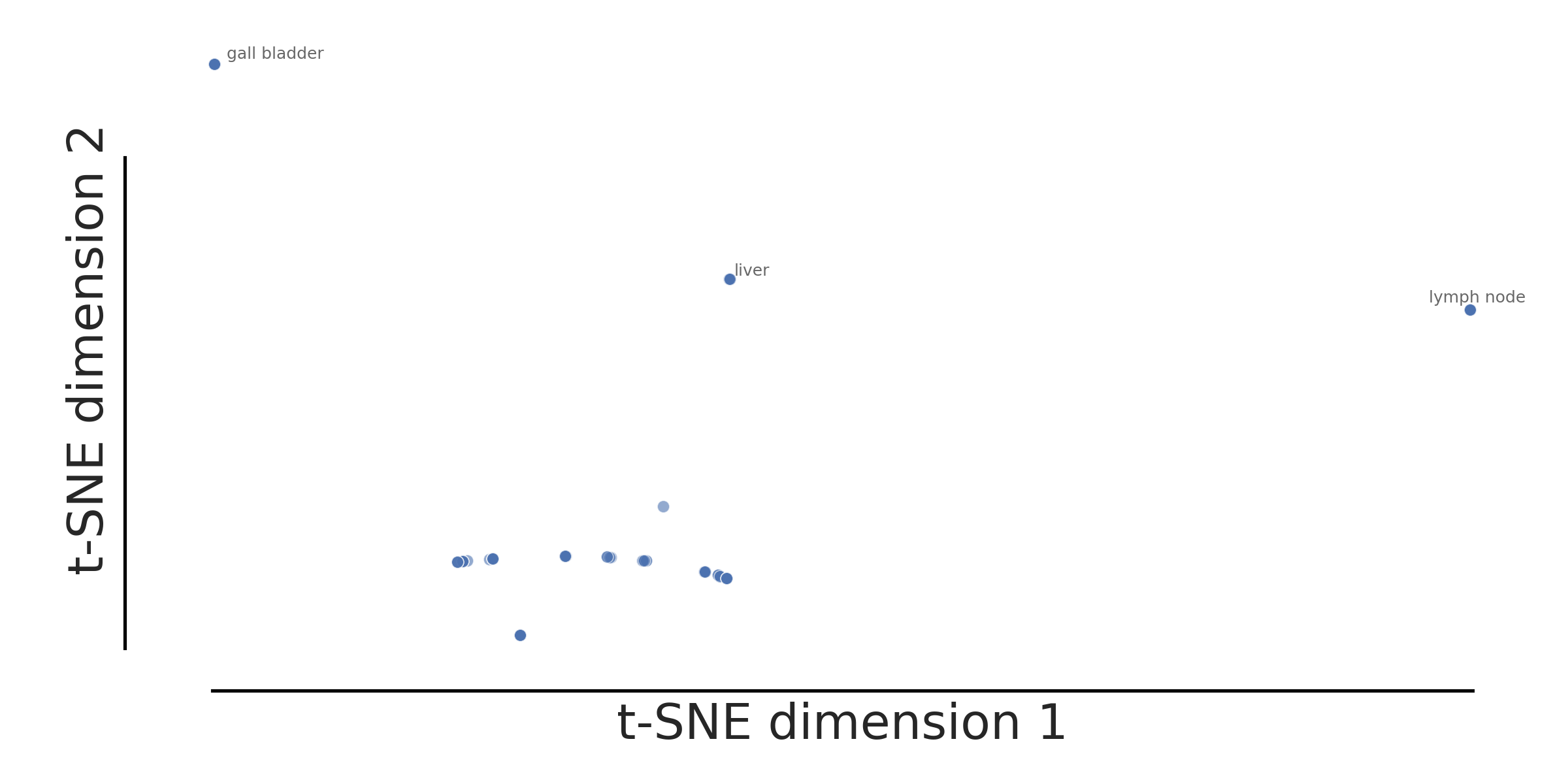}} \\
			
			Method of tissue retrieval & Retrieval method used to obtain the specimen & Categorical (balanced accuracy) & Biopsy, Excision, Resection, Brush, Not specified & 
			\raisebox{-\totalheight}{\includegraphics[width=\linewidth]{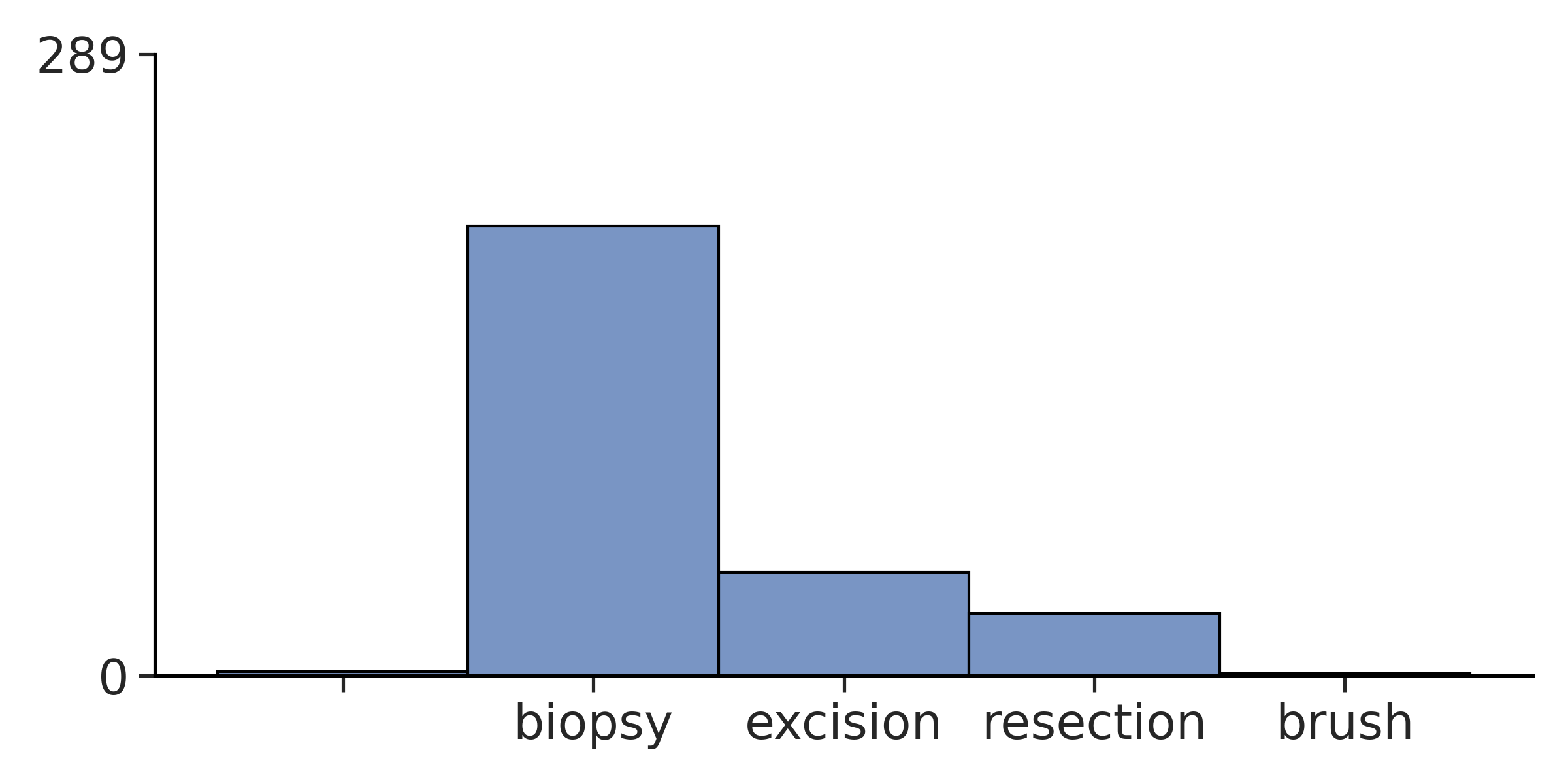}} \\
			
			Is it an explant? & Indicates whether the specimen is an explantation (entire organ removed) & Binary (balanced accuracy) & Yes, Not specified & 
			\raisebox{-\totalheight}{\includegraphics[width=\linewidth]{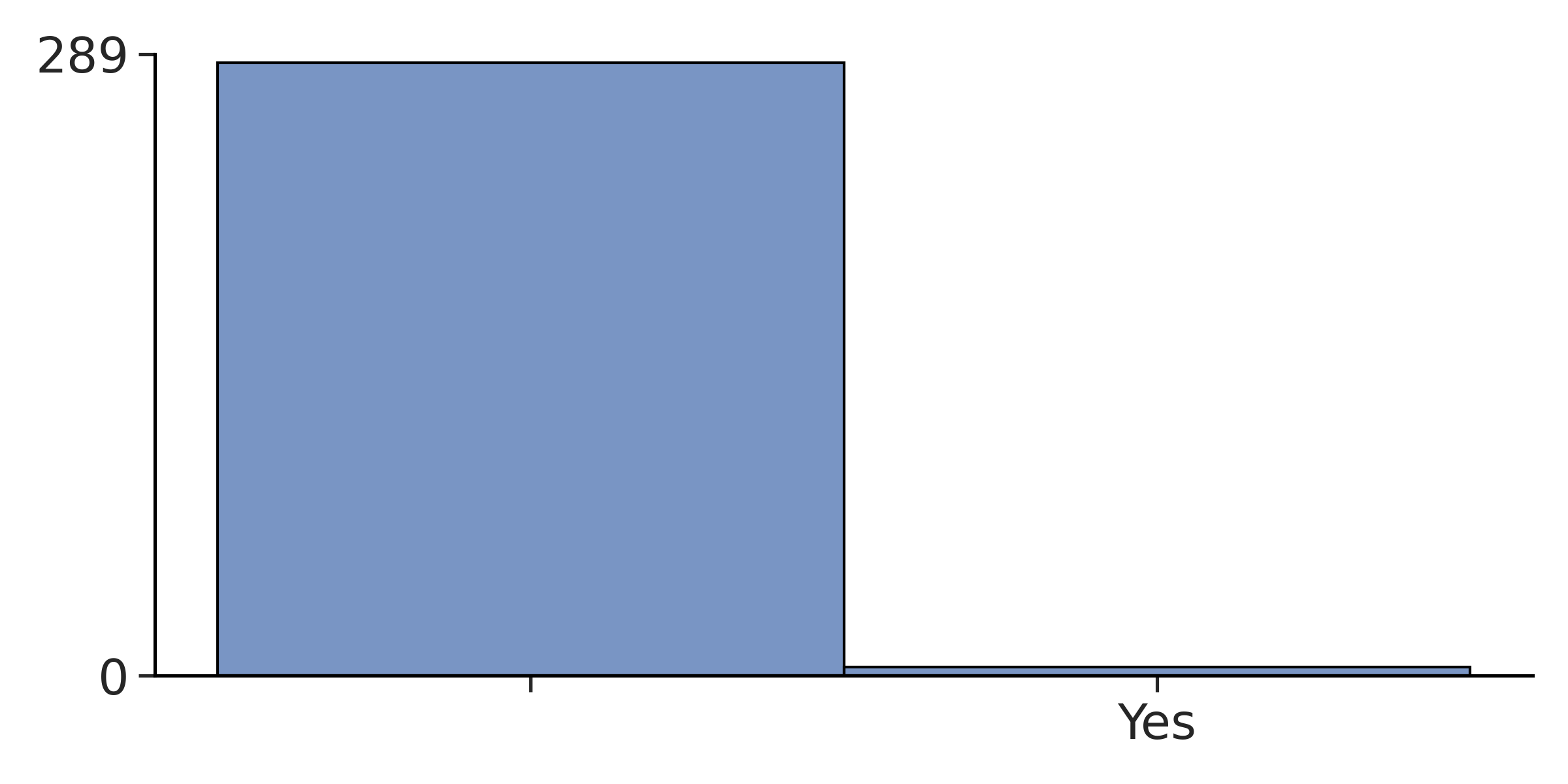}} \\
			
			Is it a frozen section? & Indicates whether a frozen section was performed & Binary (balanced accuracy) & Yes, Not specified & 
			\raisebox{-\totalheight}{\includegraphics[width=\linewidth]{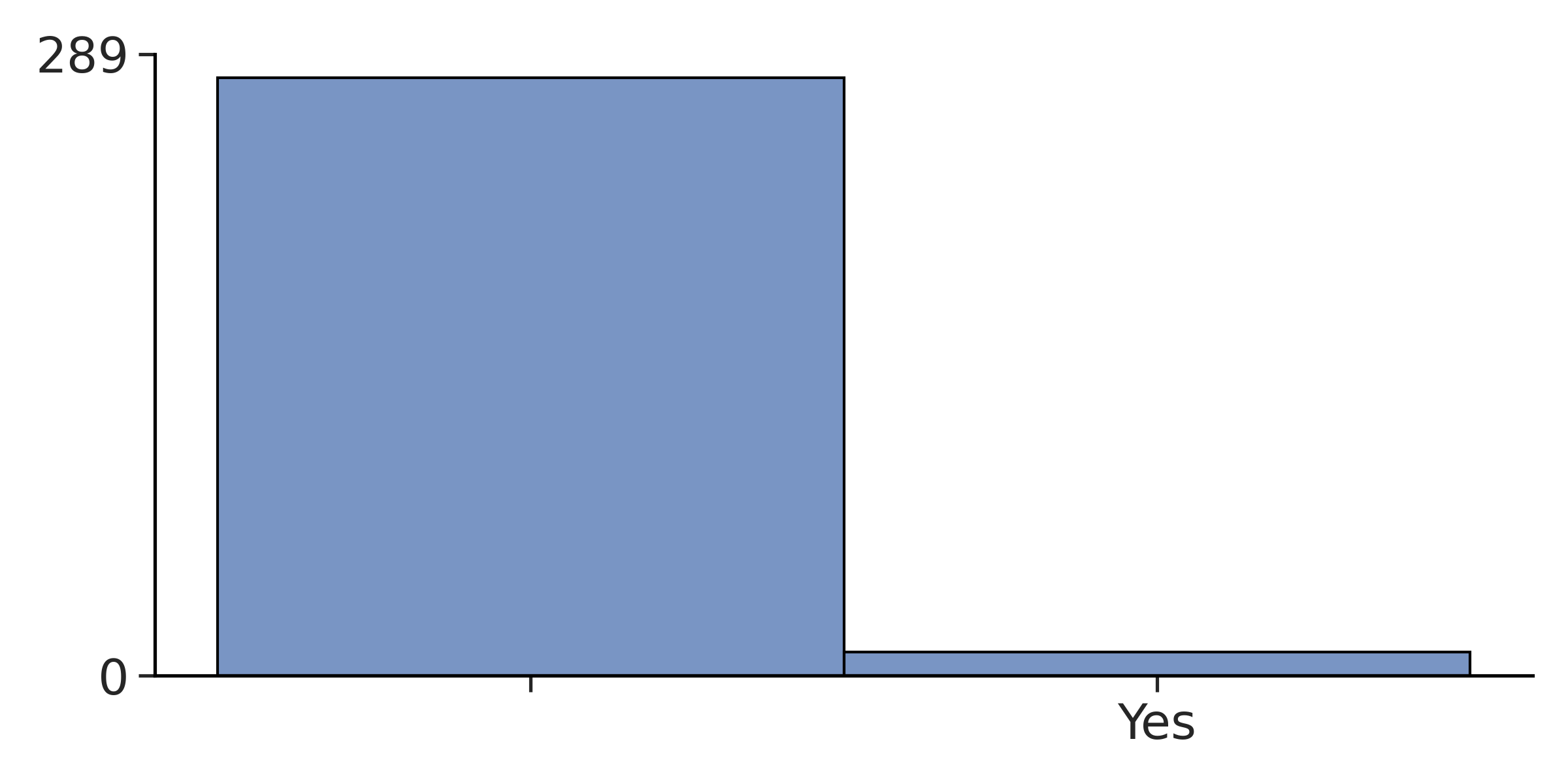}} \\
			
			Number of liver lesion types & Number of distinct liver lesion types described & Numeric (accuracy) & Any non-negative integer $\leq$ number of tissue samples & 
			\raisebox{-\totalheight}{\includegraphics[width=\linewidth]{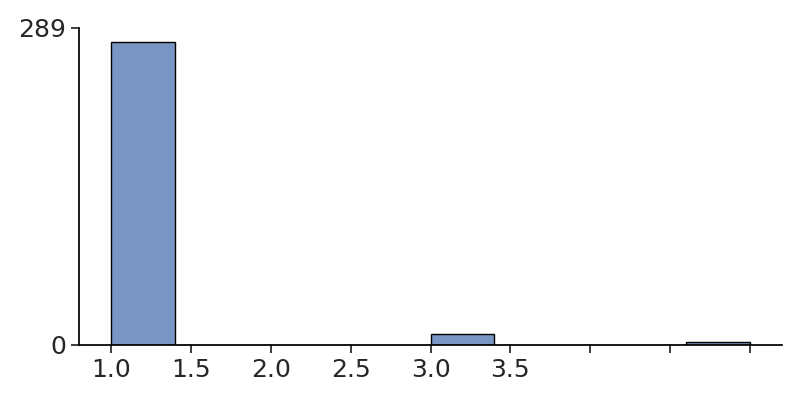}} \\
			
			Locations of tissue samples extracted from liver & Locations of liver tissue samples & List (symmetric similarity) & - & 
			\raisebox{-\totalheight}{\includegraphics[width=\linewidth]{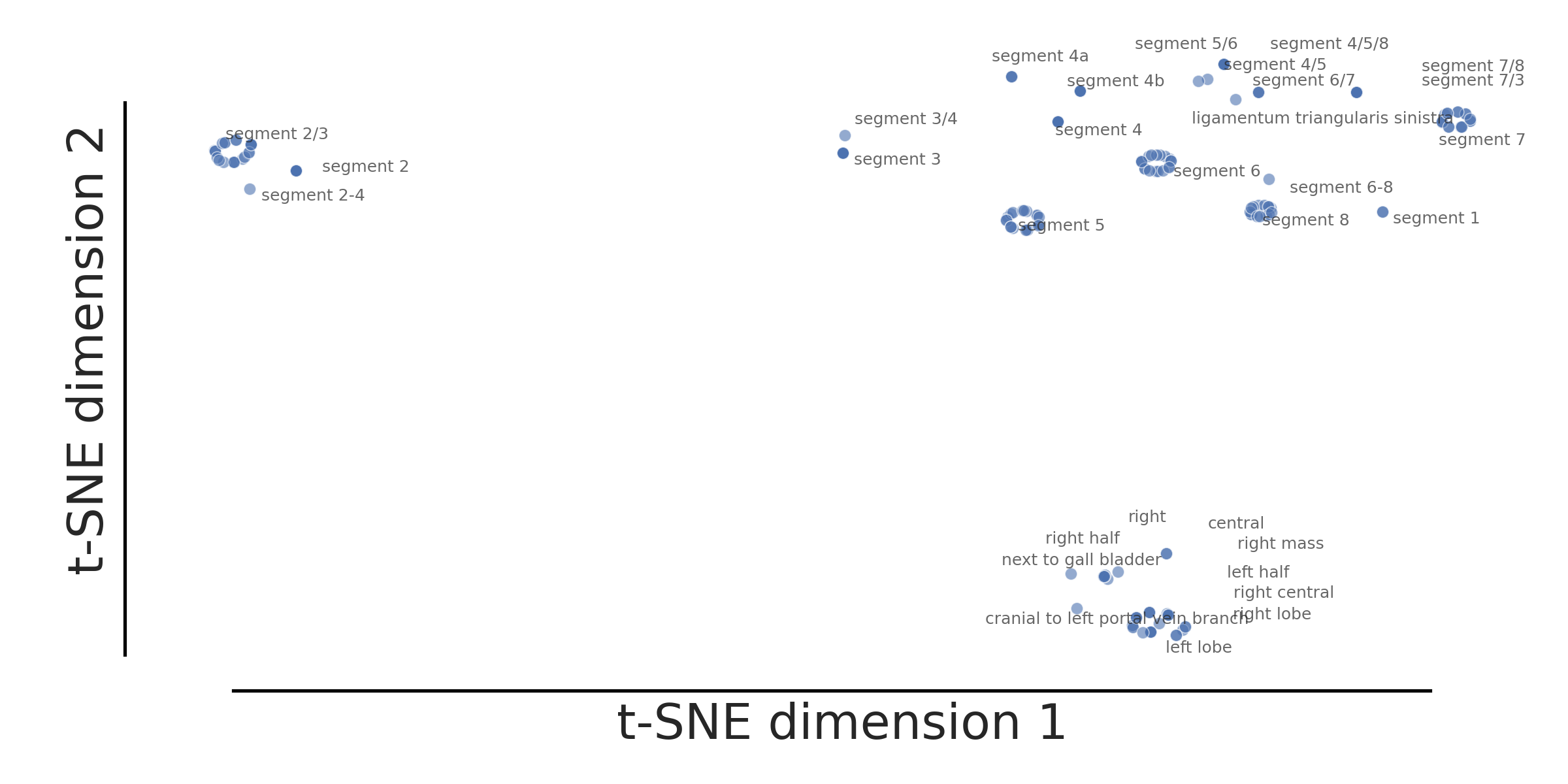}} \\
			
			Lesion phenotype & Tumour or lesion phenotype(s) described in the report & List (symmetric similarity) & - & 
			\raisebox{-\totalheight}{\includegraphics[width=\linewidth]{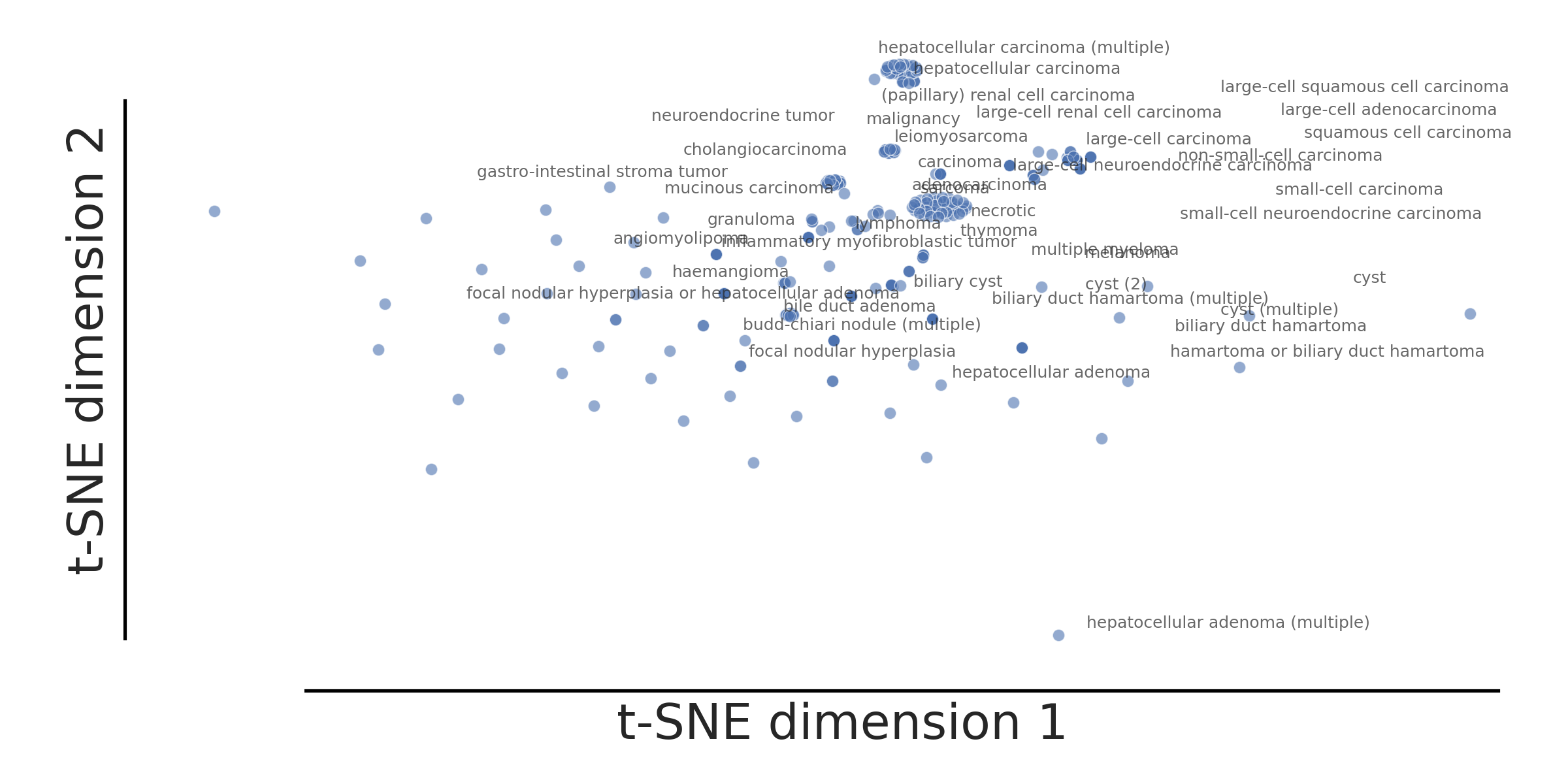}} \\
			
			Is there metastasis? & Indicates whether any lesion is a metastasis & Binary (balanced accuracy) & Yes, Not specified & 
			\raisebox{-\totalheight}{\includegraphics[width=\linewidth]{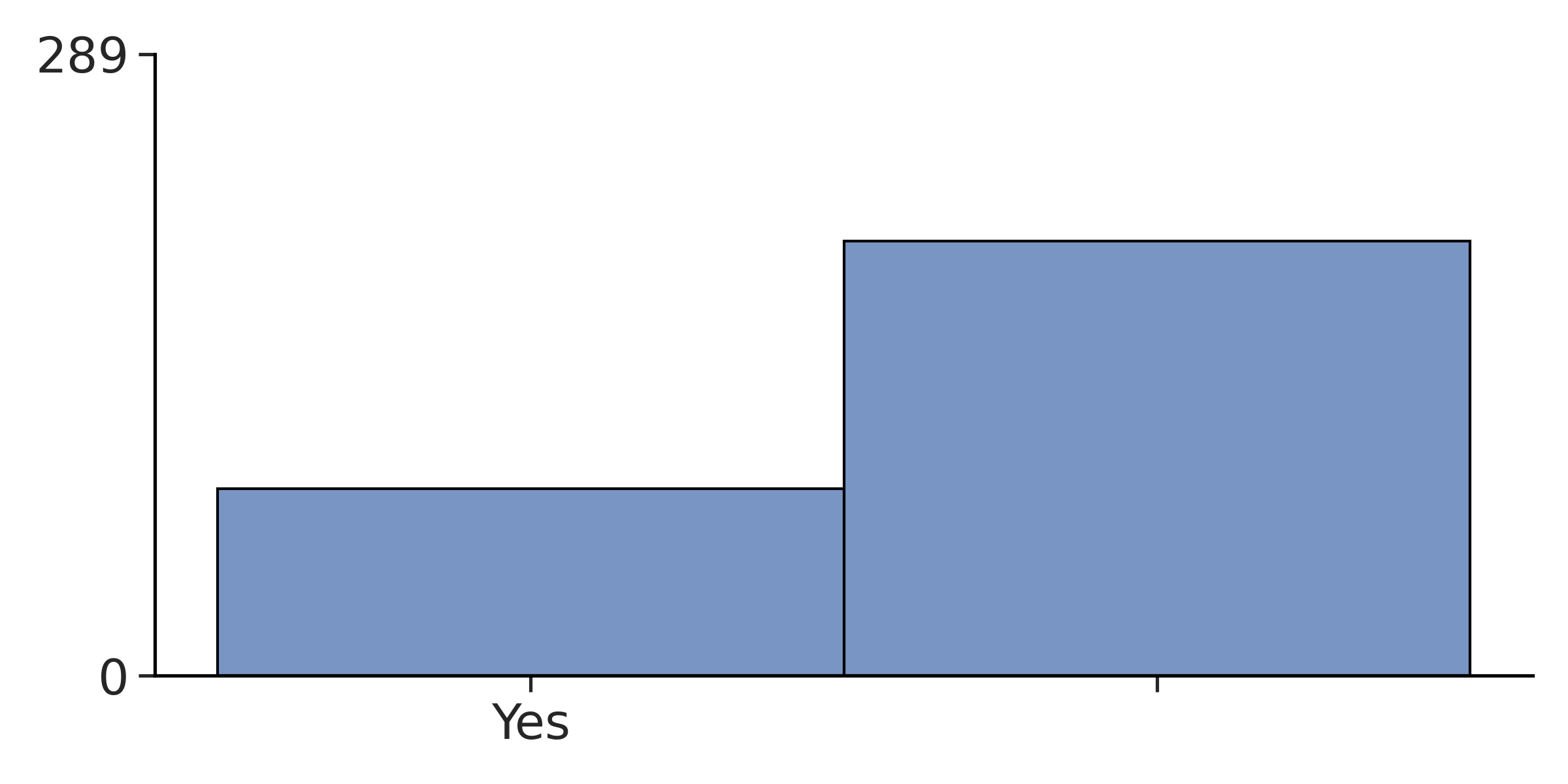}} \\
			
			Primary site of metastasis & Primary site of the metastasis, if present & Free-text (cosine similarity) & - & 
			\raisebox{-\totalheight}{\includegraphics[width=\linewidth]{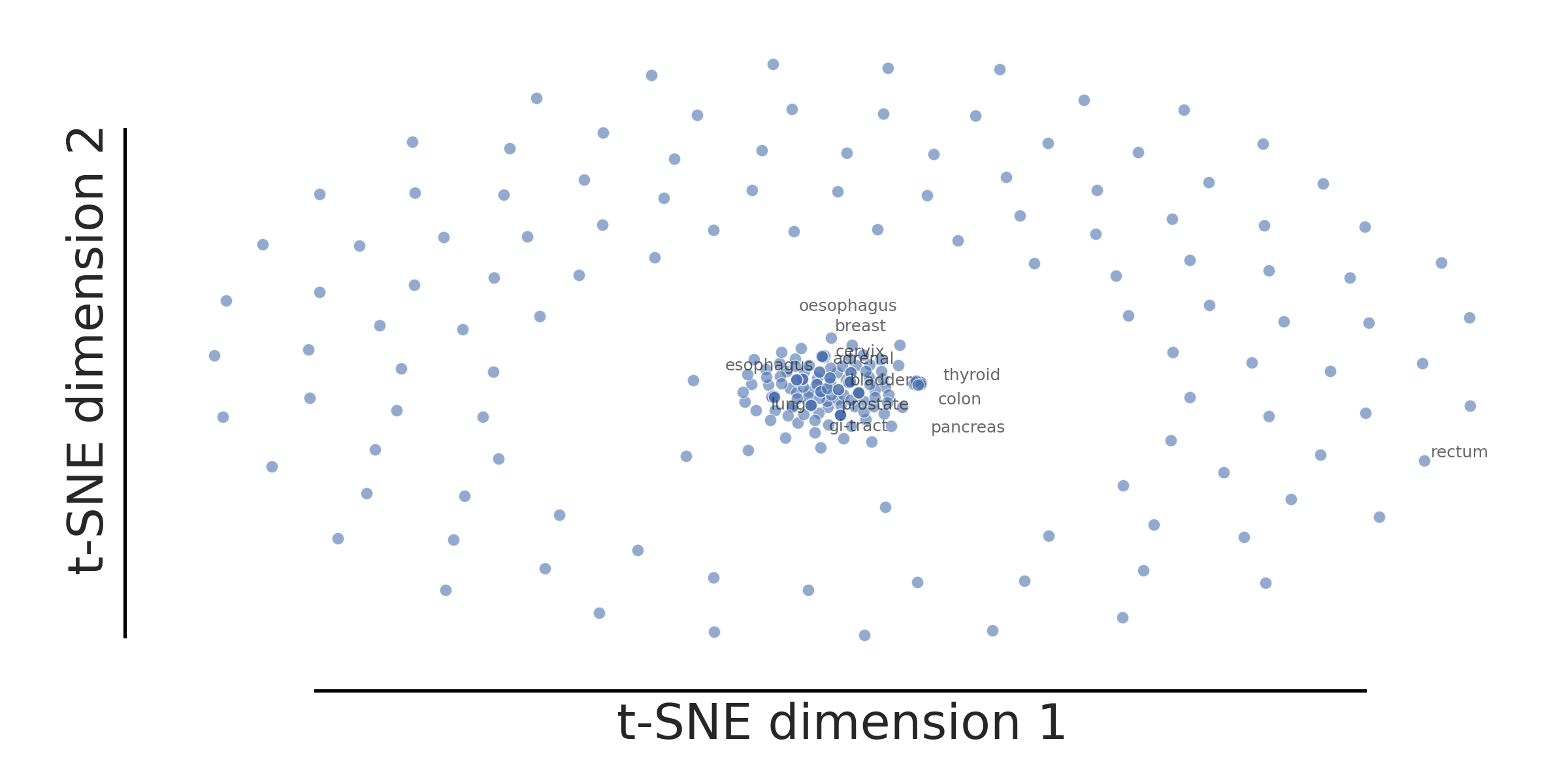}} \\
			
			Steatosis & Degree of steatosis if reported & Free-text (cosine similarity) & - & 
			\raisebox{-\totalheight}{\includegraphics[width=\linewidth]{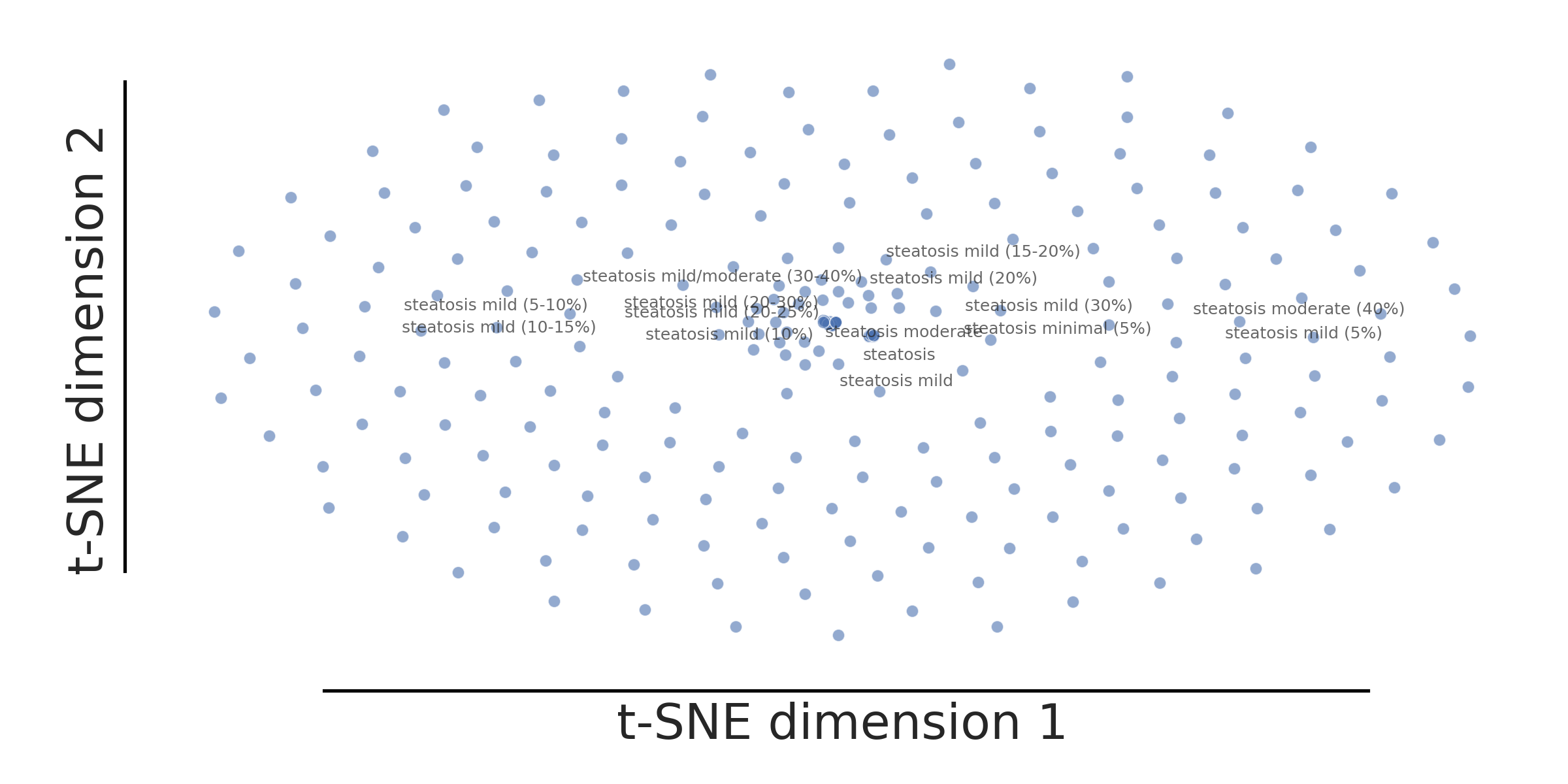}} \\
			
			Presence of fibrosis & Indicates whether fibrosis is present & Binary (balanced accuracy) & Yes, Not specified & 
			\raisebox{-\totalheight}{\includegraphics[width=\linewidth]{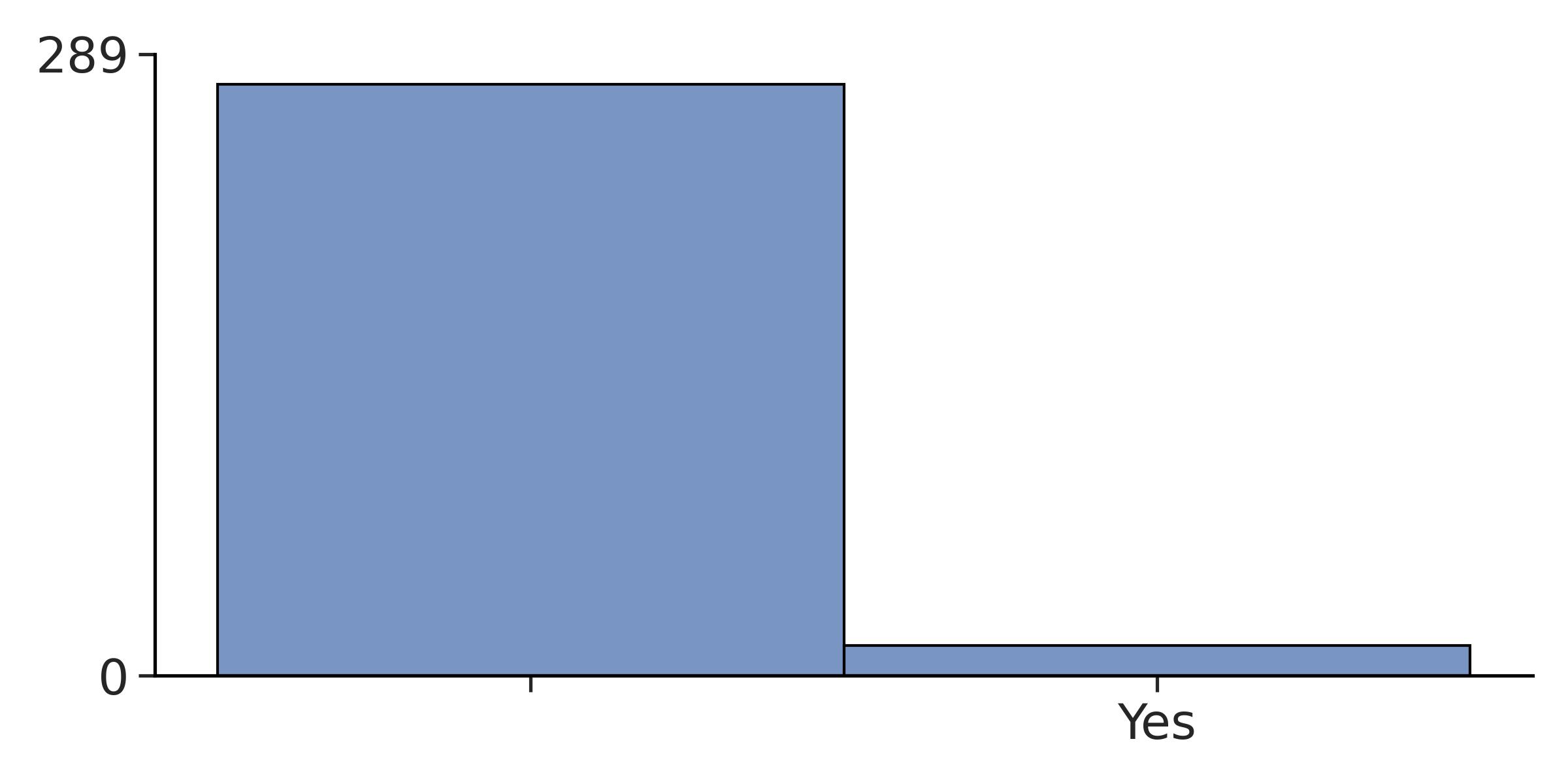}} \\
			
			Inflammation & Presence and description of inflammation & Free-text (cosine similarity) & - & 
			\raisebox{-\totalheight}{\includegraphics[width=\linewidth]{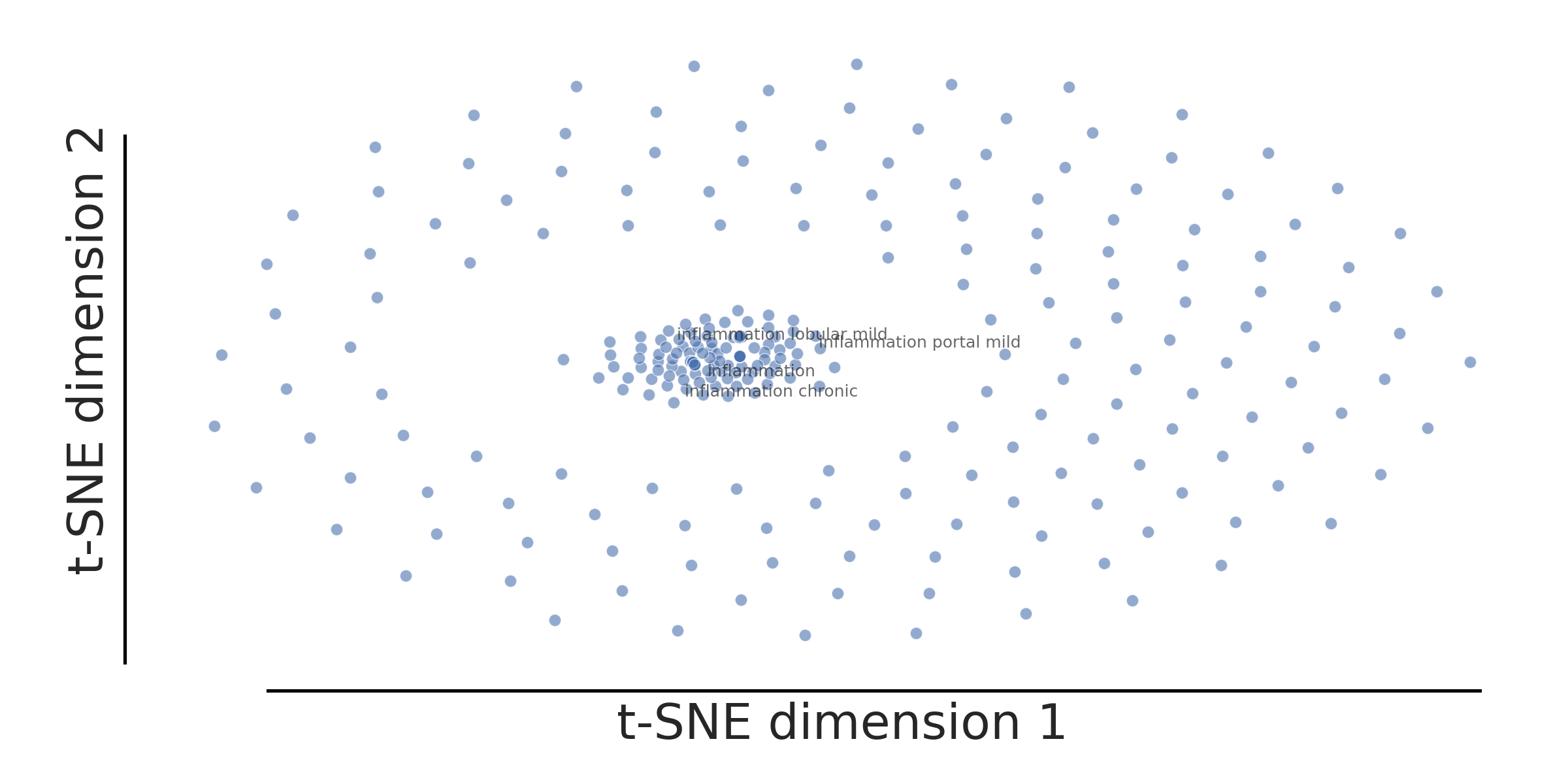}} \\
		\end{longtable}
	\end{scriptsize}
	
	\subsubsection{Prompt}
	For this use case, a structured prompt was created to ensure consistent extraction and enforce strict JSON formatting. 
	The prompt is divided into role-based instructions (System, Human, AI). A full example of the exact prompt used is shown in \autoref{tab:prompt-Liver}.
	
	\begin{scriptsize}
		\setlength{\tabcolsep}{4pt}
		\begin{longtable}{>{\scriptsize\raggedright\arraybackslash}m{0.2\linewidth} >{\scriptsize\raggedright\arraybackslash}m{0.37\linewidth} >{\scriptsize\raggedright\arraybackslash}m{0.37\linewidth}}
			\multicolumn{3}{c}{\parbox{\textwidth}{
					\normalsize \tablename~\thetable{} -- Full structured prompt used for the liver tumours extraction task. The items in brackets indicate the role of the message (System, Human, AI), while the text provides the corresponding content.\\}} \\
			\toprule
			\textbf{Section} & \multicolumn{2}{c}{\textbf{Content} (Depending on prompt strategy)} \\
			& Zero-Shot, One-Shot and Few-Shot & CoT, Self-Consistency and Graph\\
			\midrule
			\endfirsthead
			
			\multicolumn{3}{c}{\parbox{\textwidth}{
					\normalsize \tablename~\thetable{} -- Continued\\}} \\
			\toprule
			\textbf{Section} & \multicolumn{2}{c}{\textbf{Content}} \\
			\midrule
			\endhead
			
			\phantomlabel{tab:prompt-Liver}
			\textbf{[System] -- System instructions} & 
			You are a medical data extraction system that ONLY outputs valid JSON. Maintain strict compliance with these rules: \newline
			1. ALWAYS begin and end your response with \verb|```json| markers \newline
			2. Use EXACT field names and structure provided \newline
			3. If a value is missing or not mentioned, use the specified default for that field. \newline
			4. NEVER add commentary, explanations, or deviate from the output structure & 
			You are a medical data extraction system that performs structured reasoning before producing output. Follow these strict rules: \newline
			1. First, reason step-by-step to identify and justify each extracted field. \newline
			2. After reasoning, output ONLY valid JSON in the exact structure provided. \newline
			3. ALWAYS begin and end the final output with \verb|```json| markers — do not include reasoning within these markers. \newline
			4. Use EXACT field names and structure as specified. \newline
			5. If a value is missing or not mentioned, use the specified default for that field. \newline
			6. NEVER include commentary, explanations, or deviate from the specified format in the final JSON. \newline \\
			\midrule
			\textbf{[Human] -- Field instructions} & \multicolumn{2}{c}{\parbox{0.75\linewidth}{
					1. \opus{"Number of tissue samples"}: \newline
					- Type: number \newline
					- The total number of tissue samples described in the report. Usually at least 1. Leave as "" if undetermined. \newline
					- Default: "" \newline
					2. \opus{"Organs"}: \newline
					- Type: list \newline
					- List of organs the tissue samples come from, in the order mentioned. \newline
					- Default: empty list (\opus{[]}) \newline
					3. \opus{"Method of tissue retrieval"}: \newline
					- Type: string \newline
					- Single most specific retrieval method. Use "" if not mentioned. \newline
					- Options: ["", "biopsy", "excision", "resection", "brush"] \newline
					- Default: "" \newline
					4. \opus{"Is it an explant?"}: \newline
					- Type: string \newline
					- "Yes" if entire organ removed (explant), else "". \newline
					- Options: ["", "Yes"] \newline
					- Default: "" \newline
					5. \opus{"Is it a frozen section?"}: \newline
					- Type: string \newline
					- "Yes" if frozen section performed/mentioned, else "". \newline
					- Options: ["", "Yes"] \newline
					- Default: "" \newline
					6. \opus{"Number of liver lesion types"}: \newline
					- Type: number \newline
					- Number of distinct liver lesion types described. Must $\leq$ Number of tissue samples. \newline
					- Default: "" \newline
					7. \opus{"Locations of tissue samples extracted from liver"}: \newline
					- Type: list \newline
					- List of liver locations in order mentioned, preserve duplicates. Include standard and broader descriptors (e.g., "segment 6/7", "left half"). Leave empty if no location or Number of liver lesion types = 0. \newline
					- Default: empty list (\opus{[]}) \newline
					8. \opus{"Lesion phenotype"}: \newline
					- Type: list \newline
					- List of lesion phenotypes in order mentioned. Preserve uncertainty ("or") and qualifiers ("(multiple)"). Leave empty if none described. \newline
					- Default: empty list (\opus{[]}) \newline
					9. \opus{"Is there metastasis?"}: \newline
					- Type: string \newline
					- "Yes" if any liver lesion described as metastasis, else "". \newline
					- Options: ["Yes", ""] \newline
					- Default: "" \newline
					10. \opus{"Primary site of metastasis"}: \newline
					- Type: string \newline
					- Specify primary site if metastasis present, else leave empty. \newline
					- Default: "" \newline
					11. \opus{"Steatosis"}: \newline
					- Type: string \newline
					- Degree of steatosis if reported, beginning with "steatosis" and followed by descriptor (e.g., "steatosis mild (10\%)"). Leave empty if not mentioned. \newline
					- Default: "" \newline
					12. \opus{"Presence of fibrosis"}: \newline
					- Type: string \newline
					- "Yes" if fibrosis reported, else "". \newline
					- Options: ["", "Yes"] \newline
					- Default: "" \newline
					13. \opus{"Inflammation"}: \newline
					- Type: string \newline
					- Presence and description of inflammation if reported, beginning with "inflammation" and followed by ordered descriptors (e.g., "inflammation chronic portal moderate"). Leave empty if not mentioned. \newline
					- Default: "" 
			}} \newline \\
			\midrule	
			\textbf{[Human] -- Task instructions} & \multicolumn{2}{c}{\parbox{0.75\linewidth}{
					Extract information into this exact JSON structure:
					\opus{```json} \newline
					\opus{\{} \newline
					\opus{\quad "Number of tissue samples": "",} \newline
					\opus{\quad "Organs": [],} \newline
					\opus{\quad "Method of tissue retrieval": "",} \newline
					\opus{\quad "Is it an explant?": "",} \newline
					\opus{\quad "Is it a frozen section?": "",} \newline
					\opus{\quad "Number of liver lesion types": "",} \newline
					\opus{\quad "Locations of tissue samples extracted from liver": [],} \newline
					\opus{\quad "Lesion phenotype": [],} \newline
					\opus{\quad "Is there metastasis?": "",} \newline
					\opus{\quad "Primary site of metastasis": "",} \newline
					\opus{\quad "Steatosis": "",} \newline
					\opus{\quad "Presence of fibrosis": "",} \newline
					\opus{\quad "Inflammation": "",} \newline
					\opus{\}} \newline
					\opus{```}%
				}%
			} \newline \\
			\midrule
			\textbf{[Human] -- Example intro} & \multicolumn{2}{c}{\parbox{0.75\linewidth}{Below are 1 example of expected input and output, followed by a new task.}} \newline \\
			\midrule
			\textbf{[Human] -- Example user} & \multicolumn{2}{c}{\parbox{0.75\linewidth}{\opus{[...CENSORED...]}}} \newline \\
			\midrule
			\textbf{[AI] -- Example assistant reasoning} & - & 
			- \opus{Number of tissue samples} - "single liver biopsy" implies 1 \newline
			- \opus{Organs} - "organ explicitly the liver" implies ["liver"] \newline
			- \opus{Method of tissue retrieval} - "naaldbiopten (needle biopsy)" implies "biopsy" \newline
			- \opus{Is it an explant?} - "not mentioned" implies "" \newline
			- \opus{Is it a frozen section?} - "not mentioned" implies "" \newline
			- \opus{Number of liver lesion types} - "one distinct lesion phenotype reported: hepatocellular carcinoma" implies 1 \newline
			- \opus{Locations of tissue samples extracted from liver} - "location not specified" implies [] \newline
			- \opus{Lesion phenotype} - "hepatocellular carcinoma" implies ["hepatocellular carcinoma"] \newline
			- \opus{Is there metastasis?} - "lesion described as metastasis" implies "Yes" \newline
			- \opus{Primary site of metastasis} - "not reported" implies "" \newline
			- \opus{Steatosis} - "minimal steatosis (‘minder dan 5\%’)" implies "steatosis minimal (5\%)" \newline
			- \opus{Presence of fibrosis} - "not mentioned" implies "" \newline
			- \opus{Inflammation} - "matige chronische portale ontsteking" implies "inflammation chronic portal moderate" \newline \\
			\midrule
			\textbf{[AI] -- Example assistant output} & \multicolumn{2}{c}{\parbox{0.75\linewidth}{
					\opus{```json} \newline
					\opus{\{} \newline
					\opus{\quad "Number of tissue samples": 1,} \newline
					\opus{\quad "Organs": ["liver"],} \newline
					\opus{\quad "Method of tissue retrieval": "biopsy",} \newline
					\opus{\quad "Is it an explant?": "",} \newline
					\opus{\quad "Is it a frozen section?": "",} \newline
					\opus{\quad "Number of liver lesion types": 1,} \newline
					\opus{\quad "Locations of tissue samples extracted from liver": [],} \newline
					\opus{\quad "Lesion phenotype": ["hepatocellular carcinoma"],} \newline
					\opus{\quad "Is there metastasis?": "Yes",} \newline
					\opus{\quad "Primary site of metastasis": "",} \newline
					\opus{\quad "Steatosis": "steatosis minimal (5\%)",} \newline
					\opus{\quad "Presence of fibrosis": "",} \newline
					\opus{\quad "Inflammation": "inflammation chronic portal moderate",} \newline
					\opus{\}} \newline
					\opus{```}%
				}%
			} \newline \\
			\midrule
			\textbf{[Human] -- Report instructions} & \multicolumn{2}{c}{\parbox{0.75\linewidth}{[file name]: \opus{[...CENSORED...]} \newline \opus{[...CENSORED...]}}} \newline \\
			\midrule
			\textbf{[Human] -- Final instructions} &
			Begin the extraction now. Your response must contain only a single valid JSON block, enclosed in triple backticks and prefixed with \verb|`json`|, like this: \verb|```json  ... ```|& 
			Begin the extraction now. First, reason step-by-step to identify and justify the value for each required field, enclosed within \verb|<think>...</think>| tags. Then, output only the final structured data as a single valid JSON block, starting with \verb|```json| and ending with \verb|```|."
			\\
			\bottomrule
		\end{longtable}
	\end{scriptsize}
	
	\subsubsection{Prompt Graph}
	The dependencies, conditional branches, and extraction order of variables are represented as a directed acyclic graph. This graph reflects how the extraction task is decomposed into smaller, sequential subtasks for the Prompt Graph prompting strategy. 
	\begin{figure}[htbp]
		\centering
		\includegraphics[width=\linewidth, height=0.5\textheight, keepaspectratio]{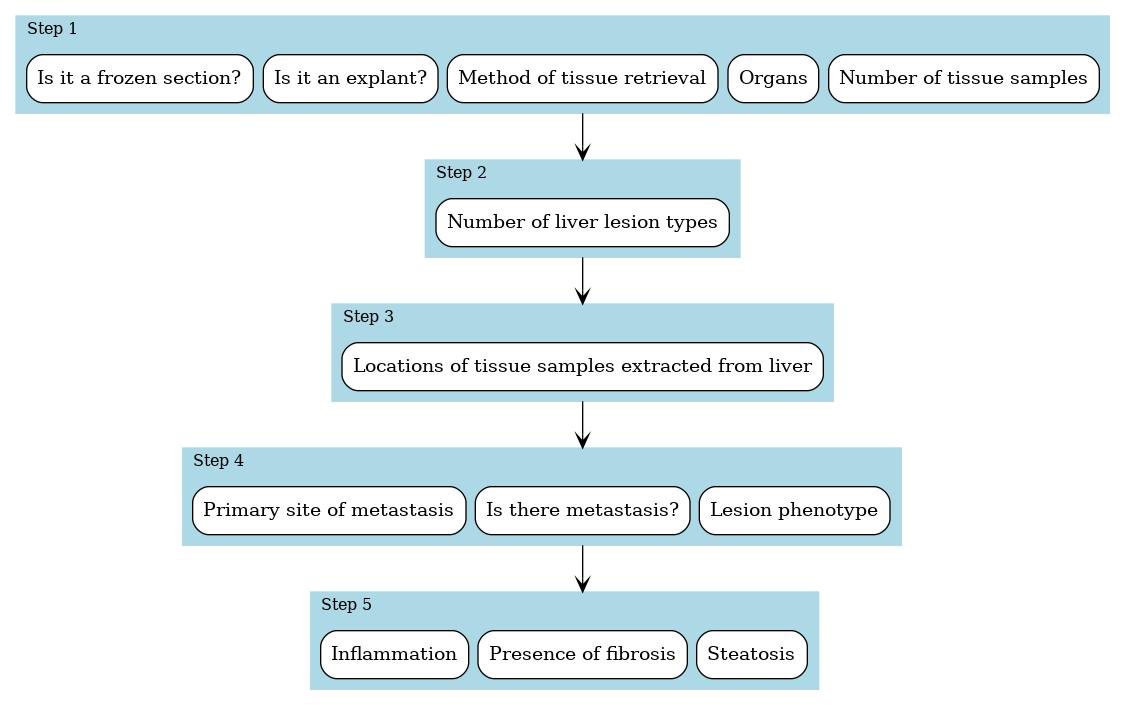}
		\caption{Directed acyclic graph showing sequential extraction order of variable extraction for liver tumours use case.}
	\end{figure}
	
	\newpage
	\subsection{Use Case - Neurodegenerative Diseases}
	
	\subsubsection{Overview}
	
	This use case focuses on extracting structured information from Dutch-language radiology reports of brain MRI examinations (\autoref{tab:use-case-Neurodegenerative-Diseases}). Extracted variables include measures of cerebral atrophy, such as the Generalized Cortical Atrophy (GCA) and Medial Temporal Atrophy (MTA) scores, as well as indicators of cerebrovascular pathology, including Fazekas scores and the presence and count of microbleeds and infarcts.
	
	\subsubsection{Inclusion and Exclusion Criteria}
	
	The dataset retrospectively comprised 947 brain MRI reports from patients evaluated at a tertiary memory clinic between 2016 and 2021. All patients who underwent MRI within three months of their initial consultation were included. Reports were generated as part of routine clinical practice by multiple radiologists at the same clinic. Of these, 152 reports followed a structured template with predefined fields, while the remaining 795 were unstructured, allowing radiologists to freely determine the report format.
	
	\subsubsection{Annotation}
	
	Each report was manually annotated for 20 variables by one of three trained final-year medical students, using a structured template with standardized guidelines specifying variable definitions, permissible values, and instructions for handling missing data. To evaluate inter-rater reliability, a random subset of 100 reports was independently annotated by two raters, with discrepancies resolved through consultation with a neuroradiologist with 15 years of experience.
	
	\subsubsection{Ethical Considerations and Funding}
	
	This study was approved by the ethical review board of Erasmus Medical Center (MEC-2016-069 and MEC-2023-0569). All patients either provided written informed consent or fell under the consent exception granted by the ethical review board, which approved the retrospective use of pseudonymized data for patients who could not be contacted about research participation. The study is part of TAP-dementia (www.tap-dementia.nl), receiving funding from ZonMw (\#10510032120003) in the context of Onderzoeksprogramma Dementie, part of the Dutch National Dementia Strategy.
	
	\begin{scriptsize}
		\setlength{\tabcolsep}{4pt}
		\begin{longtable}{@{}
				>{\scriptsize\raggedright\arraybackslash}p{0.10\textwidth} % Variable name
				>{\scriptsize\raggedright\arraybackslash}p{0.15\textwidth} % Description
				>{\scriptsize\raggedright\arraybackslash}p{0.10\textwidth} % Variable type
				>{\scriptsize\raggedright\arraybackslash}p{0.11\textwidth} % Variable options
				>{\scriptsize\raggedleft\arraybackslash}p{0.11\textwidth} % Annotator Agreement
				>{\scriptsize\centering\arraybackslash}p{0.30\textwidth} @{}} % Distribution with image
			\multicolumn{6}{c}{\parbox{\textwidth}{
					\normalsize \tablename~\thetable{} -- Neurodegenerative diseases radiology report variable definitions with reference standard distribution.\\}} \\
			\toprule
			\textbf{Variable Name} & 
			\textbf{Description} & 
			\textbf{Type (Metric)} & 
			\textbf{Variable Options} & 
			\textbf{Inter-rater Agreement} & 
			\textbf{Reference Standard Distribution} \\
			\midrule
			\endfirsthead
			
			\multicolumn{6}{c}{\parbox{\textwidth}{
					\normalsize \tablename~\thetable{} -- Continued\\}} \\
			\toprule
			\textbf{Variable Name} & 
			\textbf{Description} & 
			\textbf{Type (Metric)} & 
			\textbf{Variable Options} & 
			\textbf{Inter-rater Agreement} & 
			\textbf{Reference Standard Distribution} \\
			\midrule
			\endhead
			
			\midrule
			\endfoot
			
			\bottomrule
			\endlastfoot
			
			\phantomlabel{tab:use-case-Neurodegenerative-Diseases}
			Fazekas score & Rates severity of white matter hyperintensities (0–3 scale) & Categorical (balanced accuracy) & 0, 0.5, 1, 1.5, 2, 2.5, 3, missing & 0.99 &
			\raisebox{-\totalheight}{\includegraphics[width=\linewidth]{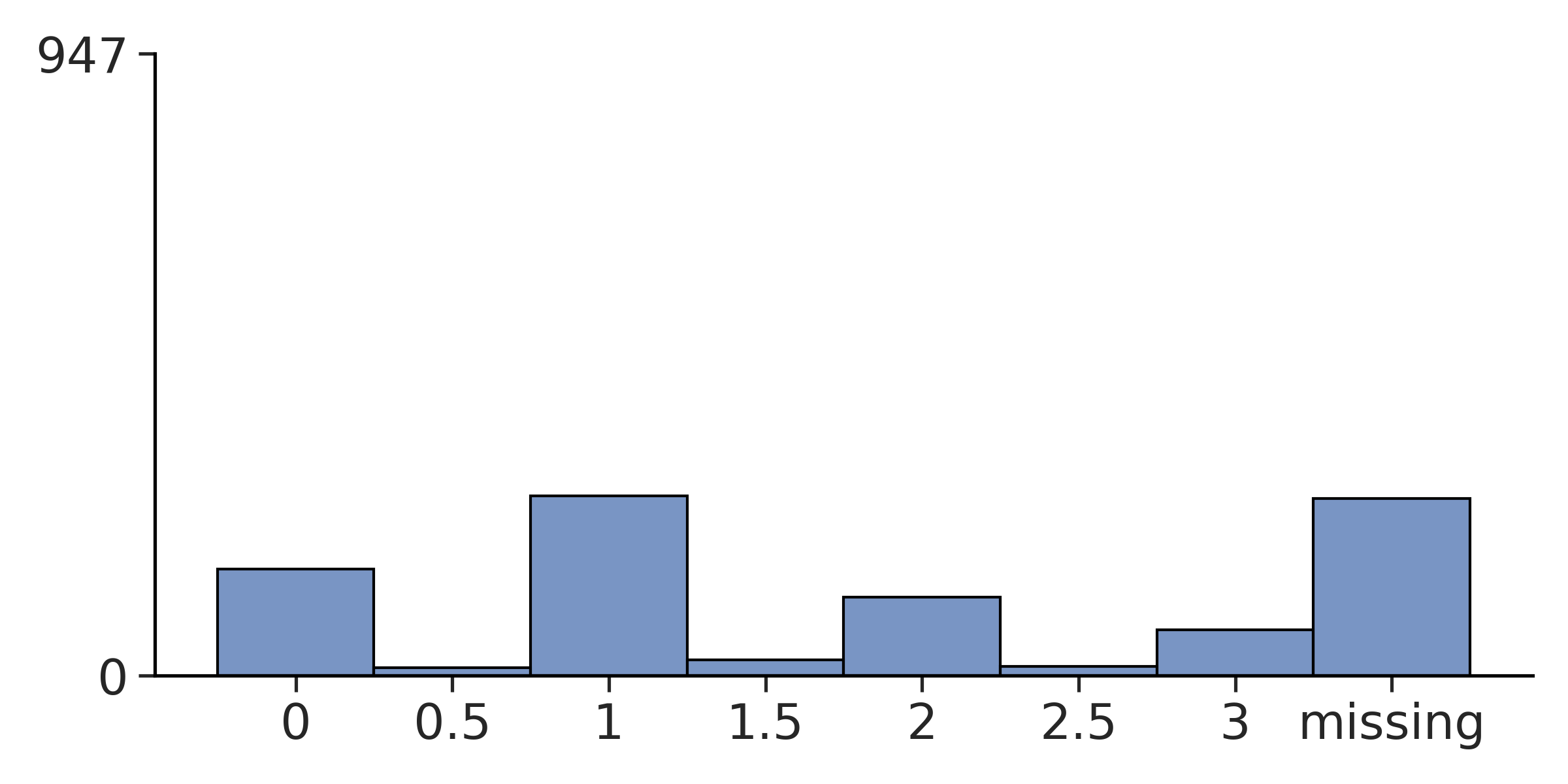}} \\
			MTA left & Medial temporal atrophy score on the left side (0–4 scale) & Categorical (balanced accuracy) & 0, 0.5, 1, 1.5, 2, 2.5, 3, 3.5, 4, missing & 0.99 &
			\raisebox{-\totalheight}{\includegraphics[width=\linewidth]{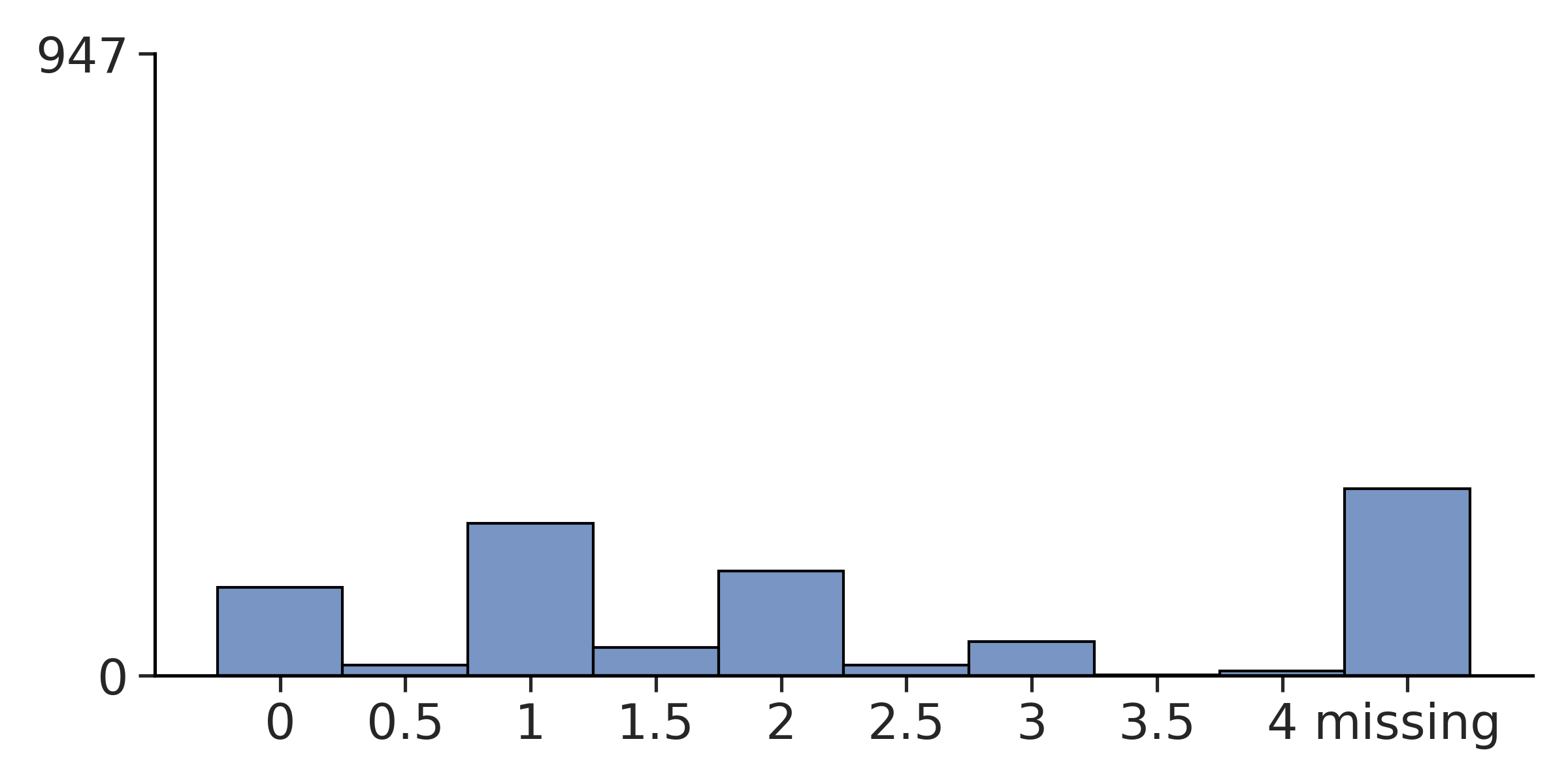}} \\
			MTA right & Medial temporal atrophy score on the right side (0–4 scale) & Categorical (balanced accuracy) & 0, 0.5, 1, 1.5, 2, 2.5, 3, 3.5, 4, missing & 0.99 &
			\raisebox{-\totalheight}{\includegraphics[width=\linewidth]{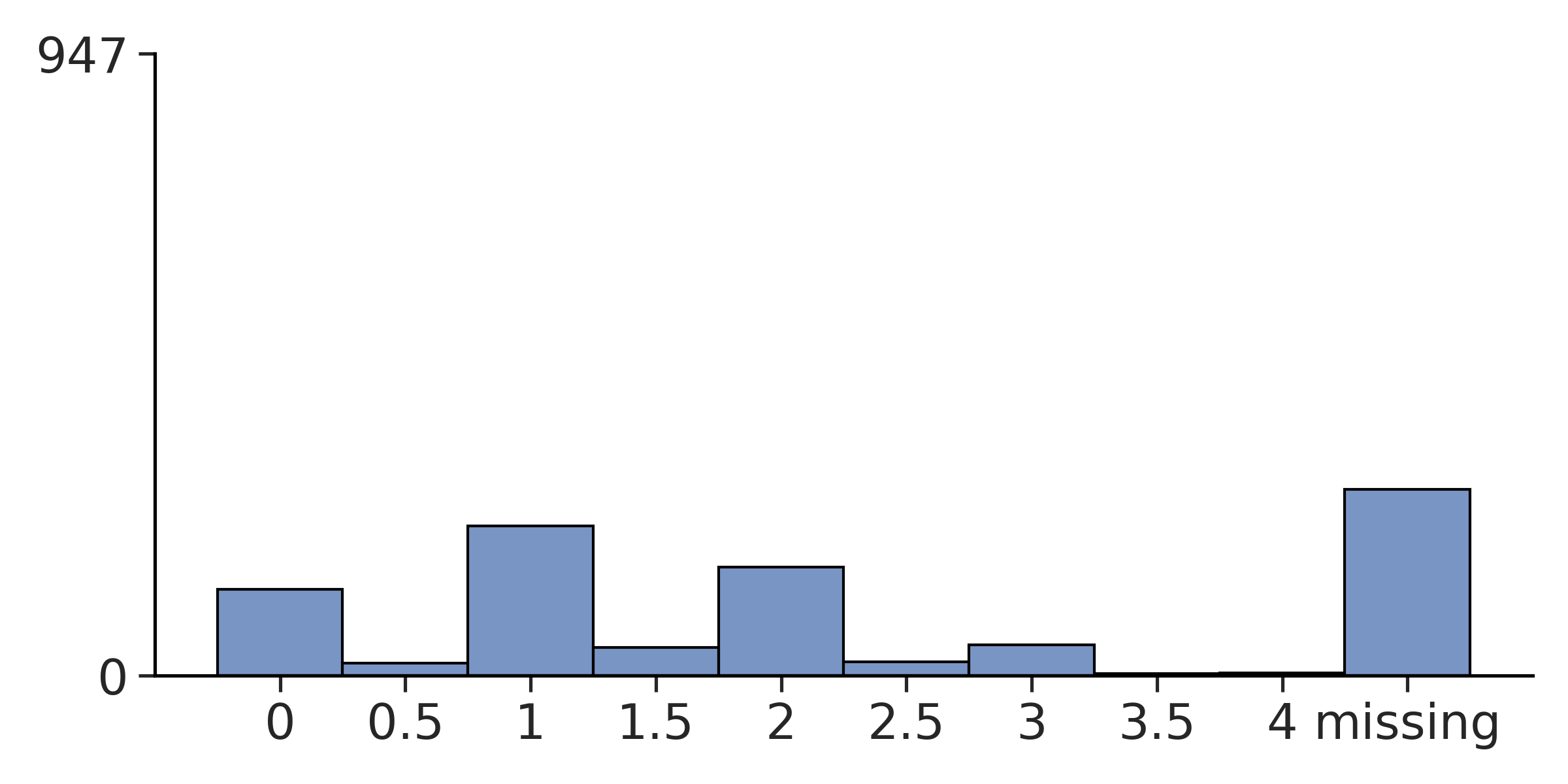}} \\
			GCA frontal & Global cortical atrophy score for the frontal lobe (0–3 scale) & Categorical (balanced accuracy) & 0, 0.5, 1, 1.5, 2, 2.5, 3, missing & 0.92 &
			\raisebox{-\totalheight}{\includegraphics[width=\linewidth]{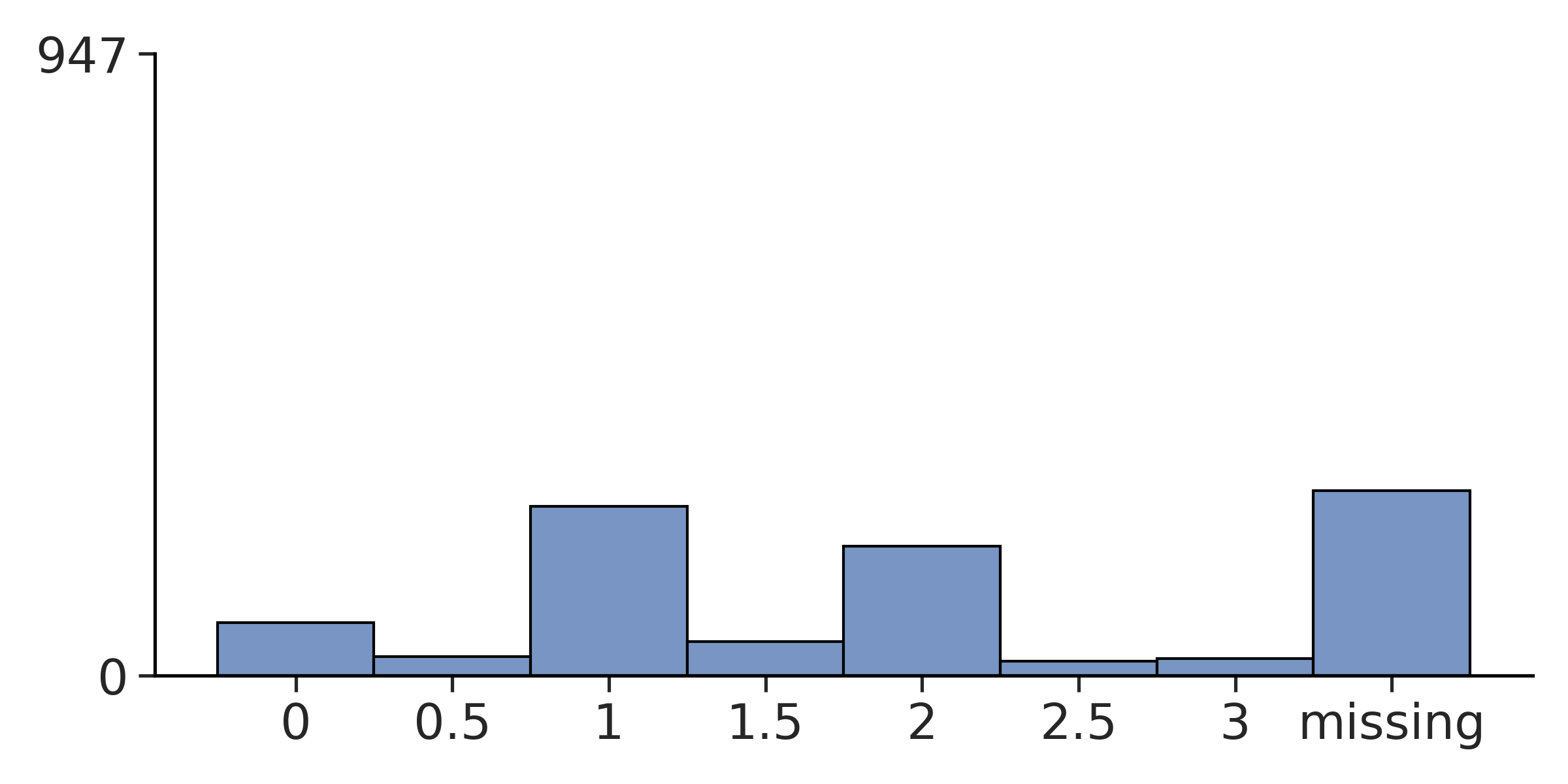}} \\
			GCA temporal & Global cortical atrophy score for the temporal lobe (0–3 scale) & Categorical (balanced accuracy) & 0, 0.5, 1, 1.5, 2, 2.5, 3, missing & 0.94 &
			\raisebox{-\totalheight}{\includegraphics[width=\linewidth]{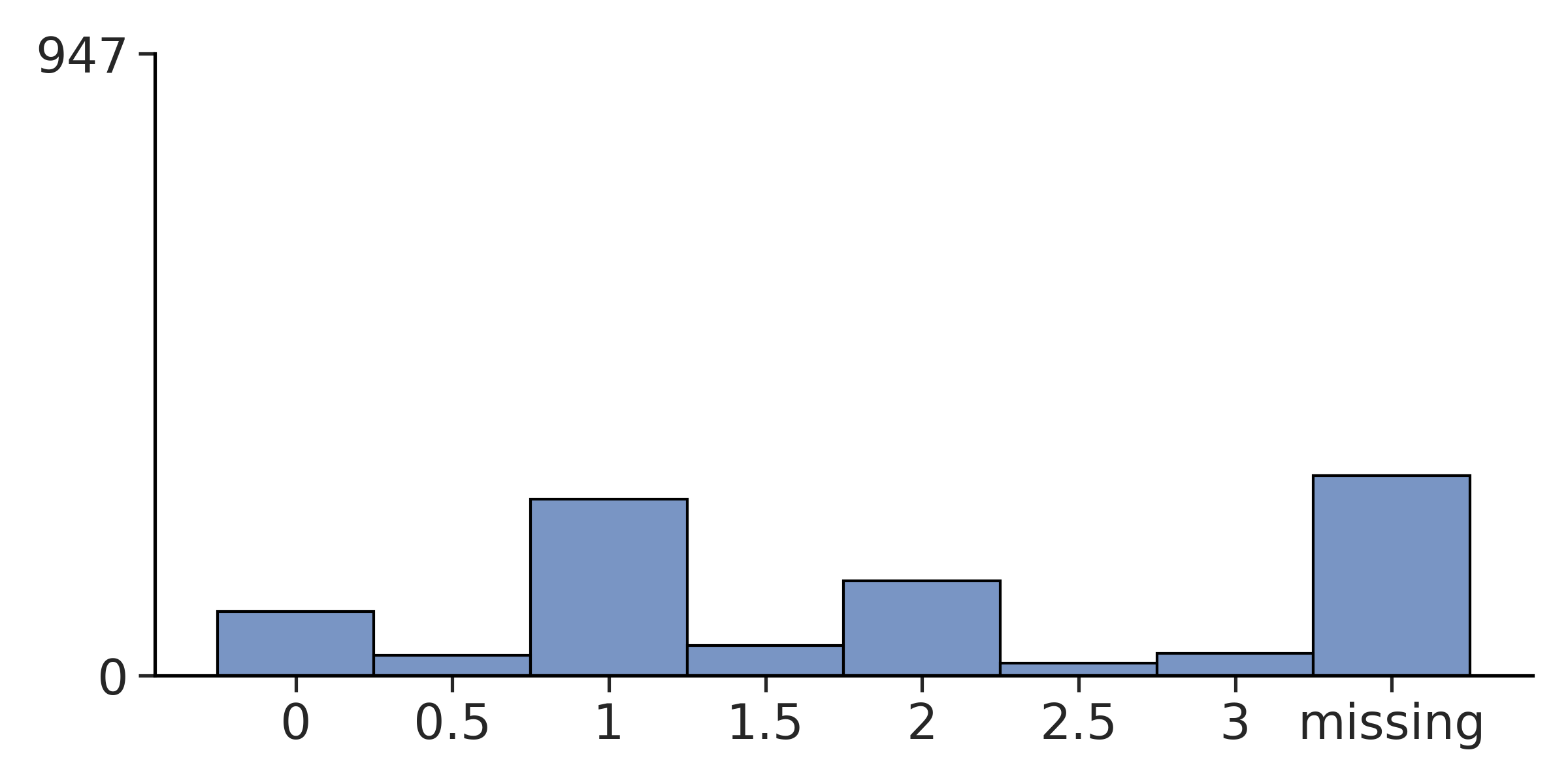}} \\
			GCA parietal & Global cortical atrophy score for the occipital lobe (0–3 scale) & Categorical (balanced accuracy) & 0, 0.5, 1, 1.5, 2, 2.5, 3, missing & 0.93 &
			\raisebox{-\totalheight}{\includegraphics[width=\linewidth]{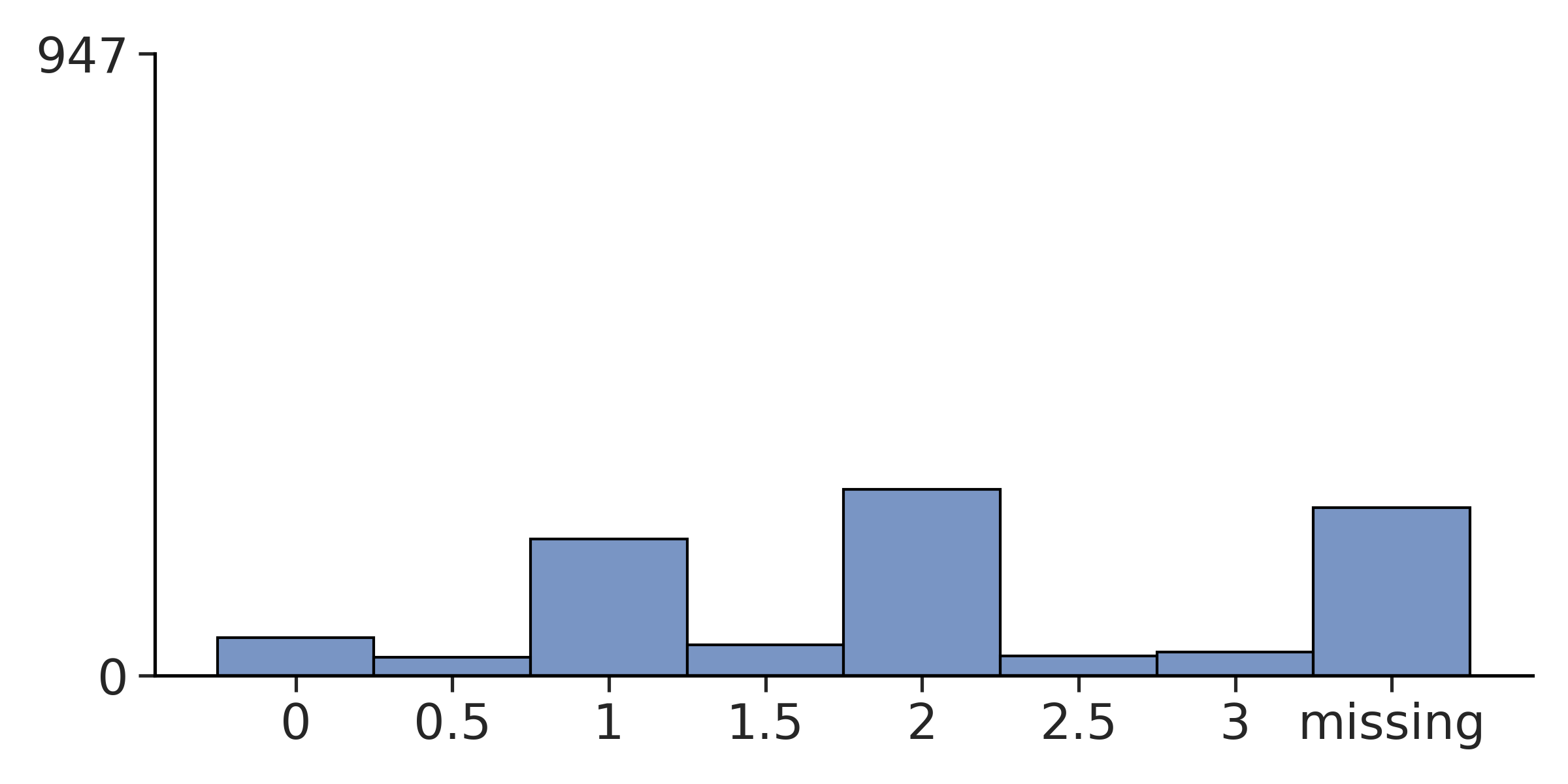}} \\
			GCA occipital & Global cortical atrophy score for the parietal lobe (0–3 scale) & Categorical (balanced accuracy) & 0, 0.5, 1, 1.5, 2, 2.5, 3, missing & 0.94 &
			\raisebox{-\totalheight}{\includegraphics[width=\linewidth]{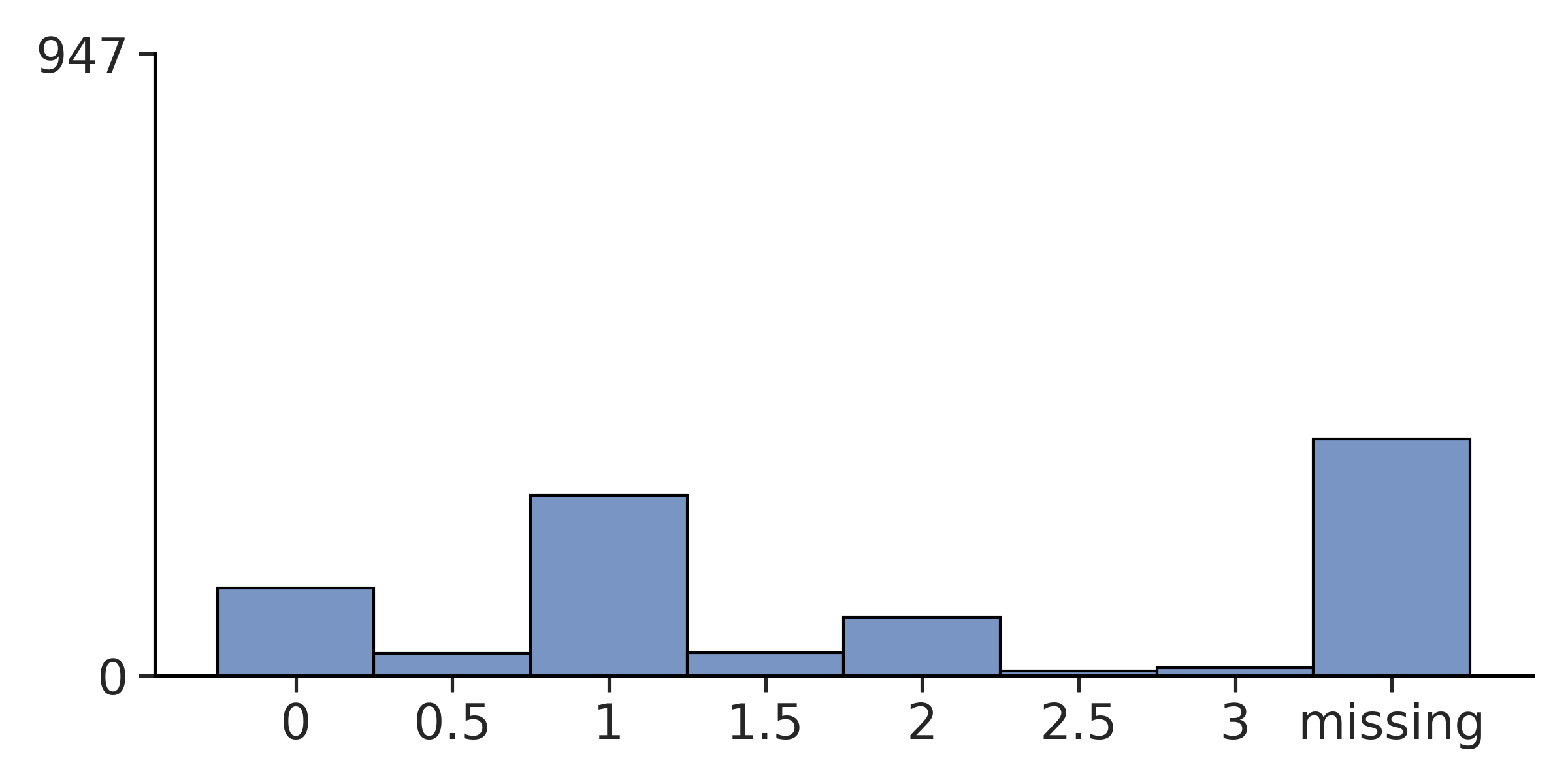}} \\
			GCA overall & Overall cortical atrophy severity (0–3, or highest lobar score) & Categorical (balanced accuracy) & 0, 0.5, 1, 1.5, 2, 2.5, 3, missing & 0.95 &
			\raisebox{-\totalheight}{\includegraphics[width=\linewidth]{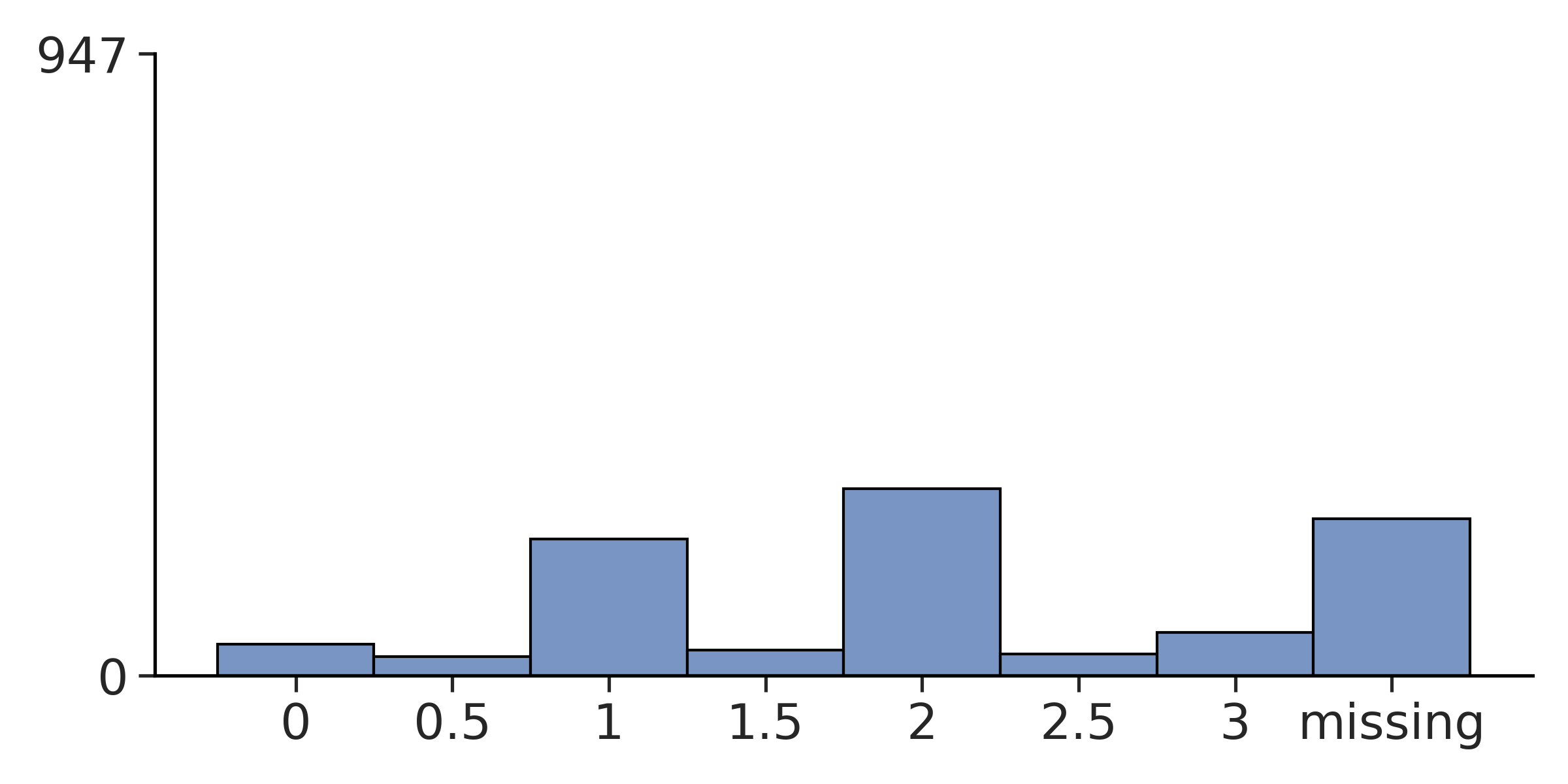}} \\
			Any brain infarct & Indicates presence of any type of brain infarct & Binary (balanced accuracy) & 0, 1, missing & 0.88 &
			\raisebox{-\totalheight}{\includegraphics[width=\linewidth]{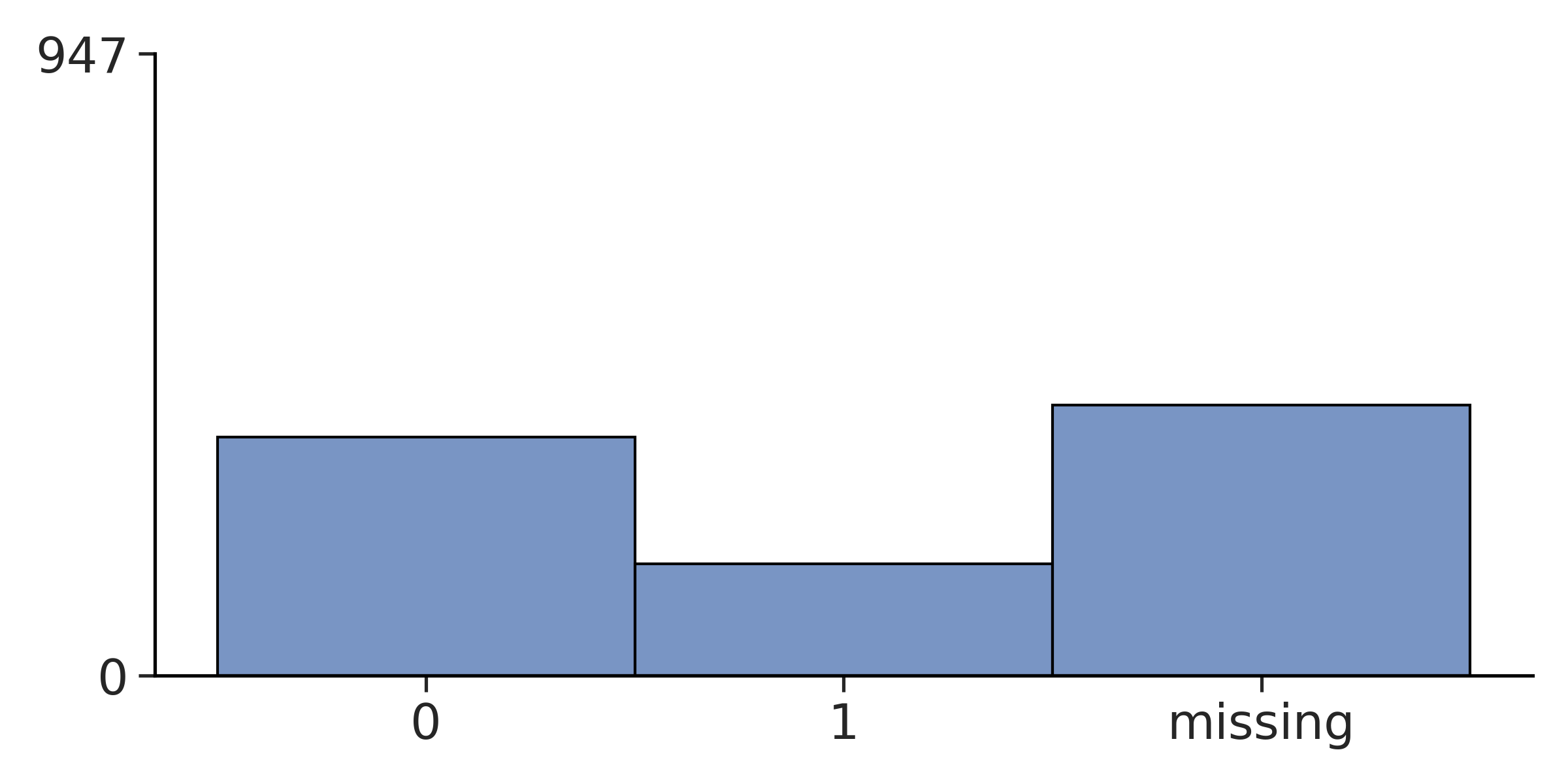}} \\
			Total number of brain infarcts & Total count of all brain infarcts & Numeric (accuracy) & 0, 1, 2 or more, missing & 0.83 &
			\raisebox{-\totalheight}{\includegraphics[width=\linewidth]{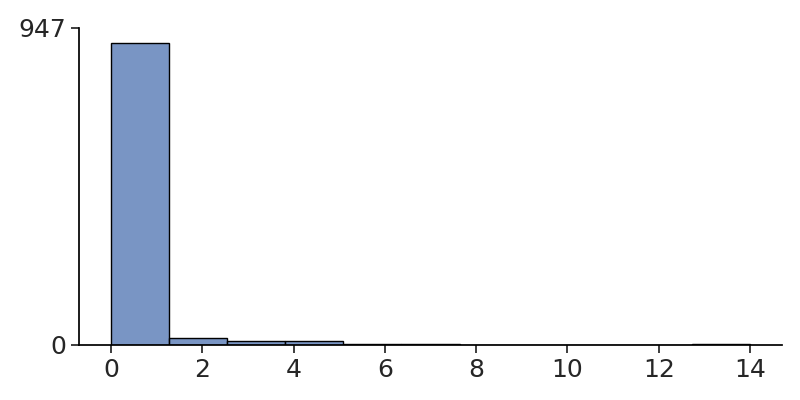}} \\
			Cortical infarcts & Indicates whether cortical infarcts are present & Binary (balanced accuracy) & 0, 1, missing & 0.87 &
			\raisebox{-\totalheight}{\includegraphics[width=\linewidth]{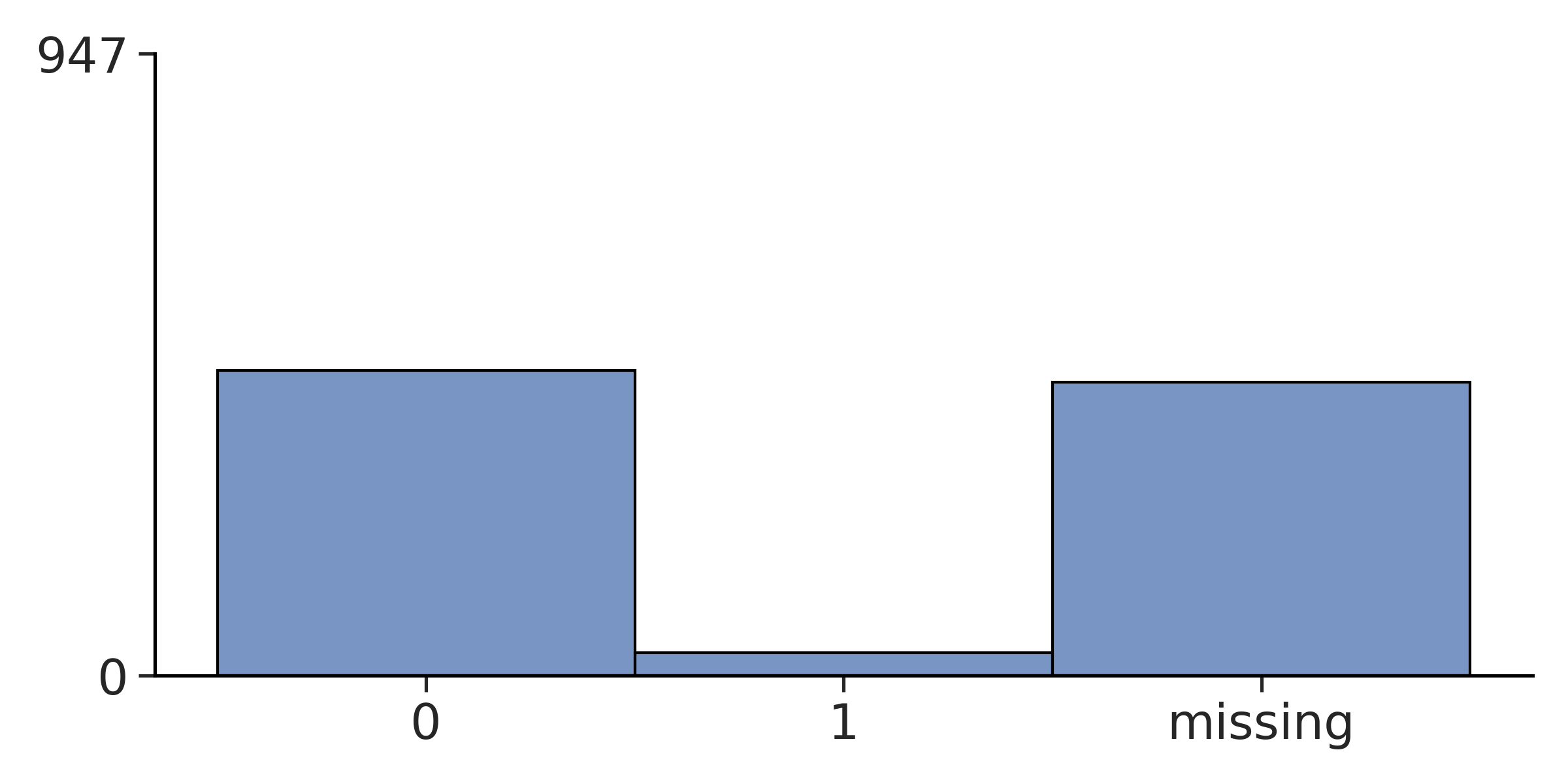}} \\
			Number of cortical infarcts & Count of cortical infarcts & Numeric (accuracy) & 0, 1, 2 or more, missing & 0.78 & 
			\raisebox{-\totalheight}{\includegraphics[width=\linewidth]{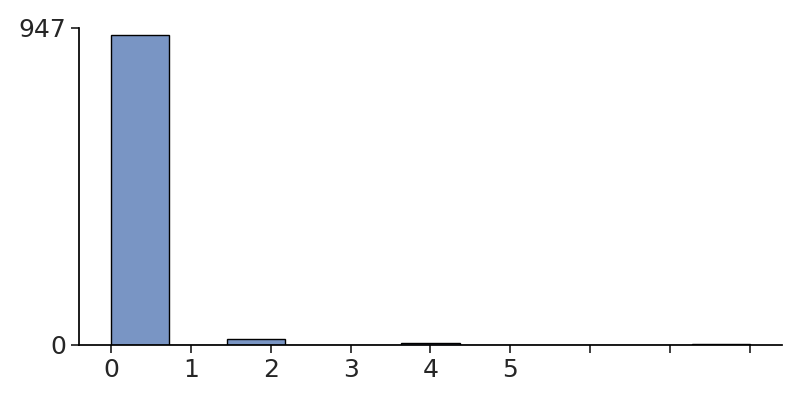}} \\
			Lacunes infarcts & Indicates whether lacunar infarcts (lacunes) are present & Binary (balanced accuracy) & 0, 1, missing & 0.92 &
			\raisebox{-\totalheight}{\includegraphics[width=\linewidth]{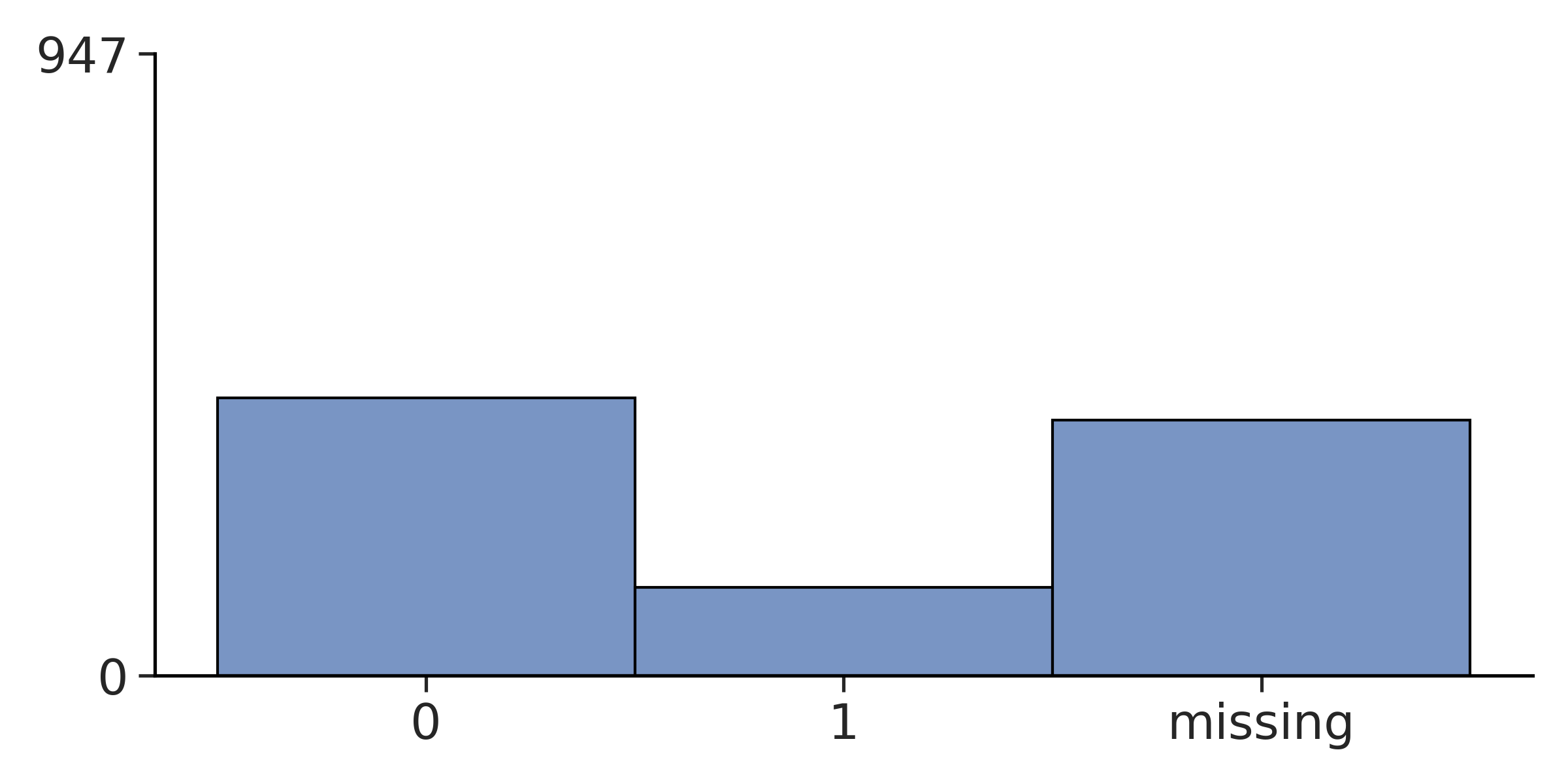}} \\
			Number of lacunar infarcts & Count of lacunar infarcts & Numeric (accuracy) & 0, 1, 2 or more, missing & 0.86 &
			\raisebox{-\totalheight}{\includegraphics[width=\linewidth]{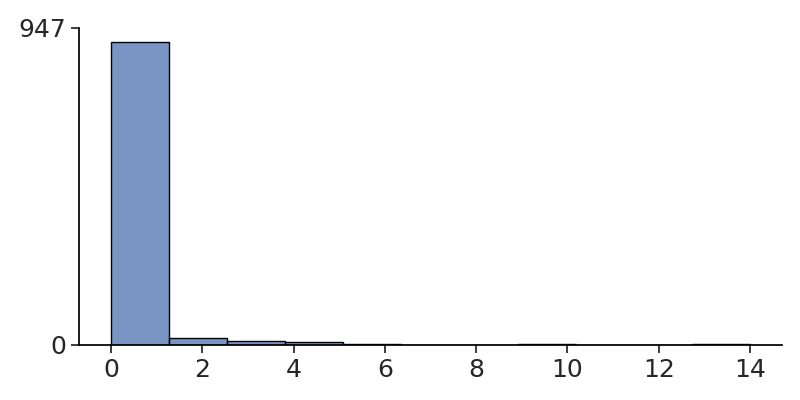}} \\
			Cerebellar infarcts & Indicates whether cerebellar infarcts are present & Binary (balanced accuracy) & 0, 1, missing & 0.80 &
			\raisebox{-\totalheight}{\includegraphics[width=\linewidth]{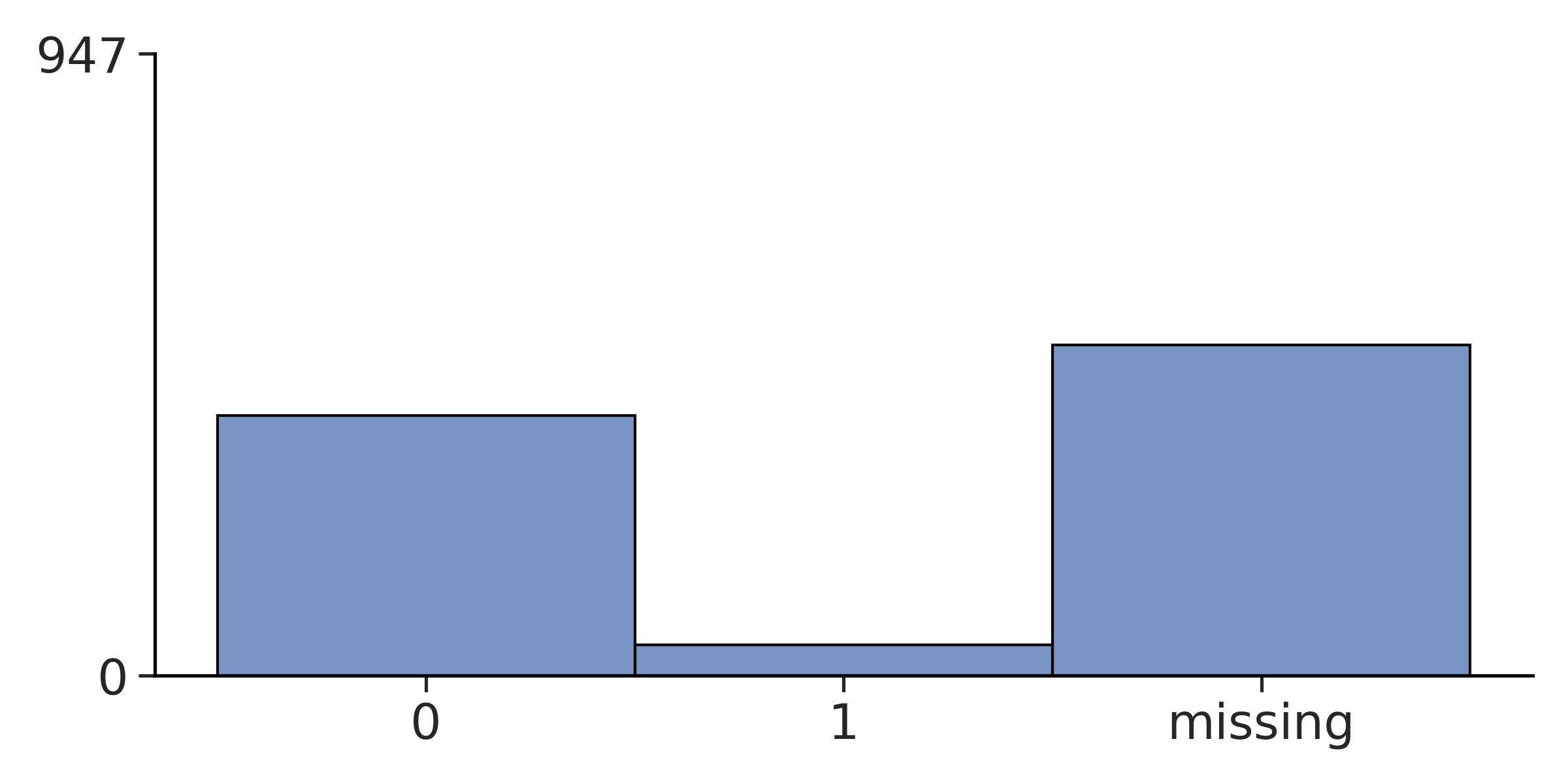}} \\
			Number of cerebellar infarcts &  Count of cerebellar infarcts& Numeric (accuracy) & 0, 1, 2 or more, missing & 0.83 &
			\raisebox{-\totalheight}{\includegraphics[width=\linewidth]{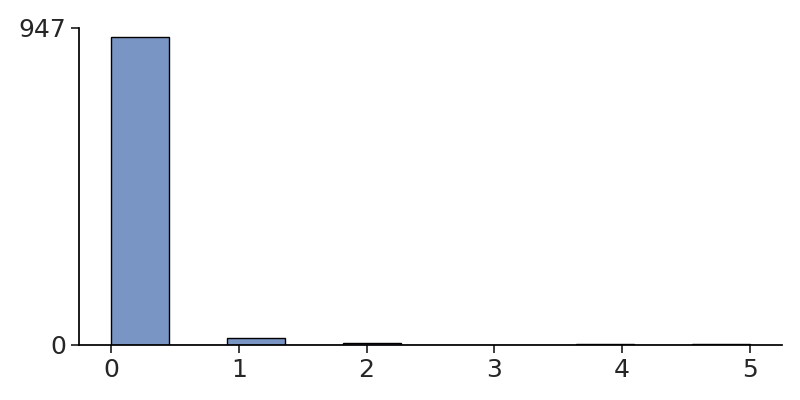}} \\
			Splinter infarcts & Indicates presence of splinter infarcts & Binary (balanced accuracy) & 0, 1, missing & 0.86 &
			\raisebox{-\totalheight}{\includegraphics[width=\linewidth]{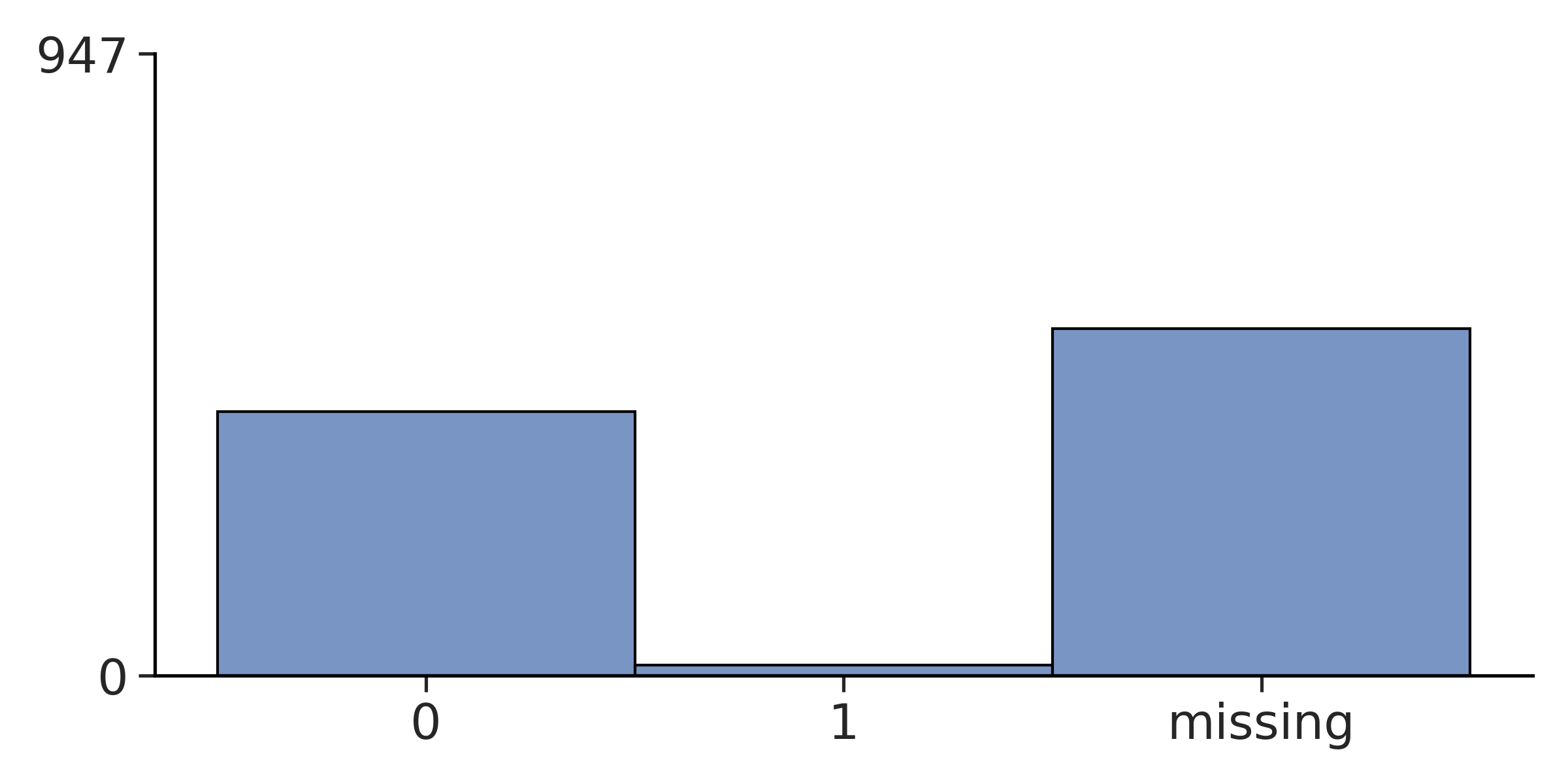}} \\
			DWI abnormalities & Flags restricted diffusion on DWI & Binary (balanced accuracy) & 0, 1, missing & 0.82 &
			\raisebox{-\totalheight}{\includegraphics[width=\linewidth]{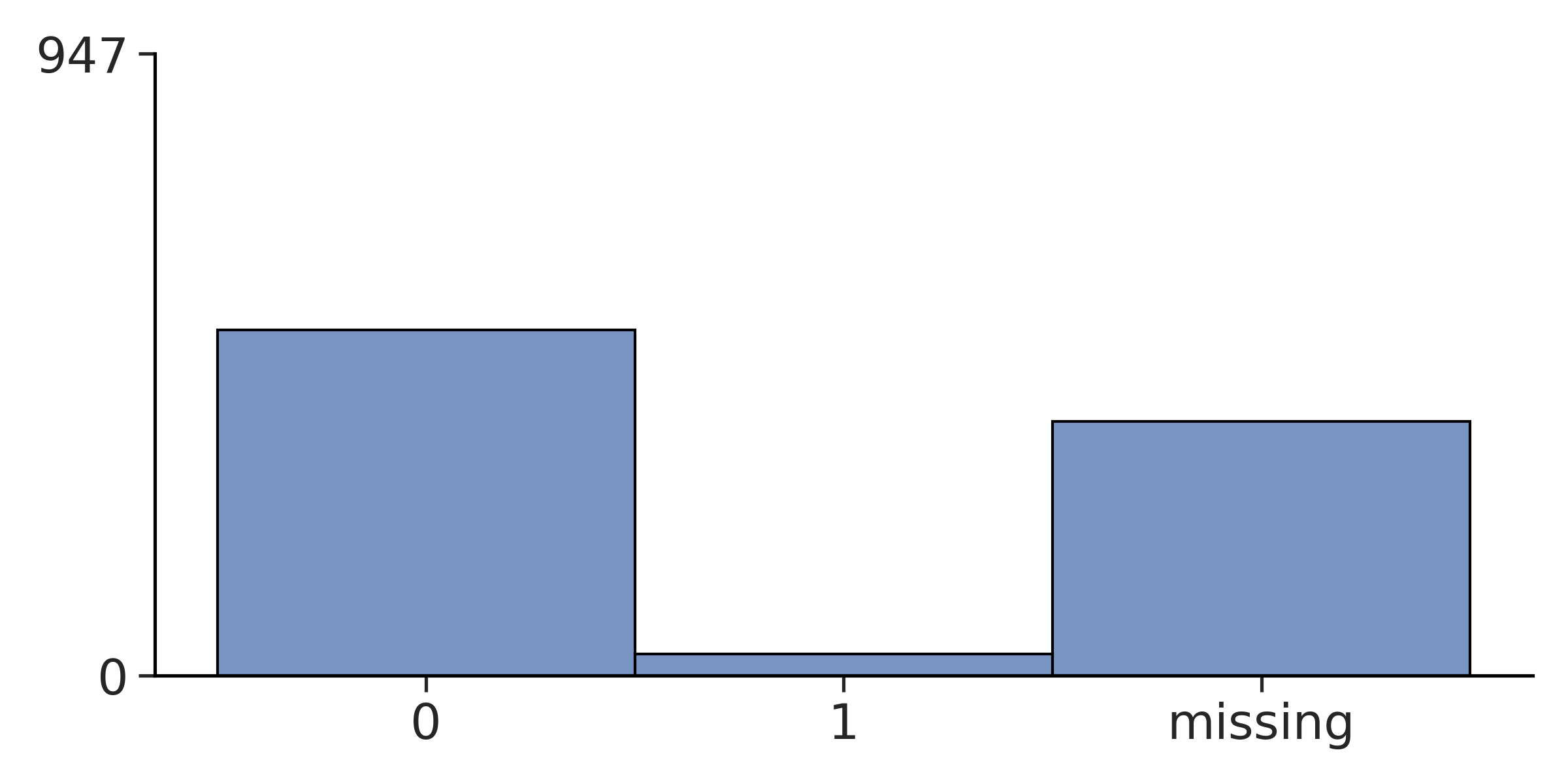}} \\
			SWI abnormalities & Flags abnormalities detected on SWI & Binary (balanced accuracy) & 0, 1, missing & 0.77 &
			\raisebox{-\totalheight}{\includegraphics[width=\linewidth]{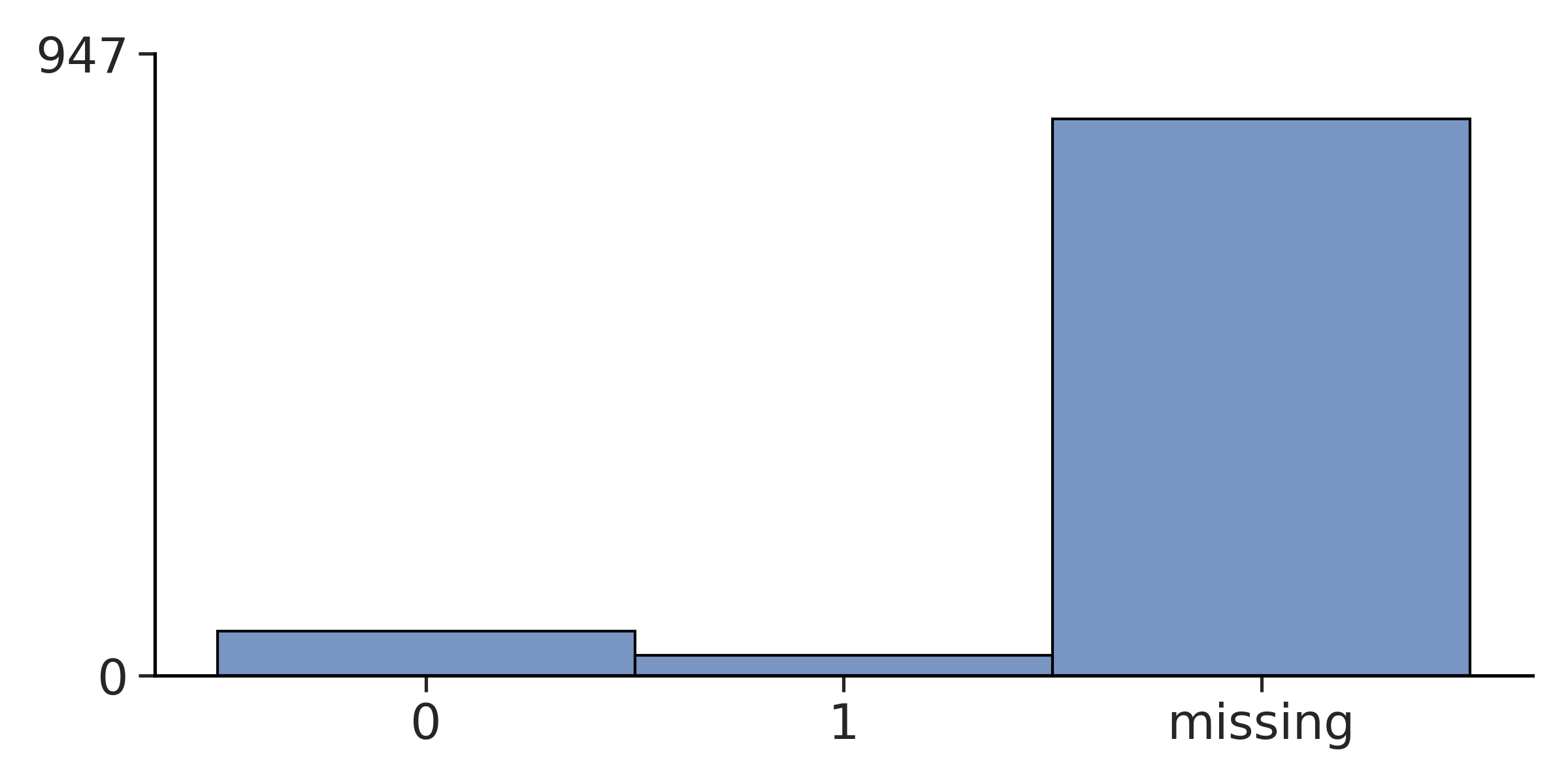}} \\
			Presence of Microbleeds & Indicates whether cerebral microbleeds are present & Binary (balanced accuracy) & 0, 1, missing & 0.85 &
			\raisebox{-\totalheight}{\includegraphics[width=\linewidth]{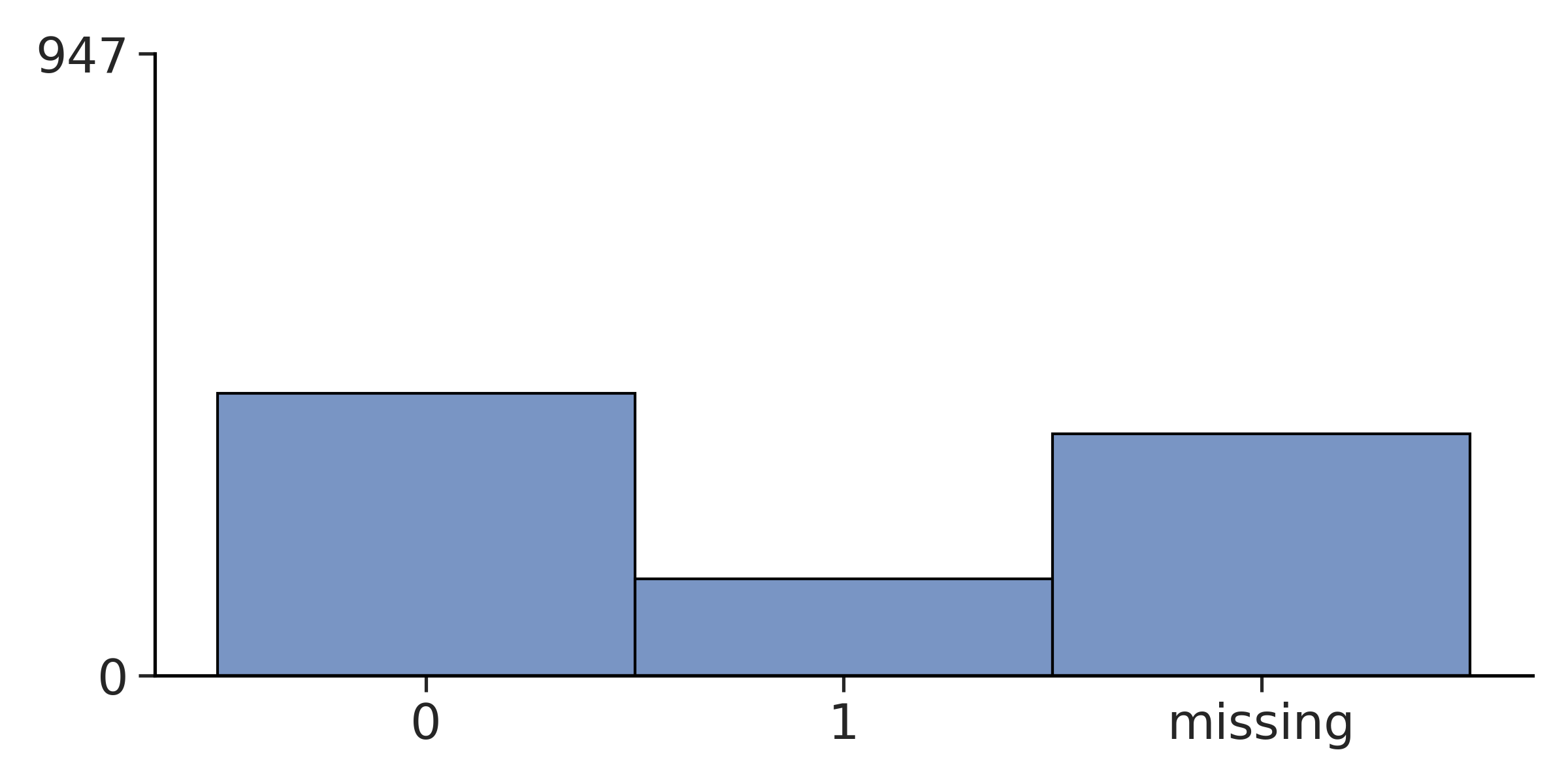}} \\
			Total number of microbleeds & Count of cerebral microbleeds & Numeric (accuracy) & 0, 1, 2 or more, missing & 0.81 &
			\raisebox{-\totalheight}{\includegraphics[width=\linewidth]{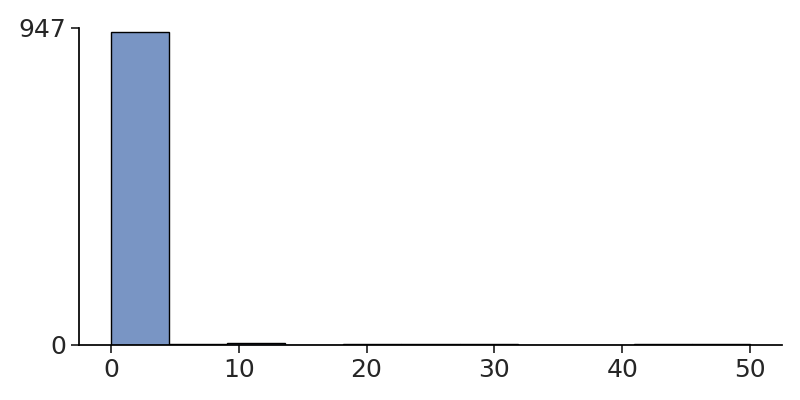}} \\
			Presence of siderosis & Indicates presence of superficial siderosis & Binary (balanced accuracy) & 0, 1 & 0.81 &
			\raisebox{-\totalheight}{\includegraphics[width=\linewidth]{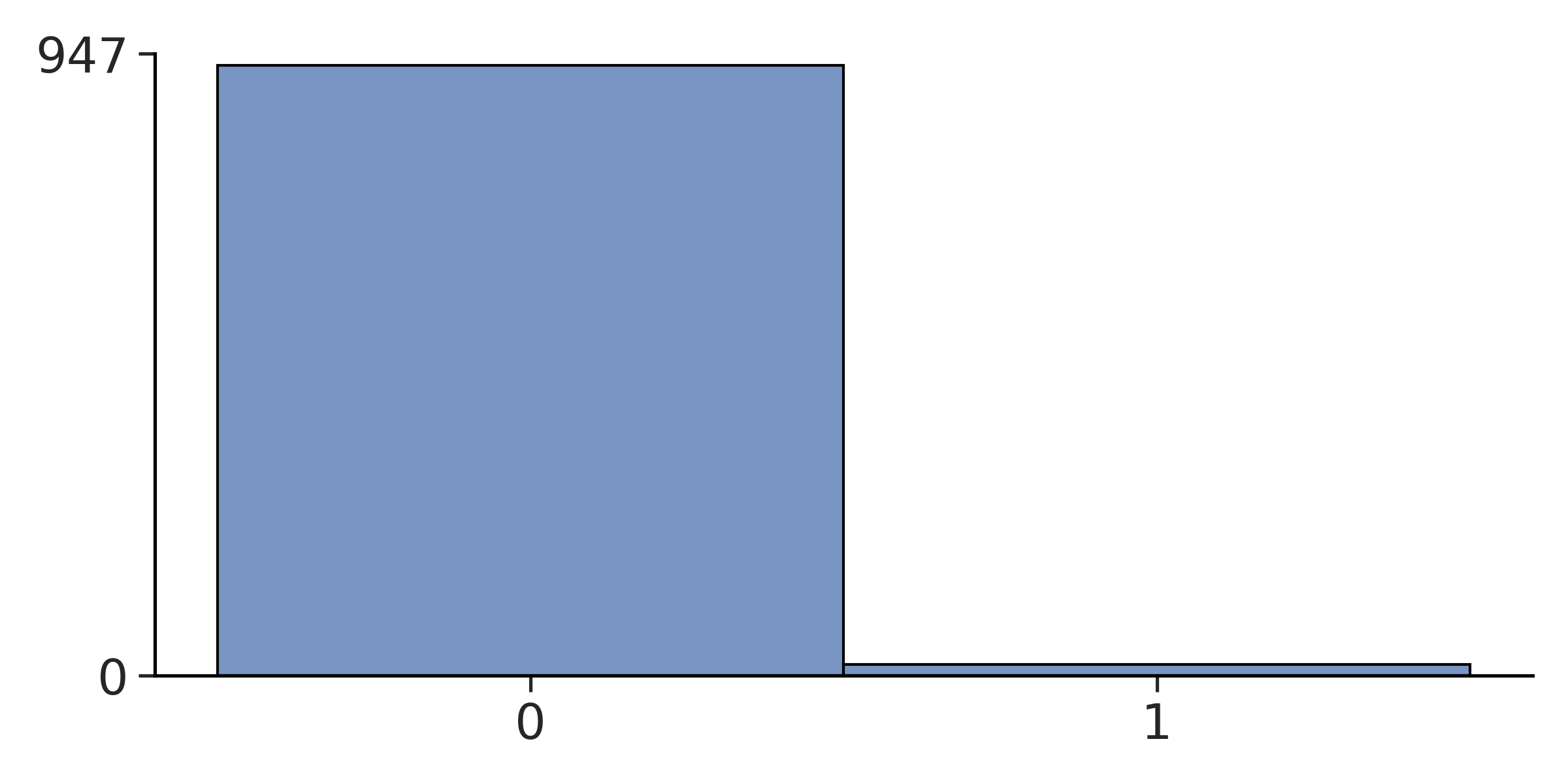}} \\
		\end{longtable}
	\end{scriptsize}
	
	\subsubsection{Prompt}
	For this use case, a structured prompt was created to ensure consistent extraction and enforce strict JSON formatting. 
	The prompt is divided into role-based instructions (System, Human, AI). A full example of the exact prompt used is shown in \autoref{tab:prompt-Dementia}.
	
	\begin{scriptsize}
		\setlength{\tabcolsep}{4pt}
		\begin{longtable}{>{\scriptsize\raggedright\arraybackslash}m{0.2\linewidth} >{\scriptsize\raggedright\arraybackslash}m{0.37\linewidth} >{\scriptsize\raggedright\arraybackslash}m{0.37\linewidth}}
			\multicolumn{3}{c}{\parbox{\textwidth}{
					\normalsize \tablename~\thetable{} -- Full structured prompt used for the neurodegenerative diseases extraction task. The items in brackets indicate the role of the message (System, Human, AI), while the text provides the corresponding content.\\}} \\
			\toprule
			\textbf{Section} & \multicolumn{2}{c}{\textbf{Content} (Depending on prompt strategy)} \\
			& Zero-Shot, One-Shot and Few-Shot & CoT, Self-Consistency and Graph\\
			\midrule
			\endfirsthead
			
			\multicolumn{3}{c}{\parbox{\textwidth}{
					\normalsize \tablename~\thetable{} -- Continued\\}} \\
			\toprule
			\textbf{Section} & \multicolumn{2}{c}{\textbf{Content}} \\
			\midrule
			\endhead
			
			\phantomlabel{tab:prompt-Dementia}
			\textbf{[System] -- System instructions} & 
			You are a medical data extraction system that ONLY outputs valid JSON. Maintain strict compliance with these rules: \newline
			1. ALWAYS begin and end your response with \verb|```json| markers \newline
			2. Use EXACT field names and structure provided \newline
			3. If a value is missing or not mentioned, use the specified default for that field. \newline
			4. NEVER add commentary, explanations, or deviate from the output structure & 
			You are a medical data extraction system that performs structured reasoning before producing output. Follow these strict rules: \newline
			1. First, reason step-by-step to identify and justify each extracted field. \newline
			2. After reasoning, output ONLY valid JSON in the exact structure provided. \newline
			3. ALWAYS begin and end the final output with \verb|```json| markers — do not include reasoning within these markers. \newline
			4. Use EXACT field names and structure as specified. \newline
			5. If a value is missing or not mentioned, use the specified default for that field. \newline
			6. NEVER include commentary, explanations, or deviate from the specified format in the final JSON. \newline \\
			\midrule
			\textbf{[Human] -- Field instructions} & \multicolumn{2}{c}{\parbox{0.75\linewidth}{
					1. \opus{"Fazekas score"}: \newline
					- Type: number\_or\_missing \newline
					- Return the Fazekas score (0-3). If given as a range (e.g., 1-2), return the average (e.g., 1.5). If not mentioned, return 'missing'. \newline
					- Options: [0, 0.5, 1, 1.5, 2, 2.5, 3, \opus{"missing"}] \newline
					- Default: \opus{"missing"} \newline
					2. \opus{"MTA left"}: \newline
					- Type: number\_or\_missing \newline
					- Return the MTA score for the left side (0-4). If given as a range, return the average. If not mentioned, return 'missing'. \newline
					- Options: [0, 0.5, 1, 1.5, 2, 2.5, 3, 3.5, 4, \opus{"missing"}] \newline
					- Default: \opus{"missing"} \newline
					3. \opus{"MTA right"}: \newline
					- Type: number\_or\_missing \newline
					- Return the MTA score for the right side (0-4). If given as a range, return the average. If not mentioned, return 'missing'. \newline
					- Options: [0, 0.5, 1, 1.5, 2, 2.5, 3, 3.5, 4, \opus{"missing"}] \newline
					- Default: \opus{"missing"} \newline
					4. \opus{"GCA frontal"}: \newline
					- Type: number\_or\_missing \newline
					- Return the GCA score of the frontal lobe (0-3). Use lobar score if available, else generalized GCA score. If not mentioned, return 'missing'. \newline
					- Options: [0, 0.5, 1, 1.5, 2, 2.5, 3, \opus{"missing"}] \newline
					- Default: \opus{"missing"} \newline
					5. \opus{"GCA temporal"}: \newline
					- Type: number\_or\_missing \newline
					- Return the GCA score of the temporal lobe (0-3). If not mentioned, return 'missing'. \newline
					- Options: [0, 0.5, 1, 1.5, 2, 2.5, 3, \opus{"missing"}] \newline
					- Default: \opus{"missing"} \newline
					6. \opus{"GCA occipital"}: \newline
					- Type: number\_or\_missing \newline
					- Return the GCA score of the occipital lobe (0-3). If not mentioned, return 'missing'. \newline
					- Options: [0, 0.5, 1, 1.5, 2, 2.5, 3, \opus{"missing"}] \newline
					- Default: \opus{"missing"} \newline
					7. \opus{"GCA parietal"}: \newline
					- Type: number\_or\_missing \newline
					- Return the GCA score of the parietal lobe (0-3). If not mentioned, return 'missing'. \newline
					- Options: [0, 0.5, 1, 1.5, 2, 2.5, 3, \opus{"missing"}] \newline
					- Default: \opus{"missing"} \newline
					8. \opus{"GCA overall"}: \newline
					- Type: number\_or\_missing \newline
					- Return the overall GCA score (0-3). If only lobar scores provided, return the highest. If not mentioned, return 'missing'. \newline
					- Options: [0, 0.5, 1, 1.5, 2, 2.5, 3, \opus{"missing"}] \newline
					- Default: \opus{"missing"} \newline
					9. \opus{"Any brain infarct"}: \newline
					- Type: binary\_or\_missing \newline
					- Return 1 if any infarct is present, 0 if explicitly absent, 'missing' if not mentioned. \newline
					- Options: [0, 1, \opus{"missing"}] \newline
					- Default: \opus{"missing"} \newline
					10. \opus{"Total number of brain infarcts"}: \newline
					- Type: number\_or\_missing \newline
					- Return total number of infarcts. If presence confirmed but not counted, return 1. Explicitly absent = 0. If not mentioned, return 'missing'. \newline
					- Default: \opus{"missing"} \newline
					11. \opus{"Cortical infarcts"}: \newline
					- Type: binary\_or\_missing \newline
					- Return 1 if cortical infarcts are present, 0 if absent, 'missing' if not mentioned. \newline
					- Options: [0, 1, \opus{"missing"}] \newline
					- Default: \opus{"missing"} \newline
			}} \newline \\
			\midrule
			\textbf{[Human] -- Field instructions} & \multicolumn{2}{c}{\parbox{0.75\linewidth}{
					12. \opus{"Number of cortical infarcts"}: \newline
					- Type: number\_or\_missing \newline
					- Return total number of cortical infarcts. Confirmed presence = 1, absent = 0, missing if not mentioned. \newline
					- Default: \opus{"missing"} \newline
					13. \opus{"Lacunes infarcts"}: \newline
					- Type: binary\_or\_missing \newline
					- Return 1 if lacunar infarcts present, 0 if absent, 'missing' if not mentioned. \newline
					- Options: [0, 1, \opus{"missing"}] \newline
					- Default: \opus{"missing"} \newline
					14. \opus{"Number of lacunar infarcts"}: \newline
					- Type: number\_or\_missing \newline
					- Return total number of lacunar infarcts. Confirmed presence = 1, absent = 0, missing if not mentioned. \newline
					- Default: \opus{"missing"} \newline
					15. \opus{"Cerebellar infarcts"}: \newline
					- Type: binary\_or\_missing \newline
					- Return 1 if cerebellar infarcts present, 0 if absent, 'missing' if not mentioned. \newline
					- Options: [0, 1, \opus{"missing"}] \newline
					- Default: \opus{"missing"} \newline
					16. \opus{"Number of cerebellar infarcts"}: \newline
					- Type: number\_or\_missing \newline
					- Return total number of cerebellar infarcts. Confirmed presence = 1, absent = 0, missing if not mentioned. \newline
					- Default: \opus{"missing"} \newline
					17. \opus{"Splinter infarcts"}: \newline
					- Type: binary\_or\_missing \newline
					- Return 1 if splinter infarcts present, 0 if absent, 'missing' if not mentioned. \newline
					- Options: [0, 1, \opus{"missing"}] \newline
					- Default: \opus{"missing"} \newline
					18. \opus{"DWI abnormalities"}: \newline
					- Type: binary\_or\_missing \newline
					- Return 1 if DWI shows diffusion restriction, 0 if normal, 'missing' if not mentioned. \newline
					- Options: [0, 1, \opus{"missing"}] \newline
					- Default: \opus{"missing"} \newline
					19. \opus{"SWI abnormalities"}: \newline
					- Type: binary\_or\_missing \newline
					- Return 1 if SWI shows abnormalities, 0 if normal, 'missing' if not mentioned. \newline
					- Options: [0, 1, \opus{"missing"}] \newline
					- Default: \opus{"missing"} \newline
					20. \opus{"Presence of Microbleeds"}: \newline
					- Type: binary\_or\_missing \newline
					- Return 1 if microbleeds present, 0 if absent, 'missing' if not mentioned. \newline
					- Options: [0, 1, \opus{"missing"}] \newline
					- Default: \opus{"missing"} \newline
					21. \opus{"Total number of microbleeds"}: \newline
					- Type: number \newline
					- Return total number of microbleeds. Confirmed presence = 1, absent = 0, 'missing' if not mentioned. \newline
					- Default: \opus{"missing"} \newline
					22. \opus{"Presence of siderosis"}: \newline
					- Type: binary \newline
					- Return 1 if superficial siderosis is mentioned, 0 otherwise. \newline
					- Options: [0, 1] \newline
					- Default: 0
			}} \newline \\
			\midrule
			\textbf{[Human] -- Task instructions} & \multicolumn{2}{c}{\parbox{0.75\linewidth}{
					Extract the neuroimaging findings into this exact JSON structure:
					\opus{```json} \newline
					\opus{\{} \newline
					\opus{\quad "Fazekas score": "",} \newline
					\opus{\quad "MTA left": "",} \newline
					\opus{\quad "MTA right": "",} \newline
					\opus{\quad "GCA frontal": "",} \newline
					\opus{\quad "GCA temporal": "",} \newline
					\opus{\quad "GCA parietal": "",} \newline
					\opus{\quad "GCA occipital": "",} \newline
					\opus{\quad "GCA overall": "",} \newline
					\opus{\quad "Any brain infarct": "",} \newline
					\opus{\quad "Total number of brain infarcts": "",} \newline
					\opus{\quad "Cortical infarcts": "",} \newline
					\opus{\quad "Number of cortical infarcts": "",} \newline
					\opus{\quad "Lacunes infarcts": "",} \newline
					\opus{\quad "Number of lacunar infarcts": "",} \newline
					\opus{\quad "Cerebellar infarcts": "",} \newline
					\opus{\quad "Number of cerebellar infarcts": "",} \newline
					\opus{\quad "Splinter infarcts": "",} \newline
					\opus{\quad "DWI abnormalities": "",} \newline
					\opus{\quad "SWI abnormalities": "",} \newline
					\opus{\quad "Presence of Microbleeds": "",} \newline
					\opus{\quad "Total number of microbleeds": "",} \newline
					\opus{\quad "Presence of siderosis": ""} \newline
					\opus{\}} \newline
					\opus{```}%
				}%
			} \newline \\
			\midrule
			\textbf{[Human] -- Example intro} & \multicolumn{2}{c}{\parbox{0.75\linewidth}{Below are 1 example of expected input and output, followed by a new task.}} \newline \\
			\midrule
			\textbf{[Human] -- Example user} & \multicolumn{2}{c}{\parbox{0.75\linewidth}{\opus{[...CENSORED...]}}} \newline \\
			\midrule
			\textbf{[AI] -- Example assistant reasoning} & - & 
			- \opus{Fazekas score} - "no score provided" implies "missing" \newline
			- \opus{MTA left} - "milde bilaterale mediale temporale atrofie, scorend 1" implies 1 \newline
			- \opus{MTA right} - "bilateral atrophy" implies 1 \newline
			- \opus{GCA frontal} - "lichte prominentie in de frontale regio's" implies 1 \newline
			- \opus{GCA temporal} - "temporal region prominence mentioned" implies 1 \newline
			- \opus{GCA parietal} - "not mentioned" implies "missing" \newline
			- \opus{GCA occipital} - "not mentioned" implies "missing" \newline
			- \opus{GCA overall} - "explicit 'Globale corticale atrofie... mild (GCA score 1)'" implies 1 \newline
			- \opus{Any brain infarct} - "explicit absence" implies 0 \newline
			- \opus{Total number of brain infarcts} - "no infarcts present" implies 0 \newline
			- \opus{Cortical infarcts} - "geen corticale infarcten" implies 0 \newline
			- \opus{Number of cortical infarcts} - "none present" implies 0 \newline
			- \opus{Lacunes infarcts} - "geen lacunaire infarcten" implies 0 \newline
			- \opus{Number of lacunar infarcts} - "none present" implies 0 \newline
			- \opus{Cerebellar infarcts} - "cerebellum not mentioned" implies "missing" \newline
			- \opus{Number of cerebellar infarcts} - "no assessment" implies "missing" \newline
			- \opus{Splinter infarcts} - "not mentioned" implies "missing" \newline
			- \opus{DWI abnormalities} - "geen diffusie restrictie" implies 0 \newline
			- \opus{SWI abnormalities} - "no microbleeds identified" implies 0 \newline
			- \opus{Presence of Microbleeds} - "matches SWI findings" implies 0 \newline
			- \opus{Total number of microbleeds} - "none detected" implies 0 \newline
			- \opus{Presence of siderosis} - "geen siderose" implies 0 \newline \\
			\midrule
			\textbf{[AI] -- Example assistant output} & \multicolumn{2}{c}{\parbox{0.75\linewidth}{
					\opus{```json} \newline
					\opus{\{} \newline
					\opus{\quad "Fazekas score": "missing",} \newline
					\opus{\quad "MTA left": 1,} \newline
					\opus{\quad "MTA right": 1,} \newline
					\opus{\quad "GCA frontal": 1,} \newline
					\opus{\quad "GCA temporal": 1,} \newline
					\opus{\quad "GCA parietal": "missing",} \newline
					\opus{\quad "GCA occipital": "missing",} \newline
					\opus{\quad "GCA overall": 1,} \newline
					\opus{\quad "Any brain infarct": 0,} \newline
					\opus{\quad "Total number of brain infarcts": 0,} \newline
					\opus{\quad "Cortical infarcts": 0,} \newline
					\opus{\quad "Number of cortical infarcts": 0,} \newline
					\opus{\quad "Lacunes infarcts": 0,} \newline
					\opus{\quad "Number of lacunar infarcts": 0,} \newline
					\opus{\quad "Cerebellar infarcts": "missing",} \newline
					\opus{\quad "Number of cerebellar infarcts": "missing",} \newline
					\opus{\quad "Splinter infarcts": "missing",} \newline
					\opus{\quad "DWI abnormalities": 0,} \newline
					\opus{\quad "SWI abnormalities": 0,} \newline
					\opus{\quad "Presence of Microbleeds": 0,} \newline
					\opus{\quad "Total number of microbleeds": 0,} \newline
					\opus{\quad "Presence of siderosis": 0} \newline
					\opus{\}} \newline
					\opus{```}%
				}%
			} \newline \\
			\midrule
			\textbf{[Human] -- Report instructions} & \multicolumn{2}{c}{\parbox{0.75\linewidth}{[file name]: \opus{[...CENSORED...]} \newline \opus{[...CENSORED...]}}} \newline \\
			\midrule
			\textbf{[Human] -- Final instructions} &
			Begin the extraction now. Your response must contain only a single valid JSON block, enclosed in triple backticks and prefixed with \verb|`json`|, like this: \verb|```json  ... ```|& 
			Begin the extraction now. First, reason step-by-step to identify and justify the value for each required field, enclosed within \verb|<think>...</think>| tags. Then, output only the final structured data as a single valid JSON block, starting with \verb|```json| and ending with \verb|```|."
			\\
			\bottomrule
		\end{longtable}
	\end{scriptsize}
	
	\subsubsection{Prompt Graph}
	The dependencies, conditional branches, and extraction order of variables are represented as a directed acyclic graph. This graph reflects how the extraction task is decomposed into smaller, sequential subtasks for the Prompt Graph prompting strategy. 
	\begin{figure}[htbp]
		\centering
		\includegraphics[width=\linewidth, height=0.5\textheight, keepaspectratio]{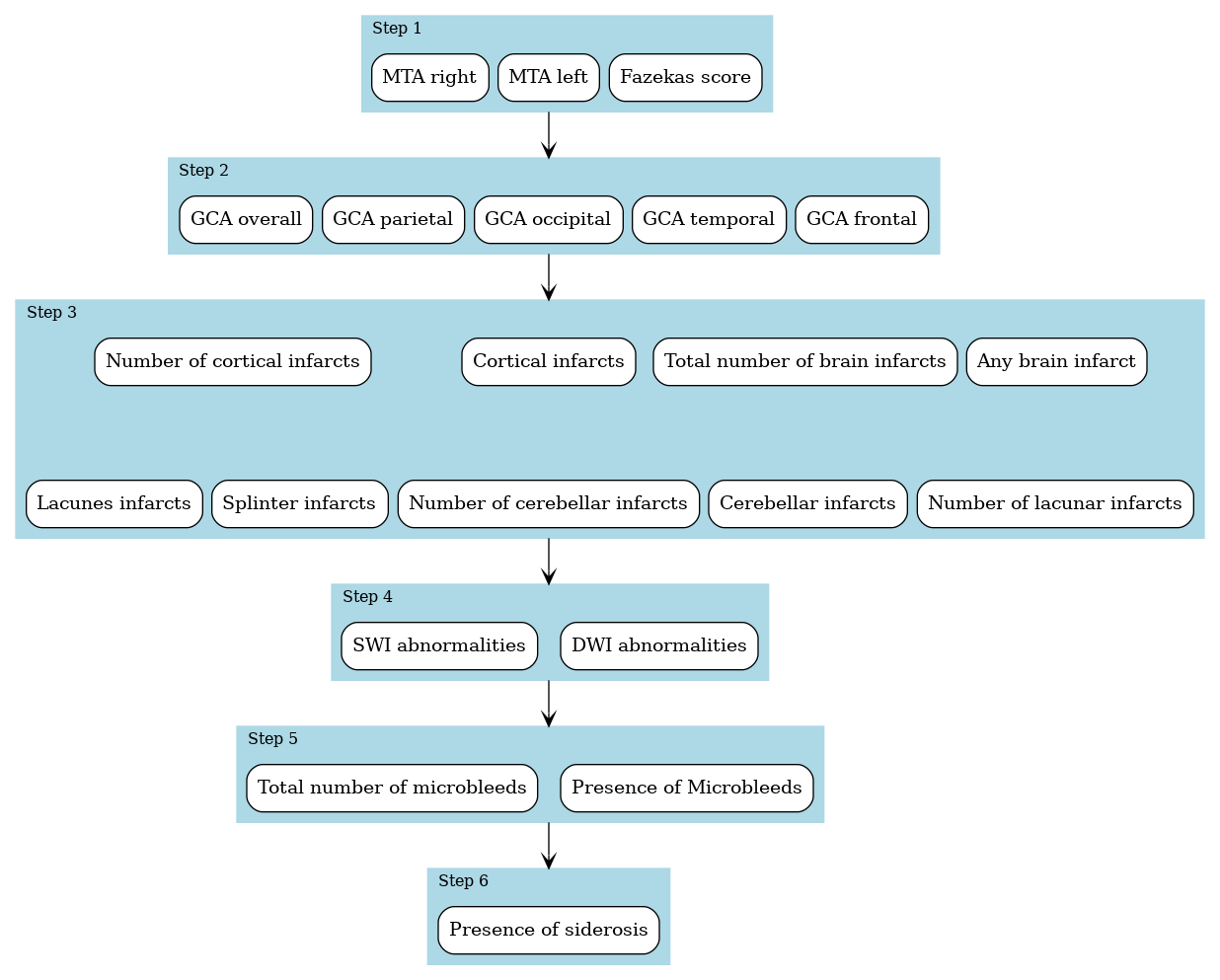}
		\caption{Directed acyclic graph showing sequential extraction order of variable extraction neurodegenerative diseases use case.}
	\end{figure}
	
	\newpage
	\subsection{Use Case - Soft Tissue Tumours}
	
	\subsubsection{Overview}
	
	This use case focuses on extracting structured information from pathology reports of patients with soft tissue tumours, specifically targeting phenotype, grade, and tumour location \autoref{tab:use-case-STT}. Notably, the task evaluates the ability to extract these variables from both English and Dutch reports using the same prompt, with the only difference being that the examples provided were language-specific (English vs. Dutch).
	
	\subsubsection{Inclusion and Exclusion Criteria}
	Pathology reports were included for patients with a suspected soft tissue tumour between 2007 and 2025 who underwent baseline imaging (CT or MRI) and received any form of pathology assessment (biopsy, resection, excision, etc.). Exclusion criteria were reports of conditions not within the differential diagnosis of soft tissue tumours, as well as cancelled pathology procedures or incomplete reporting. In total, 2021 English reports and 6,754 Dutch reports were collected. From these 350 reports were randomly sampled, ensuring that each patient was represented only once. After applying exclusion criteria during annotation, the final dataset included 300 English and 327 Dutch pathology reports.
	
	\subsubsection{Annotation}
	The English reports were annotated by a single medical professional. The Dutch reports were divided into two subsets and annotated by two PhD students. For both a subset of 50 patients was randomly selected and annotated by an additional medical professional in order to calculate inter-rater agreement.
	
	\subsubsection{Ethical Considerations and Funding}
	The collection of the English dataset was funded by Health Education England (HEE) and the National Institute for Health Research (NIHR) for this research project. The views expressed in this publication are those of the authors and do not necessarily reflect the positions of the NIHR, NHS, or the UK Department of Health and Social Care. For the collection of the Dutch data this study was approved by the ethical review board of Erasmus Medical Center (MEC-2021-0650), and was supported by an unrestricted grant from Stichting Hanarth Fonds, The Netherlands. 
	
	\begin{scriptsize}
		\setlength{\tabcolsep}{4pt}
		\begin{longtable}{@{}
				>{\scriptsize\raggedright\arraybackslash}p{0.10\textwidth} % Variable name
				>{\scriptsize\raggedright\arraybackslash}p{0.15\textwidth} % Description
				>{\scriptsize\raggedright\arraybackslash}p{0.10\textwidth} % Variable type
				>{\scriptsize\raggedright\arraybackslash}p{0.11\textwidth} % Variable options
				>{\scriptsize\raggedleft\arraybackslash}p{0.11\textwidth} % Annotator Agreement
				>{\scriptsize\centering\arraybackslash}p{0.30\textwidth} @{}} % Distribution with image
			\multicolumn{6}{c}{\parbox{\textwidth}{
					\normalsize \tablename~\thetable{} -- Soft tissue tumours pathology report variable definitions with reference standard distribution for the 300 English reports.\\}} \\
			\toprule
			\textbf{Variable Name} & 
			\textbf{Description} & 
			\textbf{Type (Metric)} & 
			\textbf{Variable Options} & 
			\textbf{Inter-rater Agreement} & 
			\textbf{Reference Standard Distribution} \\
			\midrule
			\endfirsthead
			
			\multicolumn{6}{c}{\parbox{\textwidth}{
					\normalsize \tablename~\thetable{} -- Continued\\}} \\
			\toprule
			\textbf{Variable Name} & 
			\textbf{Description} & 
			\textbf{Type (Metric)} & 
			\textbf{Variable Options} & 
			\textbf{Inter-rater Agreement} & 
			\textbf{Reference Standard Distribution} \\
			\midrule
			\endhead
			
			\midrule
			\endfoot
			
			\bottomrule
			\endlastfoot
			
			\phantomlabel{tab:use-case-STT}
			Specimen type & Type of tissue specimen obtained for analysis & Categorical (balanced accuracy) & Biopsy, Resection, Cytology, Other, Not specified & 0.66 &
			\raisebox{-\totalheight}{\includegraphics[width=\linewidth]{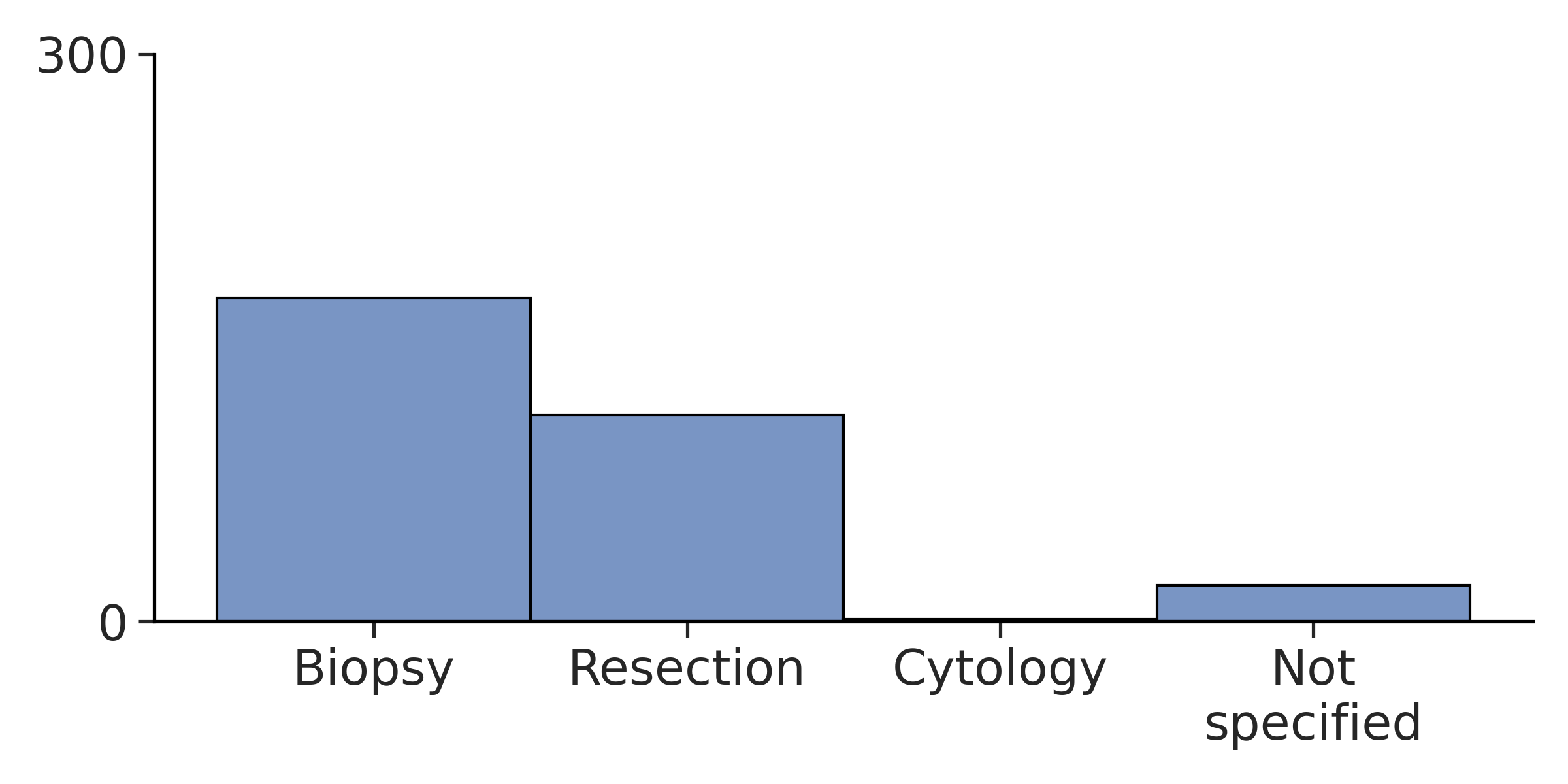}} \\
			
			Pathology request reason & Reason for pathology examination request & Categorical (balanced accuracy) & Suspicion of tumour, Metastasis, Follow-up, Recurrent tumour, Reassessment, Not specified & 0.38 &
			\raisebox{-\totalheight}{\includegraphics[width=\linewidth]{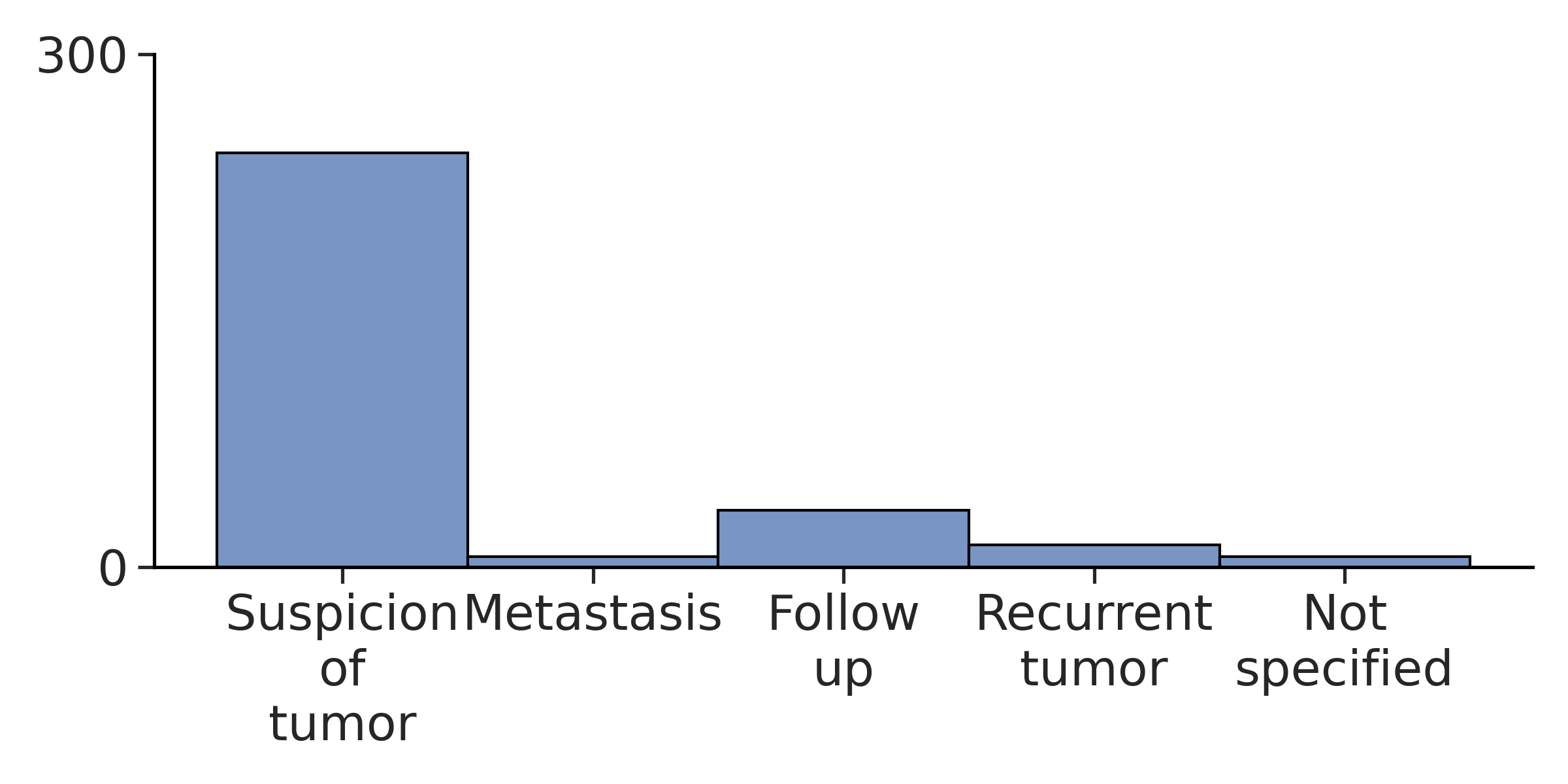}} \\
			
			Disease type & Specific disease classification based on UMLS & Free-text (cosine similarity) & - & 0.89 &
			\raisebox{-\totalheight}{\includegraphics[width=\linewidth]{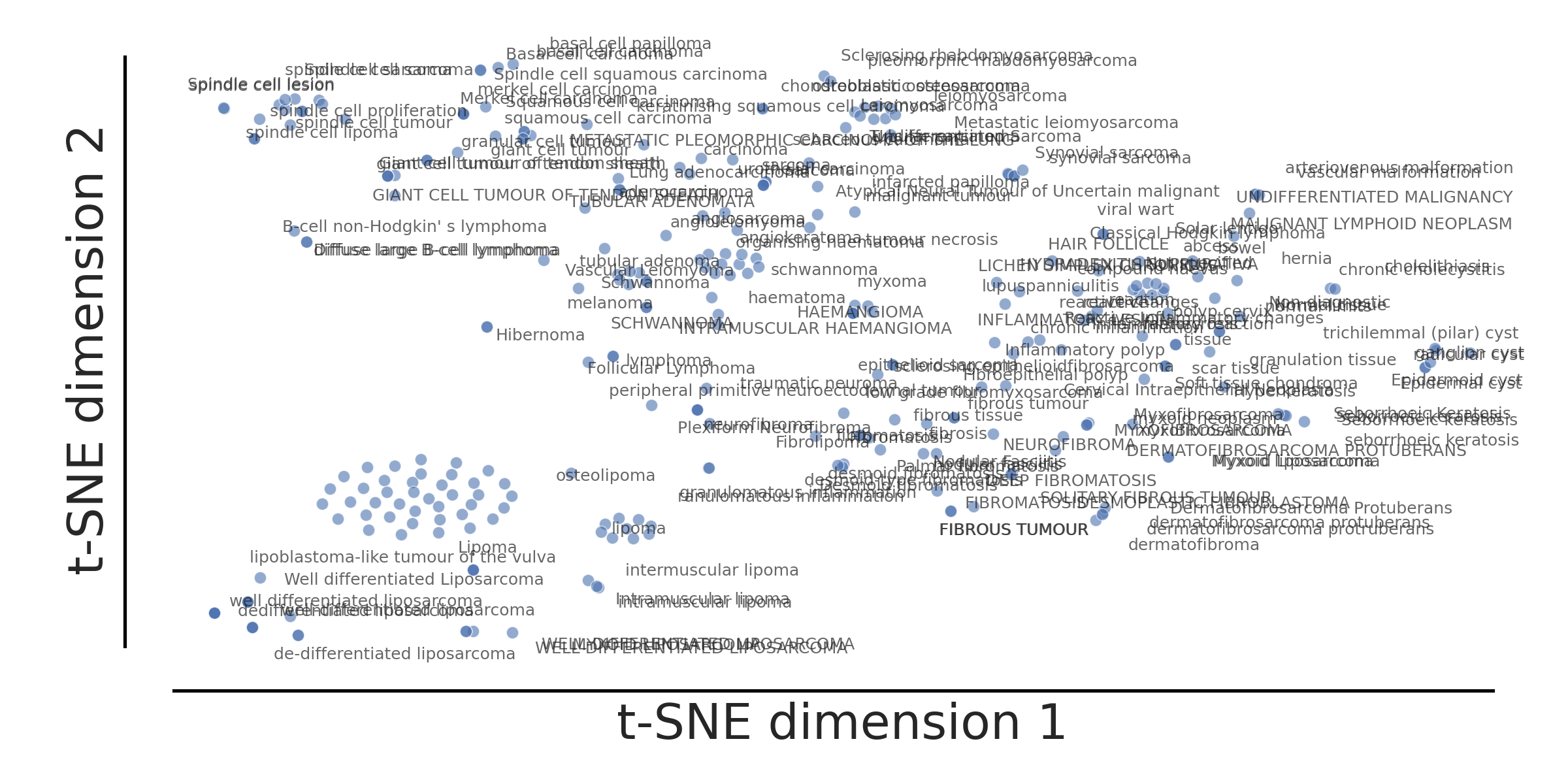}} \\
			
			Disease type differential diagnosis & Array of possible differential diagnoses & List (symmetric similarity) & - & 0.69 &
			\raisebox{-\totalheight}{\includegraphics[width=\linewidth]{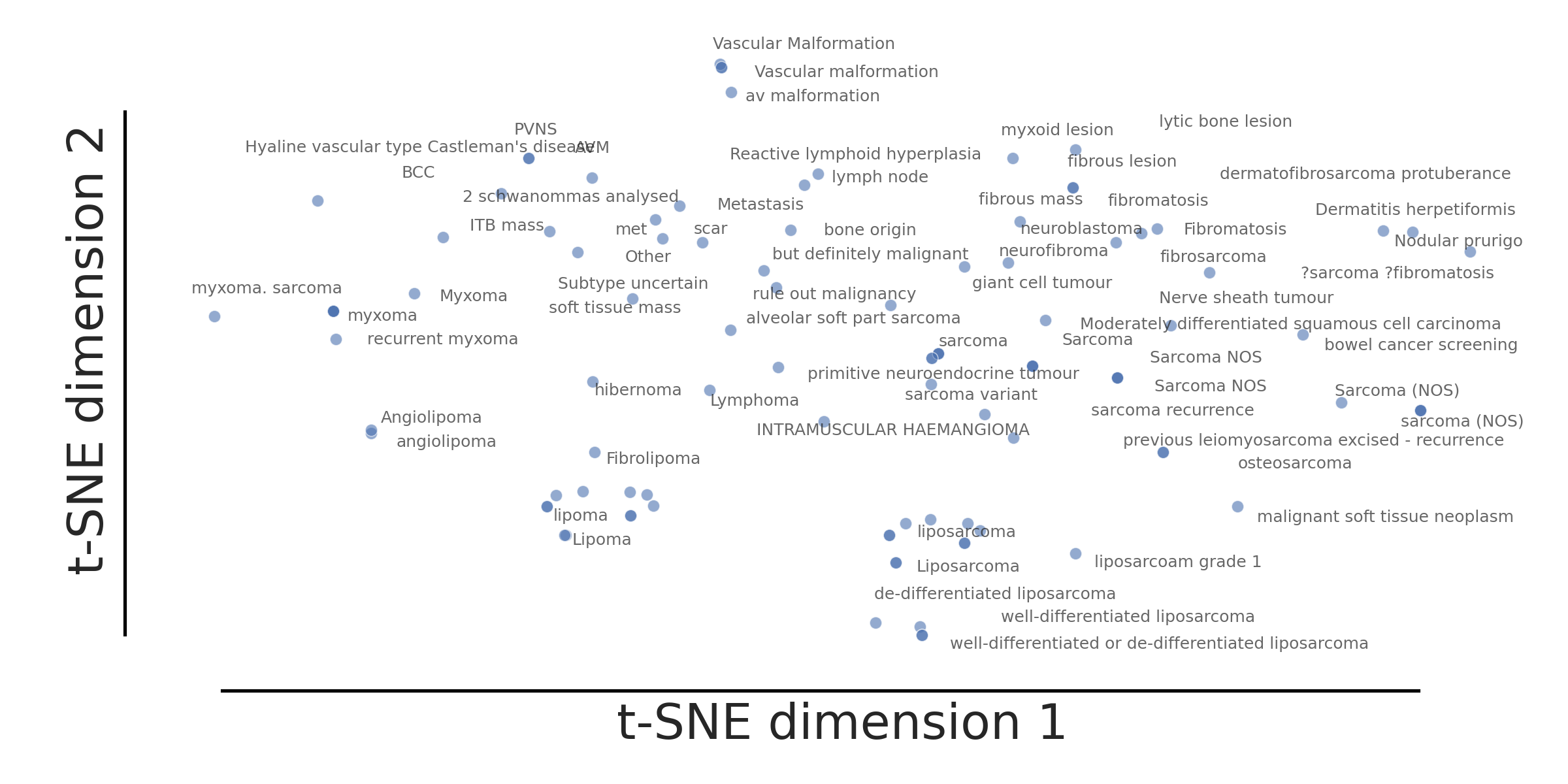}} \\
			
			Is it a soft tissue tumour? & Indicates if the specimen is a soft tissue tumour & Binary (balanced accuracy) & Yes, No, Not specified & 0.71 &
			\raisebox{-\totalheight}{\includegraphics[width=\linewidth]{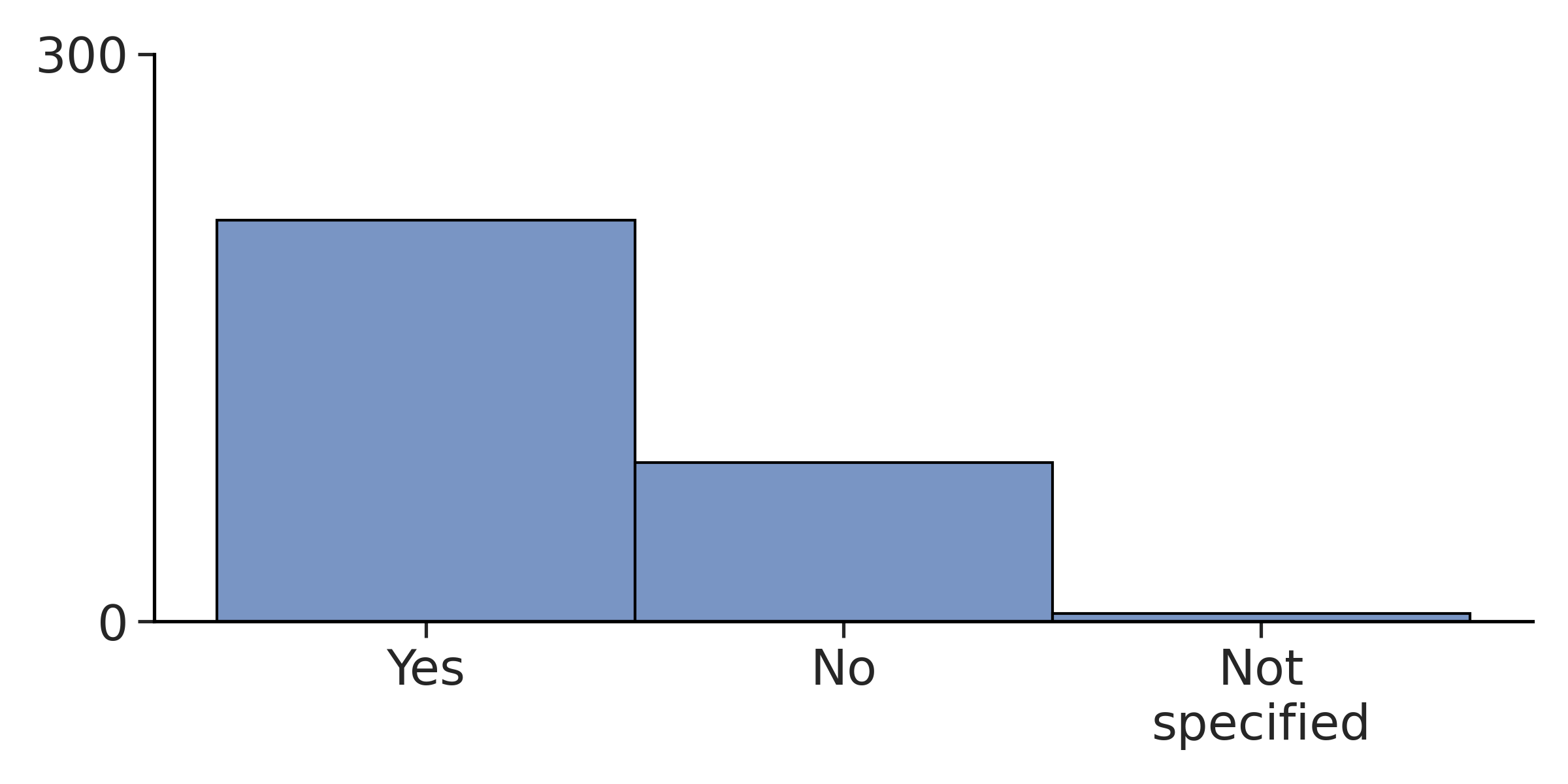}} \\
			
			Is it suspected or confirmed? & Diagnostic certainty level & Categorical (balanced accuracy) & Suspected, Confirmed, Not specified & 0.72 &
			\raisebox{-\totalheight}{\includegraphics[width=\linewidth]{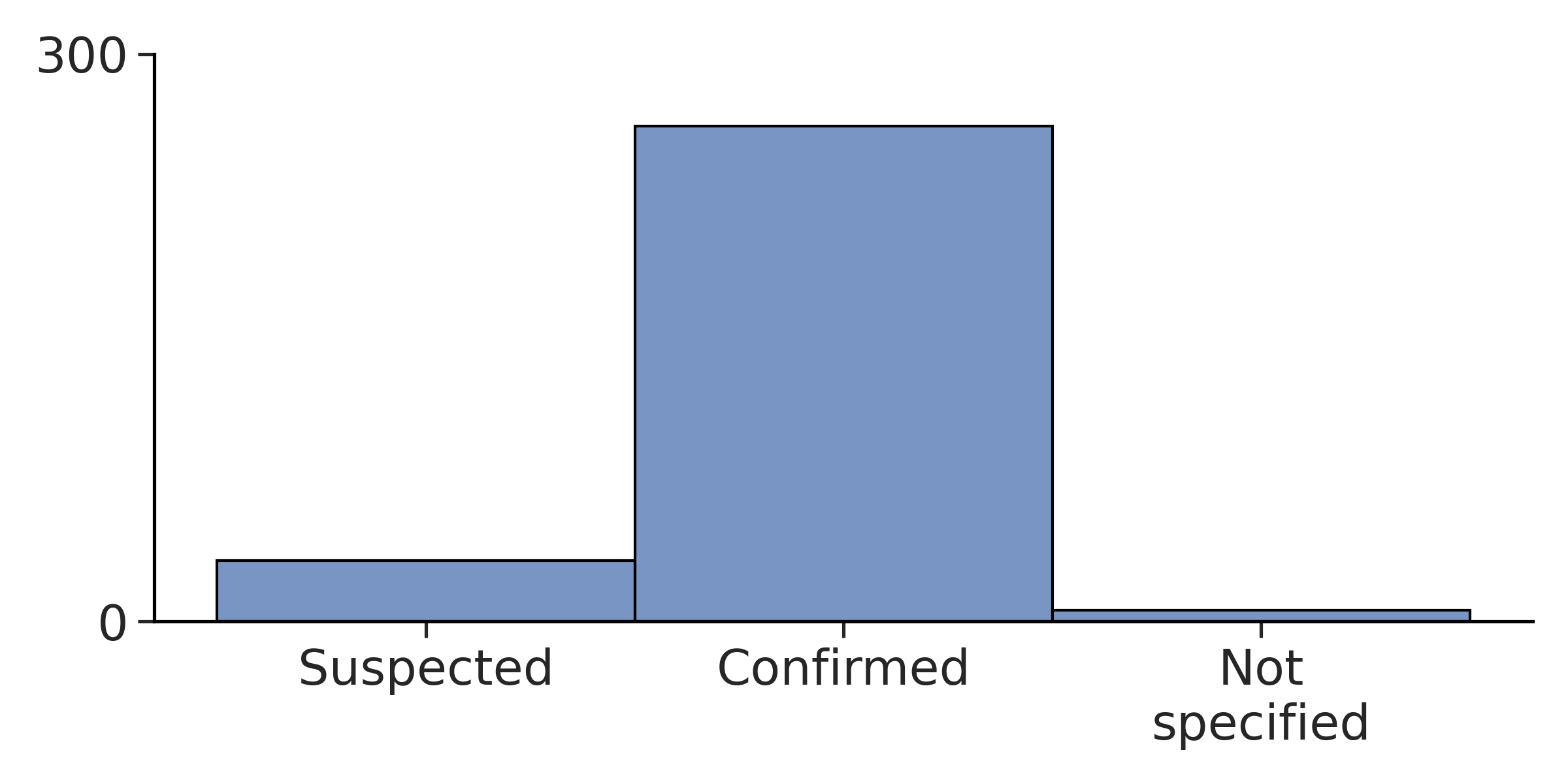}} \\
			
			Is it benign or malignant? &  behavior classification & Categorical (balanced accuracy) & Benign, Intermediate, Malignant, Not specified & 0.85 &
			\raisebox{-\totalheight}{\includegraphics[width=\linewidth]{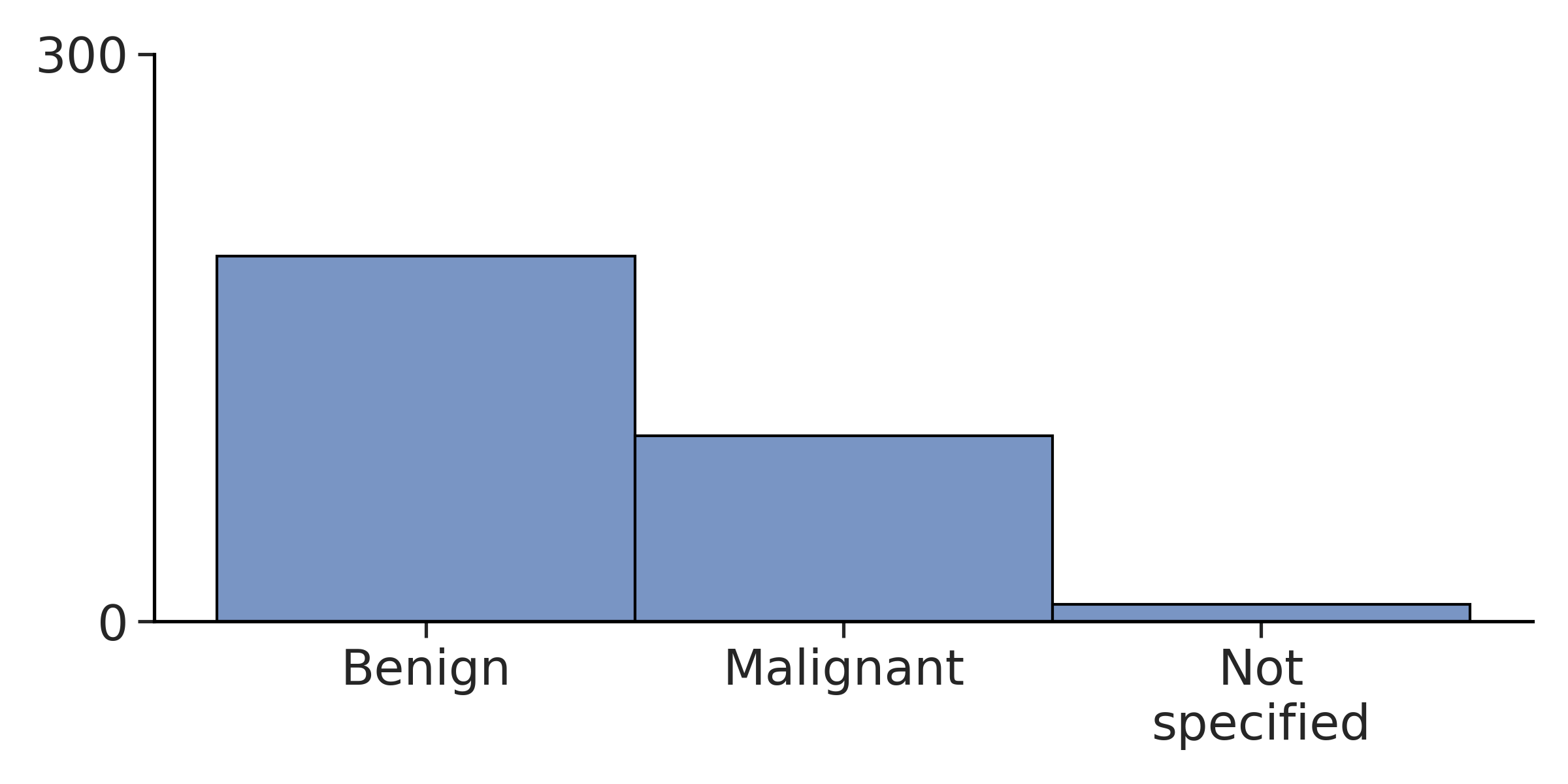}} \\
			
			grade & Histological grading of the  & Categorical (balanced accuracy) & G1, G2, G3, Low-grade, High-grade, No grade, Not specified & 0.73 &
			\raisebox{-\totalheight}{\includegraphics[width=\linewidth]{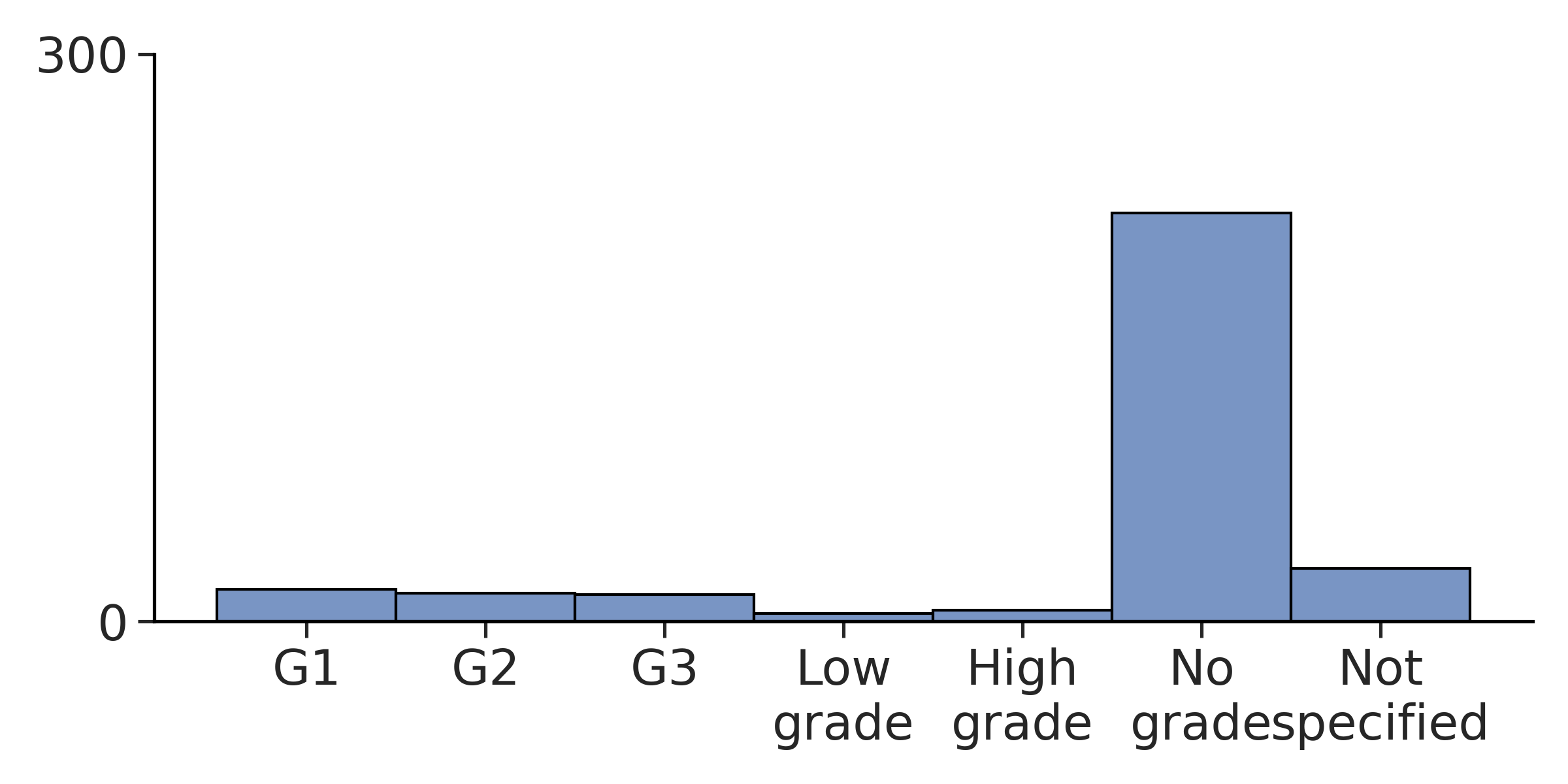}} \\
			
			Tumour location & Anatomical location of the tumour & Free-text (cosine similarity) & - & 0.92 &
			\raisebox{-\totalheight}{\includegraphics[width=\linewidth]{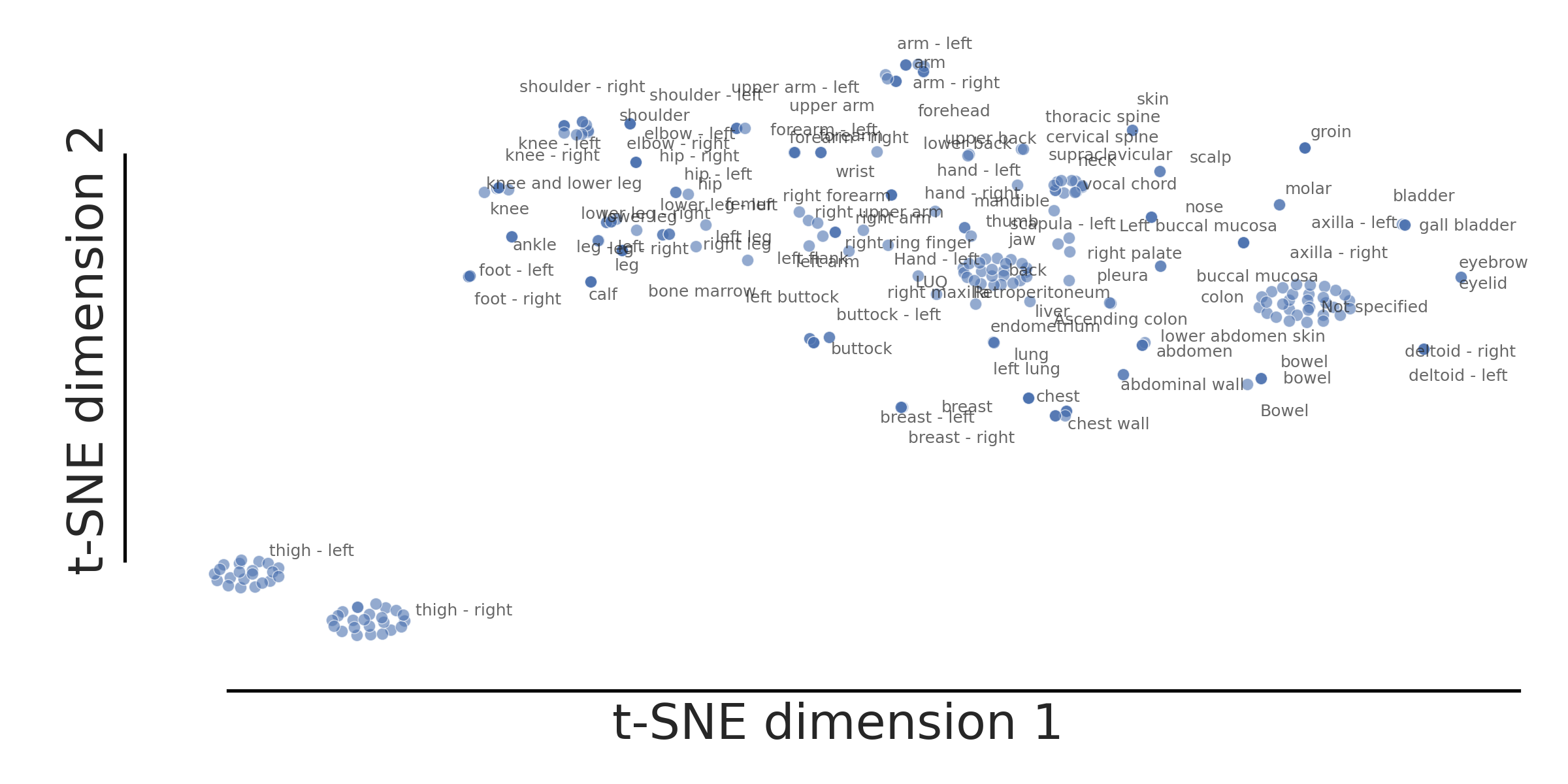}} \\
			
		\end{longtable}
	\end{scriptsize}

	\begin{scriptsize}
		\setlength{\tabcolsep}{4pt}
		\begin{longtable}{@{}
				>{\scriptsize\raggedright\arraybackslash}p{0.10\textwidth} % Variable name
				>{\scriptsize\raggedright\arraybackslash}p{0.15\textwidth} % Description
				>{\scriptsize\raggedright\arraybackslash}p{0.10\textwidth} % Variable type
				>{\scriptsize\raggedright\arraybackslash}p{0.11\textwidth} % Variable options
				>{\scriptsize\raggedleft\arraybackslash}p{0.11\textwidth} % Annotator Agreement
				>{\scriptsize\centering\arraybackslash}p{0.30\textwidth} @{}} % Distribution with image
			\multicolumn{6}{c}{\parbox{\textwidth}{
					\normalsize \tablename~\thetable{} -- Soft tissue tumours pathology report variable definitions with reference standard distribution for the 327 Dutch reports.\\}} \\
			\toprule
			\textbf{Variable Name} & 
			\textbf{Description} & 
			\textbf{Type (Metric)} & 
			\textbf{Variable Options} & 
			\textbf{Inter-rater Agreement} & 
			\textbf{Reference Standard Distribution} \\
			\midrule
			\endfirsthead
			
			\multicolumn{6}{c}{\parbox{\textwidth}{
					\normalsize \tablename~\thetable{} -- Continued\\}} \\
			\toprule
			\textbf{Variable Name} & 
			\textbf{Description} & 
			\textbf{Type (Metric)} & 
			\textbf{Variable Options} & 
			\textbf{Inter-rater Agreement} & 
			\textbf{Reference Standard Distribution} \\
			\midrule
			\endhead
			
			\midrule
			\endfoot
			
			\bottomrule
			\endlastfoot
			
			\phantomlabel{tab:use-case-STT-Dutch}
			Specimen type & Type of tissue specimen obtained for analysis & Categorical (balanced accuracy) & Biopsy, Resection, Cytology, Other, Not specified & 0.41 &
			\raisebox{-\totalheight}{\includegraphics[width=\linewidth]{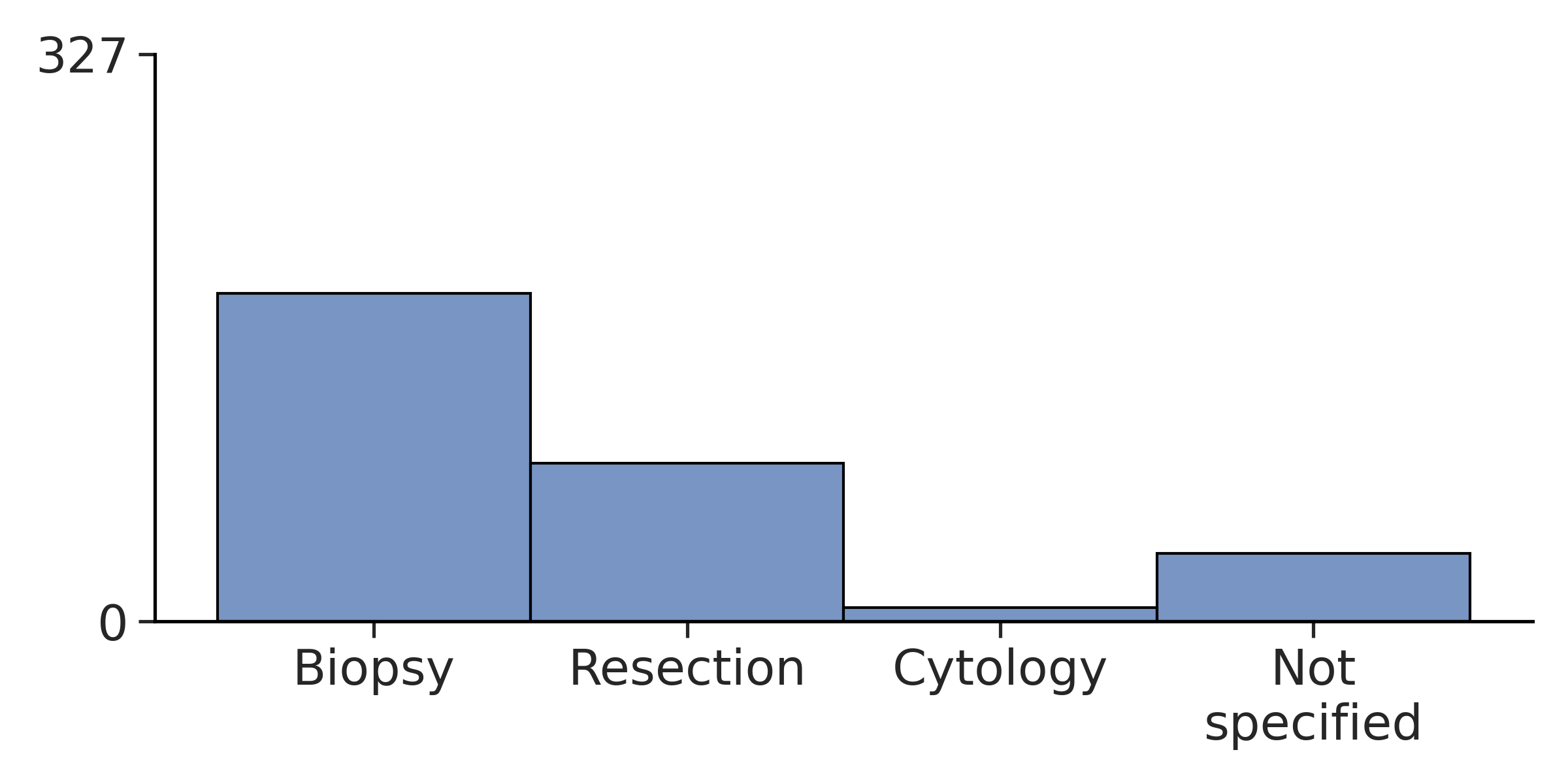}} \\
			
			Pathology request reason & Reason for pathology examination request & Categorical (balanced accuracy) & Suspicion of tumour, Metastasis, Follow-up, Recurrent tumour, Reassessment, Not specified & 0.47 &
			\raisebox{-\totalheight}{\includegraphics[width=\linewidth]{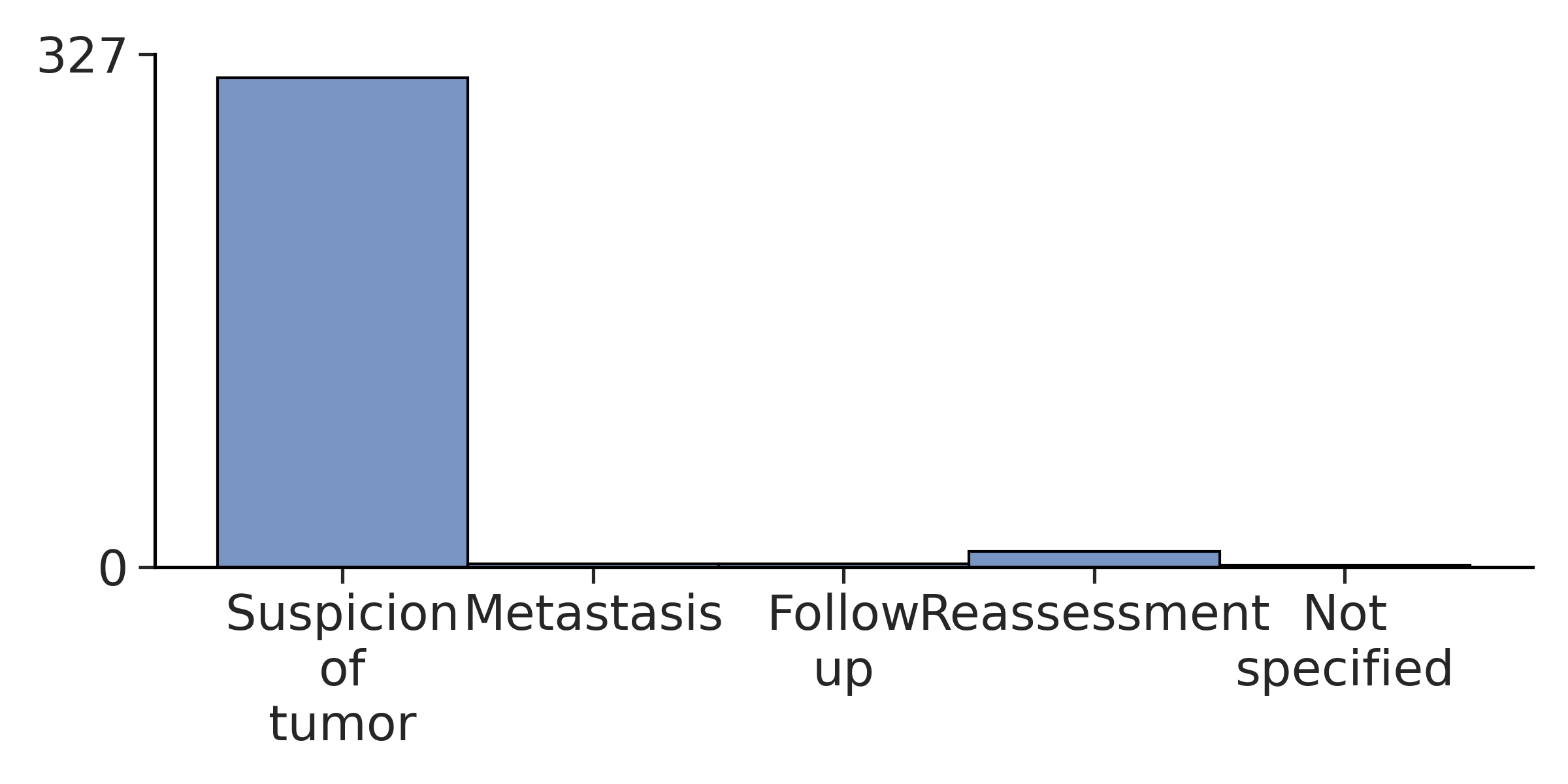}} \\
			
			Disease type & Specific disease classification based on UMLS & Free-text (cosine similarity) & - & 0.91 &
			\raisebox{-\totalheight}{\includegraphics[width=\linewidth]{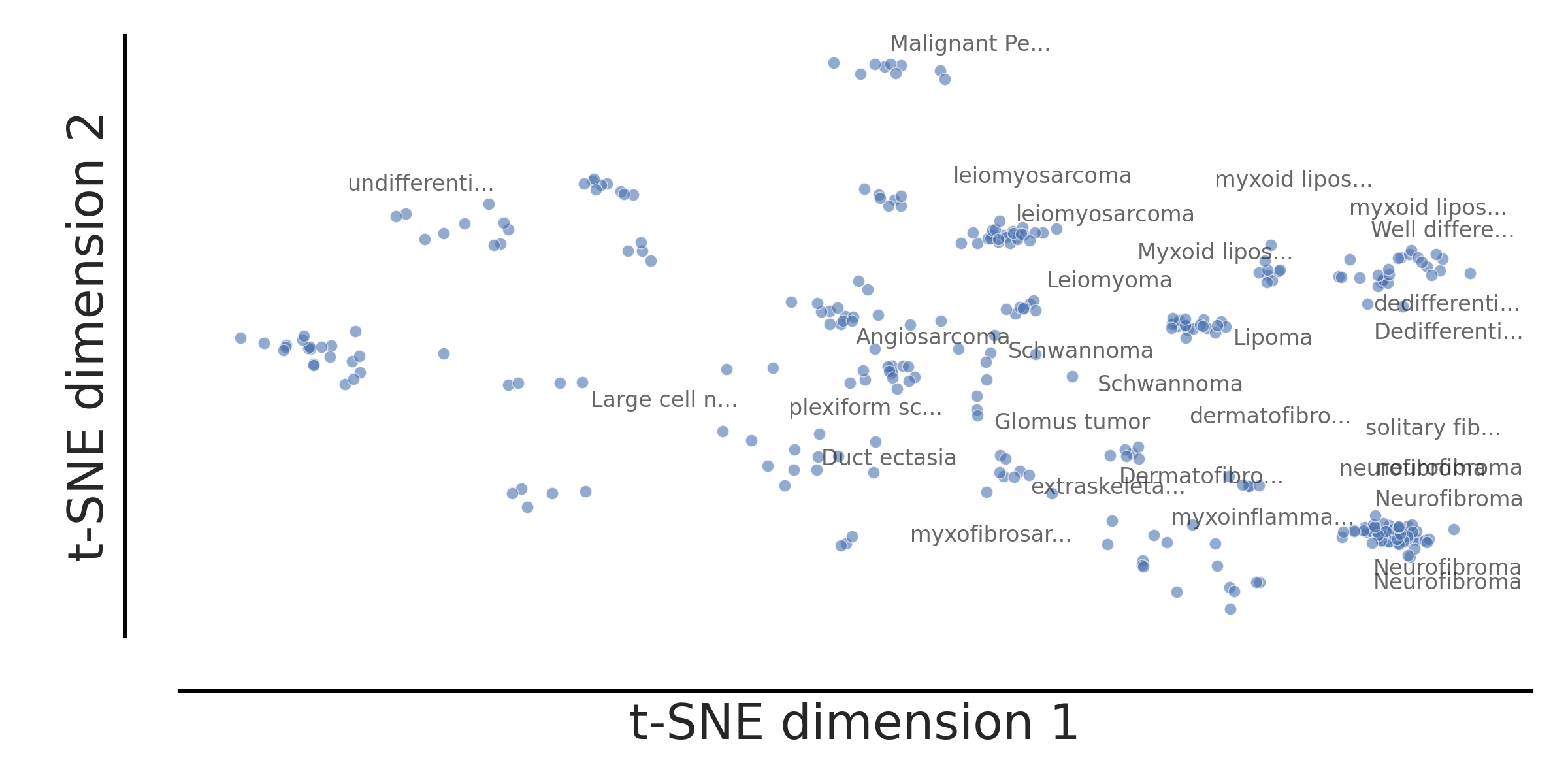}} \\
			
			Disease type differential diagnosis & Array of possible differential diagnoses & List (symmetric similarity) & - & 0.86 &
			\raisebox{-\totalheight}{\includegraphics[width=\linewidth]{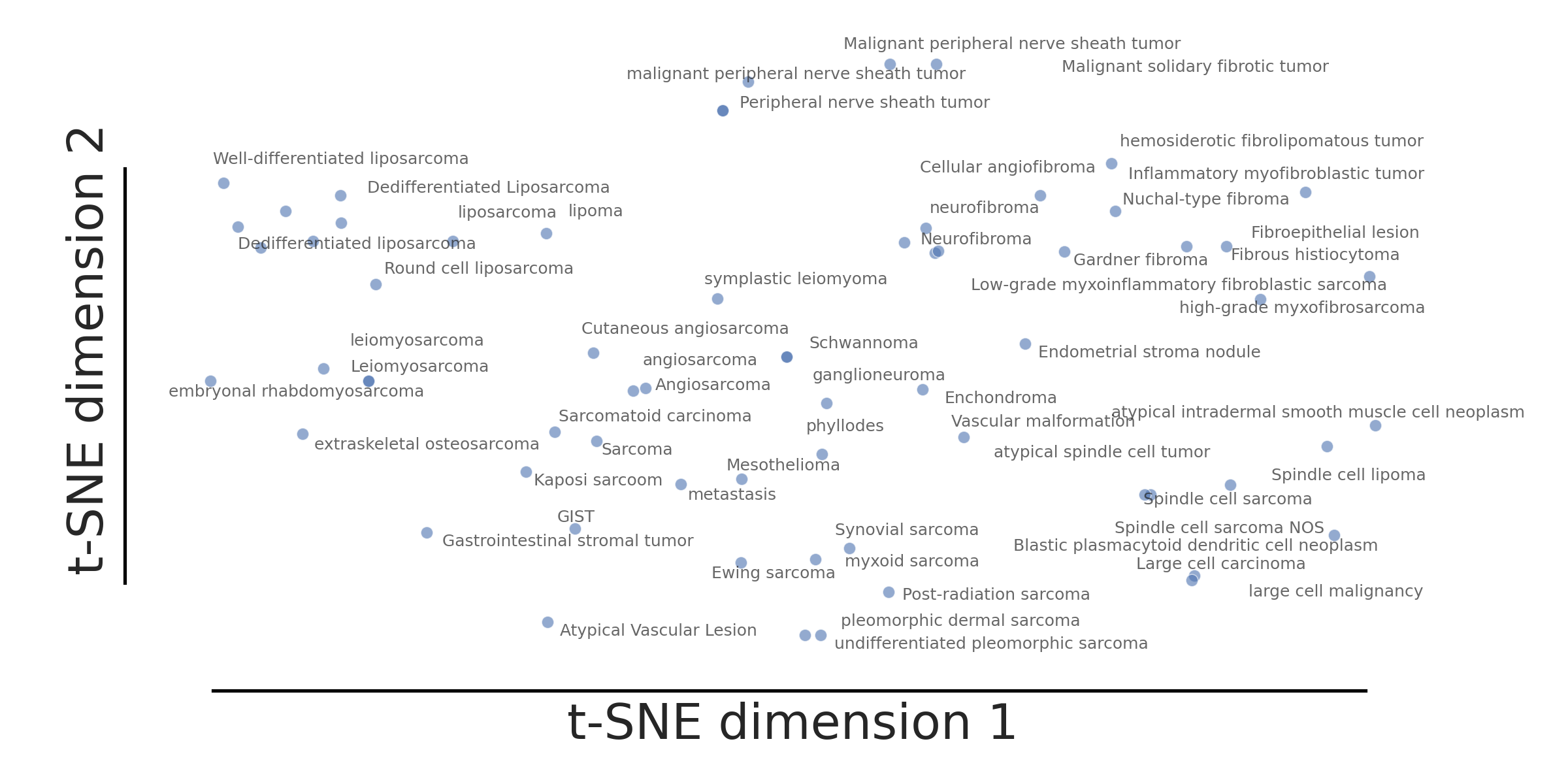}} \\
			
			Is it a soft tissue tumour? & Indicates if the specimen is a soft tissue tumour & Binary (balanced accuracy) & Yes, No, Not specified & 0.52 &
			\raisebox{-\totalheight}{\includegraphics[width=\linewidth]{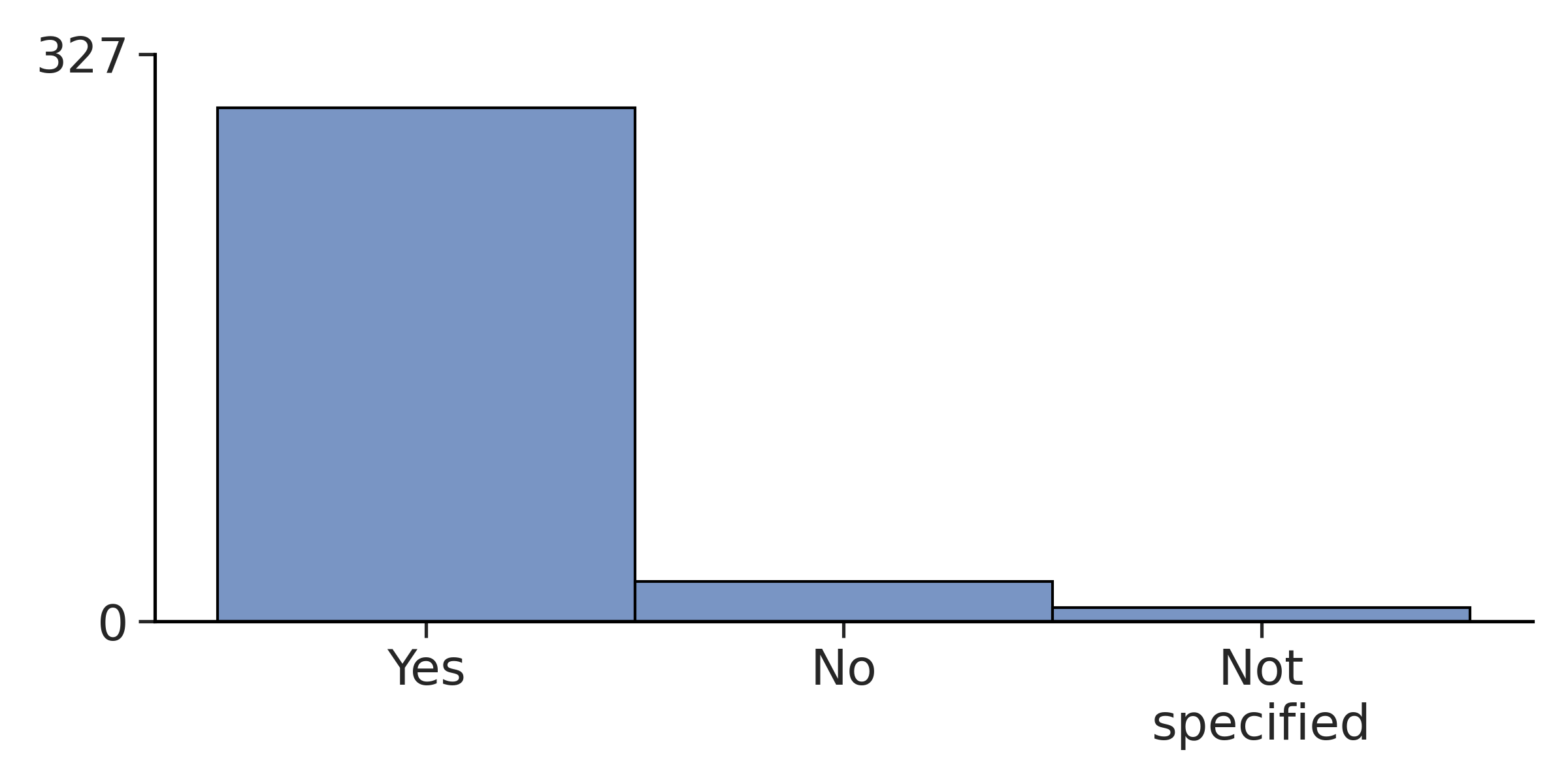}} \\
			
			Is it suspected or confirmed? & Diagnostic certainty level & Categorical (balanced accuracy) & Suspected, Confirmed, Not specified & 0.69 &
			\raisebox{-\totalheight}{\includegraphics[width=\linewidth]{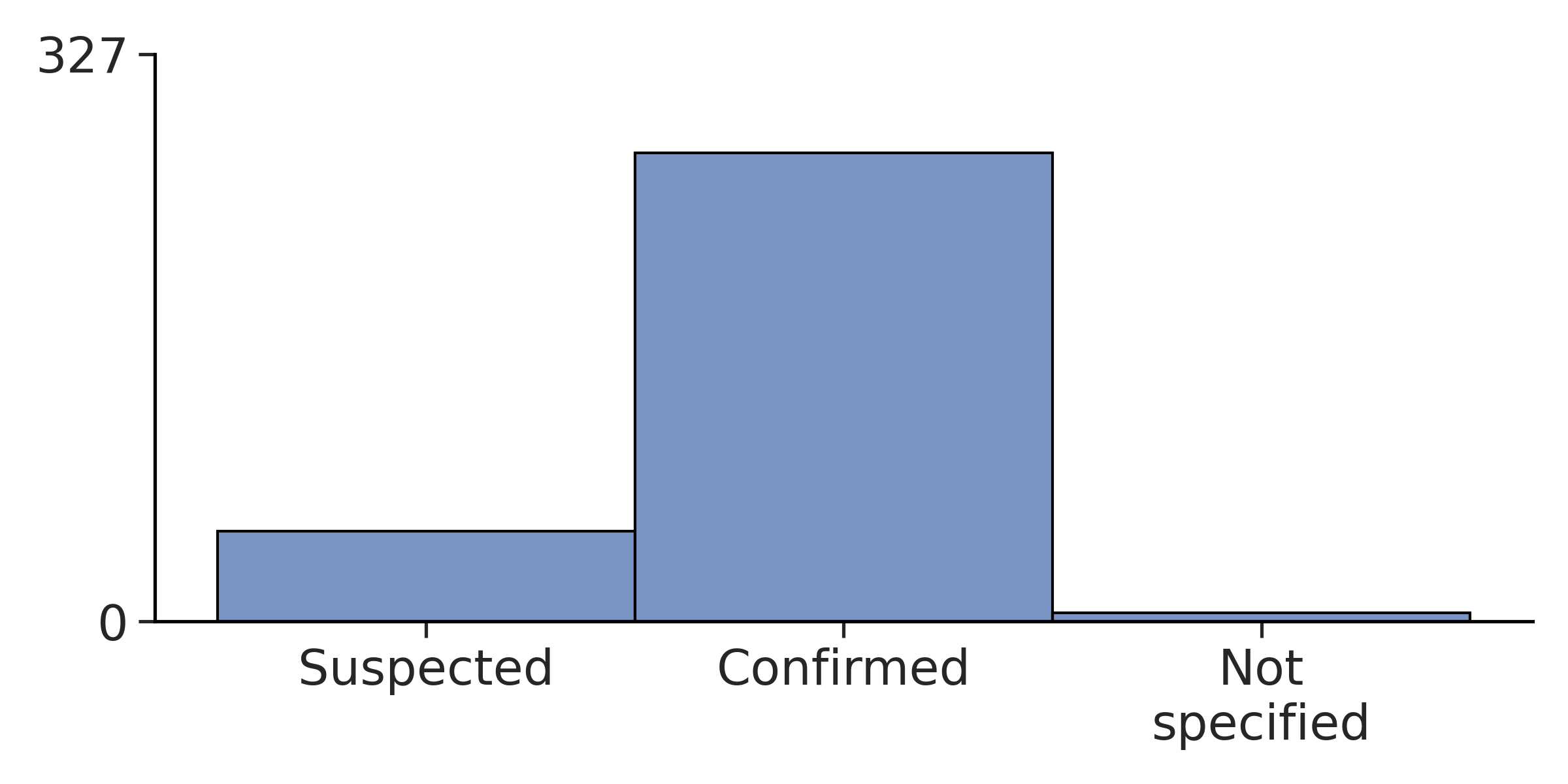}} \\
			
			Is it benign or malignant? & Tumour behavior classification & Categorical (balanced accuracy) & Benign, Intermediate, Malignant, Not specified & 0.61 &
			\raisebox{-\totalheight}{\includegraphics[width=\linewidth]{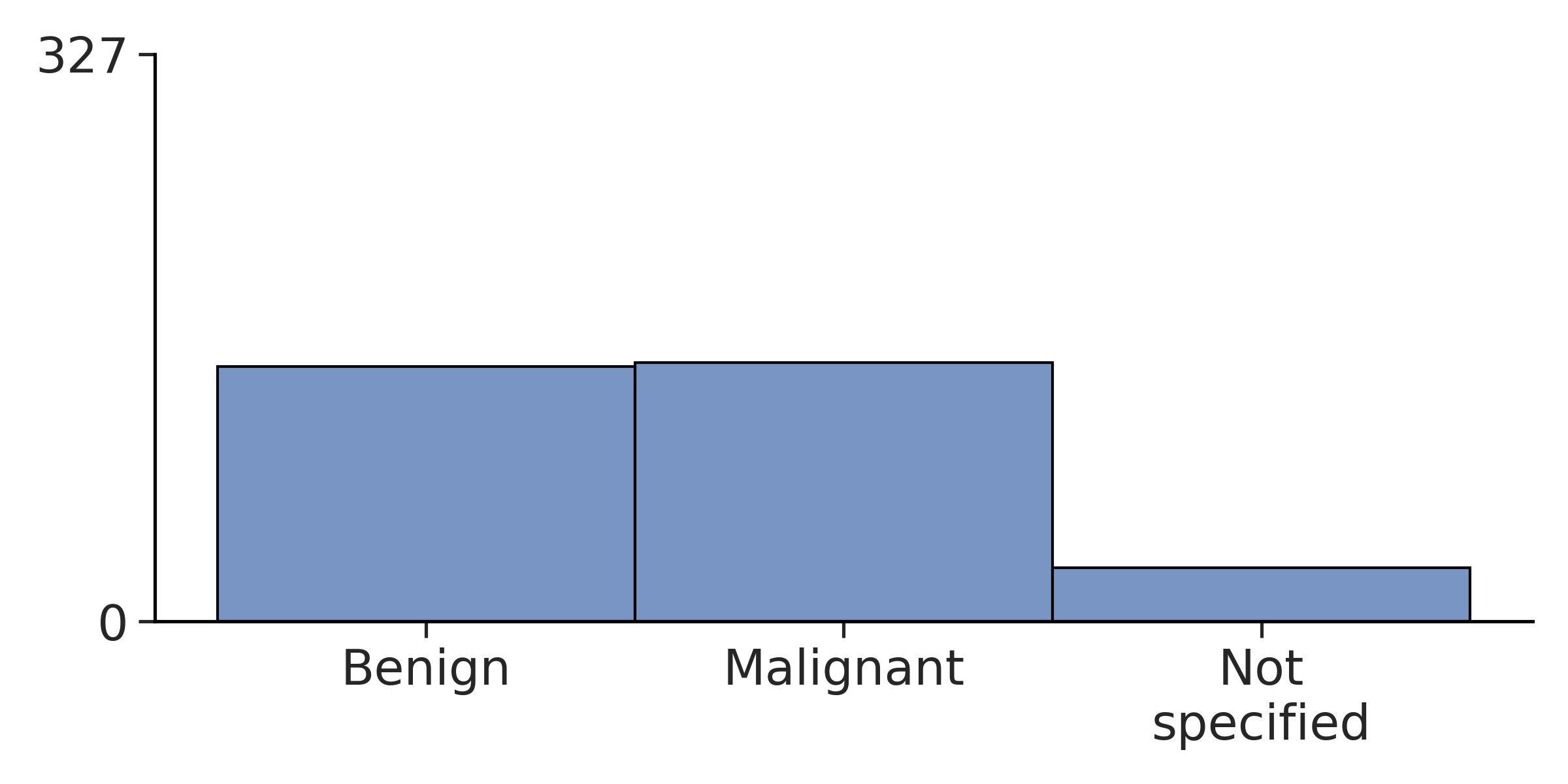}} \\
			
			Tumour grade & Histological grading of the tumour & Categorical (balanced accuracy) & G1, G2, G3, Low-grade, High-grade, No grade, Not specified & 0.72 &
			\raisebox{-\totalheight}{\includegraphics[width=\linewidth]{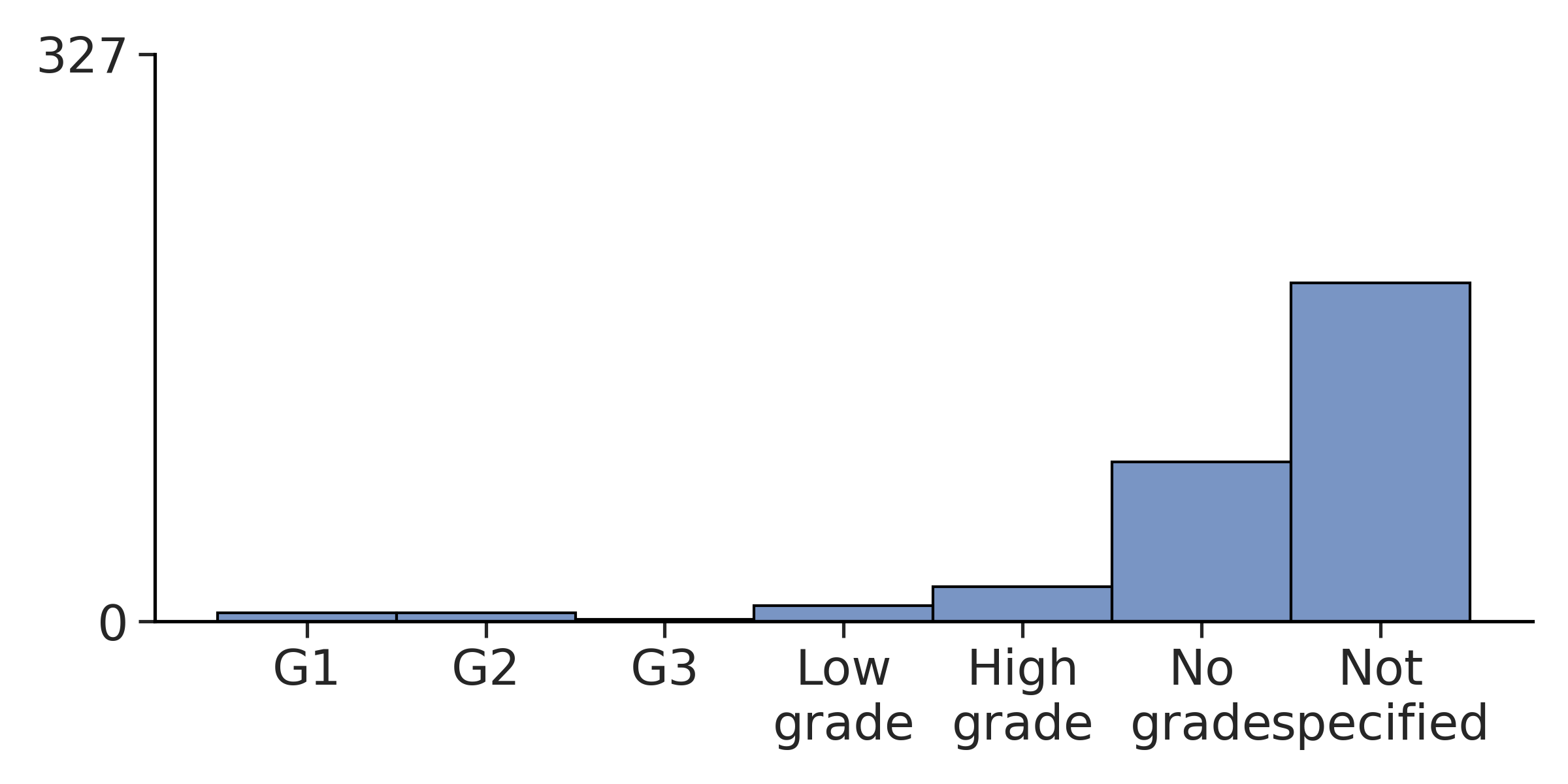}} \\
			
			Tumour location & Anatomical location of the tumour & Free-text (cosine similarity) & - & 0.89 &
			\raisebox{-\totalheight}{\includegraphics[width=\linewidth]{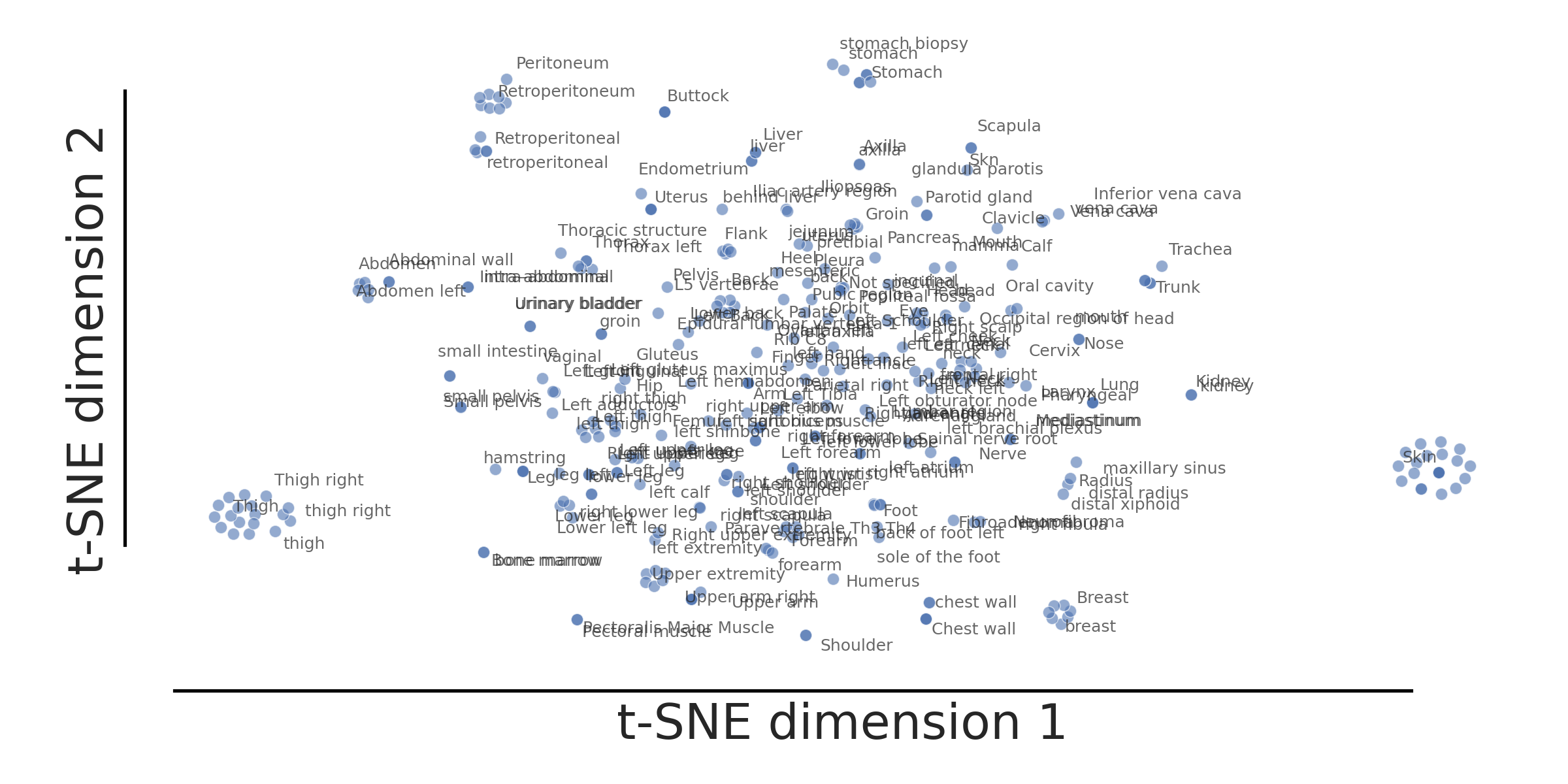}} \\
			
		\end{longtable}
	\end{scriptsize}
	
	\subsubsection{Prompt}
	For this use case, a structured prompt was created to ensure consistent extraction and enforce strict JSON formatting. 
	The prompt is divided into role-based instructions (System, Human, AI). A full example of the exact prompt used is shown in \autoref{tab:prompt-STT}.
	
	\begin{scriptsize}
		\setlength{\tabcolsep}{4pt}
		\begin{longtable}{>{\scriptsize\raggedright\arraybackslash}m{0.2\linewidth} >{\scriptsize\raggedright\arraybackslash}m{0.37\linewidth} >{\scriptsize\raggedright\arraybackslash}m{0.37\linewidth}}
			\multicolumn{3}{c}{\parbox{\textwidth}{
					\normalsize \tablename~\thetable{} -- Full structured prompt used for the soft tissue tumours (English) extraction task. The items in brackets indicate the role of the message (System, Human, AI), while the text provides the corresponding content.\\}} \\
			\toprule
			\textbf{Section} & \multicolumn{2}{c}{\textbf{Content} (Depending on prompt strategy)} \\
			& Zero-Shot, One-Shot and Few-Shot & CoT, Self-Consistency and Graph\\
			\midrule
			\endfirsthead
			
			\multicolumn{3}{c}{\parbox{\textwidth}{
					\normalsize \tablename~\thetable{} -- Continued\\}} \\
			\toprule
			\textbf{Section} & \multicolumn{2}{c}{\textbf{Content}} \\
			\midrule
			\endhead
			
			\phantomlabel{tab:prompt-STT}
			\textbf{[System] -- System instructions} & 
			You are a medical data extraction system that ONLY outputs valid JSON. Maintain strict compliance with these rules: \newline
			1. ALWAYS begin and end your response with \verb|```json| markers \newline
			2. Use EXACT field names and structure provided \newline
			3. If a value is missing or not mentioned, use the specified default for that field. \newline
			4. NEVER add commentary, explanations, or deviate from the output structure & 
			You are a medical data extraction system that performs structured reasoning before producing output. Follow these strict rules: \newline
			1. First, reason step-by-step to identify and justify each extracted field. \newline
			2. After reasoning, output ONLY valid JSON in the exact structure provided. \newline
			3. ALWAYS begin and end the final output with \verb|```json| markers — do not include reasoning within these markers. \newline
			4. Use EXACT field names and structure as specified. \newline
			5. If a value is missing or not mentioned, use the specified default for that field. \newline
			6. NEVER include commentary, explanations, or deviate from the specified format in the final JSON. \newline \\
			\midrule
			\textbf{[Human] -- Field instructions} & \multicolumn{2}{c}{\parbox{0.75\linewidth}{
					1. \opus{"Specimen Type"}: \newline
					- Type: string \newline
					- Options: [\opus{"Biopsy"}, \opus{"Resection"}, \opus{"Cytology"}, \opus{"Other"}] \newline
					- Default: \opus{"Not specified"} \newline
					2. \opus{"Pathology Request Reason"}: \newline
					- Type: string \newline
					- Options: [\opus{"Suspicion of tumour"}, \opus{"Metastasis"}, \opus{"Follow-up"}, \opus{"Recurrent tumour"}, \opus{"Reassessment"}, \opus{"Not specified"}] \newline
					- Default: \opus{"Not specified"} \newline
					3. \opus{"Disease Type"}: \newline
					- Type: string \newline
					- Short name based on Unified Medical Language System (UMLS). \newline
					- Default: \opus{"Not specified"} \newline
					4. \opus{"Disease Type differential diagnosis"}: \newline
					- Type: list \newline
					- Array of standardized possible differential diagnoses (based on UMLS). \newline
					- Default: empty list (\opus{[]}) \newline
					5. \opus{"Is it a soft tissue tumour?"}: \newline
					- Type: string \newline
					- Options: [\opus{"Yes"}, \opus{"No"}] \newline
					- Default: \opus{"Not specified"} \newline
					6. \opus{"Is it suspected or confirmed?"}: \newline
					- Type: string \newline
					- Options: [\opus{"Suspected"}, \opus{"Confirmed"}] \newline
					- Default: \opus{"Not specified"} \newline
					7. \opus{"Is it benign or malignant?"}: \newline
					- Type: string \newline
					- Options: [\opus{"Benign"}, \opus{"Intermediate"}, \opus{"Malignant"}] \newline
					- Default: \opus{"Not specified"} \newline
					8. \opus{"Tumour Grade"}: \newline
					- Type: string \newline
					- If benign: always use "No grade". Preferred: FNCLCC grading (G1, G2, G3). Fallback: "Low-grade" or "High-grade" if FNCLCC unavailable. \newline
					- Options: [\opus{"G1"}, \opus{"G2"}, \opus{"G3"}, \opus{"Low-grade"}, \opus{"High-grade"}, \opus{"No grade"}, \opus{"Not specified"}] \newline
					- Default: \opus{"Not specified"} \newline
					9. \opus{"Tumour Location"}: \newline
					- Type: string \newline
					- Use a standard UMLS-based anatomical term (e.g., "Lung", "Breast"). \newline
					- Default: \opus{"Not specified"} \newline
			}} \newline \\
			\midrule
			\textbf{[Human] -- Task instructions} & \multicolumn{2}{c}{\parbox{0.75\linewidth}{
					Extract information into this exact JSON structure:
					\opus{```json} \newline
					\opus{\{} \newline
					\opus{\quad "Specimen Type": "",} \newline
					\opus{\quad "Pathology Request Reason": "",} \newline
					\opus{\quad "Disease Type": "",} \newline
					\opus{\quad "Disease Type differential diagnosis": [],} \newline
					\opus{\quad "Is it a soft tissue tumour?": "",} \newline
					\opus{\quad "Is it suspected or confirmed?": "",} \newline
					\opus{\quad "Is it benign or malignant?": "",} \newline
					\opus{\quad "Tumour Grade": "",} \newline
					\opus{\quad "Tumour Location": "",} \newline
					\opus{\}} \newline
					\opus{```}%
				}%
			} \newline \\
			\midrule
			\textbf{[Human] -- Example intro} & \multicolumn{2}{c}{\parbox{0.75\linewidth}{Below are 1 example of expected input and output, followed by a new task.}} \newline \\
			\midrule
			\textbf{[Human] -- Example user} & \multicolumn{2}{c}{\parbox{0.75\linewidth}{\opus{[...CENSORED...]}}} \newline \\
			\midrule
			\textbf{[AI] -- Example assistant reasoning} & - & 
			- \opus{Specimen Type} - "Method of acquisition: biopsy" implies "Biopsy" \newline
			- \opus{Pathology Request Reason} - "Clinical question: metastasis or sarcoma?" implies "Metastasis" \newline
			- \opus{Disease Type} - "sarcoma with myxoid stroma" implies "Myxoid sarcoma" \newline
			- \opus{Disease Type differential diagnosis} - "myxoinflammatory fibroblastic sarcoma, myxofibrosarcoma, pleomorphic undifferentiated" implies ["Myxoinflammatory fibroblastic sarcoma", "Myxofibrosarcoma", "Pleomorphic undifferentiated sarcoma"] \newline
			- \opus{Is it a soft tissue tumour?} - "sarcoma arises in soft tissue" implies "Yes" \newline
			- \opus{Is it suspected or confirmed?} - "differential diagnosis stated" implies "Suspected" \newline
			- \opus{Is it benign or malignant?} - "sarcoma, necrosis and mitotic activity present" implies "Malignant" \newline
			- \opus{Tumour Grade} - "not provided" implies "Not specified" \newline
			- \opus{Tumour Location} - "intramuscular right thigh" maps to "Right leg" \newline \\
			\midrule
			\textbf{[AI] -- Example assistant output} & \multicolumn{2}{c}{\parbox{0.75\linewidth}{
					\opus{```json} \newline
					\opus{\{} \newline
					\opus{\quad "Specimen Type": "Biopsy",} \newline
					\opus{\quad "Pathology Request Reason": "Metastasis",} \newline
					\opus{\quad "Disease Type": "Myxoid sarcoma",} \newline
					\opus{\quad "Disease Type differential diagnosis": ["Myxoinflammatory fibroblastic sarcoma", "Myxofibrosarcoma", "Pleomorphic undifferentiated sarcoma"],} \newline
					\opus{\quad "Is it a soft tissue tumour?": "Yes",} \newline
					\opus{\quad "Is it suspected or confirmed?": "Suspected",} \newline
					\opus{\quad "Is it benign or malignant?": "Malignant",} \newline
					\opus{\quad "Tumour Grade": "Not specified",} \newline
					\opus{\quad "Tumour Location": "Right leg",} \newline
					\opus{\}} \newline
					\opus{```}%
				}%
			} \newline \\
			\midrule
			\textbf{[Human] -- Report instructions} & \multicolumn{2}{c}{\parbox{0.75\linewidth}{[file name]: \opus{[...CENSORED...]} \newline \opus{[...CENSORED...]}}} \newline \\
			\midrule
			\textbf{[Human] -- Final instructions} &
			Begin the extraction now. Your response must contain only a single valid JSON block, enclosed in triple backticks and prefixed with \verb|`json`|, like this: \verb|```json  ... ```|& 
			Begin the extraction now. First, reason step-by-step to identify and justify the value for each required field, enclosed within \verb|<think>...</think>| tags. Then, output only the final structured data as a single valid JSON block, starting with \verb|```json| and ending with \verb|```|."
			\\
			\bottomrule
		\end{longtable}
	\end{scriptsize}
	
	\subsubsection{Prompt Graph}
	The dependencies, conditional branches, and extraction order of variables are represented as a directed acyclic graph. This graph reflects how the extraction task is decomposed into smaller, sequential subtasks for the Prompt Graph prompting strategy. 
	\begin{figure}[htbp]
		\centering
		\includegraphics[width=\linewidth, height=0.5\textheight, keepaspectratio]{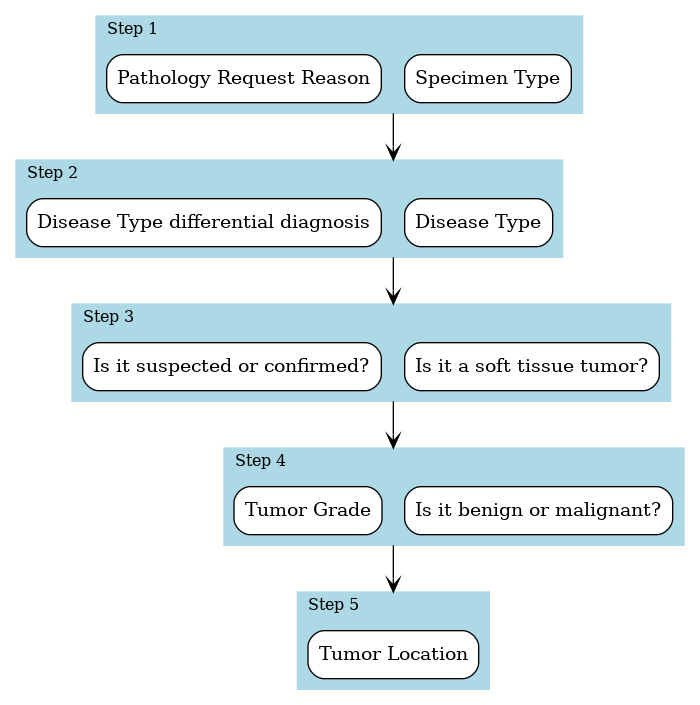}
		\caption{Directed acyclic graph showing sequential extraction order of variable extraction for soft tissue tumours use case.}
	\end{figure}
	
	\newpage
	\subsection{Use Case - Melanomas (Dutch)}
	
	\subsubsection{Overview}
	
	This use case focuses on the extraction of structured information from pathology reports of patients diagnosed with melanoma. Specifically, the study targets the identification of pathological features of primary melanoma, sentinel lymph node biopsy, and recurrence data (\autoref{tab:use-case-Melanoma}).
	
	\subsubsection{Inclusion and Exclusion Criteria}
	
	All pathology reports from patients with a confirmed pathology diagnosis of melanoma between 2002 and 2023 were included. A total of 308,213 pathology reports were collected. From these, 500 patients were randomly selected, resulting in 1,252 pathology reports included in the study.
	
	\subsubsection{Annotation}
	
	All reports were annotated by a medical student, with a random subset (120) independently reviewed by a PhD student for quality control.
	
	\subsubsection{Ethical Considerations and Funding}
	
	The study protocol was approved by the Medical Ethics Committee of the Erasmus Medical Centre (MEC 2024–0206), which waived the requirement for informed consent. This research did not receive any specific funding from public, commercial, or not-for-profit agencies.
	
	\begin{scriptsize}
		\setlength{\tabcolsep}{4pt}
		\begin{longtable}{@{}
				>{\scriptsize\raggedright\arraybackslash}p{0.10\textwidth} % Variable name
				>{\scriptsize\raggedright\arraybackslash}p{0.15\textwidth} % Description
				>{\scriptsize\raggedright\arraybackslash}p{0.10\textwidth} % Variable type
				>{\scriptsize\raggedright\arraybackslash}p{0.11\textwidth} % Variable options
				>{\scriptsize\raggedleft\arraybackslash}p{0.11\textwidth} % Annotator Agreement
				>{\scriptsize\centering\arraybackslash}p{0.30\textwidth} @{}} % Distribution with image
			\multicolumn{6}{c}{\parbox{\textwidth}{
					\normalsize \tablename~\thetable{} -- Melanomas Dutch pathology report variable definitions with reference standard distributions.\\}} \\
			\toprule
			\textbf{Variable Name} & 
			\textbf{Description} & 
			\textbf{Type (Metric)} & 
			\textbf{Variable Options} & 
			\textbf{Annotator Agreement} & 
			\textbf{Reference Standard Distribution} \\
			\midrule
			\endfirsthead
			
			\multicolumn{6}{c}{\parbox{\textwidth}{
					\normalsize \tablename~\thetable{} -- Continued\\}} \\
			\toprule
			\textbf{Variable Name} & 
			\textbf{Description} & 
			\textbf{Type (Metric)} & 
			\textbf{Variable Options} & 
			\textbf{Annotator Agreement} & 
			\textbf{Reference Standard Distribution} \\
			\midrule
			\endhead
			
			\midrule
			\endfoot
			
			\bottomrule
			\endlastfoot
			
			\phantomlabel{tab:use-case-Melanoma}
			Primary melanoma & Determination of whether this is a primary melanoma case & Categorical (balanced accuracy) & Yes, No, Not specified & 0.76 &  
			\raisebox{-\totalheight}{\includegraphics[width=\linewidth]{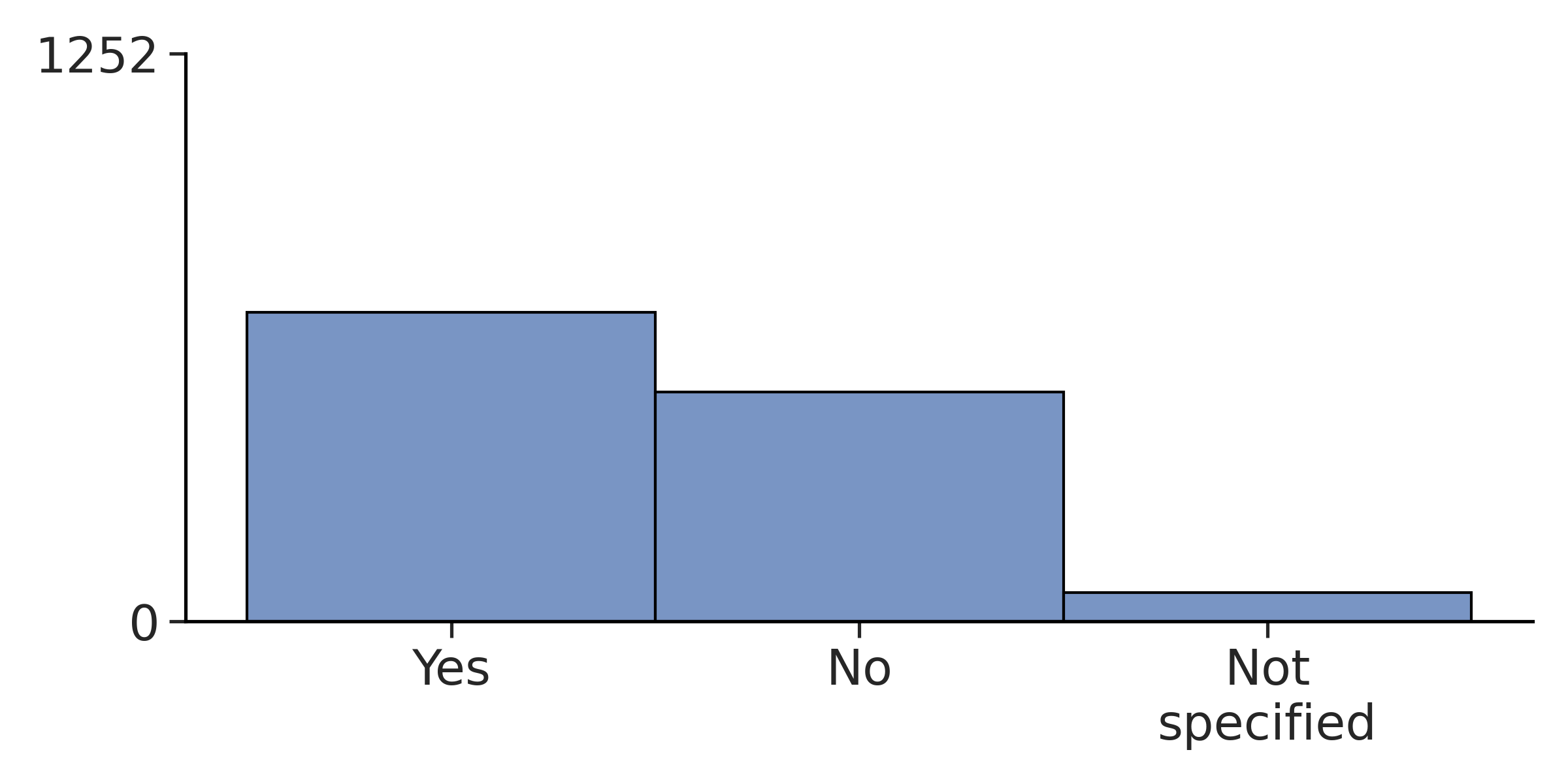}} \\
			
			Report a revision & Indicates if this report is a revision of a previous report & Categorical (balanced accuracy) & Yes, No, Not specified & 0.59 & 
			\raisebox{-\totalheight}{\includegraphics[width=\linewidth]{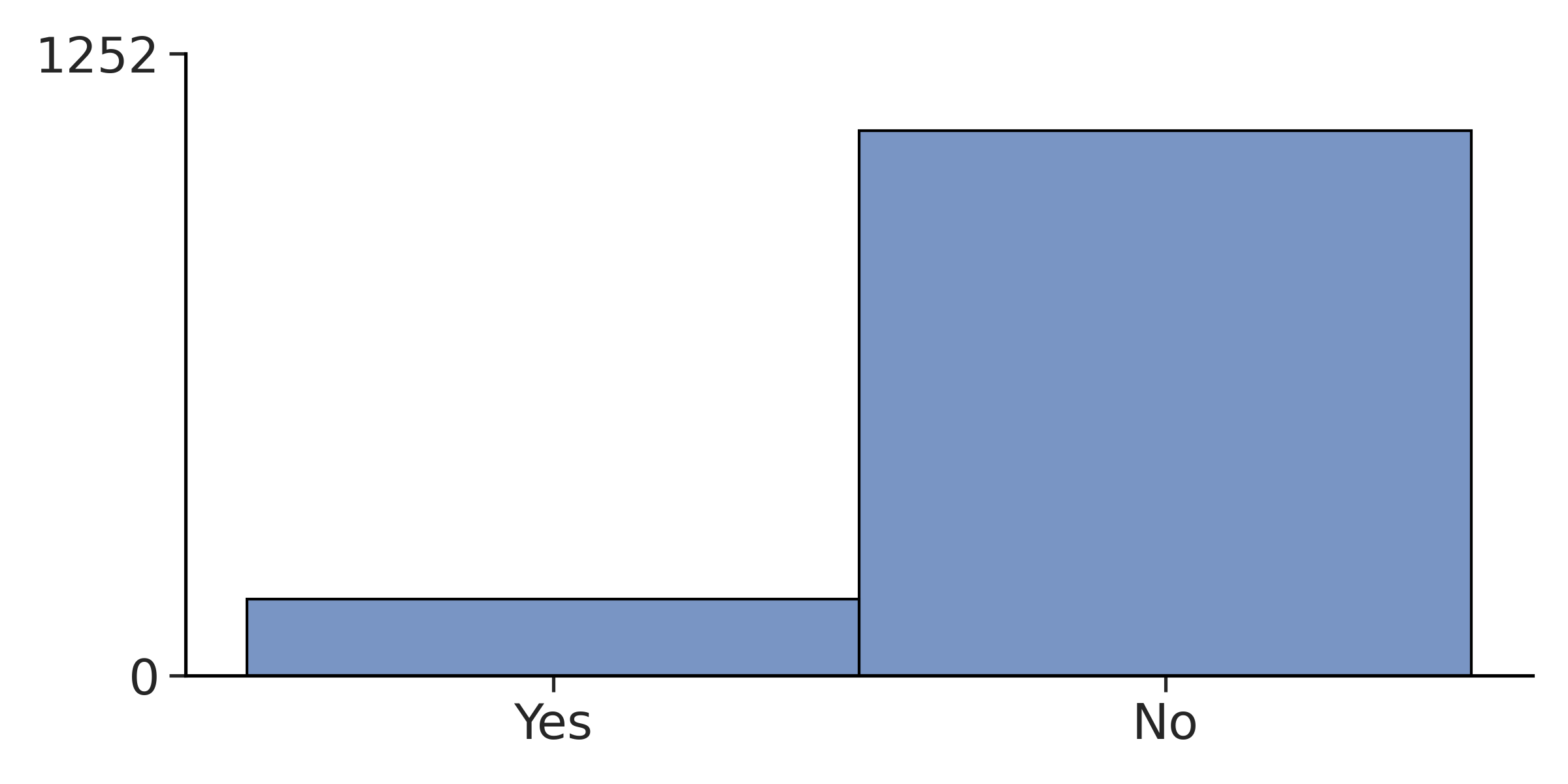}} \\
			
			Multiple primary melanoma & Indicates presence of multiple primary melanomas (only if Primary melanoma is "Yes") & Categorical (balanced accuracy) & Yes, No, Not specified & 0.80 & 
			\raisebox{-\totalheight}{\includegraphics[width=\linewidth]{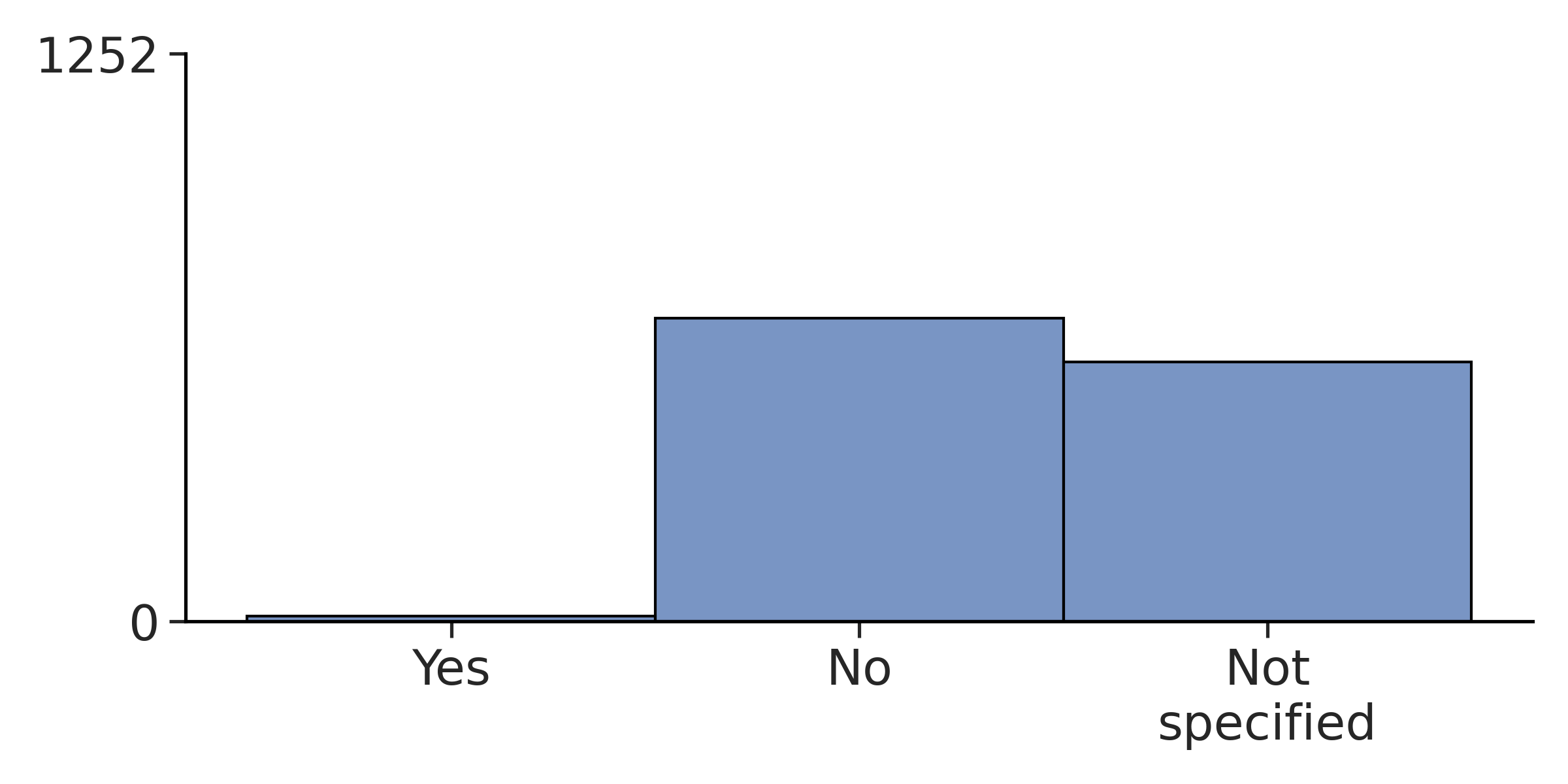}} \\
			
			In situ melanoma & Indicates if melanoma is in situ (non-invasive) & Categorical (balanced accuracy) & Yes, No, Not specified & 0.80 & 
			\raisebox{-\totalheight}{\includegraphics[width=\linewidth]{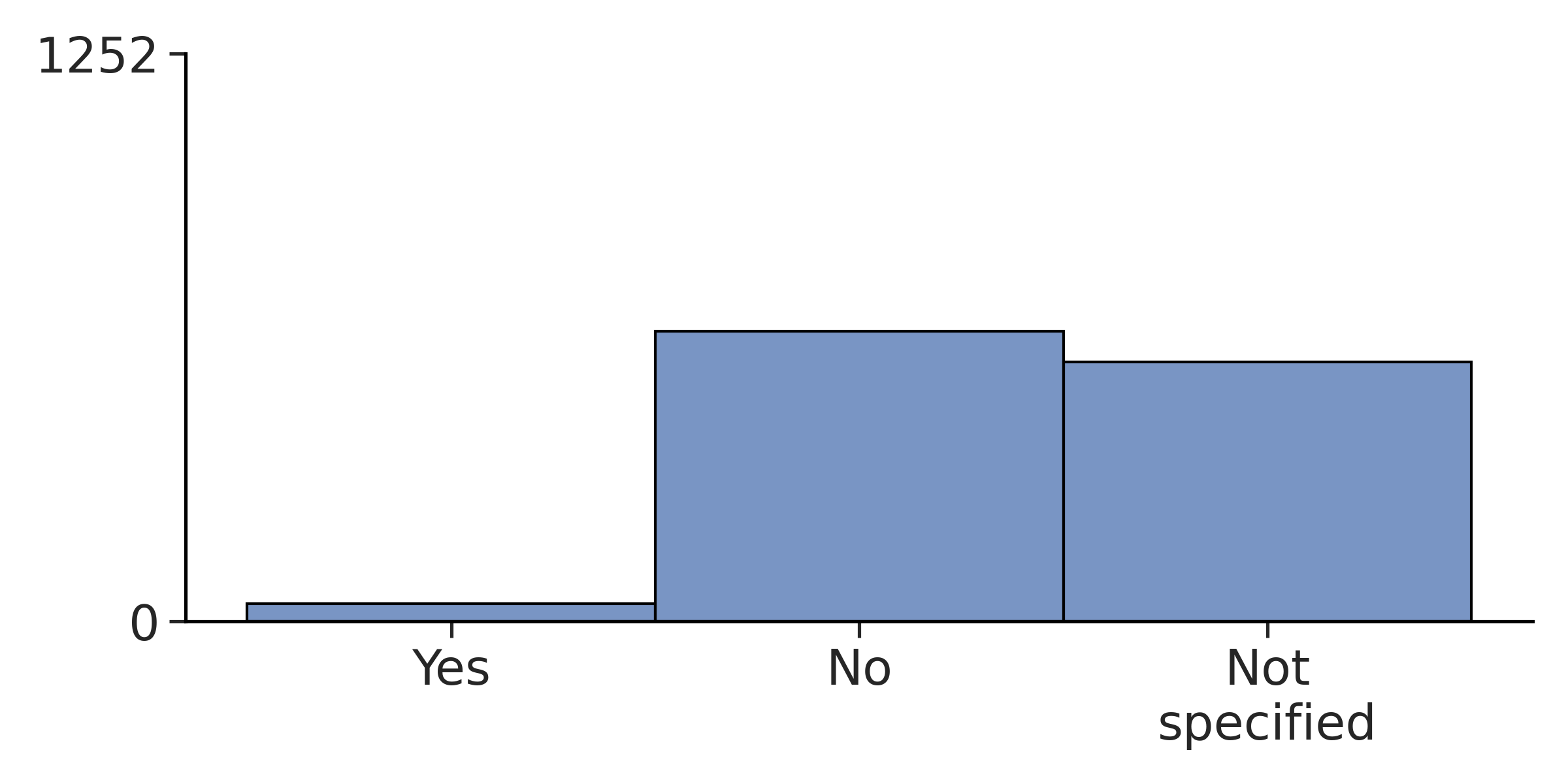}} \\
			
			Breslow thickness & Measured thickness of melanoma in millimeters (0-400) & Numeric (accuracy) & - & 0.85 & 
			\raisebox{-\totalheight}{\includegraphics[width=\linewidth]{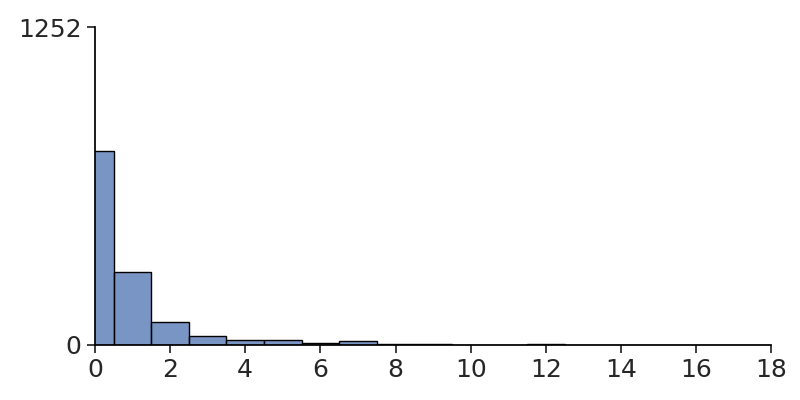}} \\
			
			Ulceration & Presence or absence of ulceration & Categorical (balanced accuracy) & Yes, No, Not specified & 0.80 & 
			\raisebox{-\totalheight}{\includegraphics[width=\linewidth]{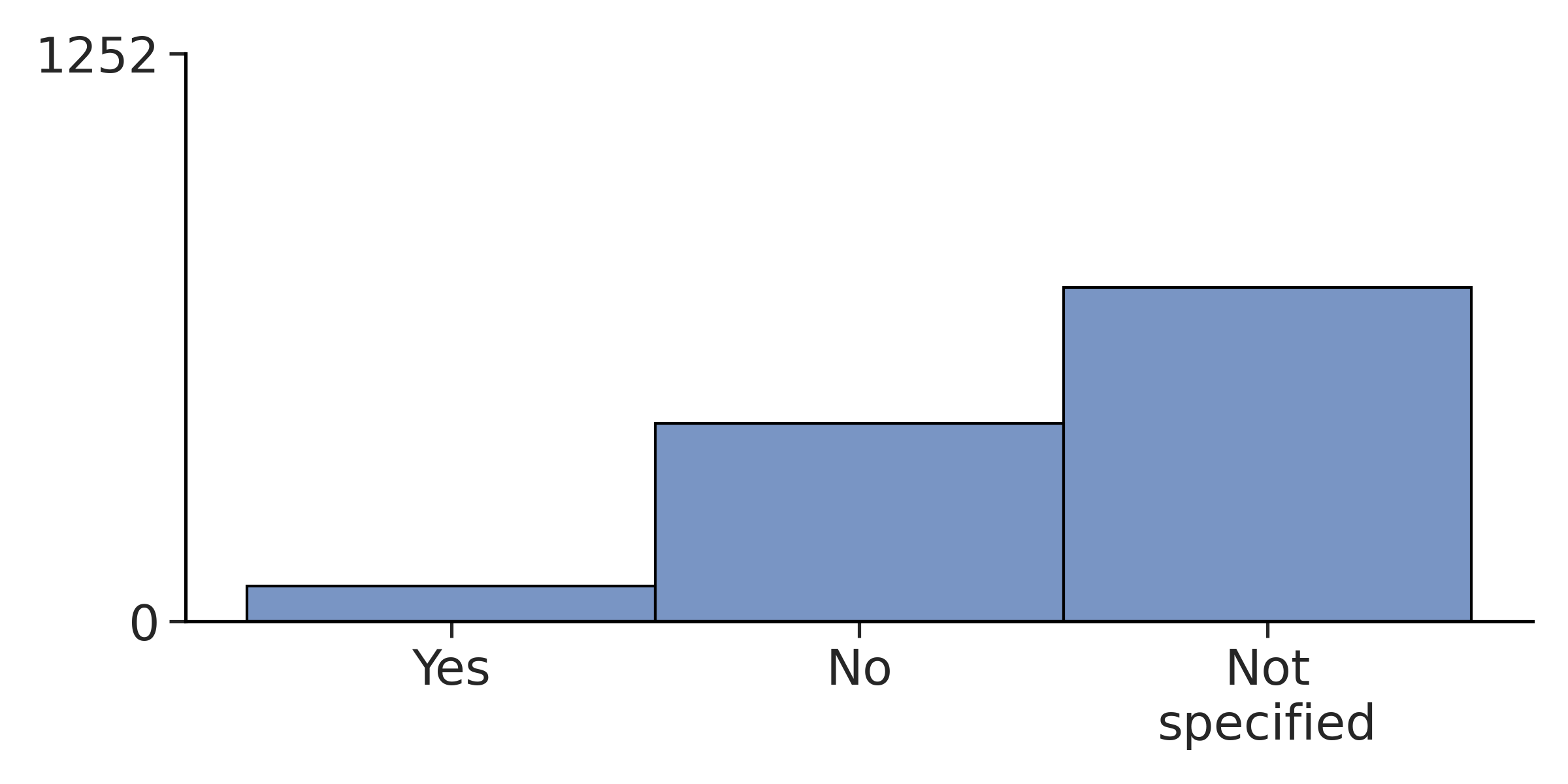}} \\
			
		Location of melanoma & Anatomical location of the primary melanoma & Categorical (balanced accuracy) & Head and Neck, Trunk, Upper extremity, Lower extremity, Not specified & 0.82 & 
		\raisebox{-\totalheight}{\includegraphics[width=\linewidth]{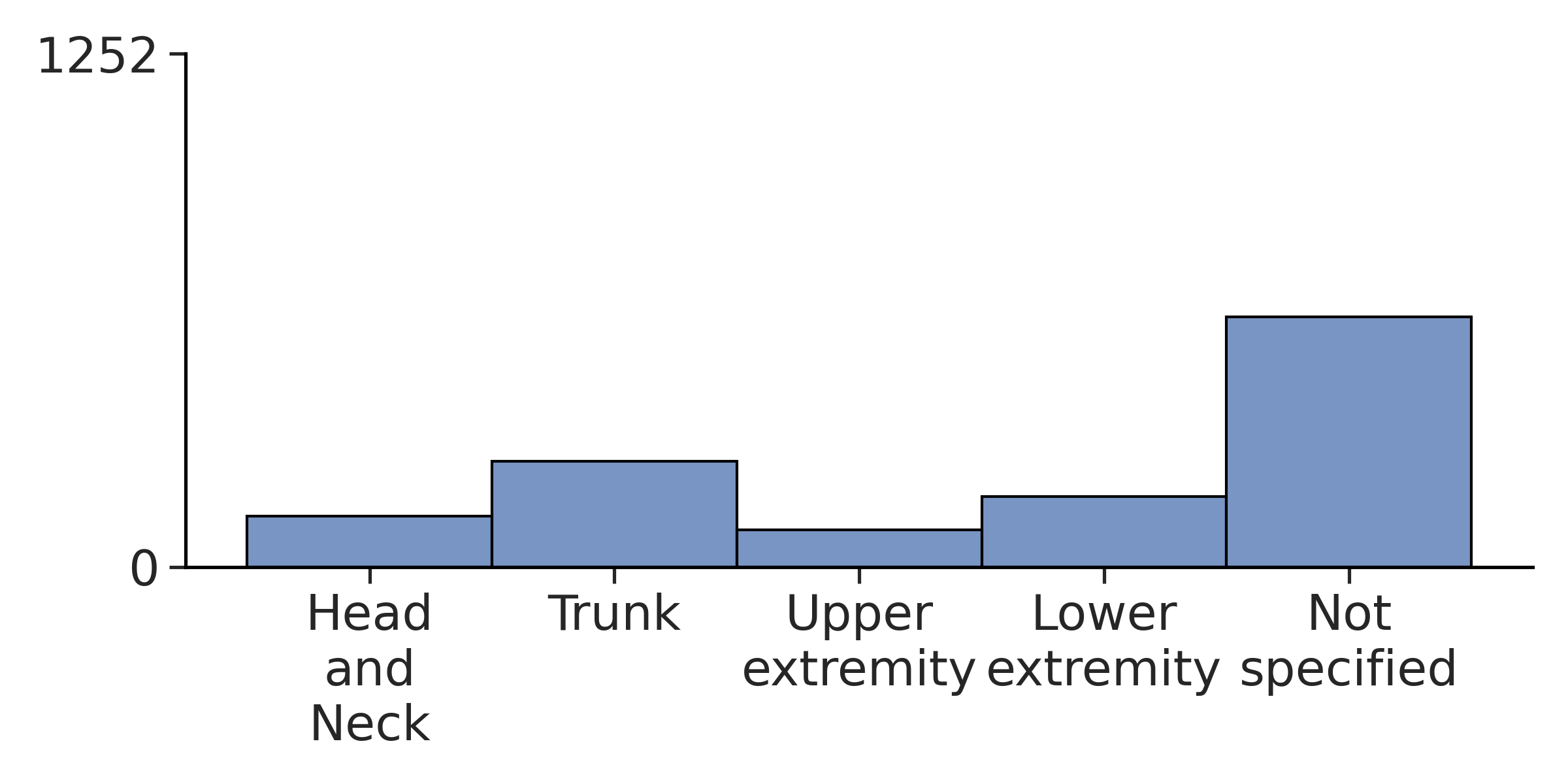}} \\
		
		Subtype of melanoma & Histological subtype classification & Categorical (balanced accuracy) & Superficial spreading melanoma, Nodular melanoma, Acral melanoma, Desmoplastic melanoma, Not specified & 0.91 & 
		\raisebox{-\totalheight}{\includegraphics[width=\linewidth]{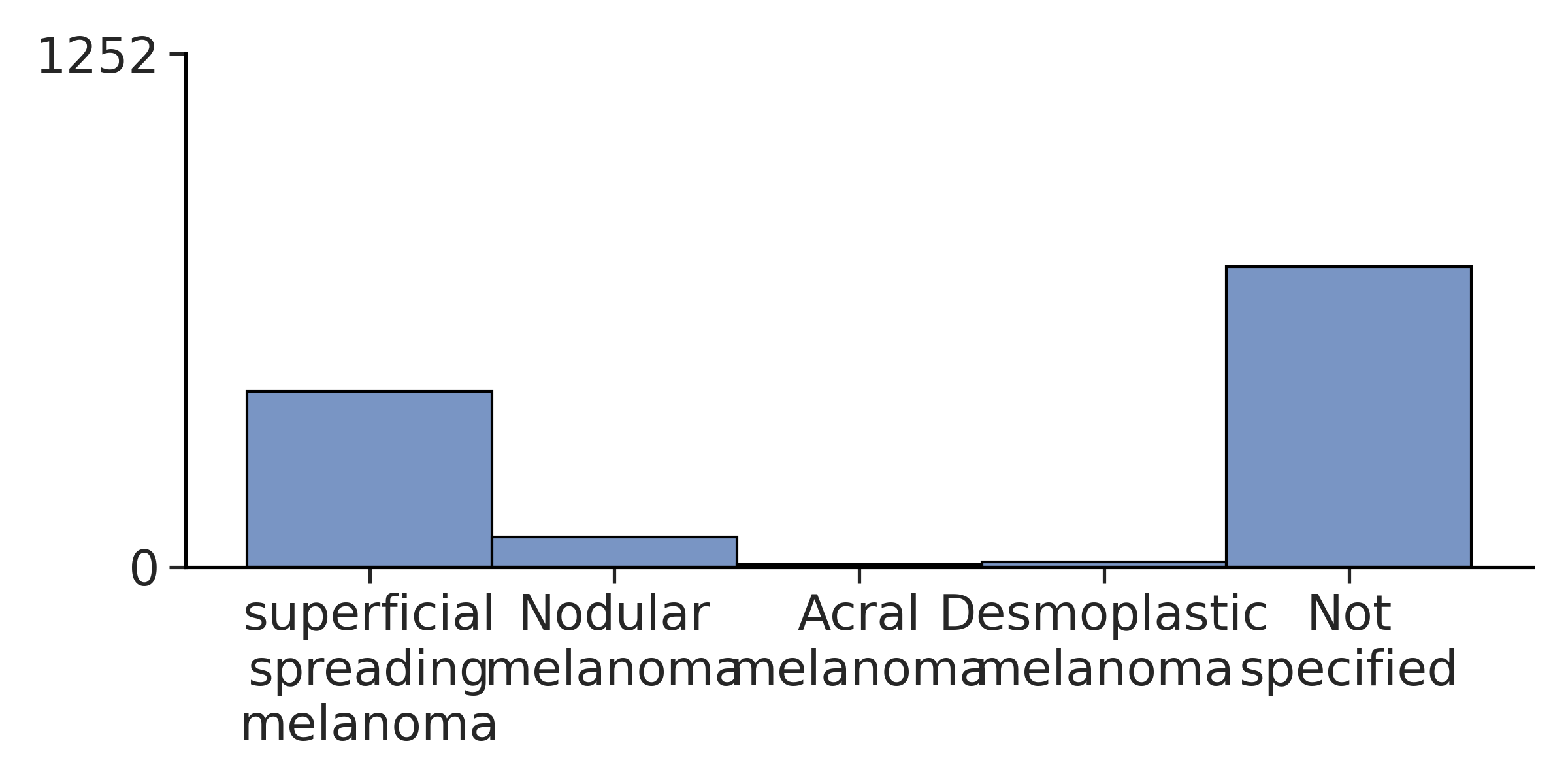}} \\
		
		Microsatellites present & Presence of microsatellite lesions & Categorical (balanced accuracy) & Yes, No, Not specified & 0.72 & 
		\raisebox{-\totalheight}{\includegraphics[width=\linewidth]{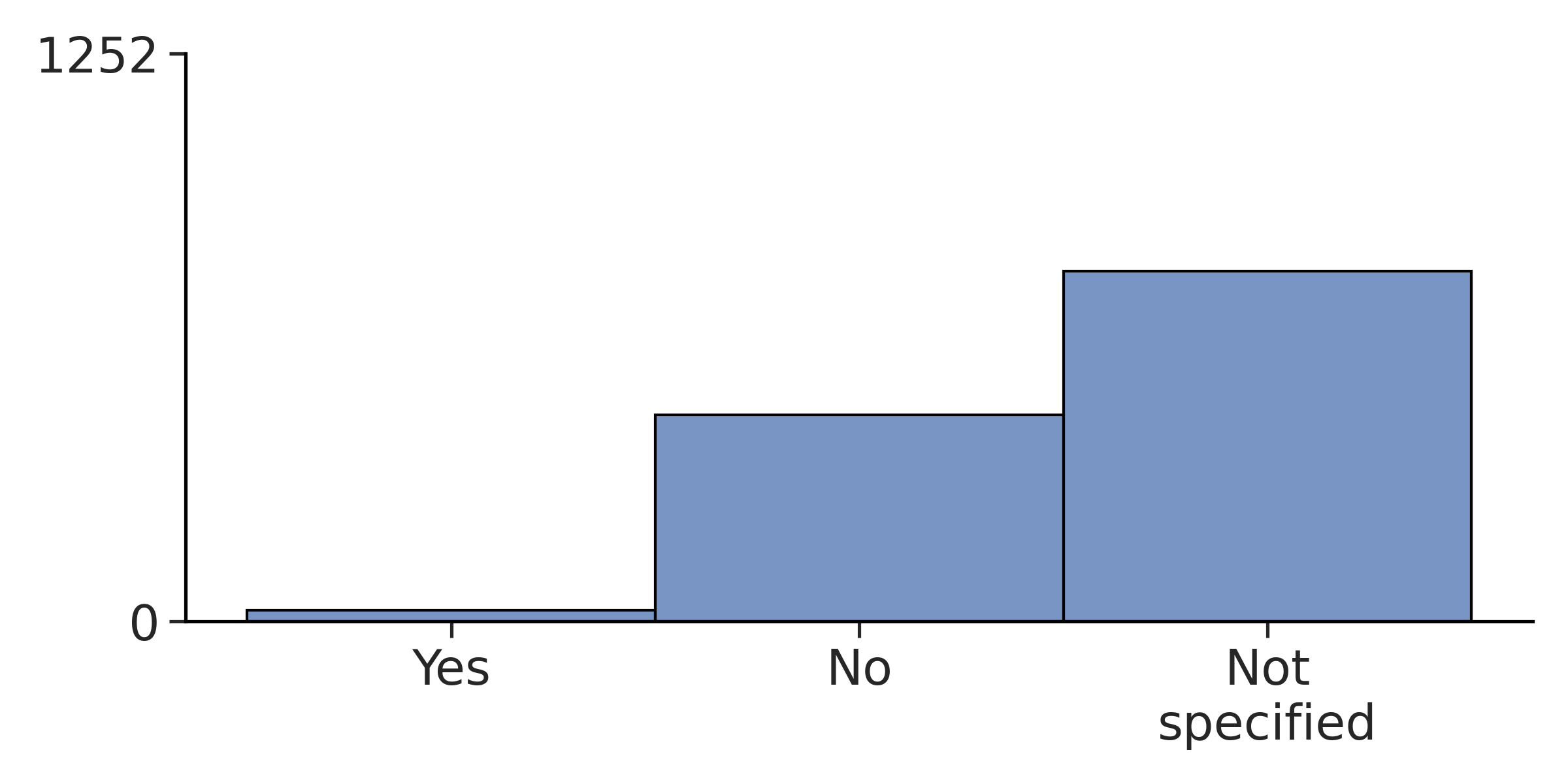}} \\
		
		Mitosis described & Whether mitotic activity is described in report & Categorical (balanced accuracy) & Yes, No, Not specified & 0.52 & 
		\raisebox{-\totalheight}{\includegraphics[width=\linewidth]{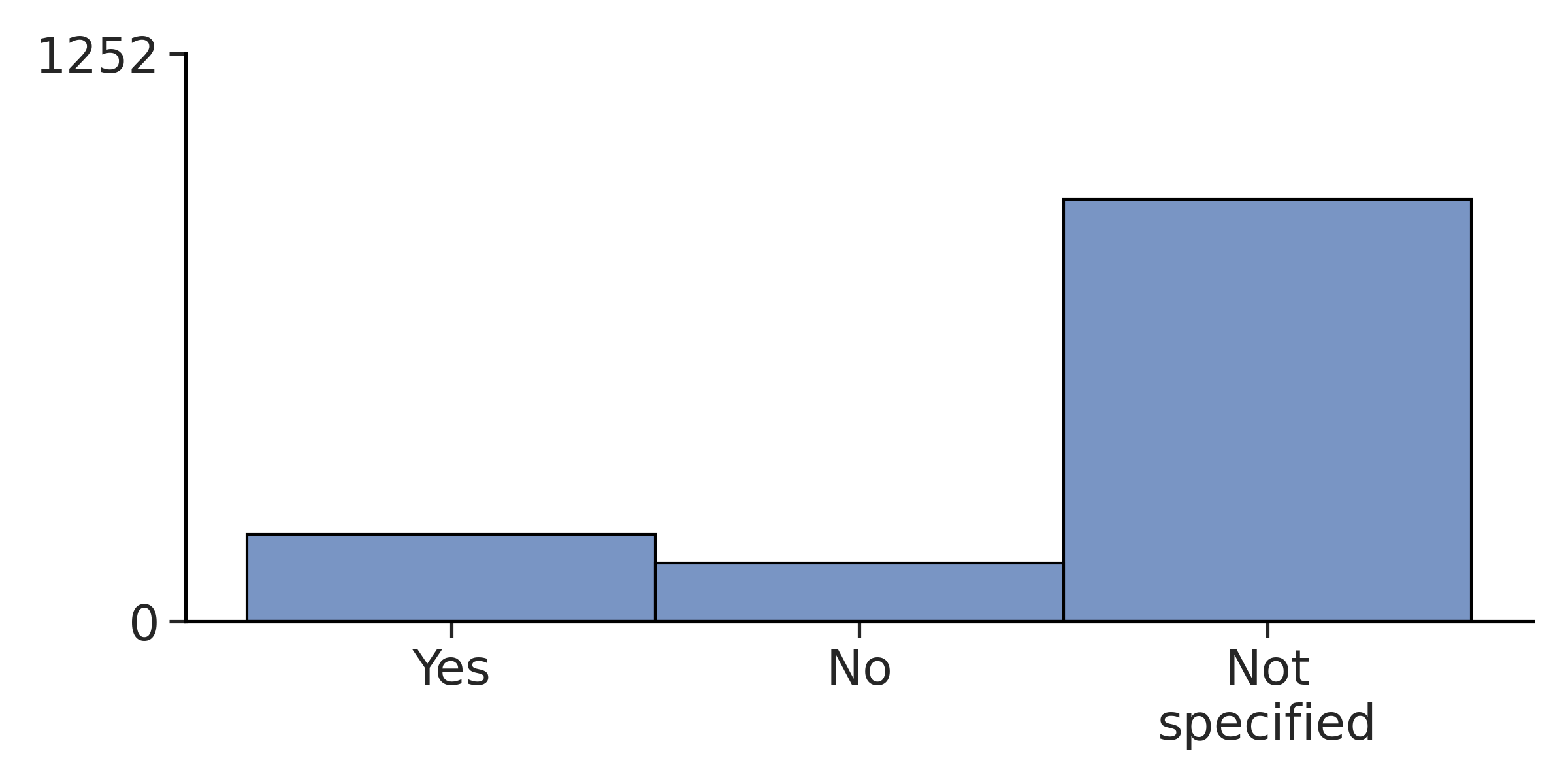}} \\
		
		Mitosis count & Number of mitoses per square millimeter (0-100, only if Mitosis described is "Yes") & Numeric (accuracy) & - & 0.99 & 
		\raisebox{-\totalheight}{\includegraphics[width=\linewidth]{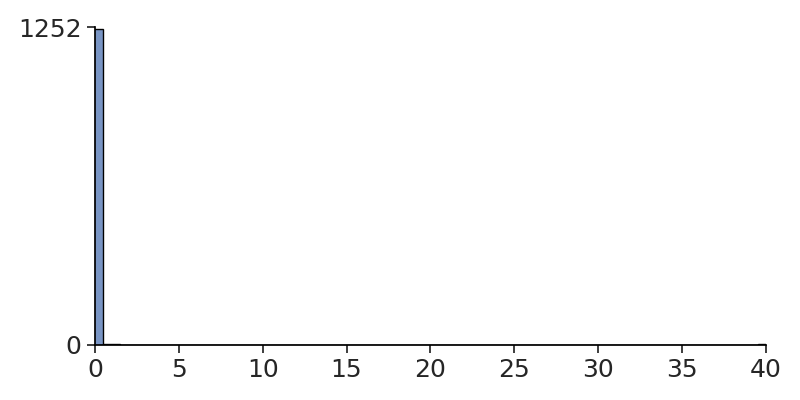}} \\
		
		Lymphatic invasion & Presence of lymphatic vessel invasion & Categorical (balanced accuracy) & Yes, No, Not specified & 0.75 & 
		\raisebox{-\totalheight}{\includegraphics[width=\linewidth]{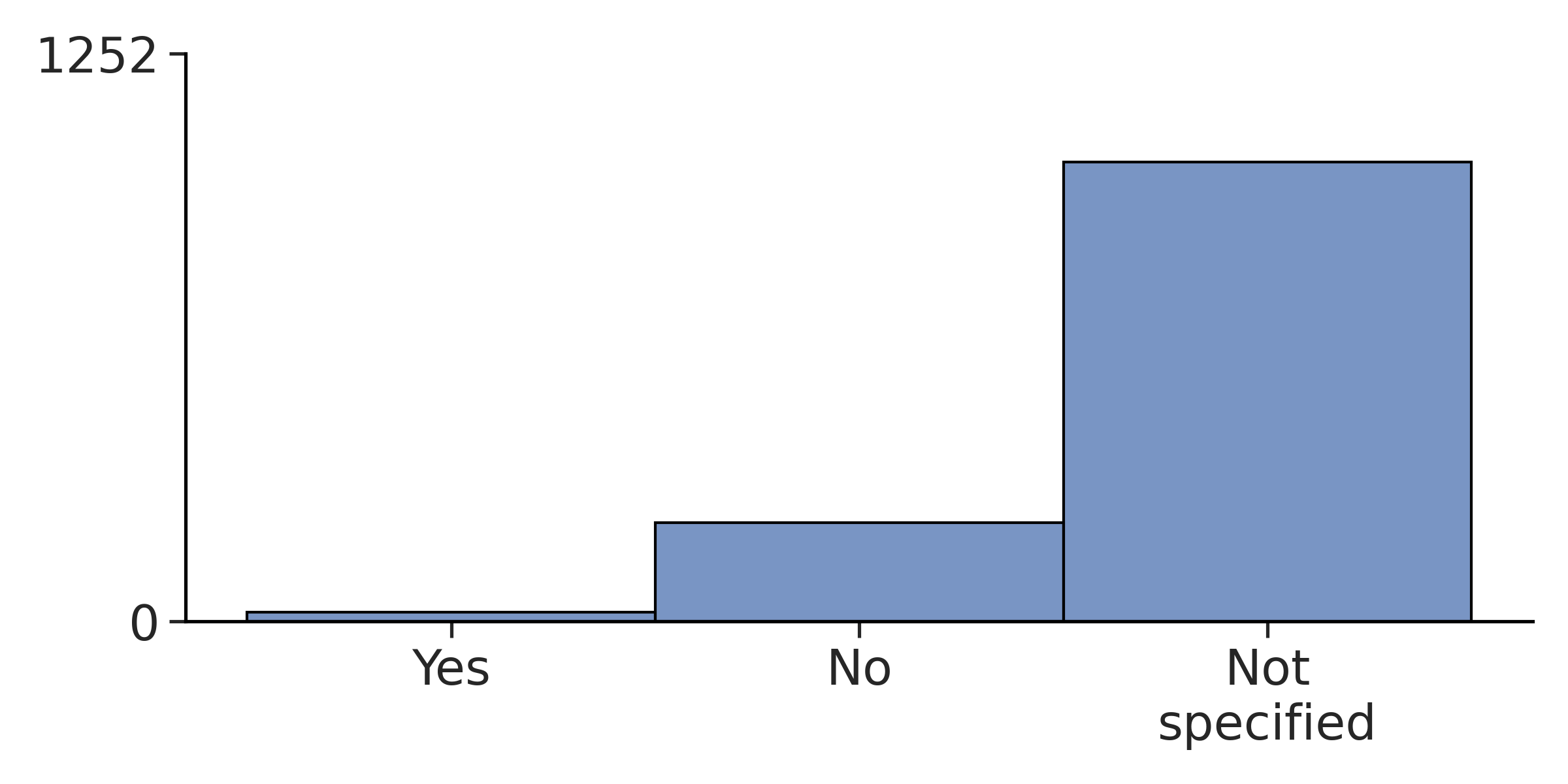}} \\
		
		Specimen type & Type of specimen when not primary melanoma & Categorical (balanced accuracy) & Both, SLNB, WLE, Not specified & 0.48 & 
		\raisebox{-\totalheight}{\includegraphics[width=\linewidth]{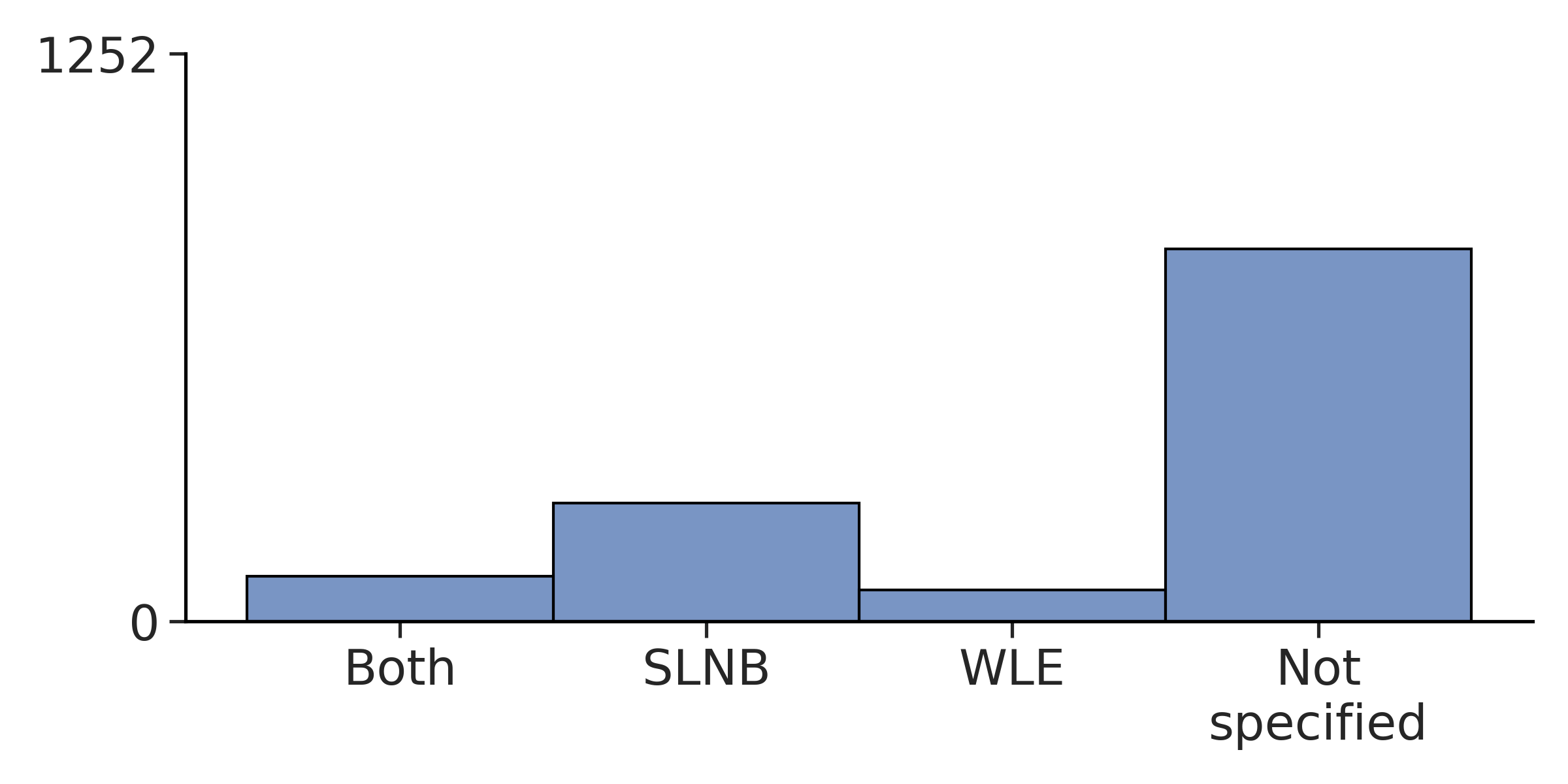}} \\
		
		SLNB metastases present & Presence of metastases in sentinel lymph node (only if specimen includes SLNB) & Categorical (balanced accuracy) & Yes, No, Not specified & 0.60 & 
		\raisebox{-\totalheight}{\includegraphics[width=\linewidth]{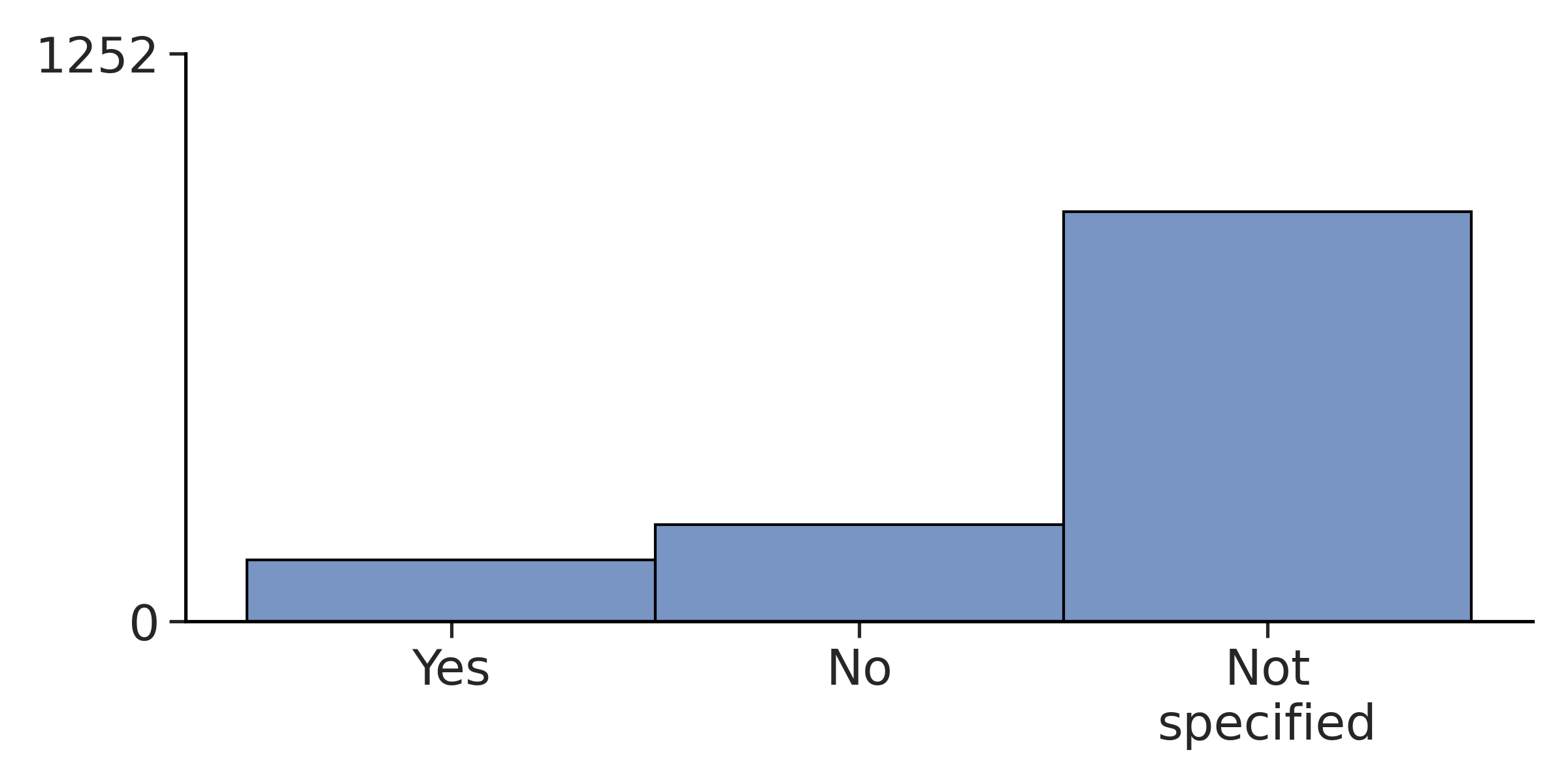}} \\
		
		SLNB tumour burden & Size of metastatic deposit in sentinel node in mm (0.01-200, only if SLNB metastases present is "Yes") & Numeric (accuracy) & - & 0.97 & 
		\raisebox{-\totalheight}{\includegraphics[width=\linewidth]{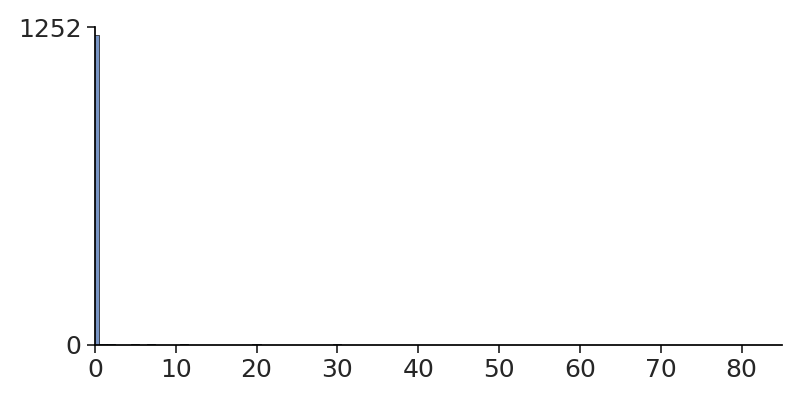}} \\
	\end{longtable}
\end{scriptsize}

\subsubsection{Prompt}
For this use case, a structured prompt was created to ensure consistent extraction and enforce strict JSON formatting. 
The prompt is divided into role-based instructions (System, Human, AI). A full example of the exact prompt used is shown in \autoref{tab:prompt-Melanoma}.

\begin{scriptsize}
	\setlength{\tabcolsep}{4pt}
	\begin{longtable}{>{\scriptsize\raggedright\arraybackslash}m{0.2\linewidth} >{\scriptsize\raggedright\arraybackslash}m{0.37\linewidth} >{\scriptsize\raggedright\arraybackslash}m{0.37\linewidth}}
		\multicolumn{3}{c}{\parbox{\textwidth}{
				\normalsize \tablename~\thetable{} -- Full structured prompt used for the melanomas extraction task. The items in brackets indicate the role of the message (System, Human, AI), while the text provides the corresponding content.\\}} \\
		\toprule
		\textbf{Section} & \multicolumn{2}{c}{\textbf{Content} (Depending on prompt strategy)} \\
		& Zero-Shot, One-Shot and Few-Shot & CoT, Self-Consistency and Graph\\
		\midrule
		\endfirsthead
		
		\multicolumn{3}{c}{\parbox{\textwidth}{
				\normalsize \tablename~\thetable{} -- Continued\\}} \\
		\toprule
		\textbf{Section} & \multicolumn{2}{c}{\textbf{Content}} \\
		\midrule
		\endhead
		
		\phantomlabel{tab:prompt-Melanoma}
		\textbf{[System] -- System instructions} & 
		You are a medical data extraction system that ONLY outputs valid JSON. Maintain strict compliance with these rules: \newline
		1. ALWAYS begin and end your response with \verb|```json| markers \newline
		2. Use EXACT field names and structure provided \newline
		3. If a value is missing or not mentioned, use the specified default for that field. \newline
		4. NEVER add commentary, explanations, or deviate from the output structure & 
		You are a medical data extraction system that performs structured reasoning before producing output. Follow these strict rules: \newline
		1. First, reason step-by-step to identify and justify each extracted field. \newline
		2. After reasoning, output ONLY valid JSON in the exact structure provided. \newline
		3. ALWAYS begin and end the final output with \verb|```json| markers — do not include reasoning within these markers. \newline
		4. Use EXACT field names and structure as specified. \newline
		5. If a value is missing or not mentioned, use the specified default for that field. \newline
		6. NEVER include commentary, explanations, or deviate from the specified format in the final JSON. \newline \\
		\midrule
		\textbf{[Human] -- Field instructions} & \multicolumn{2}{c}{\parbox{0.75\linewidth}{
				1. \opus{"Primary melanoma"}: \newline
				- Type: string \newline
				- Determination of whether this is a primary melanoma case. \newline
				- Options: [\opus{"Yes"}, \opus{"No"}, \opus{"Not specified"}] \newline
				- Default: \opus{"Not specified"} \newline
				2. \opus{"Report a revision"}: \newline
				- Type: string \newline
				- Indicates if this report is a revision of a previous report. \newline
				- Options: [\opus{"Yes"}, \opus{"No"}, \opus{"Not specified"}] \newline
				- Default: \opus{"Not specified"} \newline
				3. \opus{"Multiple primary melanoma"}: \newline
				- Type: string \newline
				- Only applicable if Primary melanoma is "Yes". Indicates presence of multiple primary melanomas. \newline
				- Options: [\opus{"Yes"}, \opus{"No"}, \opus{"Not specified"}] \newline
				- Default: \opus{"Not specified"} \newline
				4. \opus{"In situ melanoma"}: \newline
				- Type: string \newline
				- Indicates if melanoma is in situ (non-invasive). \newline
				- Options: [\opus{"Yes"}, \opus{"No"}, \opus{"Not specified"}] \newline
				- Default: \opus{"Not specified"} \newline
				5. \opus{"Breslow thickness"}: \newline
				- Type: float \newline
				- Measured thickness of melanoma in millimeters. Must be between 0.0 and 400.0 if specified. \newline
				- Default: \opus{"Not specified"} \newline
				6. \opus{"Ulceration"}: \newline
				- Type: string \newline
				- Presence or absence of ulceration. \newline
				- Options: [\opus{"Yes"}, \opus{"No"}, \opus{"Not specified"}] \newline
				- Default: \opus{"Not specified"} \newline
				7. \opus{"Location of melanoma"}: \newline
				- Type: string \newline
				- Anatomical location of the primary melanoma. \newline
				- Options: [\opus{"Head and Neck"}, \opus{"Trunk"}, \opus{"Upper extremity"}, \opus{"Lower extremity"}, \opus{"Not specified"}] \newline
				- Default: \opus{"Not specified"} \newline
				8. \opus{"Subtype of melanoma"}: \newline
				- Type: string \newline
				- Histological subtype classification. \newline
				- Options: [\opus{"superficial spreading melanoma"}, \opus{"Nodular melanoma"}, \opus{"Acral melanoma"}, \opus{"Desmoplastic melanoma"}, \opus{"Not specified"}] \newline
				- Default: \opus{"Not specified"} \newline
				9. \opus{"Microsatellites present"}: \newline
				- Type: string \newline
				- Presence of microsatellite lesions. \newline
				- Options: [\opus{"Yes"}, \opus{"No"}, \opus{"Not specified"}] \newline
				- Default: \opus{"Not specified"} \newline
				10. \opus{"Mitosis described"}: \newline
				- Type: string \newline
				- Whether mitotic activity is described in report. \newline
				- Options: [\opus{"Yes"}, \opus{"No"}, \opus{"Not specified"}] \newline
				- Default: \opus{"Not specified"} \newline
				11. \opus{"Mitosis count"}: \newline
				- Type: float \newline
				- Number of mitoses per square millimeter. Only applicable if 'Mitosis described' is "Yes". Must be between 0.0 and 100.0 if specified. \newline
				- Default: \opus{"Not specified"} \newline
				12. \opus{"Lymphatic invasion"}: \newline
				- Type: string \newline
				- Presence of lymphatic vessel invasion. \newline
				- Options: [\opus{"Yes"}, \opus{"No"}, \opus{"Not specified"}] \newline
				- Default: \opus{"Not specified"} \newline
		}} \newline \\
		\midrule
		\textbf{[Human] -- Field instructions} & \multicolumn{2}{c}{\parbox{0.75\linewidth}{
				13. \opus{"Specimen type"}: \newline
				- Type: string \newline
				- Type of specimen when not primary melanoma. SLNB = Sentinel Lymph Node Biopsy, WLE = Wide Local Excision. \newline
				- Options: [\opus{"Both"}, \opus{"SLNB"}, \opus{"WLE"}, \opus{"Not specified"}] \newline
				- Default: \opus{"Not specified"} \newline
				14. \opus{"SLNB metastases present"}: \newline
				- Type: string \newline
				- Presence of metastases in sentinel lymph node. Only applicable if specimen includes SLNB. \newline
				- Options: [\opus{"Yes"}, \opus{"No"}, \opus{"Not specified"}] \newline
				- Default: \opus{"Not specified"} \newline
				15. \opus{"SLNB tumour burden"}: \newline
				- Type: float \newline
				- Size of metastatic deposit in sentinel node in mm. Only applicable if 'SLNB metastases present' is "Yes". Must be between 0.01 and 200.00 if specified. \newline
				- Default: \opus{"Not specified"}
		}} \newline \\
		\midrule
		\textbf{[Human] -- Task instructions} & \multicolumn{2}{c}{\parbox{0.75\linewidth}{
				Extract information into JSON with these fields. The output must look like:
				\opus{```json} \newline
				\opus{\{} \newline
				\opus{\quad "Primary melanoma": "",} \newline
				\opus{\quad "Report a revision": "",} \newline
				\opus{\quad "Multiple primary melanoma": "",} \newline
				\opus{\quad "In situ melanoma": "",} \newline
				\opus{\quad "Breslow thickness": "",} \newline
				\opus{\quad "Ulceration": "",} \newline
				\opus{\quad "Location of melanoma": "",} \newline
				\opus{\quad "Subtype of melanoma": "",} \newline
				\opus{\quad "Microsatellites present": "",} \newline
				\opus{\quad "Mitosis described": "",} \newline
				\opus{\quad "Mitosis count": "",} \newline
				\opus{\quad "Lymphatic invasion": "",} \newline
				\opus{\quad "Specimen type": "",} \newline
				\opus{\quad "SLNB metastases present": "",} \newline
				\opus{\quad "SLNB tumour burden": "",} \newline
				\opus{\}} \newline
				\opus{```}%
			}%
		} \newline \\
		\midrule	
		\textbf{[Human] -- Example intro} & \multicolumn{2}{c}{\parbox{0.75\linewidth}{Below are 1 example of expected input and output, followed by a new task.}} \\
		
		\textbf{[Human] -- Example user} & \multicolumn{2}{c}{\parbox{0.75\linewidth}{\opus{[...CENSORED...]}}} \newline \\
		\midrule
		\textbf{[AI] -- Example assistant reasoning} & - & 
		- \opus{Primary melanoma} - "maligne melanoom" implies "Yes" \newline
		- \opus{Report a revision} - "no indication of revised report or second opinion" implies "Not specified" \newline
		- \opus{Multiple primary melanoma} - "only describes one lesion" implies "Not specified" \newline
		- \opus{In situ melanoma} - "dermal invasion present" implies "No" \newline
		- \opus{Breslow thickness} - "explicitly reported as 3 mm" implies 3.0 \newline
		- \opus{Ulceration} - "explicitly mentioned in conclusion and microscopy" implies "Yes" \newline
		- \opus{Location of melanoma} - "bovenarm links" maps to "Upper extremity" \newline
		- \opus{Subtype of melanoma} - "spoelcellig (spindle cell melanoma)" implies "spindle cell melanoma" \newline
		- \opus{Microsatellites present} - "Geen satellitose" implies "No" \newline
		- \opus{Mitosis described} - "meer dan 1 mitose per vierkante millimeter" implies "Yes" \newline
		- \opus{Mitosis count} - "more than 1/mm²; exact count not provided" implies ">1" \newline
		- \opus{Lymphatic invasion} - "Geen angio-invasieve groei" implies "No" \newline
		- \opus{Specimen type} - "huidexcisie" implies "Excision" \newline
		- \opus{SLNB metastases present} - "sentinel node procedure recommended but not yet performed" implies "Not specified" \newline
		- \opus{SLNB tumour burden} - "no SLNB biopsy performed yet" implies "Not specified" \newline
		- \opus{Quality Check} - "all findings consistent; spindle cell morphology, ulceration, high mitotic rate, positive deep margin" implies "No issues detected" \newline \\
		\midrule
		\textbf{[AI] -- Example assistant output} & \multicolumn{2}{c}{\parbox{0.75\linewidth}{
				\opus{```json} \newline
				\opus{\{} \newline
				\opus{\quad "Primary melanoma": "Yes",} \newline
				\opus{\quad "Report a revision": "Not specified",} \newline
				\opus{\quad "Multiple primary melanoma": "Not specified",} \newline
				\opus{\quad "In situ melanoma": "No",} \newline
				\opus{\quad "Breslow thickness": 3.0,} \newline
				\opus{\quad "Ulceration": "Yes",} \newline
				\opus{\quad "Location of melanoma": "Upper extremity",} \newline
				\opus{\quad "Subtype of melanoma": "spindle cell melanoma",} \newline
				\opus{\quad "Microsatellites present": "No",} \newline
				\opus{\quad "Mitosis described": "Yes",} \newline
				\opus{\quad "Mitosis count": ">1",} \newline
				\opus{\quad "Lymphatic invasion": "No",} \newline
				\opus{\quad "Specimen type": "Excision",} \newline
				\opus{\quad "SLNB metastases present": "Not specified",} \newline
				\opus{\quad "SLNB tumour burden": "Not specified",} \newline
				\opus{\quad "Quality Check": "No issues detected - Diagnosis consistent across all reported features: spindle cell morphology, ulceration, high mitotic rate, and positive deep margin"} \newline
				\opus{\}} \newline
				\opus{```}%
			}%
		} \newline \\
		\midrule
		\textbf{[Human] -- Report instructions} & \multicolumn{2}{c}{\parbox{0.75\linewidth}{[file name]: \opus{[...CENSORED...]} \newline \opus{[...CENSORED...]}}} \newline \\
		\midrule
		\textbf{[Human] -- Final instructions} &
		Begin the extraction now. Your response must contain only a single valid JSON block, enclosed in triple backticks and prefixed with \verb|`json`|, like this: \verb|```json  ... ```|& 
		Begin the extraction now. First, reason step-by-step to identify and justify the value for each required field, enclosed within \verb|<think>...</think>| tags. Then, output only the final structured data as a single valid JSON block, starting with \verb|```json| and ending with \verb|```|."
		\\
		\bottomrule
	\end{longtable}
\end{scriptsize}

\subsubsection{Prompt Graph}
The dependencies, conditional branches, and extraction order of variables are represented as a directed acyclic graph, including paths for dependent fields. This graph reflects how the extraction task is decomposed into smaller, sequential subtasks for the Prompt Graph prompting strategy, where the output of one field constrains the extraction of subsequent fields. 
\begin{figure}[htbp]
	\centering
	\includegraphics[width=\linewidth, height=0.5\textheight, keepaspectratio]{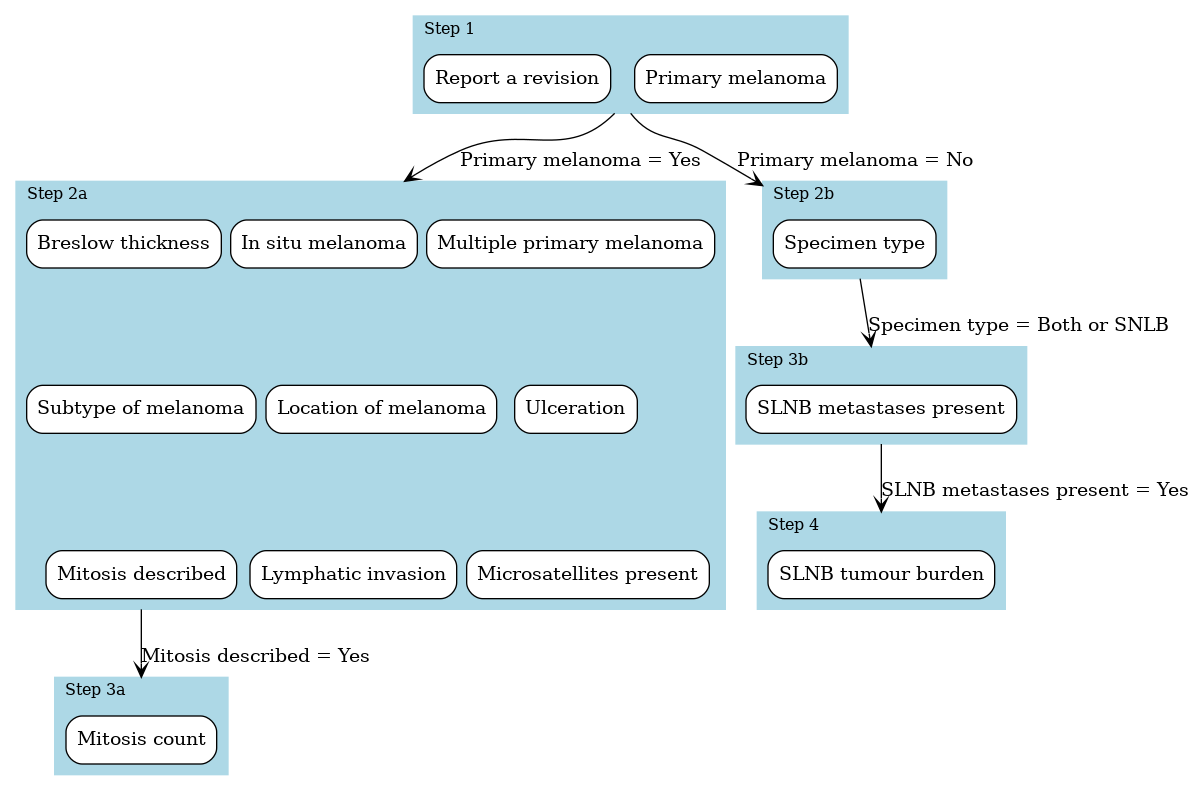}
	\caption{Directed acyclic graph showing the dependencies, conditional branches, and extraction order of variables for melanomas use case.}
\end{figure}

\newpage
\subsection{Use Case - Sarcomas (Czech)}

\subsubsection{Overview}
This use case focuses on extracting structured information from Czech pathology reports of sarcoma cancer patients. The entities of interest include primary tumour size, mitotic activity (in count/mm$^2$, and count/10 HPF), vascular invasion, invasion into fascia, cytologic atypia degree and presence of necrosis (\autoref{tab:use-case-Sarcoma}). This task assesses the extensibility of the tool and different LLMs to support Czech, a relatively low-resource Slavic language.

\subsubsection{Inclusion and Exclusion Criteria}
Pathology reports were included for sarcoma patients treated at the Masaryk Memorial Cancer Institute (MMCI) between 2008 and 2023. No exclusion criteria were applied. In total, 80 clinical notes from 80 unique patients were annotated for this use case.

\subsubsection{Annotation}
The reports were annotated by nine clinical professionals from MMCI and Motol University Hospital. Each note was span-annotated independently by three clinicians, and the resulting annotations were consolidated into a single, unified annotation per report through consensus.

\subsubsection{Ethical Considerations and Funding}

The use of Czech pathology reports was approved by the MMCI Ethics Committee (approval no. 2022/3528/MOU).  
Dataset collection and study support were provided by:
\begin{itemize}
	\item European Union Horizon Research and Innovation Programme (grant no.\@ \newline 101057048, IDEA4RC)  
	\item Grant Agency of Masaryk University (grant no.\@ MUNI/A/1638/2024, SV25-AI4Data)  
	\item Ministry of Health of the Czech Republic (grant no.\@ NW25-09-00465, EMPOWER)  
\end{itemize}

\noindent Computational resources were provided by the e-INFRA CZ project (ID: 90254), supported by the Ministry of Education, Youth and Sports of the Czech Republic.

\begin{scriptsize}
	\setlength{\tabcolsep}{4pt}
	\begin{longtable}{@{}
			>{\scriptsize\raggedright\arraybackslash}p{0.10\textwidth} % Variable name
			>{\scriptsize\raggedright\arraybackslash}p{0.15\textwidth} % Description
			>{\scriptsize\raggedright\arraybackslash}p{0.10\textwidth} % Variable type
			>{\scriptsize\raggedright\arraybackslash}p{0.11\textwidth} % Variable options
			>{\scriptsize\raggedleft\arraybackslash}p{0.11\textwidth} % Annotator Agreement
			>{\scriptsize\centering\arraybackslash}p{0.30\textwidth} @{}} % Distribution with image
		\multicolumn{6}{c}{\parbox{\textwidth}{
				\normalsize \tablename~\thetable{} -- Sarcomas Czech pathology report variable definitions with reference standard distribution.\\}} \\
		\toprule
		\textbf{Variable Name} & 
		\textbf{Description} & 
		\textbf{Type (Metric)} & 
		\textbf{Variable Options} & 
		\textbf{Annotator Agreement} & 
		\textbf{Reference Standard Distribution} \\
		\midrule
		\endfirsthead
		
		\multicolumn{6}{c}{\parbox{\textwidth}{
				\normalsize \tablename~\thetable{} -- Continued\\}} \\
		\toprule
		\textbf{Variable Name} & 
		\textbf{Description} & 
		\textbf{Type (Metric)} & 
		\textbf{Variable Options} & 
		\textbf{Annotator Agreement} & 
		\textbf{Reference Standard Distribution} \\
		\midrule
		\endhead
		
		\midrule
		\endfoot
		
		\bottomrule
		\endlastfoot
		
		\phantomlabel{tab:use-case-Sarcoma}
		Cytologic atypia & Highest degree of cytologic atypia detected & Categorical (balanced accuracy) & Mild, Moderate, Severe, missing & 0.57 & 
		\raisebox{-\totalheight}{\includegraphics[width=\linewidth]{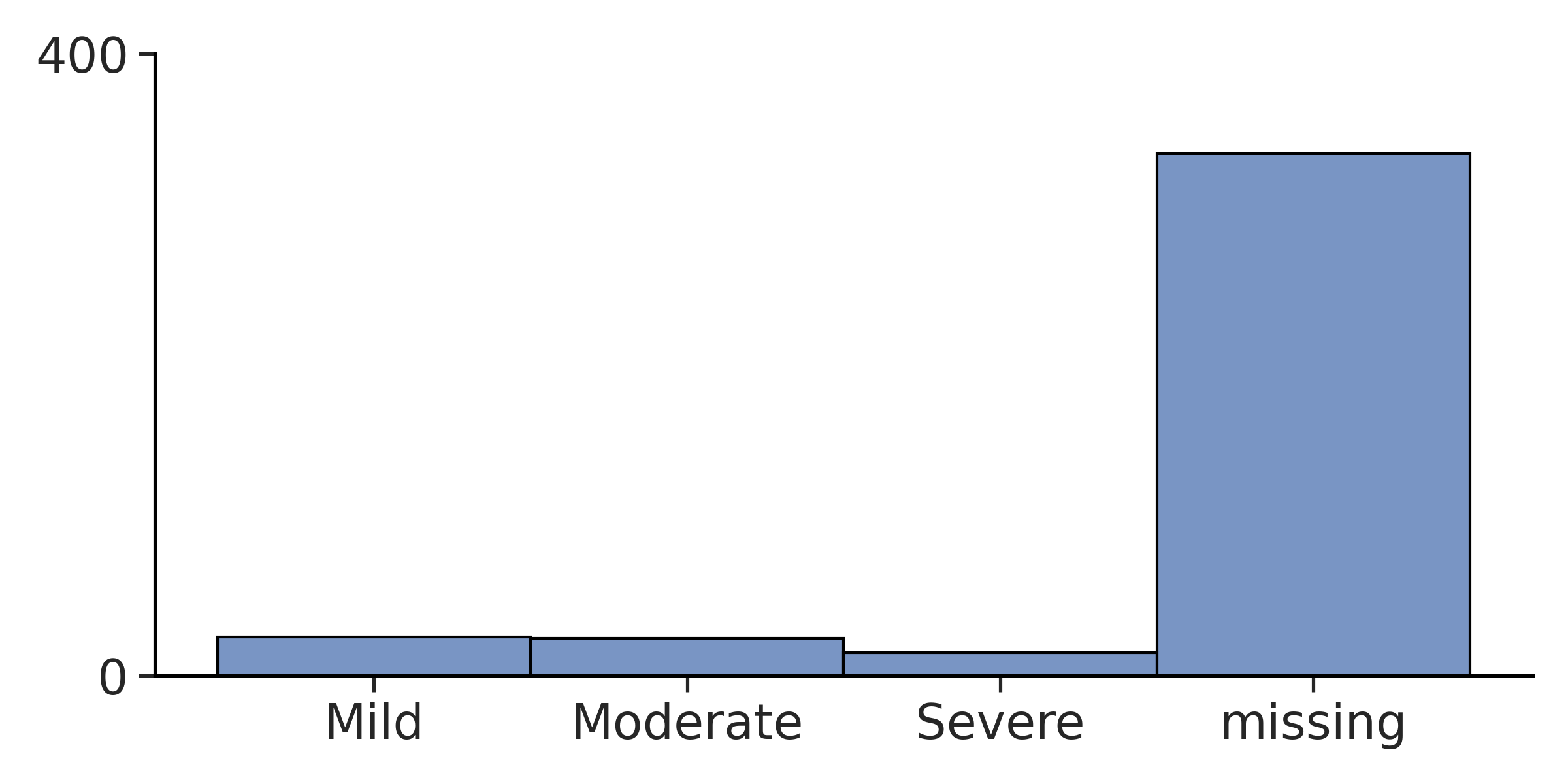}} \\
		
		Invasion into fascia & Indicates whether invasion into fascia was detected & Binary (balanced accuracy) & Yes, No, missing & 0.64 & 
		\raisebox{-\totalheight}{\includegraphics[width=\linewidth]{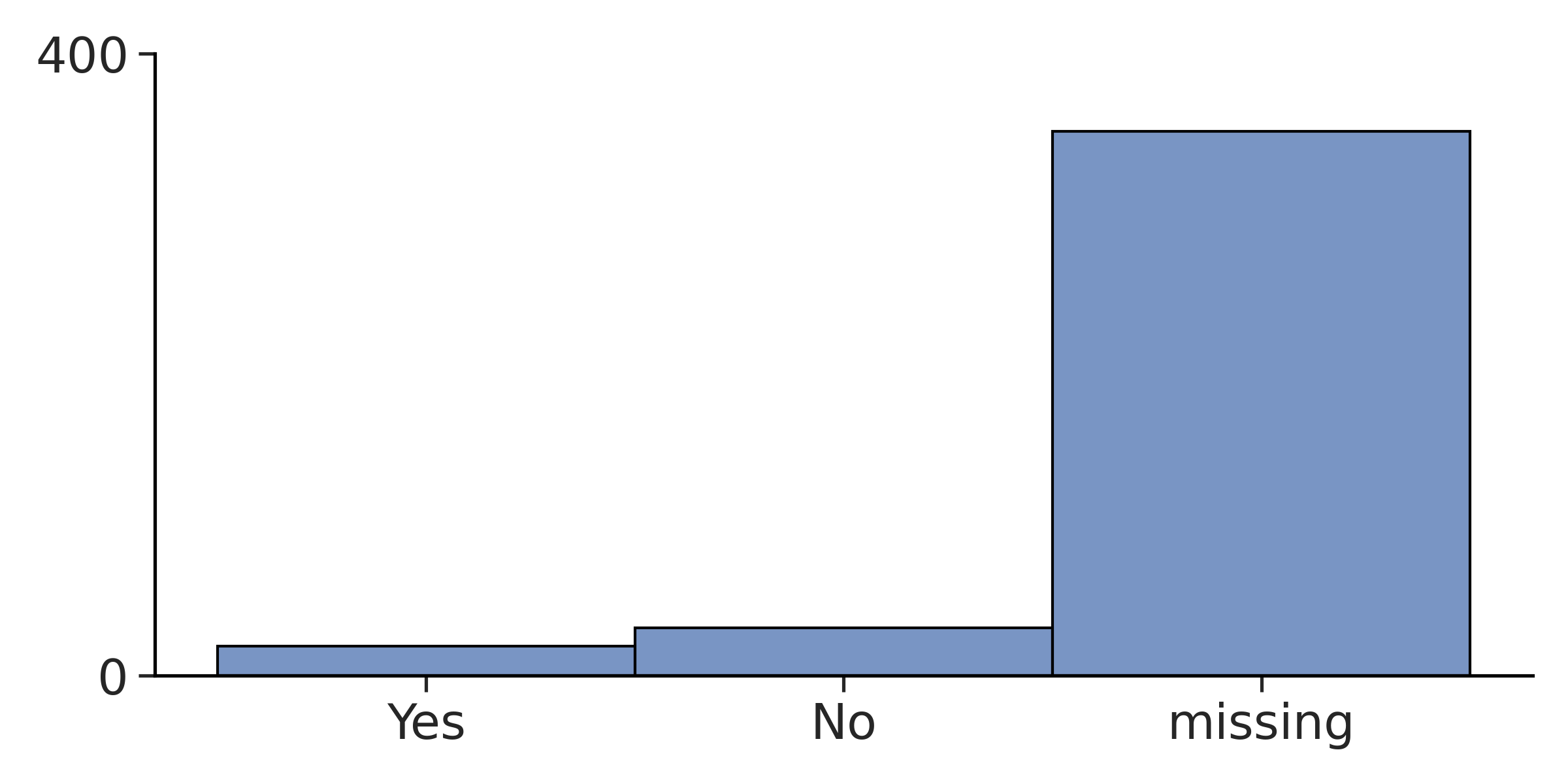}} \\
		
		Vascular invasion & Indicates whether vascular invasion was detected & Binary (balanced accuracy) & Yes, No, missing & 0.87 & 
		\raisebox{-\totalheight}{\includegraphics[width=\linewidth]{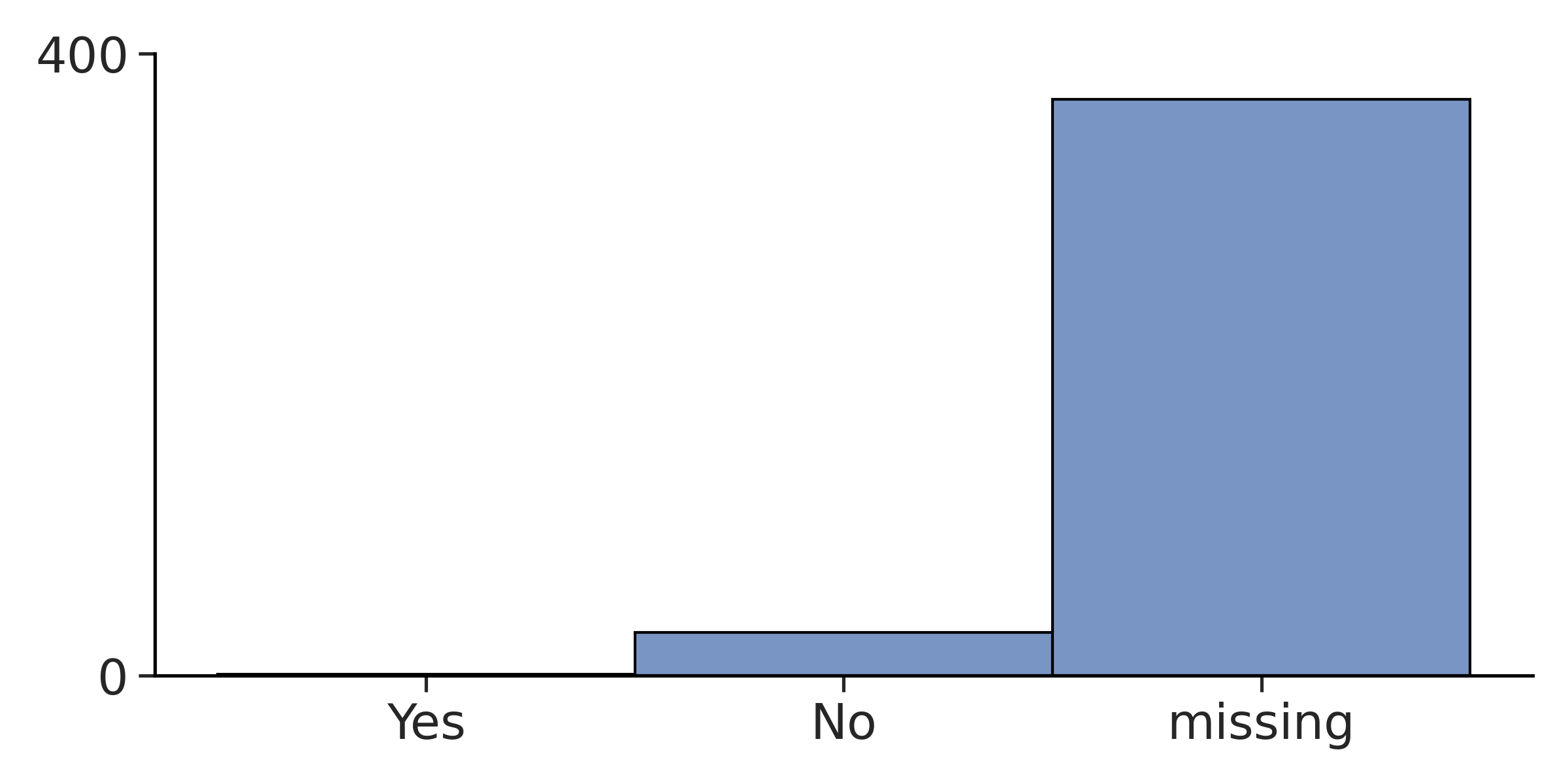}} \\
		
		Necrosis & Indicates whether necrosis was detected & Binary (balanced accuracy) & Yes, No, missing & 0.88 & 
		\raisebox{-\totalheight}{\includegraphics[width=\linewidth]{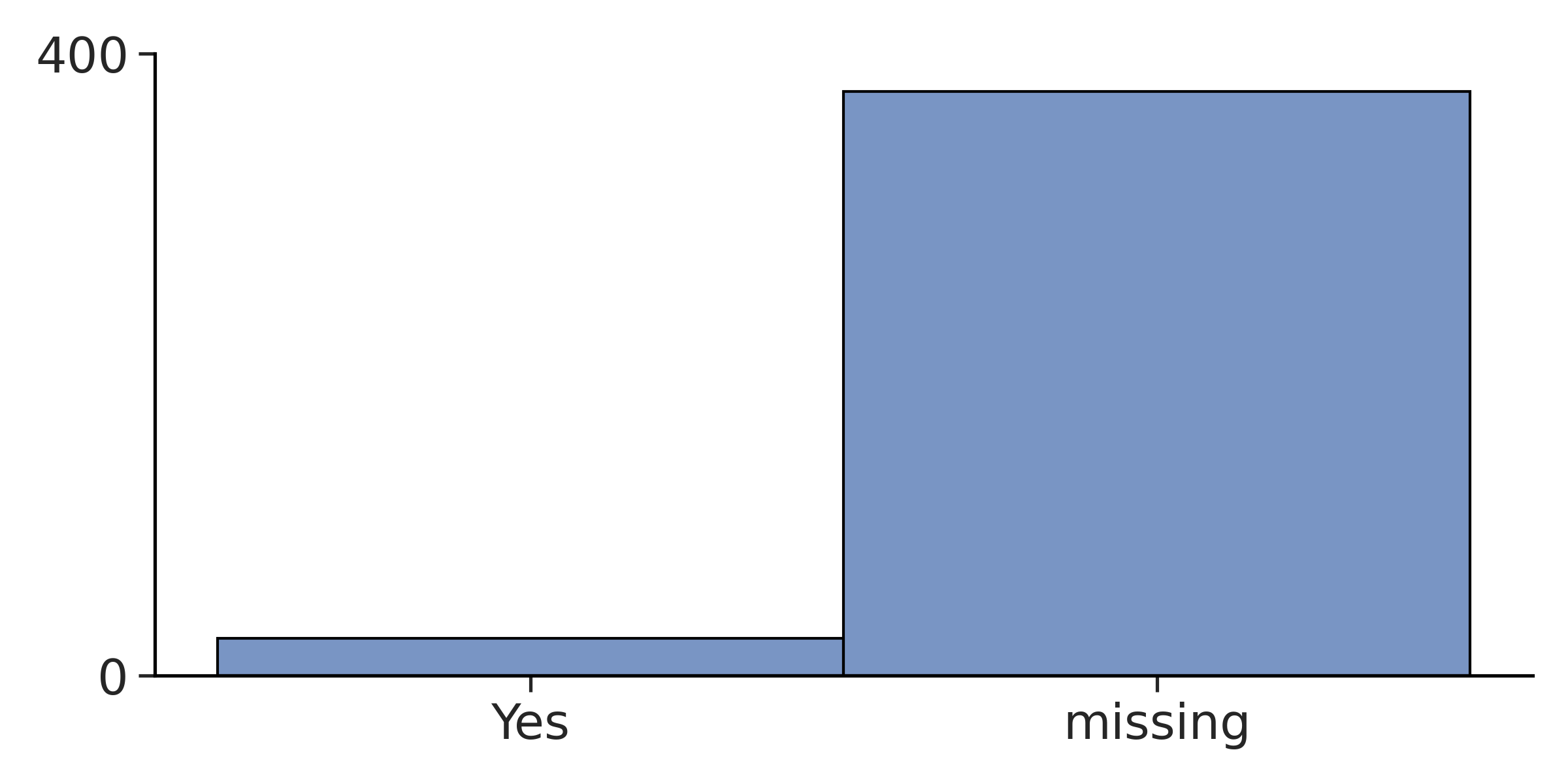}} \\
		
		Tumour size & Largest dimension of the tumour in millimeters & Numeric (accuracy) & Any non-negative integer (mm) or missing & 0.84 & 
		\raisebox{-\totalheight}{\includegraphics[width=\linewidth]{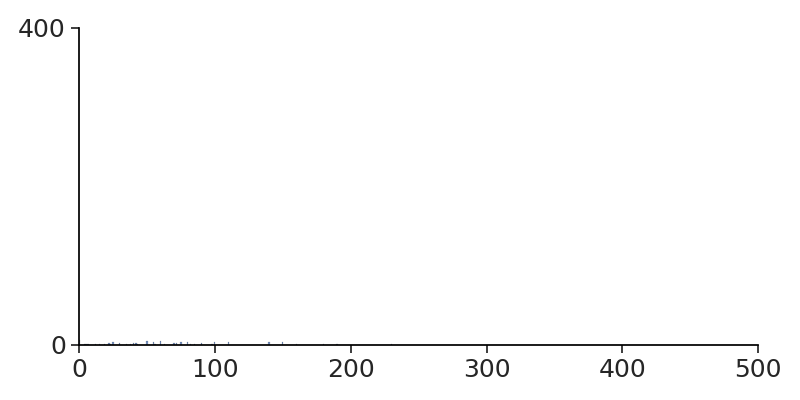}} \\
		
		Mitotic activity (per HPF) & Highest measurement of mitotic activity per high power field & String (exact match) & Reported string value (e.g., “3/HPF”), or missing & 0.76 & 
		\raisebox{-\totalheight}{\includegraphics[width=\linewidth]{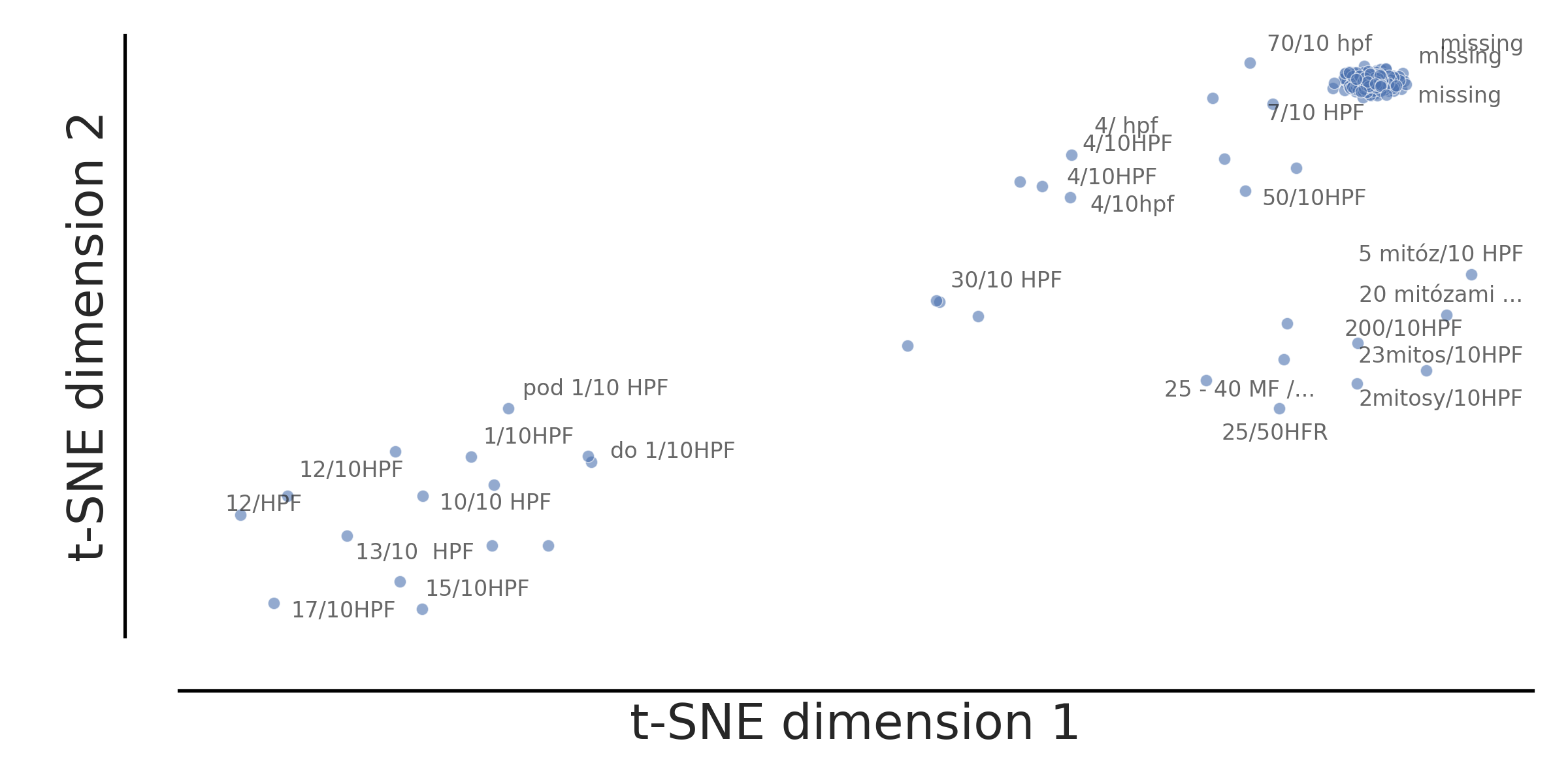}} \\
		
		Mitotic activity (per mm²) & Highest measurement of mitotic activity per mm² & String (exact match) & Reported string value (e.g., “8/mm²”), or missing & 0.93 & 
		\raisebox{-\totalheight}{\includegraphics[width=\linewidth]{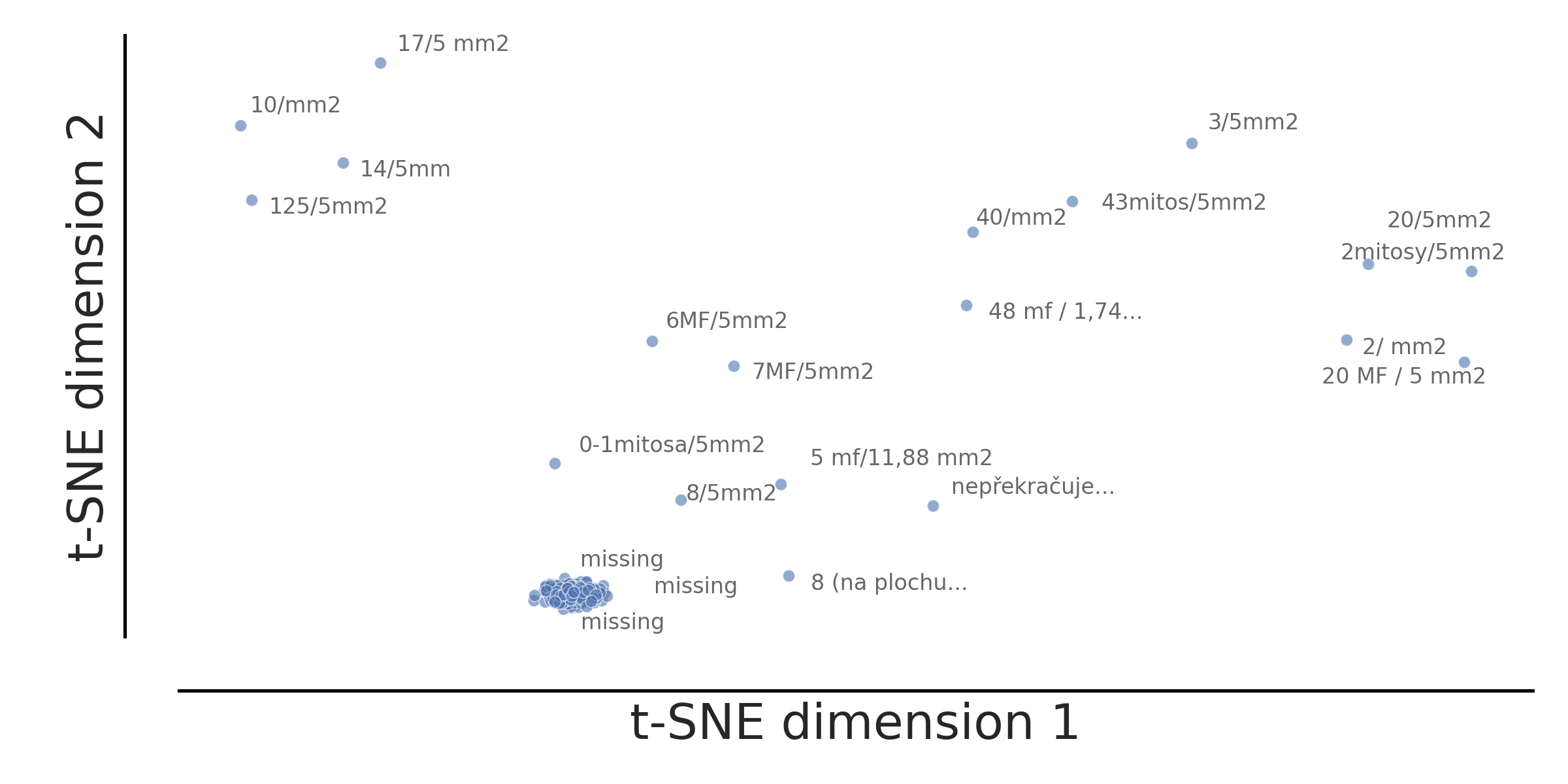}} \\
	\end{longtable}
\end{scriptsize}

\subsubsection{Prompt}
For this use case, a structured prompt was created to ensure consistent extraction and enforce strict JSON formatting. 
The prompt is divided into role-based instructions (System, Human, AI). A full example of the exact prompt used is shown in \autoref{tab:prompt-Sarcoma}.

\begin{scriptsize}
	\setlength{\tabcolsep}{4pt}
	\begin{longtable}{>{\scriptsize\raggedright\arraybackslash}m{0.2\linewidth} >{\scriptsize\raggedright\arraybackslash}m{0.37\linewidth} >{\scriptsize\raggedright\arraybackslash}m{0.37\linewidth}}
		\multicolumn{3}{c}{\parbox{\textwidth}{
				\normalsize \tablename~\thetable{} -- Full structured prompt used for the sarcomas extraction task. The items in brackets indicate the role of the message (System, Human, AI), while the text provides the corresponding content.\\}} \\
		\toprule
		\textbf{Section} & \multicolumn{2}{c}{\textbf{Content} (Depending on prompt strategy)} \\
		& Zero-Shot, One-Shot and Few-Shot & CoT, Self-Consistency and Graph\\
		\midrule
		\endfirsthead
		
		\multicolumn{3}{c}{\parbox{\textwidth}{
				\normalsize \tablename~\thetable{} -- Continued\\}} \\
		\toprule
		\textbf{Section} & \multicolumn{2}{c}{\textbf{Content}} \\
		\midrule
		\endhead
		
		\phantomlabel{tab:prompt-Sarcoma}
		\textbf{[System] -- System instructions} & 
		You are a medical data extraction system that ONLY outputs valid JSON. Maintain strict compliance with these rules: \newline
		1. ALWAYS begin and end your response with \verb|```json| markers \newline
		2. Use EXACT field names and structure provided \newline
		3. If a value is missing or not mentioned, use the specified default for that field. \newline
		4. NEVER add commentary, explanations, or deviate from the output structure & 
		You are a medical data extraction system that performs structured reasoning before producing output. Follow these strict rules: \newline
		1. First, reason step-by-step to identify and justify each extracted field. \newline
		2. After reasoning, output ONLY valid JSON in the exact structure provided. \newline
		3. ALWAYS begin and end the final output with \verb|```json| markers — do not include reasoning within these markers. \newline
		4. Use EXACT field names and structure as specified. \newline
		5. If a value is missing or not mentioned, use the specified default for that field. \newline
		6. NEVER include commentary, explanations, or deviate from the specified format in the final JSON. \\
		
		\textbf{[Human] -- Field instructions} & \multicolumn{2}{c}{\parbox{0.75\linewidth}{
				1. \opus{"cytologic atypia"}: \newline
				- Type: string \newline
				- Highest degree of cytologic atypia detected. \opus{"missing"} if not specified. \newline
				- Options: [\opus{"Mild"}, \opus{"Moderate"}, \opus{"Severe"}, \opus{"missing"}] \newline
				- Default: \opus{"missing"} \newline
				2. \opus{"invasion into fascia"}: \newline
				- Type: string \newline
				- Was invasion into fascia detected? \opus{"missing"} if not specified. \newline
				- Options: [\opus{"Yes"}, \opus{"No"}, \opus{"missing"}] \newline
				- Default: \opus{"missing"} \newline
				3. \opus{"vascular invasion"}: \newline
				- Type: string \newline
				- Was vascular invasion detected? \opus{"missing"} if not specified. \newline
				- Options: [\opus{"Yes"}, \opus{"No"}, \opus{"missing"}] \newline
				- Default: \opus{"missing"} \newline
				4. \opus{"necrosis"}: \newline
				- Type: string \newline
				- Was necrosis detected? \opus{"missing"} if not specified. \newline
				- Options: [\opus{"Yes"}, \opus{"No"}, \opus{"missing"}] \newline
				- Default: \opus{"missing"} \newline
				5. \opus{"tumour size"}: \newline
				- Type: number\_or\_missing \newline
				- Largest dimension in mm. \opus{"missing"} if not specified. \newline
				- Default: \opus{"missing"} \newline
				6. \opus{"mitotic hpf"}: \newline
				- Type: string\_exact\_match \newline
				- Highest mitotic activity in \_/HPF. Use exact string. \opus{"missing"} if not specified. \newline
				- Default: \opus{"missing"} \newline
				7. \opus{"mitotic mm"}: \newline
				- Type: string\_exact\_match \newline
				- Highest mitotic activity in \_/mm2. Use exact string. \opus{"missing"} if not specified. \newline
				- Default: \opus{"missing"}}} \newline \\
		\midrule
		\textbf{[Human] -- Task instructions} & \multicolumn{2}{c}{\parbox{0.75\linewidth}{
				Extract information from the following Czech clinical note into JSON with these fields. The output must look like:
				\opus{```json} \newline
				\opus{\{} \newline
				\opus{\quad "cytologic atypia": "",} \newline
				\opus{\quad "invasion into fascia": "",} \newline
				\opus{\quad "vascular invasion": "",} \newline
				\opus{\quad "necrosis": "",} \newline
				\opus{\quad "tumour size": 0,} \newline
				\opus{\quad "mitotic hpf": "",} \newline
				\opus{\quad "mitotic mm": ""} \newline
				\opus{\}} \newline
				\opus{```}%
			}%
		} \newline \\
		\midrule
		\textbf{[Human] -- Example intro} & \multicolumn{2}{c}{\parbox{0.75\linewidth}{Below are 1 example of expected input and output, followed by a new task.}} \newline \\
		\midrule
		\textbf{[Human] -- Example user} & \multicolumn{2}{c}{\parbox{0.75\linewidth}{\opus{[...CENSORED...]}}} \newline \\
		\midrule
		\textbf{[AI] -- Example assistant reasoning} & - & - \opus{cytologic atypia} - "monomorfní neoplasie" and "Jádra jsou uniformní" implies Mild \newline
		- \opus{invasion into fascia} - "tumour infiltruje svalovinu žaludku a tukovou tkáň omenta" and "v recidivách je přítomna infiltrace tukové tkáně" implies Yes \newline
		- \opus{vascular invasion} - no mention implies missing \newline
		- \opus{necrosis} - "Nekrózy nejsou přítomny" implies No \newline
		- \opus{tumour size} - Mentioned tumour sizes are: "160x130x90 mm", "40 mm", "100 mm" and "160 mm". The largest dimension is 160 mm \newline
		- \opus{mitotic hpf} - Mentions of mitotic activity in HPF: "2mit/50HPF", "2mit/50HPF". Highest is 2mit/50HPF \newline
		- \opus{mitotic mm} - Mentions of mitotic activity in mm2: none implies missing \newline \\
		\midrule
		\textbf{[AI] -- Example assistant output} & \multicolumn{2}{c}{\parbox{0.75\linewidth}{
				\opus{```json} \newline
				\opus{\{} \newline
				\opus{\quad "cytologic atypia": "Moderate", "invasion into fascia": "Yes",} \newline
				\opus{\quad "vascular invasion": "No", "necrosis": "Yes",} \newline
				\opus{\quad "tumour size": 75.0, "mitotic hpf": "missing",} \newline
				\opus{\quad "mitotic mm": "0-1mitosa/5mm2"} \newline
				\opus{\}} \newline
				\opus{```}%
			}%
		} \newline \\
		\midrule
		\textbf{[Human] -- Report instructions} & \multicolumn{2}{c}{\parbox{0.75\linewidth}{[file name]: \opus{[...CENSORED...]} \newline \opus{[...CENSORED...]}}} \newline \\
		\midrule
		\textbf{[Human] -- Final instructions} &
		Begin the extraction now. Your response must contain only a single valid JSON block, enclosed in triple backticks and prefixed with \verb|`json`|, like this: \verb|```json  ... ```|& 
		Begin the extraction now. First, reason step-by-step to identify and justify the value for each required field, enclosed within \verb|<think>...</think>| tags. Then, output only the final structured data as a single valid JSON block, starting with \verb|```json| and ending with \verb|```|."
		\\
		\bottomrule
	\end{longtable}
\end{scriptsize}

\subsubsection{Prompt Graph}
The dependencies and extraction order of variables are represented as a directed acyclic graph. This graph reflects how the extraction task is decomposed into smaller, sequential subtasks for the Prompt Graph prompting strategy. 

\begin{figure}[htbp]
	\centering
	\includegraphics[width=\linewidth, height=0.5\textheight, keepaspectratio]{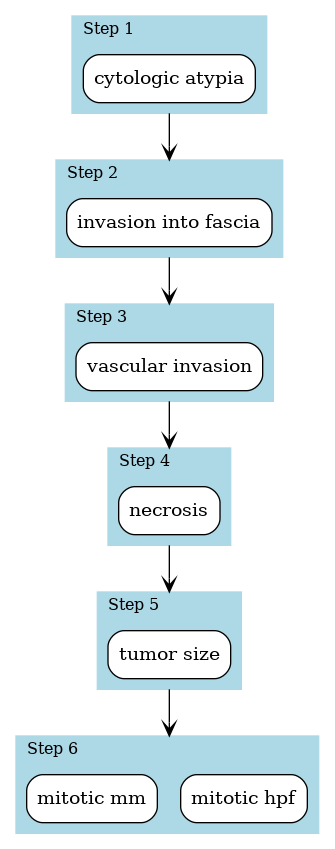}
	\caption{Directed acyclic graph showing sequential extraction order of variable extraction for sarcomas use case.}
\end{figure}

\newpage
\suppsection{Evaluation Metrics}\label{sec:LLMSup2}
To evaluate outputs generated by large language models (LLMs), we compare the model predictions against reference standard annotations using metrics that are defined separately for each variable. These variable-specific metrics are specified in a structured configuration file in YAML format, which serves to standardize the evaluation across all variables.

\subsection{Per-variable Metrics}

These metrics assess model performance at the level of individual variables, ensuring that categorical, numerical, and free-text outputs are evaluated using methods suited to their respective data types.

\subsubsection{Categorical Fields}

For categorical fields, we compute \opus{balanced accuracy}, defined as the average of recall across all categories:
\begin{equation}
\text{BalancedAccuracy}_v = \frac{1}{C_v} \sum_{c=1}^{C_v} 
\frac{TP_c}{TP_c + FN_c}
\end{equation}

\noindent where:
\begin{itemize}
	\item $C_v$ is the number of categories for variable $v$,
	\item $TP_c$ is the number of true positives for category $c$,
	\item $FN_c$ is the number of false negatives for category $c$.
\end{itemize}

\subsubsection{Numeric Fields}

For numeric fields, we compute \opus{accuracy} based on exact matches between predictions and reference standard labels, which is the average of recall (true positive rate):

\begin{equation}
\text{Accuracy}_v = \frac{1}{N} \sum_{i=1}^{N} \mathbf{1}\!\left(y_{v}^{(i)} = \hat{y}_{v}^{(i)}\right)
\end{equation}

\noindent where:
\begin{itemize}
	\item $y_{v}^{(i)}$ is the reference standard for sample $i$ and variable $v$,
	\item $\hat{y}_{v}^{(i)}$ is the predicted value,
	\item $\mathbf{1}(\cdot)$ is the indicator function,
	\item $N$ is the number of samples.
\end{itemize}

\subsubsection{Free-text Fields}

For free-text fields, we use \opus{cosine similarity} between sentence embeddings generated by a pre-trained sentence transformer model:

\begin{equation}
\text{Similarity}_v = \frac{1}{N} \sum_{i=1}^{N} 
\cos\big( E(y_{v}^{(i)}), E(\hat{y}_{v}^{(i)}) \big)
\end{equation}

\noindent where:
\begin{itemize}
	\item $E(\cdot)$ denotes the embedding function (SBERT model \newline \opus{embeddinggemma-300m-medical})\citeSup{reimers_sentence-bert_2019}. More details on this embedding function are described in \autoref{sec:text_embedding}.
	\item $\cos(u,v)$ is the cosine similarity:
\end{itemize}

\begin{equation}
\cos(u, v) = \frac{u \cdot v}{\|u\| \, \|v\|}
\end{equation}

\subsubsection{List Fields}

For list-type variables, we employ a \opus{symmetric similarity} metric that balances recall (coverage of ground-truth items) and precision (penalization of hallucinated items). The per-sample score is the average of two directional measures: the mean maximum similarity from reference standard to predicted items, and the mean maximum similarity from predicted to reference standard items.

\begin{equation}
\begin{split}
\text{SymmetricSimilarity}_{v}^{(i)} = \tfrac{1}{2} \Bigg( 
&\frac{1}{|G^{(i)}|} \sum_{g \in G^{(i)}} 
\max_{p \in P^{(i)}} \cos(E(g), E(p)) \\
+\;&\frac{1}{|P^{(i)}|} \sum_{p \in P^{(i)}} 
\max_{g \in G^{(i)}} \cos(E(p), E(g)) 
\Bigg)
\end{split}
\end{equation}

\noindent where for patient $i$:
\begin{itemize}
	\item $G^{(i)}$ is the set of ground-truth items for variable $v$,
	\item $P^{(i)}$ is the set of predicted items for variable $v$.
\end{itemize}

\noindent Finally, the \opus{symmetric similarity} for variable $v$ across the dataset is obtained by averaging over all $N$ samples:

\begin{equation}
\text{SymmetricSimilarity}_{v} = \frac{1}{N} \sum_{i=1}^{N} \text{Symmetric similarity}_{v}^{(i)}.
\end{equation}

\subsection{Aggregated Metrics}

To provide an overall view of model performance across all variables and scenarios, we compute aggregated metrics that combine the per-variable scores into summary measures.

\subsubsection{Macro Average Score}

To summarize model performance across all variables (denoted $V$), we compute the macro average:

\begin{equation}
\text{MacroAvg} = \frac{1}{V} \sum_{v=1}^{V} S_v
\end{equation}

\noindent where:
\begin{itemize}
	\item $V$ is the total number of variables,
	\item $S_v$ is the evaluation metric for variable $v$ (e.g., accuracy, cosine similarity, or symmetric similarity).
\end{itemize}

\subsubsection{Rank Aggregation}

To identify the best-performing model, we used the Kemeny--Young rank aggregation method \citeSup{kemeny_mathematics_1959_sup}. This method combines several individual rankings, here, the rankings of models across different evaluation metrics and use cases—into one overall consensus ranking. Each metric or use case acts as a ``voter'' that provides its own ranking of the models. The goal is to find the ordering of models that disagrees least with these individual rankings.

For a given candidate ordering $\pi$, the disagreement score is defined as:
\begin{equation}
\text{Score}(\pi) = \sum_{v \in \mathcal{V}} \sum_{i < j} \mathbf{1}\!\Big( \pi(i) < \pi(j) \;\neq\; (v(i) < v(j)) \Big),
\end{equation}
\noindent where:
\begin{itemize}
	\item $\mathcal{V}$ is the set of voters (evaluation metrics and use cases),
	\item $v(i)$ is the rank position of model $i$ according to voter $v$,
	\item $\pi(i)$ is the position of model $i$ in the aggregated ordering $\pi$,
	\item $\mathbf{1}(\cdot)$ is the indicator function, equal to 1 when the pairwise order between $i$ and $j$ disagrees between $\pi$ and $v$.
\end{itemize}

\noindent The final consensus ranking $\pi^*$ minimizes the total disagreement:
\begin{equation}
\pi^* = \arg\min_{\pi} \text{Score}(\pi).
\end{equation}

\noindent Since the search space grows factorially with the number of models, we used a hybrid optimization approach:
\begin{itemize}
	\item For small numbers of models ($n! \le 10^9$), all permutations were exhaustively evaluated (\emph{brute-force search}).
	\item For larger sets, we applied a \emph{simulated annealing} procedure with random segment reversals to explore the space of permutations efficiently \citeSup{kirkpatrick_optimization_1983}.
\end{itemize}

\noindent The algorithm stops early if an exact consensus (score = 0) is reached. For simulated annealing, we used \opus{max\_iter = 50{,}000}, \opus{initial\_temp = 100{,}000}, and \opus{cooling\_rate = 0.99}, following common practice where a high initial temperature and gradual cooling improve exploration and convergence \citeSup{johnson_optimization_1989}.

\subsection{Text embedding and similarity index} \label{sec:text_embedding}

To compute similarities between free-text fields, we represent each string as a dense embedding vector using pre-trained models from the Sentence Transformers library \citeSup{reimers_sentence-bert_2019}. We evaluated three embedding functions:

\begin{itemize}
	\item \opus{all-MiniLM-L6-v2}: a lightweight, fast model optimized for general semantic similarity tasks \citeSup{reimers_sentence-bert_2019}.
	\item \opus{all-mpnet-base-v2}: a higher-capacity model that captures richer semantic information, often yielding stronger performance on sentence-level similarity benchmarks \citeSup{reimers_sentence-bert_2019}.
	\item \opus{embeddinggemma-300m-medical}: a domain-specific model trained on medical corpora to better capture biomedical terminology and context \citeSup{reimers_sentence-bert_2019,gao2021scaling}.
\end{itemize}

\noindent Similarity between two strings is then computed as the cosine similarity between their embeddings. \autoref{tab:embedding_scores} presents representative examples comparing the performance of the three evaluated embedding functions across different types of variable pairs. Based on these results, we selected \opus{embeddinggemma-300m-medical} as the primary embedding function for computing free-text similarity, as it consistently aligns embeddings with the semantic meaning of medical terms. However, caution is warranted in cases where high semantic similarity does not correspond to clinical equivalence. For example, embeddings may produce high similarity scores between strings such as “No evidence of tumour” and “Evidence of tumour” which convey opposite clinical meanings. In such situations, we recommend using rule-based categorical variables to correctly handle negations, uncertainty expressions, and conditional statements (e.g., “if metastasis is suspected”).

\begin{table}[H]
	\centering
	\caption{Representative examples illustrating the performance of different embedding functions. Values indicate the cosine similarity between the reference standard and predicted free-text fields. Bold numbers highlight the best-performing embedding for each example (and the embedding function chosen). For certain cases, such as “Medical, semantically different” pairs, the lowest similarity is preferable, as high similarity would indicate a failure to distinguish distinct concepts.}
	\label{tab:embedding_scores}
	\begin{tabular}{>{\scriptsize\raggedright\arraybackslash}p{0.13\linewidth}
			>{\scriptsize\raggedright\arraybackslash}p{0.13\linewidth}
			>{\scriptsize\raggedright\arraybackslash}p{0.15\linewidth}
			>{\scriptsize\raggedright\arraybackslash}p{0.12\linewidth}
			>{\scriptsize\raggedright\arraybackslash}p{0.12\linewidth}
			>{\scriptsize\raggedright\arraybackslash}p{0.12\linewidth}}
		\hline
		Example reference standard & Example prediction & Problem type & all-MiniLM-L6-v2 & all-mpnet-base-v2 & \textbf{embedding-gemma-300m-medical} \\
		\hline
		Resection & Resectie & Multilingual (Dutch) & 0.50 & 0.56 & \textbf{0.97} \\
		Moderate & Střední & Multilingual (Czech) & 0.23 & 0.22 & \textbf{0.81} \\
		Leukaemia & Leukemia & Spelling variation / typo & 0.83 & 0.84 & \textbf{0.93} \\
		GIST & Gastro-intestinal stromal tumour & Medical abbreviation & 0.19 & 0.12 & \textbf{0.71} \\
		Renal cell carcinoma & RCC & Medical abbreviation & 0.06 & 0.25 & \textbf{0.74} \\
		CHF & Congestive heart failure & Medical abbreviation & 0.20 & 0.27 & \textbf{0.74} \\
		Heart attack & Myocardial infarction & Synonyms & 0.59 & 0.64 & \textbf{0.88} \\
		Liver metastases & Metastatic lesion in liver & Synonyms & 0.86 & 0.88 & \textbf{0.79} \\
		Cold & Common cold & Ambiguous term / polysemy & 0.63 & 0.55 & \textbf{0.91} \\
		Bladder & Head-and-neck & Medical, semantically different & \textbf{0.19} & 0.23 & 0.77 \\
		Suspicion of tumour & Metastasis & Medical, semantically different & \textbf{0.44} & 0.54 & 0.67 \\
		Lipoma & Atypical lipomatous tumour & Medical, semantically close & 0.76 & \textbf{0.79} & 0.78 \\
		Arm & Upper extremity & Medical, semantically close & 0.41 & 0.58 & \textbf{0.90} \\
		No evidence of tumour & Evidence of tumour & Negation / modifiers & \textbf{0.80} & 0.87 & 0.96 \\
		History of diabetes & No history of diabetes & Negation / modifiers & 0.82 & \textbf{0.56} & 0.97 \\
		Stage II cancer & Stage 2 cancer & Numerical entities & 0.95 & \textbf{0.97} & 0.92 \\
		\hline
	\end{tabular}
\end{table}

\newpage

\suppsection{Tables \& Figures}\label{sec:LLMSup3}

\begin{figure}[p]
	\centering
	\includegraphics[width=\textwidth, height=0.9\textheight, keepaspectratio]{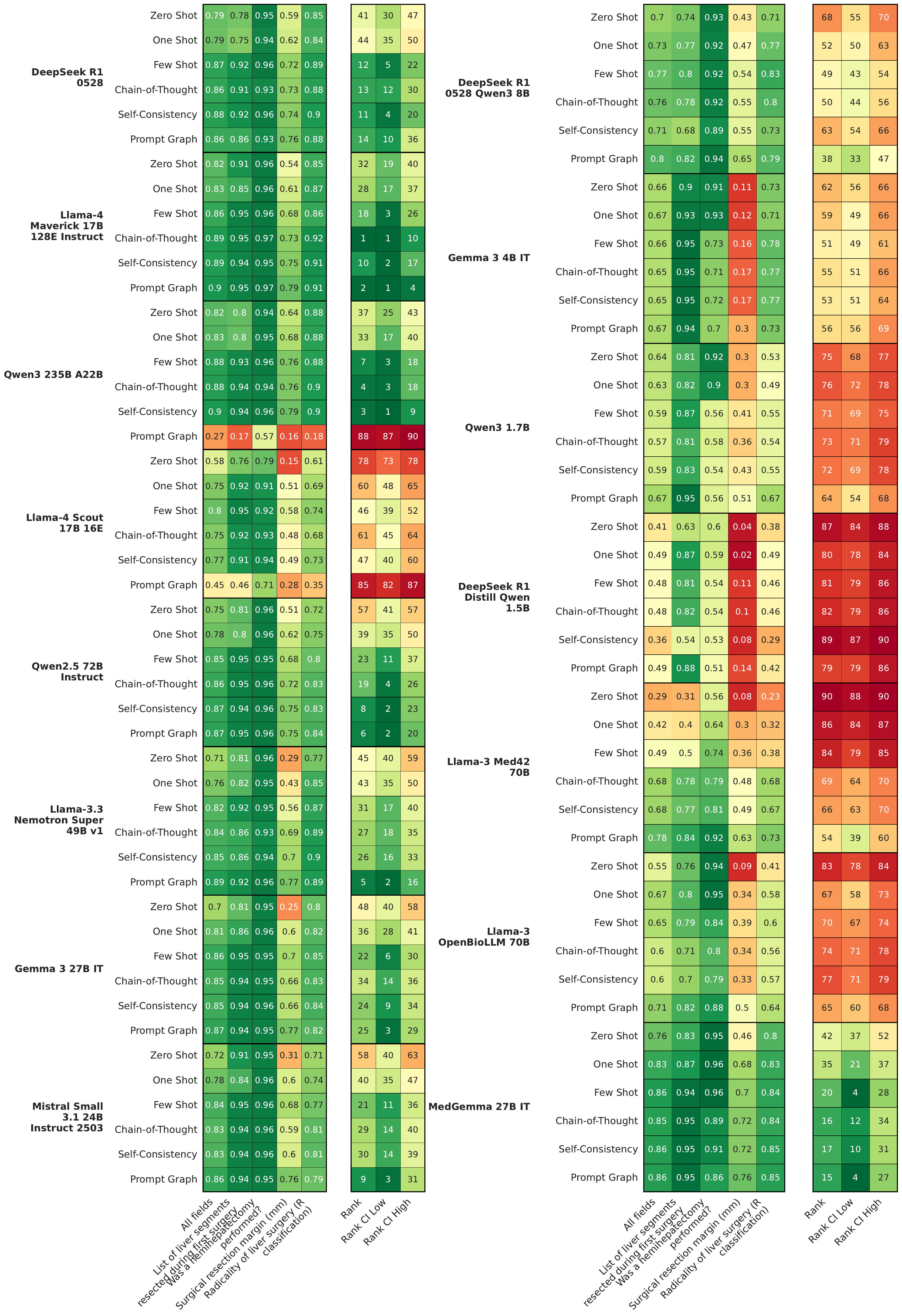}
	\caption{Heatmap of large language model (LLM) performance across prompting strategies for the colorectal liver metastases use case. Each row represents a specific LLM–prompting strategy pair, and each column corresponds to an evaluation field. The first column shows the macro average across all fields, while the final columns presents the aggregated rank including confidence interval based on the Kemeny–Young rank aggregation method.}
	\label{fig:LLMfigS1}
\end{figure}

\begin{figure}[htbp]
	\centering
	\includegraphics[width=\textwidth, height=0.9\textheight, keepaspectratio]{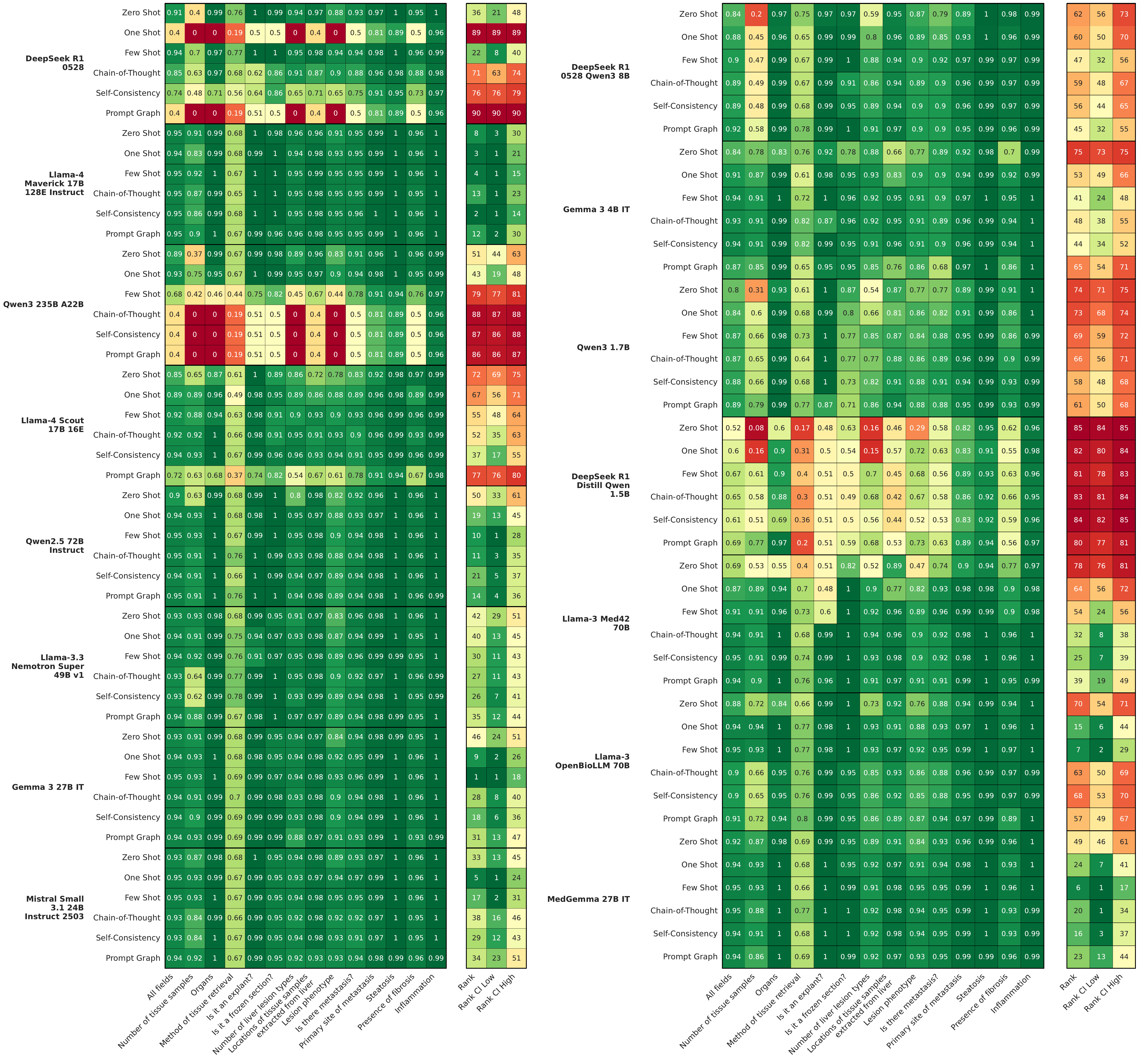}
	\caption{Heatmap of large language model (LLM) performance across prompting strategies for the liver tumour use case. Each row represents a specific LLM–prompting strategy pair, and each column corresponds to an evaluation field. The first column shows the macro average across all fields, while the final columns presents the aggregated rank including confidence interval based on the Kemeny–Young rank aggregation method.}
	\label{fig:LLMfigS2}
\end{figure}

\begin{figure}[htbp]
	\centering
	\includegraphics[width=\textwidth, height=0.9\textheight, keepaspectratio]{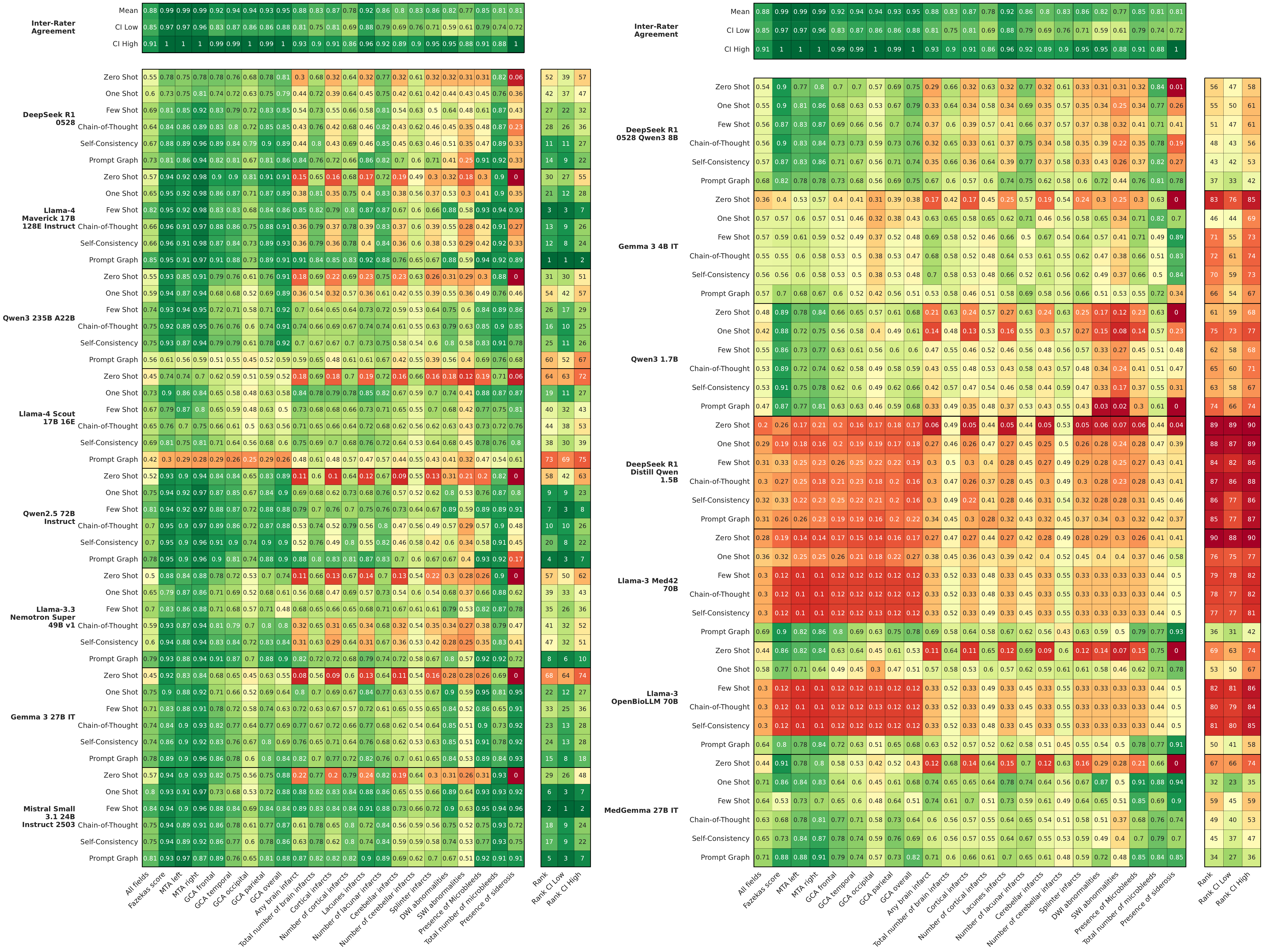}
	\caption{Heatmap of large language model (LLM) performance across prompting strategies for the neurodegenerative diseases use case. Each row represents a specific LLM–prompting strategy pair, and each column corresponds to an evaluation field. The first column shows the macro average across all fields, while the final columns presents the aggregated rank including confidence interval based on the Kemeny–Young rank aggregation method. The first rows present the inter-rater agreement including confidence interval.}
	\label{fig:LLMfigS3}
\end{figure}

\begin{figure}[htbp]
	\centering
	\includegraphics[width=\textwidth, height=0.9\textheight, keepaspectratio]{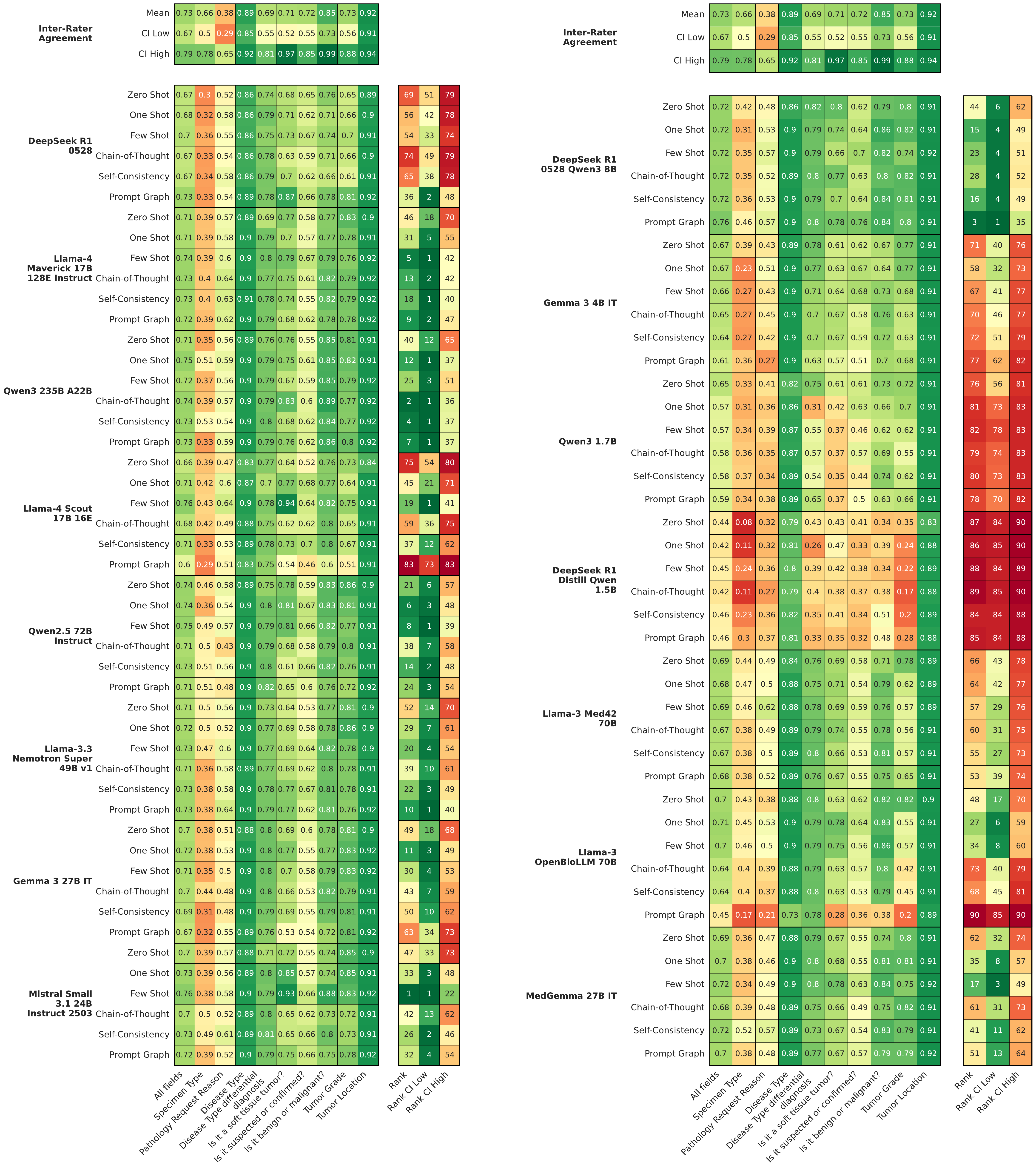}
	\caption{Heatmap of large language model (LLM) performance across prompting strategies for the English soft tissue tumour use case. Each row represents a specific LLM–prompting strategy pair, and each column corresponds to an evaluation field. The first column shows the macro average across all fields, while the final columns presents the aggregated rank including confidence interval based on the Kemeny–Young rank aggregation method. The first rows present the inter-rater agreement including confidence interval.}
	\label{fig:LLMfigS4}
\end{figure}

\begin{figure}[htbp]
	\centering
	\includegraphics[width=\textwidth, height=0.9\textheight, keepaspectratio]{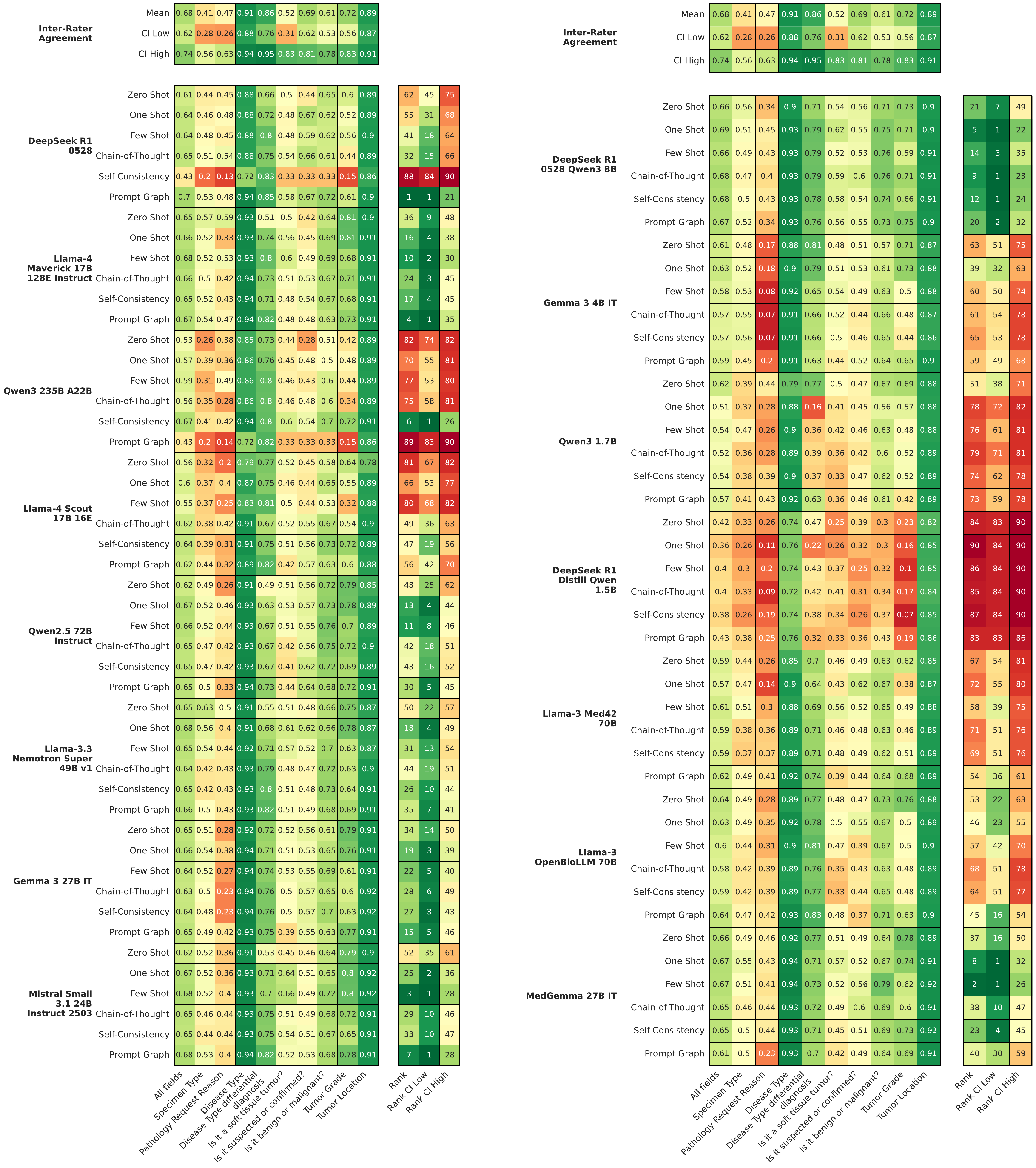}
	\caption{Heatmap of large language model (LLM) performance across prompting strategies for the Dutch soft tissue tumour use case. Each row represents a specific LLM–prompting strategy pair, and each column corresponds to an evaluation field. The first column shows the macro average across all fields, while the final columns presents the aggregated rank including confidence interval based on the Kemeny–Young rank aggregation method. The first rows present the inter-rater agreement including confidence interval.}
	\label{fig:LLMfigS5}
\end{figure}

\begin{figure}[htbp]
	\centering
	\includegraphics[width=\textwidth, height=0.9\textheight, keepaspectratio]{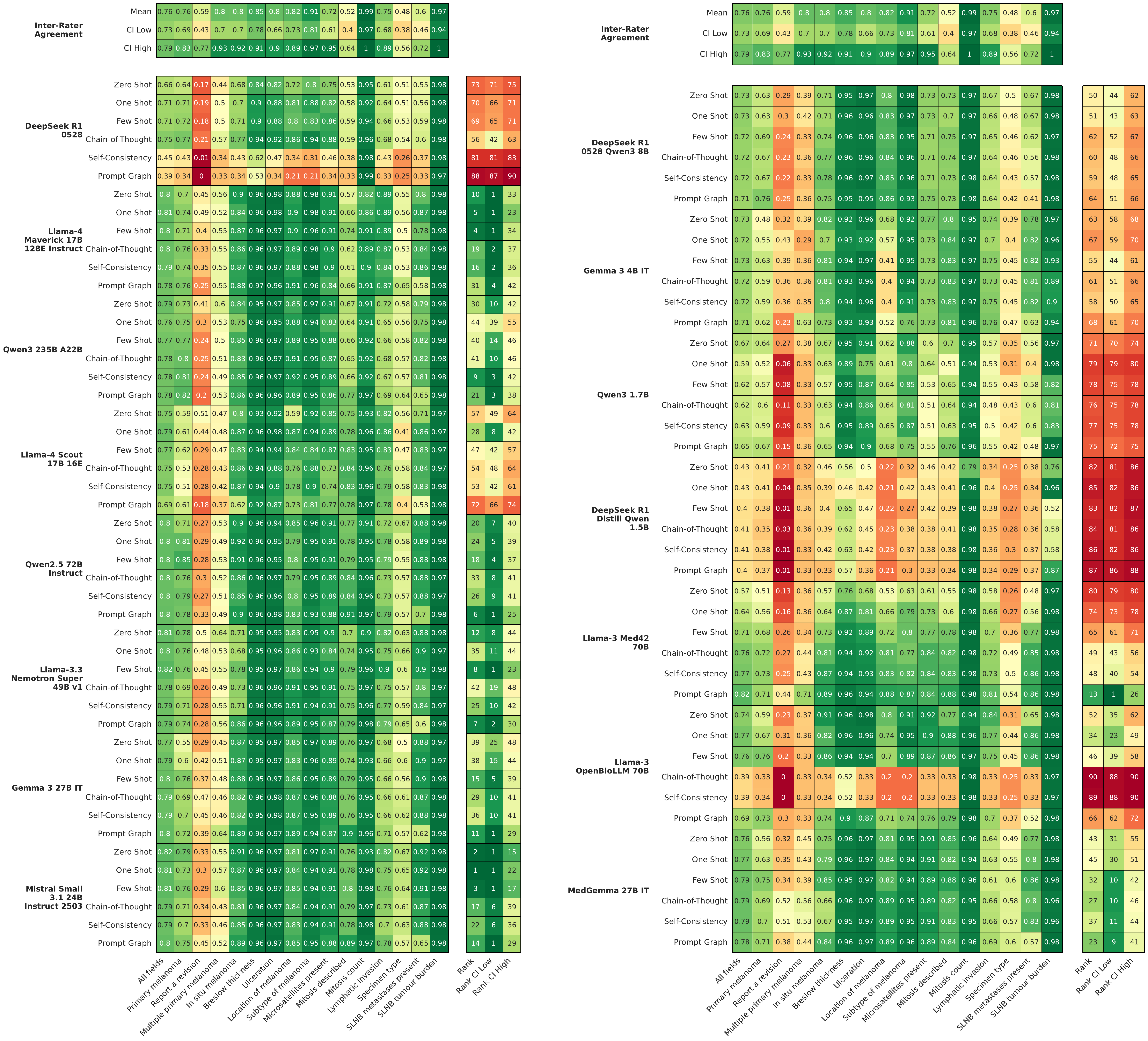}
	\caption{Heatmap of large language model (LLM) performance across prompting strategies for the melanomas use case. Each row represents a specific LLM–prompting strategy pair, and each column corresponds to an evaluation field. The first column shows the macro average across all fields, while the final columns presents the aggregated rank including confidence interval based on the Kemeny–Young rank aggregation method. The first rows present the inter-rater agreement including confidence interval.}
	\label{fig:LLMfigS6}
\end{figure}

\begin{figure}[htbp]
	\centering
	\includegraphics[width=\textwidth, height=0.9\textheight, keepaspectratio]{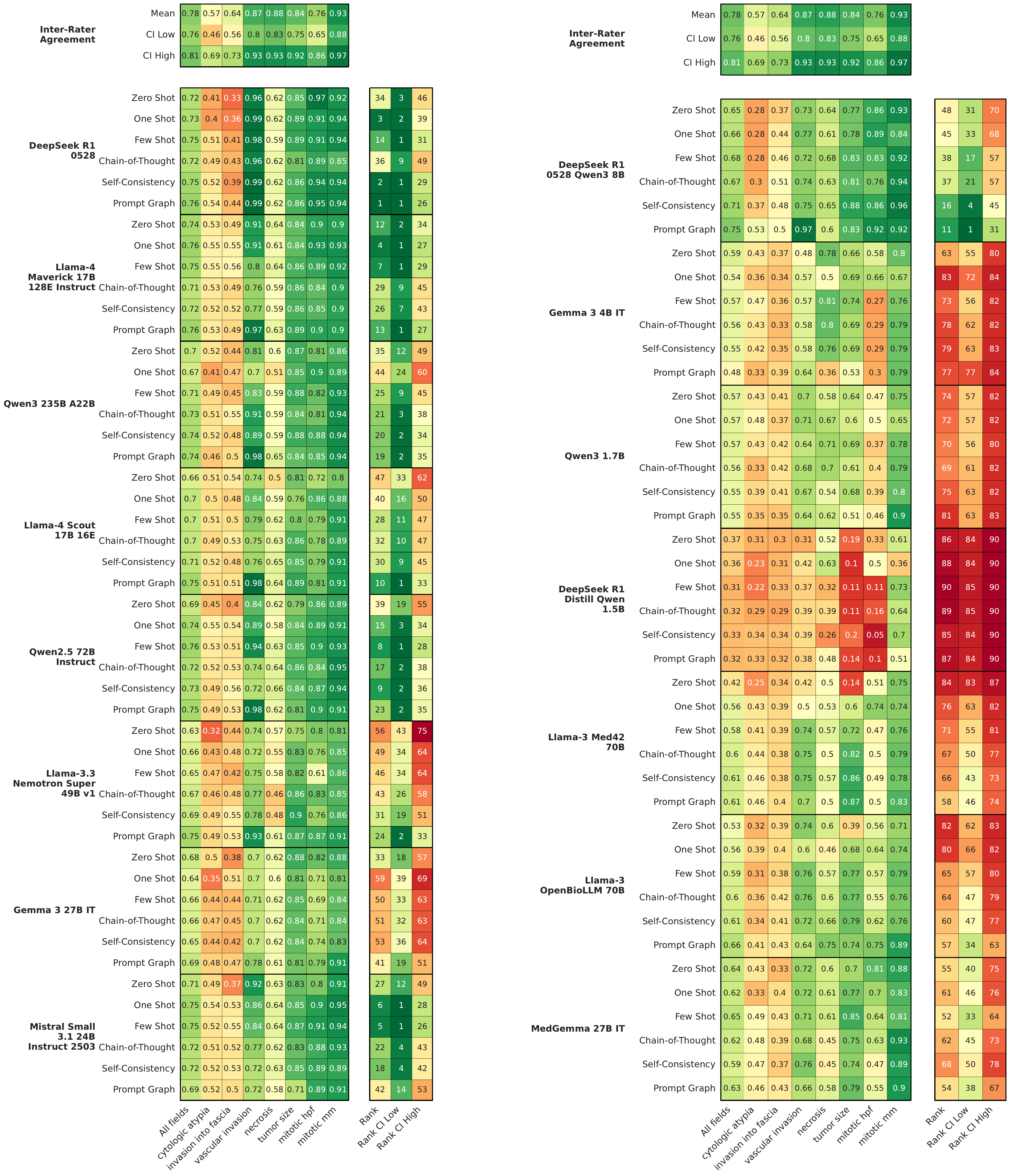}
	\caption{Heatmap of large language model (LLM) performance across prompting strategies for the sarcomas use case. Each row represents a specific LLM–prompting strategy pair, and each column corresponds to an evaluation field. The first column shows the macro average across all fields, while the final columns presents the aggregated rank including confidence interval based on the Kemeny–Young rank aggregation method. The first rows present the inter-rater agreement including confidence interval.}
	\label{fig:LLMfigS7}
\end{figure}

\begin{table}[h!]
	\centering
	\caption{Hyperparameters for the evaluated large language models (LLMs). Values were taken from the official Hugging Face model pages or the corresponding GitHub repositories.}
	\label{tab:llm_hyperparameters}
	\begin{tabular}{>{\scriptsize\raggedright\arraybackslash}m{0.37\linewidth} >{\scriptsize\raggedleft\arraybackslash}m{0.16\linewidth} >{\scriptsize\raggedleft\arraybackslash}m{0.16\linewidth}>{\scriptsize\raggedleft\arraybackslash}m{0.16\linewidth}}
		\hline
		Model Name & Temperature & Top-P & Top-K \\
		\hline
		DeepSeek R1 0528 & 0.60 & 0.95 & 50 \\
		Llama-4-Maverick & 0.60 & 0.95 & 50 \\
		Qwen3 235B A22B & 0.60 & 0.95 & 50 \\
		Llama 4 Scout & 0.60 & 0.95 & 50 \\
		Qwen2.5 72B Instruct & 0.60 & 0.95 & 50 \\
		NVIDIA Nemotron-Super-49B & 0.60 & 0.95 & 50 \\
		Gemma 3 27B & 0.60 & 0.95 & 50 \\
		Mistral Small 3.1 24B Instruct 2503 & 0.15 & 0.95 & 50 \\
		DeepSeek R1 0528 Qwen3 8B & 0.60 & 0.95 & 50 \\
		Gemma3 4B & 0.60 & 0.95 & 50 \\
		Qwen3 1.7B & 0.60 & 0.95 & 50 \\
		DeepSeek R1 Distill Qwen 1.5B & 0.60 & 0.95 & 50 \\
		Llama3-Med42-70B & 0.40 & 0.95 & 50 \\
		Llama3‑OpenBioLLM‑70B & 0.00 & 0.95 & 50 \\
		MedGamma 3 27B & 0.60 & 0.95 & 50 \\
		\hline
	\end{tabular}
\end{table}

\begin{table}[h!]
	\centering
	\caption{Variance partitioning of LLM performance across use cases. Mixed-effects models were used with LLM identity as the grouping factor and prompting strategy as a random effect. The table shows the variance components and the proportion of variance explained by each factor for all evaluated use cases.}
	\label{tab:variance_partitioning}
	\begin{tabular}{>{\scriptsize\raggedright\arraybackslash}p{0.37\linewidth} >{\scriptsize\raggedleft\arraybackslash}p{0.16\linewidth} >{\scriptsize\raggedleft\arraybackslash}p{0.16\linewidth}>{\scriptsize\raggedleft\arraybackslash}p{0.16\linewidth}}
		\hline
		Use Case & LLM contribution (\%) & Prompting strategy contribution (\%) & Residual (\%) \\
		\hline
		Colorectal Liver Metastases (Dutch) & 36.7 & 0.4 & 62.9 \\
		Liver Tumours (Dutch) & 23.9 & 30.8 & 45.4 \\
		Neurodegenerative Diseases (Dutch) & 18.6 & 35.0 & 46.5 \\
		Soft Tissue Tumours (English) & 20.8 & 0.7 & 78.6 \\
		Soft Tissue Tumours (Dutch) & 28.4 & 1.9 & 69.7 \\
		Melanomas (Dutch) & 17.9 & 1.7 & 80.5 \\
		Sarcomas (Czech) & 21.5 & 0.2 & 78.3 \\
		\hline
	\end{tabular}
\end{table}

% ----------------------------
% Separate Supplementary Bibliography
% ----------------------------
\bibliographystyleSup{elsarticle-num}
\bibliographySup{references_sup}

\end{appendices}
\end{document}